\documentclass[a4paper,plainchapterheads,yschapters,twoside,truedoublelespace,openright]{iitbthesis}
\usepackage{epsfig}
\usepackage[T1]{fontenc}
\usepackage[utf8]{inputenc}
\usepackage{mathptmx}
\usepackage{amssymb}
\usepackage{xcolor}
\usepackage{wrapfig}
\usepackage{amsmath,epsfig}
\usepackage{graphicx,graphics}
\usepackage{url}
\usepackage{hyperref}
\usepackage[capitalise]{cleveref}
\usepackage{textcomp}
\usepackage{enumitem}
\usepackage{natbib}
\usepackage{pdflscape}
\renewcommand\bibname{References}
\usepackage{titlesec}
\usepackage{bm,bbm}
\usepackage{float,lscape}
\usepackage{subfig}

\usepackage{epstopdf}
\usepackage{tabu}
\usepackage{multirow}
\usepackage{array}
\usepackage{booktabs}
\usepackage{graphicx}
\usepackage{caption}
\usepackage{lettrine}
\usepackage{enumitem}
\usepackage{lscape}
\usepackage{rotating}
\usepackage{booktabs}
\usepackage{longtable}
\usepackage{mathptmx}
\usepackage{anyfontsize}
\usepackage{t1enc}
\usepackage{relsize}
\usepackage{placeins}
\usepackage{nomencl}
\usepackage{rotating}
\usepackage{glossaries}
\usepackage{scrlayer}
\usepackage{enumitem}
\usepackage{datetime}
\newdateformat{monthyeardate}{%
  \monthname[\THEMONTH], \THEYEAR}

\usepackage{gensymb}
\usepackage{amssymb}
\usepackage{pifont}
\newcommand{\cmark}{\ding{51}}
\newcommand{\xmark}{\ding{55}}

\usepackage[titletoc]{appendix}
\usepackage[nottoc]{tocbibind}

\begin{document}

\abovedisplayskip=8mm
\abovedisplayshortskip=8mm
\belowdisplayskip=8mm
\belowdisplayshortskip=8mm



\pagenumbering{gobble}

\title{Hyperspectral Image Analysis in Single-Modal and Multimodal setting using Deep Learning Techniques}
\author{Shivam Pande}
\date{2023}

\rollnum{184314002} 

\iitbdegree{Doctor of Philosophy}

\reporttype{}

\department{Centre of Studies in Resources Engineering}

\setguide{Prof. Biplab Banerjee}

\maketitle


\begin{dedication}
\large{\textit{Dedicated to the researcher's spirit and the perpetual quest for knowledge}}
\end{dedication}


\chapter*{}
\thispagestyle{empty}
\begin{center}
{\Huge  {\bf Thesis Approval}}
\end{center}

\begin{figure*}[h!]
  \centering
  \centerline{\includegraphics[width=\textwidth]{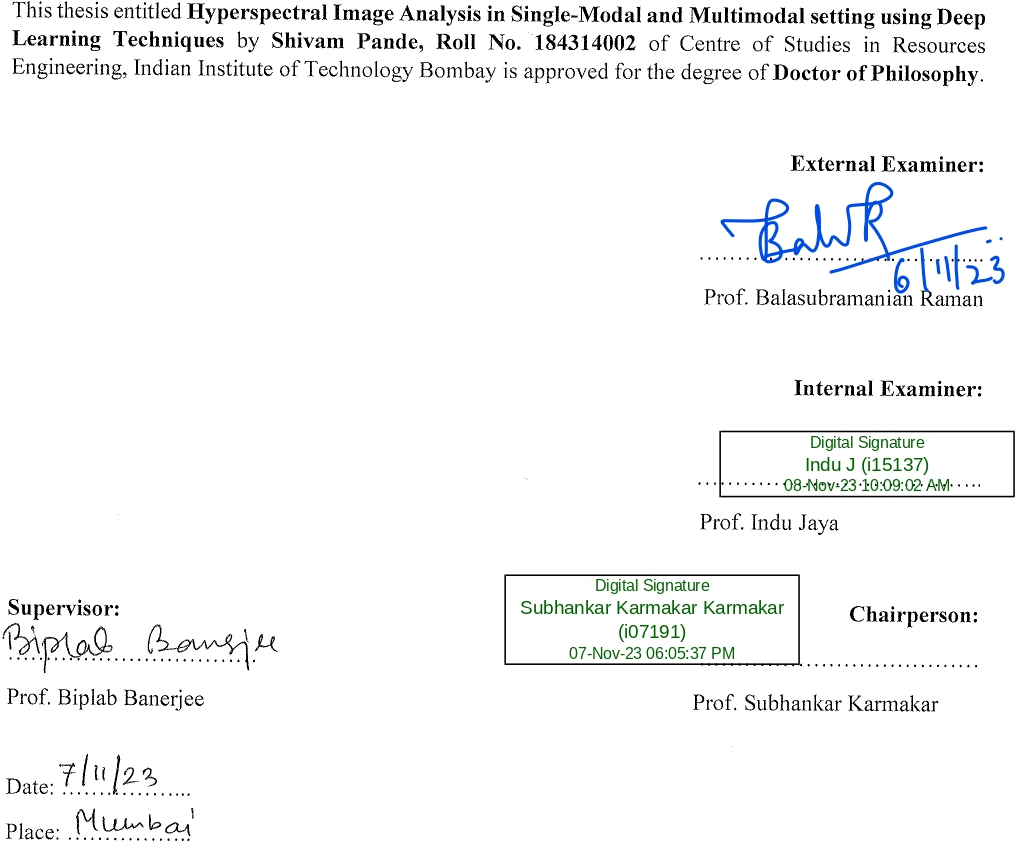}}
\end{figure*}




\newpage
\begin{center}
{\Huge \textbf{Declaration}} 
\end{center}
\vspace*{1cm}  

\begin{figure*}[h!]
  \centering
  \centerline{\includegraphics[width=\textwidth]{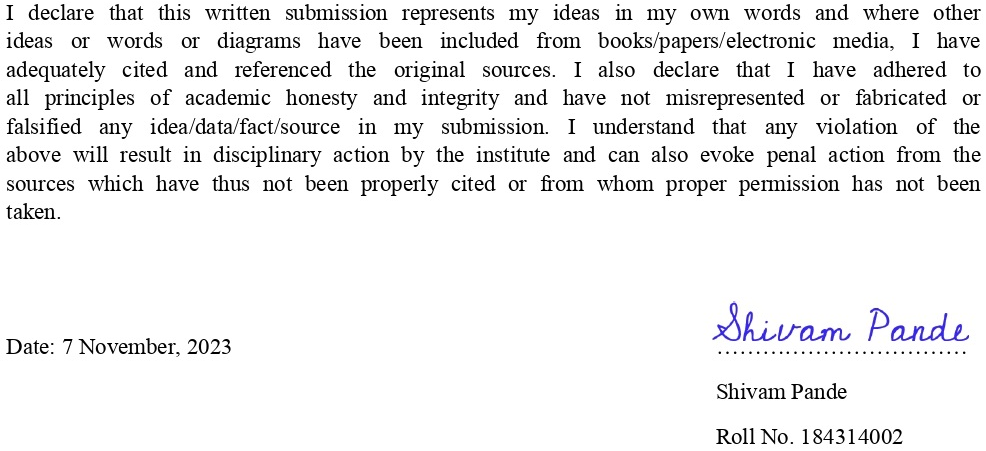}}
\end{figure*}

\joiningdate{12 July 2018}
\begin{coursecertificate}
\addcourse{CS 725}{Foundations of Machine Learning}{6}
\addcourse{GNR 607}{Principles of Satellite Image Processing }{6}
\addcourse{GNR 627}{Geospatial Predictive Modelling }{6}
\addcourse{GNR 653}{Data Analysis Methods for Geospatial Applications }{6}
\addcourse{GNR S01}{Seminar}{4}
\addppcourse{HS 791}{Communication Skills-I}{PP}
\addppcourse{ES 792}{Communication Skills-II}{PP}
\addppcourse{GC 101}{Gender in the workplace }{PP}
\addppcourse{GNR 638}{ Machine Learning for Remote Sensing - II}{AU}
\end{coursecertificate}


\clearpage
\pagenumbering{roman}
\begin{abstract}
  \renewcommand{\thepage}{\roman{page}} \setcounter{page}{1}

The field of remote sensing has undergone rapid development in recent years, driven by advancements in sensing technologies. This has led to the acquisition of data from various modalities, including multispectral images (MSI), hyperspectral images (HSI), synthetic aperture radar (SAR), light detection and ranging (LiDAR), and others. Each modality possesses distinct characteristics and focuses on different aspects of earth observation. In this research, the primary emphasis lies on HSI and their processing using deep learning techniques, particularly for land-use/land-cover classification tasks. HSI exhibits remarkable spectral resolution compared to other modalities, thanks to its narrow sampling interval. Consequently, it is well-suited for precise colour/spectral identification in spectrally overlapping classes, such as vegetation and crops, roads and roofs, and more. However, the high spectral resolution of HSI means a large number of channels, increasing the dimensionality of the images. This poses difficulties for supervised tasks like classification and detection, as they require a substantial number of training samples. Furthermore, the higher spectral resolution often comes at the expense of reduced spatial resolution due to the division of the electromagnetic spectrum among multiple wavelengths. Moreover, there are additional challenges of a more general nature. For instance, disparities in the geographical distribution between training and test sets can result in distribution differences during evaluation. Additionally, there may be sensor-related issues, where a sensor utilized during training may not be available during test time.

To address these challenges, the research employs deep learning techniques for efficient data processing and feature extraction, with a primary focus on end-to-end feature extraction and classification. To mitigate the limitations imposed by low spatial resolution, a multimodal learning approach is adopted, leveraging complementary information from other modalities such as LiDAR and SAR for joint feature extraction and classification. Simultaneously, the research work tackles the issue of domain differences and missing modalities by incorporating adversarial learning and knowledge distillation. It is worth noting that even with the utilization of deep learning techniques, it is crucial to tailor these techniques to the specific characteristics of the data. For example, the continuous nature of the spectral dimension in HSI necessitates the utilization of sequential methods such as 1D convolutional neural networks (CNNs) and recurrent neural networks (RNNs) for effective processing. Furthermore, techniques from the visual attention domain and feedback connections are employed to facilitate superior and robust feature extraction. These techniques effectively reduce the number of parameters in neural network architectures while focusing on the relevant parts of the HSI.

Lastly, to overcome the limitation of limited training samples, experiments are conducted in a self-supervised learning scenario. Various types of autoencoders are developed to enable deep dimensionality reduction in HSI. Moreover, label-guided and contrastive learning-based semi-supervised techniques are explored to enhance feature representation using unlabeled samples and enable cross-modal and cross-sample reconstruction. The proposed methods are evaluated on multiple HSI datasets, including Indian Pines 1992, Indian Pines 2010, Salinas Valley, Botswana, Pavia Centre and University datasets, HSI and LiDAR datasets such as Houston datasets (from data fusion contests in 2013 and 2018), Trento and MUUFL Gulfport datasets, as well as HSI-SAR datasets like TU Berlin. In all cases, the proposed methods demonstrate significant performance improvements, surpassing several state-of-the-art techniques.
\end{abstract}

\tableofcontents
\listoftables
\listoffigures



%
%


\chapter*{List of Abbreviations}
\label{ch:ListOfAbr}
\addcontentsline{toc}{chapter}{\nameref{ch:ListOfAbr}}

\makeatletter
\newcommand{\tocfill}{\cleaders\hbox{}\hfill}
\makeatother
\newcommand{\abbrlabel}[1]{\makebox[3cm][l]{\textbf{#1}\ \tocfill}}
\newenvironment{abbreviations}{\begin{list}{}{\renewcommand{\makelabel}{\abbrlabel}
                                              \setlength{\itemsep}{0pt}}}{\end{list}}
\begin{abbreviations}

\item[AttAE] Attention based Autoencoder
\item[BiGRU] Bidirectional Gated Recurrent Units
\item[CCF]  Canonical Correlation Forests 
\item[CNN]  Convolutional Neural Network
\item[DFC]  Data Fusion Contest
\item[DSM]  Digital Surface Model
\item[FBAE] Feedback Connections based Autoencoder
\item[GAN]  Generative Adversarial Network
\item[GAP]  Global Average Pooling
\item[GNN]  Graph Neural Network 
\item[GRU]  Gated Recurrent Units
\item[HRW]  Hierarchical Random Walk
\item[HSI]  Hyperspectral Images
\item[LiDAR]  Light Detection and Ranging
\item[LR]  Low Resolution
\item[LSTM]  Long Short Term Memory
\item[LULC]  Land Use/Land Cover
\item[MAHiDFNet]  Multi-Attentive Hierarchical Dense Fusion Net
\item[MCSL] Multiscale Coupled Self-Looping
\item[MLP] Multilayer Perceptron
\item[OA] Overall Accuracy 
\item[PA] Producer's Accuracy 
\item[ReLU]  Rectified Linear Unit
\item[RF]  Random Forests 
\item[RNN]  Recurrent Neural Network 
\item[SAR]  Synthetic Aperture Radar 
\item[SSL]  Self-supervised Learning 
\item[UDA]  Unsupervised Domain Adaptation
\item[VHR]  Very High Resolution

\end{abbreviations}

\setlength{\parskip}{2.5mm}
\titlespacing{\chapter}{0cm}{55mm}{10mm}
\titleformat{\chapter}[display]
  {\normalfont\huge\bfseries\centering}
  {\chaptertitlename\ \thechapter}{20pt}{\Huge}
  
  \titlespacing*{\section}
  {0pt}{8mm}{8mm}
  \titlespacing*{\subsection}
  {0pt}{8mm}{8mm}
\pagebreak
\pagenumbering{arabic}

\makeatletter
\def\cleardoublepage{\clearpage\if@twoside \ifodd\c@page\else
	\hbox{}
	\vspace*{\fill}
	\begin{center}
		This page was intentionally left blank.
	\end{center}
	\vspace{\fill}
	\thispagestyle{empty}
	\newpage
	\if@twocolumn\hbox{}\newpage\fi\fi\fi}
\makeatother

\newpage
\pagebreak
\cleardoublepage
\chapter{Introduction}

\begin{figure}
    \centering
    \includegraphics[width=0.8\textwidth]{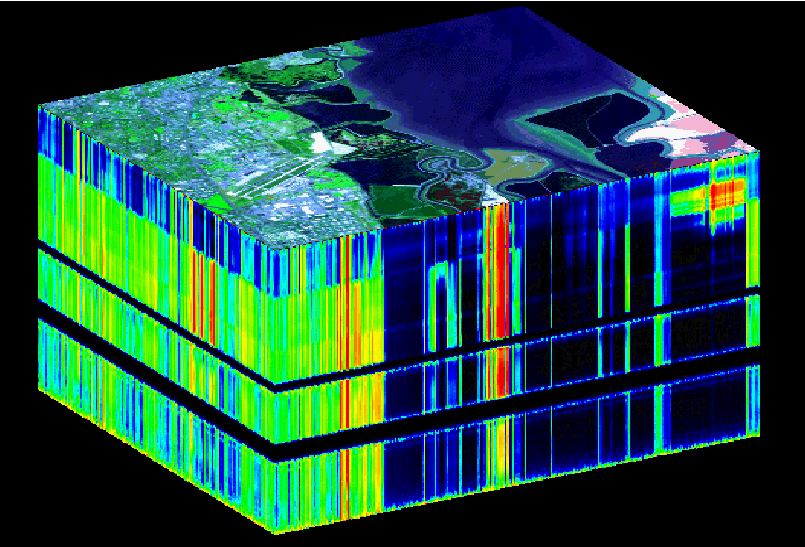}
    \caption{Illustration of a Hyperspectral Cube to showcase the contiguous spectrum of the multiple channels \citep{yusuf2018survey}.}
    \label{fig:HypCube}
\end{figure}

Hyperspectral imaging has been one of the most prominent remote sensing techniques where the imagery of terrain and its features are obtained through hyperspectral sensors fitted in imaging satellites or drones. The idea behind the hyperspectral imaging is to utilize the maximum available wavelengths of the electromagnetic spectrum so as to get more information in the obtained image (at least in the spectral domain). Owing to this, the sampling difference between the different wavelengths is very small resulting in a large number of bands (even upto 2500) that are contiguous in nature. Thus, each pixel in the image represents a multidimensional vector with dimensions equal to number of bands \citep{prasad2020hyperspectral}. Figure \ref{fig:HypCube} \citep{yusuf2018survey} shows a conventional example of a hyperspectral cube.

Though, the massive spectral information contained in the HSI imagery is of significant value in several tasks such as per pixel image classification, there are several associated challenges as well such as \citep{li2019deep}:

\begin{itemize}
    \item The contiguous nature of bands might result in very high correlation among the bands.
    \item The training pixels might be very few leading to an underdetermined problem in classification.
    \item There is large spatial variability among the spectral signatures.
    \item The spatial resolution of the pixels in generally not as high as corresponding multispectral/ panchromatic counterpart.
    \item There may be inherent domain differences between the training and test domains. 
\end{itemize}

In the initial stages, there were several efforts to exploit the spectral variability of the bands using conventional machine learning techniques (Gaussian Maximum Likelihood (GML), Support Vector Machines (SVMs), Random Forest (RF) etc.) to carry out efficient image classification. In addition to using the spectral information, there were several approaches to tackle the problem of high dimensionality. Techniques such as Principal Component Analysis (PCA), Linear Discriminant Analysis (LDA), Decision Boundary Feature Extraction (DBFE) and Non-Weighted Feature Extraction (NWFE) have been used the dimensionality of the HSI data. However, the spectral component of the imagery can only do so much since it does not take into account the placement of a pixel with respect to other pixels and hence does not include any context information.  Hence, there arose the need to somehow incorporate the contextual information in the image as well along with the spectral information. The spatial/contextual information is derived either as morphological profiles, attribute profiles, extinction profiles or textural profiles that are created by subjecting the original image to several convolution operations. The other method of achieving the same is to create superpixels by segmenting the original image into several homogeneous areas for each class. These features when combined with spectral features, give relatively higher accuracy and better performance \citep{ablin2013survey}. 

Besides dealing with the problem of high dimensionality, the problem of low spatial resolution is also tackled by combing the features obtained from a high resolution MS/PAN image or data from other modality such as LiDAR or SAR. The features could be combined either through concatenation or synergistically through fusion \citep{mohla2020fusatnet}. The resulting features thus created exhibit relatively higher spectral and spatial resolutions and provide better classification accuracies.  

However, most of the techniques mentioned above require expert supervision and detailed knowledge of the geographical area. Since a lot of dependency is on the hand crafted features/ human intervention, the low intraclass and high interclass variations are not captured efficiently among the classes. Hence, to make the entire process automatic, the deep learning models are brought into the scenario. The models are capable of extracting the low, medium and high level features from the imageries through various hierarchical layers. Since the process is automatic, it is not limited to specific task and can be applied to a wide variety of situations.

Deep learning (DL) is now being used in large number of applications pertaining to HSI analysis. A large number of algorithms derived from the field of computer vision is actively applied on the HS images for the tasks included but not limited to image classification, scene classification, image captioning, image fusion, multimodal learning and segmentation. In addition, DL is also being applied to tackle the problem of dimensionality reduction (autoencoders), data generation and data augmentation (using generative adversarial networks (GANs), variational autencoders (VAEs) etc.) \citep{li2019deep}.

Though there has been phenomenal and extensive work in most of the fields of HSI analysis, there is still a wide research gap in the areas such as classification using unsupervised domain adaptation, multimodal learning for missing band prediction and classification, and multimodal/multisource fusion. Additionally, due to limited label scenarios, it is also required to research on the semi-supervised and self-supervised learning based techniques. In the subsequent section, a thorough literature review is presented for the multiple techniques used in remote sensing domain, primarily for HSI analysis. 

\section{Literature review}

This section presents a review of the significant available literature for hyperspectral image analysis from single-modal, multimodal and self-supervised settings. In this section and the subsequent subsections, we will see an overview of landuse/land cover (LULC) classification in remote sensing domain. Additionally, we will also see well defined architectures such as CNNs and their application in the analysis of remotely sensed images, particularly hyperspectral images. Additionally, we will discuss the incorporation of domain adaptation frameworks, feedback connections, and attention mechanisms into neural networks for various applications. From classification perspective, the focus will also be on the concepts of domain adaptation. We will also observe their significant contributions to the computer vision community. Additionally, we will also discuss the importance of loss functions for classification, as prevalent in the existing literature. 

From multimodal learning perspective, there is discussion presented over the existing techniques to combine the information from multiple remote sensing modalities, with the focus on the existing research gaps. Additionally, we also highlight the limited research work in the domain of missing and sparse modalities. In addition to this, the paradigm of self-supervised learning (SSL) is simultaneously discussed in conjunction with the remote sensing image classification. The review presents a vast array of techniques from generative, predictive and contrastive SSL perspectives applied over different remote sensing modalities. 

\subsection{LULC classification with remotely sensed data}

One of the central applications that is generally addressed using remote sensing data is LULC classification. It is owing to the fact that classification in itself is an independent task for multiple applications like disease identification in crops, finding regions affected by hazards and several others. It is simultaneously a prerequisite in multiple other tasks such as object detection, instance and panoptic segmentation \citep{kirillov2019panoptic}, change detection \citep{saha2020change}, image and shape retrieval \citep{chaudhuri2019siamese, pal2024domain} and several others \citep{zhang2022artificial}. Historically, there have been several techniques from conventional machine learning domain, that tackle LULC classification using single or multiple modalities from remote sensing domain. For instance \cite{pal2005support} presented one of the initial works, where they classified optical (Landset 7) and hyperspectral (DAIS) data using support vector machines (SVM) and compared their methods against maximum likelihood method and artificial neural networks. Similarly, \cite{sukawattanavijit2017ga} proposed a genetic algorithm coupled with CNNs for joint classification of SAR data (from THEOSAT and RADARSAT-2) and optical data (Landsat 8 and RS2). Another technique, namely random forest was introduced by \cite{pal2005random} for LULC classification of Landsat ETM+ data, and the classification was compared against SVM classifier. From, hyperspectral images domain, \cite{xia2017random} introduces a method of LULC classification on hyperspectral images (HSI), where random forests were used on derived HSI features using multi-extinction profiles for classification task. \cite{xia2017hyperspectral} extended the idea of ensembling further for HSI classification, and utilised the concept of oblique decision trees, where the transformation factors were calculated using partial least squares, and the ensembling was carried out. However, given the limited feature extraction capacity of traditional methods along with excessive feature handcrafting, deep learning techniques entered the foray. In the subsequent sections, we will see several such techniques from image processing and remote sensing perspectives. 

\subsection{Convolutional architectures}

Since the emergence of deep learning, particularly for image analysis, there have been significant advancements in network engineering. Traditional CNN architectures like VGGNet \citep{wang2015places205} and AlexNet \citep{alom2018history} have evolved to include networks that support wider and multi-scale features, such as GoogleNet \citep{ballester2016performance} and Inception Networks \citep{szegedy2017inception}. These networks have demonstrated exceptional performance and have been widely adopted in various application domains, including medical image analysis \citep{suzuki2017overview} and remote sensing \citep{zhu2017deep}. In the field of remote sensing, extensive research has focused on image classification, object detection, and change detection. For instance, \citep{zhang2018hybrid} proposed a hybrid approach that combines multilayer perceptron (MLP) and CNN for image classification from aerial photographs, leveraging the spatial aspect through CNN and spectral features through MLP. Similarly, \citep{wang2018change} employed faster R-CNNs for change detection in remote sensing images, achieving a significant reduction in false positives. \citep{liu2020multiscale} introduced a multiscale U-shaped CNN for building extraction from high-resolution RGB remote sensing images, successfully capturing buildings of different sizes using multiscale features. CNNs have also found applications in the hyperspectral image (HSI) domain. \citep{chen2016deep} conducted comprehensive research on the use of 1D, 2D, and 3D CNNs for HSI classification. Furthermore, efforts have been made to combine different CNN architectures, as demonstrated in \citep{roy2019hybridsn}, where 3D and 2D CNNs are sequentially employed within the same network for HSI classification. Various studies, such as \citep{he2018feature, gong2019cnn}, have explored the impact of network width by introducing parallel convolution filters of different sizes to encompass multiple spatial levels of HSI feature representations. Additionally, \citep{xu2017multisource} proposed a dual branch architecture that integrates information from two modalities, hyperspectral and LiDAR data, and jointly trains the network for pixel classification. To improve efficiency, \citep{meng2021lightweight} introduced a network with lightweight modules consisting of depthwise separable convolutions for HSI classification, enabling faster inference and reduced memory consumption. 

\subsection{Convolutional architectures with residual connections}

To address the challenges of vanishing and exploding gradients when constructing deeper models, various modifications have been introduced. For instance, \citep{srivastava2015highway} proposed highway networks that utilize gating mechanisms, similar to long short-term memory (LSTM), to control the information flow across the network. Similarly, models like ResNet and ResNext were developed, incorporating skip connections to ensure gradient flow \citep{szegedy2017inception}. Another approach is FractalNet, introduced by \citep{larsson2016fractalnet}, where multiple networks of variable depth are used within a single network to stabilize gradient flow, instead of skip connections. Building upon the ResNet concept, a newer architecture called DenseNet was introduced in \citep{huang2017densely}, replacing identity mappings with concatenation to reuse previous features. These advancements have been extensively applied in the field of remote sensing. For example, \citep{huang2019ship} proposed a ship detection method using a feature pyramid network enhanced by skip connections and a squeeze and excitation module \citep{hu2018squeeze}. Similarly, \citep{li2019adaptive} introduced an adaptive feature fusion network for remote sensing image classification, where individual streams in the network extract features at different scales using residual convolution modules, which are adaptively combined based on their contribution to image classification. Residual architectures have also been utilized in hyperspectral image (HSI) analysis. \citep{zhong2017deep} introduced a simple 4-layered CNN with skip connections for HSI classification. Taking it further, \citep{paoletti2018deep} combined residual connections in a pyramidal CNN, jointly leveraging the network for HSI classification, generating features at multiple spatial levels, and ensuring convergence. Skip connections with 3D CNNs have also been explored. \citep{jiang2019hyperspectral} introduced a 3D CNN framework with skip connections and transfer learning for HSI classification. 3D CNNs offer the advantage of jointly extracting spectral and spatial features from HSIs. Transfer learning was applied to avoid overfitting, where the network was initially trained on one dataset and fine-tuned on another dataset. Similarly, \citep{wu2020three} designed an end-to-end 3D ResNext architecture with label smoothing for HSI classification, addressing class imbalance. In a similar vein, \citep{zhang2021spectral} employed 3D FractalNets with residual connections for spectral-spatial feature extraction in HSIs, while \citep{bai2019ssdc} used 3D DenseNets for HSI classification with the same purpose.

\subsection{Attention mechanisms in CNNs}

\textcolor{black}{While exploring architectural modifications in CNNs, researchers also sought to incorporate attention mechanisms within these architectures.} The objective is to emulate the functioning of the visual cortex by focusing on significant portions of the image while disregarding less significant or background areas \citep{itti2001computational}. Various studies, such as \citep{wang2017residual, nam2017dual, fu2017look, woo2018cbam, songara2023visual}, introduced attention mechanisms in the CNN community for applications such as image recognition, reasoning and matching, and visual-question answering. \textcolor{black}{In most cases, attention units or networks act as independent modules (e.g., squeeze and excitation module \citep{hu2018squeeze}, convolutional block attention module \citep{woo2018cbam}, triple attention module \citep{misra2021rotate}, and others) that can be incorporated into networks to enhance their performance. This concept has also been extended to the field of remote sensing. For instance, \citep{wang2018multiscale} proposed a multiscale attention model for object detection in very high-resolution (VHR) remote sensing images, pioneering the combination of an encoder-decoder framework for this task. \citep{zhang2019description} introduced a novel approach to generating image descriptions/captions for remote sensing images, where the network utilizes CNN blocks and an attention module to generate captions based on the output features. Attention mechanisms have also been extensively employed in HSI classification.} In \citep{mou2019learning}, a study enforced a spectral attention mask on the input features of the CNN to highlight the most representative bands for classification. \citep{haut2019visual} demonstrated the use of attention mechanisms in a ResNet-like architecture to highlight the spectral-spatial properties of the HSI cube. \citep{hang2020hyperspectral} proposed a dual-branch deep learning attention model, where each branch focuses on spectral and spatial layers individually, and the results are adaptively combined before classification. \citep{pande2021adaptive} proposed a hybrid approach that employs 1D and 2D CNNs to extract spectral and spatial attention masks, respectively, and the attentive features are then fed into a 3D CNN for classification. \textcolor{black}{Expanding on the idea of parallel connections, \citep{qu2021triple} introduced a deep HSI classifier that not only had forward connections between layers but also incorporated sideways connections to enhance the model's robustness.} Furthermore, \citep{tang2020hyperspectral} proposed a framework that utilized octave convolutions \citep{chen2019drop}, which decompose spatial information into high and low frequencies, along with attention mechanisms (extending in both spatial and spectral dimensions) for hyperspectral image classification. The concept of attention has also been extended to multimodal learning in remote sensing. This is demonstrated in \citep{mohla2020fusatnet}, where an attention-based method is used to combine HSI features with spatial/depth information from LiDAR for joint classification. Taking a step further, \citep{bose2021two} incorporated the concept of Transformers (a modification of attention mechanisms) for HSI and LiDAR fusion. The model consists of subsequent encoder-decoder blocks, each implementing multihead attention. The features from all the blocks are concatenated and sent for classification. \citep{yang2018convolutional} argues that feedback architectures have the potential to refine spatial attention in features, thereby increasing classification accuracy, making them a subject of research in this field.

\subsection{Feedback connections in CNNs}

The concept of attention has also spurred the utilization of feedback mechanisms in neural networks. For example, \citep{zamir2017feedback} proposed a feedback network for object detection, employing a convolutional-LSTM model that iteratively enhances the model's classification ability by progressing from coarse identification to finer prediction. \citep{li2019feedback} utilized a similar convolutional-RNN network for super-resolution, gradually generating a higher-resolution image guided by previous states through successive feedback loops. In the application of crowd counting, \citep{sam2018top} applied feedback connections in CNNs by employing two simultaneous CNNs—one for feature generation and the other for feedback generation. The enhanced features from the feedback are then used to generate a crowd density map. Feedback-based CNNs have also been adopted in the field of remote sensing. \citep{cheng2019multi} proposed a network for multiclass object detection, where feedback from the last layer is provided to each of the previous layers. This allows the inclusion of features from relevant neurons based on the posterior probability of the object classes. \citep{wang2021lightweight} introduced feedback CNNs for remote sensing image super-resolution, employing the feedback mechanism with ResNets to refine low-level features using high-level features to achieve higher resolution. Similarly, \citep{fu2020two} utilized a two-path CNN with a feedback mechanism, employing an RNN-like network for PAN-sharpening of multispectral images. In the HSI domain, feedback mechanisms have not been extensively explored. \citep{li2021recurrent} and \citep{yu2021feedback} are recent works that propose feedback-based CNN models for HSI classification. In the former, an attention block in the network (a sub-network of convolutional layers) is reinforced with a feedback connection, and the outputs from different attention blocks are then used for classification. In the latter, feedback connections are employed to derive attention masks, which are then used to highlight relevant features obtained from dense CNN layers.

\subsection{Effect of loss functions in classification}

Even though different model architectures possess distinct characteristics that lead to the realization of various types of abstractions, it is equally important to comprehend the behavior of network parameters under different types of losses, depending on the application \citep{janocha2017loss}. This holds true for classification tasks as well. Traditionally, classification-based deep learning frameworks are trained using cross-entropy loss \citep{de2005tutorial} and divergence-based losses \citep{goldberger2003efficient, wang2006groupwise}. However, there has been a recent trend towards the use of metric learning and distance-based losses for classification purposes as well \citep{chen2020low}, which were initially used for reconstruction \citep{masci2011stacked}. For example, in HSI classification, least-squares loss \citep{zhang2019discriminative} and center loss \citep{dong2020cooperative} have been employed. Furthermore, several research studies have explored the use of distribution-based losses over sample-based losses to achieve more realistic abstractions in the transformed feature subspace \citep{zhou2020learning}. A novel loss function called ``magnet loss'' is proposed in \citep{rippel2015metric}, which models the distribution of classes in the transformed space and minimizes the distribution overlap to enhance local class discrimination. Similarly, \citep{frogner2015learning} introduces the concept of learning with Wasserstein loss, which utilizes the earthmover distance \citep{andoni2008earth} to map class distributions from the sample space in the context of multilabel classification. Due to its high discriminatory ability, Wasserstein loss has also been incorporated into generative models, such as Wasserstein GANs \citep{arjovsky2017wasserstein} and Wasserstein auto-encoders \citep{tolstikhin2017wasserstein}.

\subsection{Domain adaptation in remote sensing}

In remote sensing images, two common scenarios arise: i) images of a specific ground area captured at different time points, and ii) images of distinct geographical areas with similar land-cover types. Generating training samples for all these images is often challenging, leading to the practice of reusing training samples obtained from images exhibiting \textit{similar} characteristics. This approach allows for supervised learning tasks on new images. Consequently, inductive transfer learning, particularly domain adaptation, has gained significant popularity.

Unsupervised domain adaptation (UDA) techniques, by definition, involve two distinct yet interconnected data domains: a source domain $\mathcal{S}$ equipped with an abundant amount of training samples, and a target domain $\mathcal{T}$ where test samples are gathered. Since the data distributions differ between these two domains ($P(\mathcal{S}) \neq P(\mathcal{T})$), a classifier trained on $\mathcal{S}$ fails to generalize effectively to $\mathcal{T}$ in accordance with the probably approximately correct (PAC) assumptions of statistical learning theory \citep{tuia2016domain} \citep{sohn2018unsupervised}.

Traditional unsupervised domain adaptation (UDA) techniques can be broadly categorized into two main types: i) classifier adaptation, and ii) learning domain invariant feature spaces. In the case of classifier adaptation, a common feature space is learned to minimize the domain divergence or to model a transformation matrix that maps samples from one domain to their counterparts in the other domain \citep{chen2015deep}, \citep{li2018domain}. Several popular ad hoc methods in this category include transfer component analysis (TCA) \citep{pan2011domain}, subspace alignment (SA) \citep{fernando2013unsupervised}, and geodesic flow kernel (GFK) \citep{gong2012geodesic} based manifold alignment. On the other hand, UDA approaches based on maximum mean discrepancy (MMD) \citep{yan2017mind} aim to learn a domain invariant space within a kernel-induced Hilbert space. 

More recently, adversarial training has gained significant popularity in UDA. These approaches involve a min-max game between two modules: a feature generator ($G$) and a discriminator ($D$). While $D$ tries to distinguish between samples from the source domain ($\mathcal{S}$) and the target domain ($\mathcal{T}$), $G$ is trained to make the target features indistinguishable from those in $\mathcal{S}$ \citep{goodfellow6572explaining}. The RevGrad algorithm is particularly noteworthy in this context as it introduces a gradient reversal layer to maximize the gradient of the $D$ loss \citep{ganin2014unsupervised}. This encourages $G$ to learn a feature space that confuses the domains, thereby significantly reducing the domain gap. Another notable approach, known as adversarial residual transform networks (ARTN) \citep{cai2018unsupervised}, utilizes adversarial learning in UDA. Additionally, generative adversarial networks (GANs) have been widely used for various cross-domain inference tasks, including image style transfer and cross-modal image generation. Examples of GAN-based endeavors in this area include DAN \citep{ganin2016domain}, CycleGAN \citep{zhu2017unpaired}, and ADDA \citep{tzeng2017adversarial}.

Since UDA problems are frequently encountered in remote sensing (RS), the aforementioned ad hoc techniques have been explored in the RS domain as well \citep{tuia2016domain}. For instance, a recent study \citep{banerjee2017hierarchical} proposes a hierarchical subspace learning strategy that considers the semantic similarity among land-cover classes at multiple levels and learns a series of domain-invariant subspaces. The use of a shared dictionary between domains is also a popular practice for hyperspectral image (HSI) pairs \citep{ye2017dictionary}. Deep learning techniques such as GANs and domain-independent convolution networks have also been investigated in this context \citep{bejiga2018gan}.

\subsection{Multimodal data fusion in remote sensing}

The need for data fusion across multiple sources or modalities arises from the limited perspective captured by each source. For instance, HSI captures high reflectance information at the expense of spatial resolution \citep{landgrebe2002hyperspectral}, LiDAR captures depth information but lacks backscatter and intensity/reflectance information \citep{liu2008airborne}, and SAR images have good penetration capabilities and backscatter information but lack spectral information and are susceptible to speckle noise \citep{curlander1991synthetic}. Therefore, combining information from multiple sources becomes crucial for achieving more accurate results in tasks such as LULC classification. In remote sensing, there are generally three levels or modes of fusing the data from individual modalities: pixel level, feature level, and decision level.

Pixel level fusion occurs at the earliest stage when the data or images are directly combined \citep{ghassemian2016review}. The typical approach is to transform the original images to a domain suitable for fusion, and after fusion, transform the image back to the original domain. Conventional examples of pixel level fusion include Gram-Schmidt fusion, Brovey transform, and Ehlers Fusion \citep{wang2005comparative}.

In feature level fusion, features are first extracted from the images, and then fusion is performed. Fusion at this level is more decisive and accurate for tasks such as image classification \citep{klein1999sensor}. For example, \citep{pedergnana2012classification} proposed a fusion method for optical and LiDAR data, where extended attribute profiles (EAPs) were extracted from both modalities and concatenated with the original modalities for classification. However, concatenation is not always a viable option as it increases the number of dimensions, leading to the ``Hughes phenomenon'' \citep{hang2020classification}.

Decision level fusion involves passing the modalities through separate classifiers, and then establishing decision rules to generate the final classification output. This fusion regime typically has more information than the other regimes \citep{kulkarni2020pixel}. For instance, \citep{kasapouglu2013decision, ma2015new} implemented fusion using majority voting, where features from different modalities were classified using multiple classifiers. The predicted classes were then combined using a majority voting rule to obtain the final prediction. Many deep learning algorithms inherently employ feature level fusion, as the data modalities are passed through a deep feature extractor before fusion. The subsequent section discusses the most prominent deep learning algorithms in this regard.

\subsection{Deep learning for remote sensing data fusion}

In the field of remote sensing, deep architectures are commonly used for feature and decision level fusion due to their ability to efficiently handle a large number of features in a data-driven manner. Data fusion can be applied to various tasks such as resolution enhancement of low-resolution (LR) images or land use and land-cover (LULC) classification \cite{pande2023land}.

For example, \citep{wei2017boosting} proposed a residual architecture for pansharpening of multispectral images. \citep{liu2020psgan} introduced a GAN-based architecture for pansharpening, incorporating a GAN framework to generate high-resolution multispectral images from low-resolution PAN images. The generated images were compared to the original high-resolution MS images using a discriminator. \citep{ma2020pan} further enhanced the pansharpening of MSI by using a GAN-based double-discriminator framework. In this approach, the fused product generated by the generator was compared to both the original MS and PAN images by separate discriminators, ensuring that the fused product retains characteristics of both types of images.

Fusion techniques have also been extended to classification tasks, particularly in the domain of HSI. \citep{li2018hyperspectral} proposed a three-stream architecture for the fusion of HSI and LiDAR images. Handcrafted features such as extinction profiles of HSI and LiDAR, along with the original HSI features, were passed through a three-stream convolution architecture and used for classification. \citep{zhao2020joint} proposed a method for HSI and LiDAR fusion using hierarchical random walk (HRW) and CNNs. This model extracted spatial and spectral features from HSI using a dual tunnel CNN, while simultaneously capturing class-specific information from LiDAR-based DSM using a parallel affinity branch. The HSI features served as a global prior in the HRW layer, and spatial consistency was enforced using pixel affinity features.

Attention mechanisms have also been employed for HSI and LiDAR fusion. \citep{mohla2020fusatnet} introduced the use of three individual attention modules for fusion. The first two modules operated in parallel to extract spectral and spatial features from the HSI using LiDAR information. The third attention module selected the most informative modality from the extracted features and used it for classification. \citep{bose2021two} further extended the idea of attention for HSI and LiDAR fusion by incorporating transformers-based self-attention modules. These modules combined HSI and LiDAR features obtained from convolutional layers, and the fused features from all attention blocks were combined and used for classification.

Another recent work utilized the concept of self-attention for HSI and LiDAR fusion \citep{wang2022multi}. In this approach, convolutional layers were initially used to extract HSI and LiDAR features, which were then combined using self-attention in a shallow feature fusion paradigm. The fused features were further processed by a modality attention module for deep fusion before being used for classification. Deep learning-based fusion techniques are not as prevalent in the SAR domain, but they still hold significance. \citep{meraner2020cloud} proposed a method for optical-SAR data fusion. They designed a deep residual network to mitigate the impact of cloud cover in optical imagery (obtained from the Sentinel-2 satellite) and used SAR-optical guided fusion to efficiently reconstruct the imagery. In the context of infrared and SAR fusion for automatic target object detection, \citep{kim2018double} introduced a decision fusion-based scheme. Their method employs a doubly weighted scheme to measure offline sensor confidence and online sensor reliability. The LeNet classifier is utilized in both cases, and the weighted sensor scores are fused either linearly or nonlinearly to generate predictions. For the fusion of HSI and PolSAR data, \citep{hu2017fusionet} proposed a method called ``FusioNet.'' This model consists of two parallel branches that take HSI and SAR features as input. The extracted features are then concatenated and used for classification. A recent work focused on HSI-SAR fusion is presented in \citep{li2022asymmetric}, called AsyFFNet. The proposed model employs residual blocks with shared weights for feature extraction. The scaling factors in batch-normalization layers are used to identify redundant features and impose a sparse constraint. To improve class identification, a self-attention-based feature calibration module is included in the model.

\subsection{Learning with privileged information (LUPI)}

The LUPI paradigm was initially introduced in \citep{vapnik2009new} specifically for support vector machine (SVM) classifiers. This concept was later expanded upon in \citep{pechyony2010theory}, where the authors introduced the concept of privileged empirical risk minimization to identify a faster learning function in the decision space. Modifications were made to the existing LUPI paradigm in \citep{vapnik2015learning}, and it was also used in conjunction with knowledge distillation in \citep{lopez2015unifying}. LUPI has since been implemented in various other domains, including unsupervised learning \citep{feyereisl2012privileged}, metric learning \citep{fouad2013incorporating, xu2015distance, fouad2013ordinal}, object localization \citep{feyereisl2014object}, face detection \citep{yang2013privileged}, expression recognition \citep{wang2018thermal}, and more. However, all these works incorporate privileged information within the conventional machine learning setting, where the goal is to either maximize the margin between classes or maximize the likelihood of an instance belonging to a certain class. More recently, \citep{zhang2018deep} introduced the idea of LUPI in the deep learning setting. Instead of a teacher-student framework, an ensemble of students is considered, which encourages cooperative learning. This is achieved by incorporating two losses: a supervised learning loss and a mimicry loss. The mimicry loss aims to match the posterior of each student network to the class probabilities of other students.

\subsubsection{LUPI with modality distillation}

The idea of LUPI has been further extended to the deep learning setting, mostly in combination with knowledge distillation frameworks. In \citep{chen2016learning}, LUPI was incorporated with a convolutional neural network (CNN) for image categorization, where the mapping difference between visual features and word embeddings was used as privileged information (referred to as ``privileged cost''). \citep{neverova2015moddrop} introduced a multimodal method for gesture recognition from video datasets by combining multiscale learning (using spatial and temporal scales) with multimodal learning. A multimodal CNN is used to fuse the different modalities and perform classification. Additionally, a regularization technique called ModDrop is presented, where weights corresponding to certain modalities are dropped during fusion for each iteration based on probabilities from Bernoulli's selector. In \citep{hoffman2016learning}, depth information is used as extra information in a convolutional RGB object detection framework, along with modality hallucination. \citep{garcia2018learning} improved upon the aforementioned method by incorporating adversarial learning in the hallucination process, where depth is used along with RGB images during the training phase but omitted during the testing phase. \citep{roheda2018cross} proposed a conditional generative adversarial network (C-GAN) based approach to generate missing modalities from the available ones. LUPI is further utilized in various other deep learning-based computer vision applications, such as brain tumor detection \citep{ye2017glioma}, action recognition using recurrent neural networks (RNNs) \citep{shi2017learning}, and multi-instance multi-label (MIML) learning \citep{yang2017miml}.

\subsection{Self-supervised learning in remote sensing}

Self-supervised learning found its way in the deep learning domain to deal with scenario of humongous data and limited to no annotations during training. The core idea behind SSL based methods is to train the models using supervision signals from the data itself, by utilizing certain `\textit{pretext tasks}'. Notably, SSL based methods pretext tasks maybe grouped into three categories, namely generative tasks, predictive tasks and contrastive tasks, or any combination of these tasks. After pretraining, the models are deployed for the different downstream tasks like classification or change detection. In the subsequent sections, we will see a few works focussing on each of these tasks \citep{wang2022self}. 

\subsubsection{Generative tasks in SSL for remote sensing}

These pretext tasks rely on generative modelling frameworks like autoencoders and generative adversarial networks for training on the pretext tasks. Hyperspectral dimensionality reduction is one of such tasks that is often approached using this method. \citep{sharma2021self} utilized autoencoders for self-supervised learning and dimensionality in high dimensional vegetation indices derived from Sentinel-2 images. \citep{madhuanand2021self} presented a method of depth estimation from monocular UAV images driven by autoencoder based self-supervision. Besides focussing on dimensionality reduction, applications such as image restoration \citep{kumar2022rsinet}, and unmixing \citep{patel2020deep} are used. \citep{chen2021hyperspectral} used SSL in for hyperspectral image superresolution using low resolution HSI and high resolution MSI. \citep{palsson2020convolutional} proposed a method based on convolutional autoencoder for HSI unmixing, where the AEs are used to get the abundance matrices in the low dimensional space. The idea of generative SSL is also realized using GAN based frameworks. \citep{cheng2021perturbation} introduced a method called perturbation seeking GAN for classification in remote sensing. For unmixing based tasks as well, the authors of \citep{ozkan2020spectral} used Wasserstein GANs for in a SSL based scenario. \citep{alvarez2020s2} proposed a change detection method, where a conditional GAN is used to generate the distribution of unchanged samples. \citep{walter2020self} investigate the used of GANs and variational autoencoders (VAEs) in tasks of content based image retrieval. 

\subsubsection{Predictive tasks in SSL for remote sensing}

Several SSL based techniques are also realized using predictive tasks such as breaking the image into patches and identifying their relative position after shuffling. Similarly, there could be tasks related to predicting the rotations by which the samples have been rotated. \citep{zhao2020self} proposed a method of SSL for RS image classification, where the pretext task was defined of predicting the rotation of the images. \citep{li2023self} presented a method of self-supervised inpainting using deep image prior. Their work leverage a low rank sparse matrix in a plug and play fashion for HSI inpainting. \citep{vincenzi2021color} introduced colourization technique as self-supervised pretext task on RGB satellite images. The model is later fintuned on classification based prediction task and then evaluated by ensembling. \citep{singh2018self} proposed a SSL method based on inpainting for semantic segmentation of overhead imageries. 

\subsubsection{Contrastive tasks in SSL for remote sensing} 

Contrastive tasks in SSL involve identifying the similarity among samples belonging to the same class while simultaneously increasing the distance between these samples and those belonging to other classes. There is a significant amount of research on contrastive SSL in the remote sensing domain. Stojnic et al. \citep{stojnic2021self} proposed a contrastive multiview coding technique for RS image classification, where similar views of the input image are matched, and the model is trained based on this criterion. This idea was extended to hyperspectral images by Liu et al. \citep{liu2020deep}. \citep{xu2021adversarial} introduced the concept of contrastive SSL from an adversarial learning perspective for SAR target detection. \citep{wang2022self_dino} presented a unified framework for SAR and optical image fusion in the multimodal remote sensing domain, where they used DINO \citep{caron2021emerging}, a recent SSL technique, with a ViT \citep{dosovitskiy2020image}-based encoder. \citep{wang2023nearest} proposed a nearest neighbor-based contrastive learning approach for hyperspectral and LiDAR fusion. In their approach, they aim to match the semantic relationship among the nearest neighbors in the spatial domain using the momentum contrast SSL technique.

\section{Research gaps and motivation} 

After the thorough literature review of the existing works with regards to hyperspectral imaging, it was realised that there is a huge scope of improvement in HSI analysis in several avenues. It is to be emphasized that the PhD research encompasses an approximate period of 5 years, and during this tenure, several new state of the art (SOTA) techniques were created and broken in the research community. However, the gaps enlisted below had been more relevant during the time of their identification and the published research works as part of this PhD had been the SOTAs contemporaneously. Following are the areas where improvement is considered as part of the research work:

\begin{enumerate}

    \item \textbf{HSIs in single-modal setting}\\
        \begin{enumerate}
            \item While processing of hyperspectral images using DL based architectures, it was realised that the existing feature extractors in most of the cases needed improvements with regards to their feature extraction capabilities. Since most of the feature extractors are designed using CNNs, the associated drawbacks creep in as well. For instance, for a long time, there had been the problem of `vanishing' or `exploding' gradients associated with the HSIs, and it had been the need of the hour to come up with the techniques that ensure efficient gradient flow. Furthermore, the contemporary feature extractors in the DL domain are mainly constructed in the forward direction. This severely limits their information flow and prevents them from learning robust representations in the initial stages. 
    
            \item Moreover, from modality perspective, given the resolution differences in the spectral and spatial domain, it becomes of prime importance to evenly weigh different kinds of information based on different tasks such as classification. 
    
            \item The research gaps also extend to different learning based settings such as that of domain adaptation, where the distribution of the training data is vastly different from that of the samples in the deployment phase.
        
            \item In addition, it is also identified that for better discrimination in the classes in supervised setting, one has to design better and more diverse loss functions that are complementary to the most used ones like categorical cross-entropy. 
            
        \end{enumerate}
    
    \item \textbf{HSIs in multimodal setting}\\

        \begin{enumerate}
    
            \item In multimodal learning scenario, there could be multiple ways of handling/addressing different modalities for a given tasks. For the joint classification with multiple modalities, the main challenge is to establish effective communication among the participating modalities. Currently most of the deep feature extractors independently handle all the modalities and then fuse them by concatenation or addition at some later stage for the tasks such as classification. 
    
            \item In addition, it is also observed that there is a need of designing modality oriented feature extractors pertaining to the participating modalities, so as to extract complementary information for the task at hand. Furthermore, for better interaction among the modalities, there was also the need of utilising annotation independent tasks for better cross-modal representations. 

            \item Simultaneously, there was a parallel problem identified in the domain of multimodal learning. Conventionally, it is assumed that all the modalitites involved in training the classifier would also be present during the deployment phase. However, in our work, we address a scenario, where a subset of modalities (and channels in case of HSIs) are missing during the deployment phase. Currently, no research work has addressed such problem in the remote sensing domain. 
            
        \end{enumerate}
    
    \item \textbf{Self-supervised learning for HSIs}\\

        \begin{enumerate}
    
            \item As is fairly established by now, hyperspectral images suffer from the challenge of high dimensionality, that interferes with several tasks in purely supervised setting, such as classification. Even though, there are several classification techniques, where models are trained and evaluated in the purely supervised setting, these techniques have high reliance on the amount of annotated data present. However, when using off the shelf classification techniques like random forests \citep{breiman2001random}, there is still a challenge of using high dimensional data in limited sample regime. Therefore, it becomes necessary to project the high dimensional HSI data to a subspace of manageable dimension so that it can be used by such classifiers. Therefore, in this research work, we also address the problem of efficient deep dimensionality reduction using feature extractors catered to capture relevant information form HSIs. 
    
            \item There is also a wide scope in HSI classification to develop the efficient self-supervised and semi-supervised techniques from contarstive learning regime. The idea is to pre-train the DL models with limited to no supervision and then deploy these models on any downstream task, such as classification. 
            
        \end{enumerate}
    
\end{enumerate}

\section{Objectives and contributions}

Based on the aforementioned gaps, the current PhD research work makes the following contributions in the area of hyperspectral image analysis. 

\begin{enumerate}

    \item \textbf{HSIs in single-modal setting}\\
        \begin{enumerate}

            \item We start our research work with the paradigm of unsupervised domain adaptation in hyperspectral images. To this end, we introduce an adversarial learning based framework to bring the distribution of source and target domains closer without use of any class labels. To enhance the feature extractors, cross-sample reconstruction within the classes and orthogonality constraint over the extracted features are incorporated as well. 
            
            \item To improve the feature extraction capabilities of the deep learning architectures, we have introduced an attention framework in LULC classification, that selectively weighs the spectral and spatial information accordingly for the classes (such as spectral information for vegetation and spatial information for urban classes). Additionally, we also introduce a distribution based novel loss function (Wasserstein loss), which complements the categorical cross-entropy loss and enhances the classification performance of the proposed model. 
    
            \item In the final research work, we introduce the idea of feedback connections in the domain of hyperspectral image classification. Here, we construct a network in a multiscale setting, where the information from multiple future states are propagated back to the initial stages for feature refinement. Such network possesses ``self-correcting'' properties akin to ``curriculum learning'' based framworks.   
            
        \end{enumerate}
    
    \item \textbf{HSIs in multimodal setting}\\

        \begin{enumerate}
    
            \item The research work takes off by tackling the problem of missing modality hallucination from the available modality. In this research work, we introduce the concept of knowledge distillation in the remote sensing to transfer the information from teacher network (with all the available modalities) to the student network (where only a subset of modalities are present). In addition, to generate the absent modalities we introduce a conditional generative adversarial learning based framework.
    
            \item In the next work, we tackle the problem of HSI-LiDAR fusion where we introduce the notion of attention mechanism to selectively highlight the information from the participating modalities for the task of LULC classification. This was the pioneer work to introduce attention mechanism for remote sensing modalities.  

            \item Finally, we introduce the notion of cross-modal self-supervision assisted fusion for HSI-LiDAR/SAR fusion. Here, to control the number of parameters, we design the a common feature extractor across the modalities, that also shares parameters in time among its layers (like a ``feedback'' loop). The notion of self-supervision is incorporated by forcing the features of one modality to generate those of other modality. 
            
        \end{enumerate}
    
    \item \textbf{Self-supervised learning for HSIs}\\

        \begin{enumerate}
    
            \item To address the issue of dimensionality reduction, we design autoencoders with different features extractors catering to the information inherent in the HSIs. For the spatial-spectral feature extraction, we design a 3D CNN based autoencoder with residual connections for efficient gradient flow. For dimensionality reduction in spectral domain, variants of 1D CNNs enhance with attention mechanism and feedback connection were used, to efficiently harness the contiguous information in the spectral bands. Alongwith 1D CNNs, feature extractors such as Bidirectional gated recurrent units were also utilised for efficient feature extraction in spectral domain. 
    
            \item The most recent research work focussed on incorporating the notion of SSL in a label guided setting for HSI classification. We (pre-)train our model with several unannotated and a few labelled samples (that are assumed to be present during training phase anyways). Here, we send supervision signals to the network by designing an auxiliary task to identify the different augmentations of the same unlabelled sample. The augmented samples are compared with the labelled sample to ge the corresponding probability distributions, and the model is trained on the closeness of the two distributions.  
            
        \end{enumerate}
    
\end{enumerate}

\section{Datasets}

The experiments have generally been conducted on multiple hyperspectral datasets, either in a single modal setting, or from a multimodal domain like LiDAR and SAR. All the datasets are benchmarks datastes, acquired from several remote sensing data repositiories online, and hence there is no onus of preprocessing involved. In addition, to test the modality specific generalisation of algorithms, other modalities apart from HSIs, like multispectral and panchromatic images are also sometimes considered. 

\subsection{Single-modal hyperspectral datasets}

\subsubsection{Indian pines 1992 dataset}
\begin{figure}[!htb]
    \centering
    \includegraphics[width=0.8\textwidth]{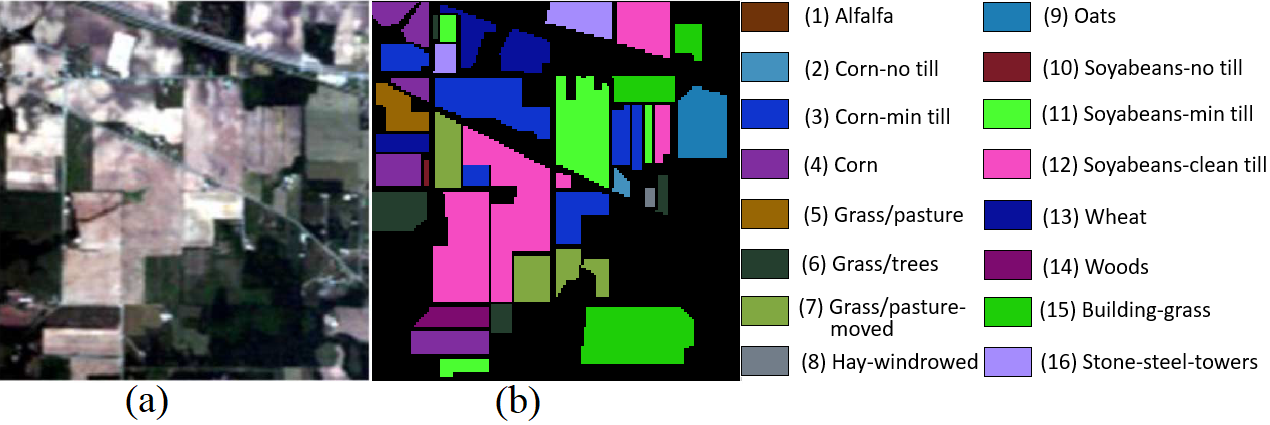}
    \caption{Indian Pines hyperspectral dataset with (a) True colour composite (b) Groundtruth map}
    \label{fig:IP_ds}
\end{figure}

Collected through the AVIRIS sensor, the dataset comprises 200 bands and possesses spatial dimensions of 145 by 145. Within the image, there are 10,249 ground truth samples that span 16 distinct land use and land cover classes (see Figure \ref{fig:IP_ds}) \citep{landgrebe1992220}. 

\subsubsection{Indian pines 2010 dataset}
\begin{figure}[!htb]
    \centering
    \includegraphics[width=0.8\textwidth]{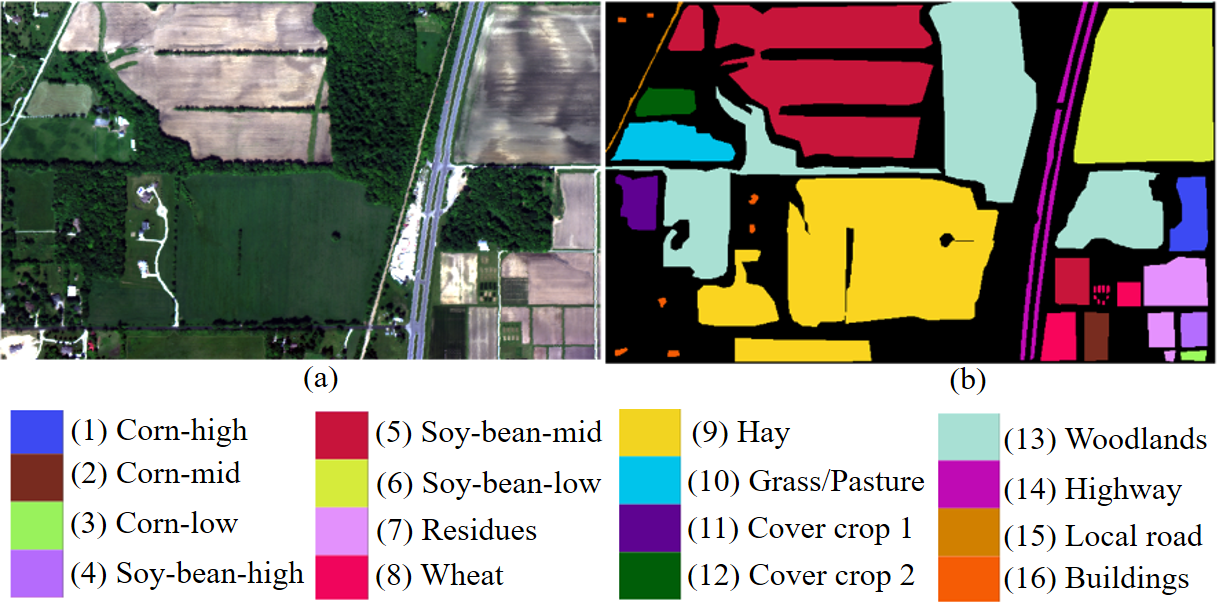}
    \caption{Indian Pines 2010 hyperspectral dataset with (a) True colour composite (b) Groundtruth map}
    \label{fig:IP10_ds}
\end{figure}

The dataset was obtained using the ProSpecTIR system in May 2010, near Purdue University, Indiana. The image encompasses dimensions of 445 by 750 pixels and includes 360 hyperspectral bands. Within the image, there are 10,954 training samples and 187,120 test samples, which are divided across 16 different classes (refer to Figure \ref{fig:IP10_ds}, \cite{rasti2020feature}).

\subsubsection{Salinas valley dataset}
\begin{figure}[!htb]
    \centering
    \includegraphics[width=0.8\textwidth]{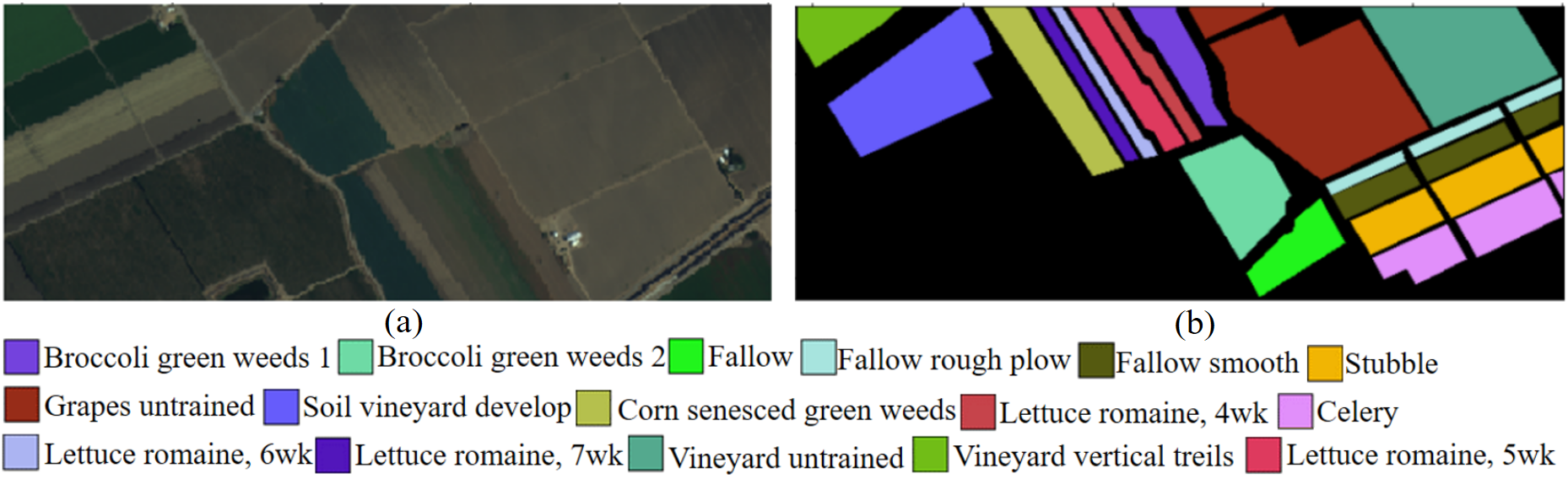}
    \caption{Salinas Valley hyperspectral dataset with (a) True colour composite (b) Groundtruth map}
    \label{fig:Sal_ds}
\end{figure}

The Salinas hyperspectral imagery is gathered using the AVIRIS sensor, covering the expanse of Salinas Valley in California. This imagery is composed of 204 bands, with each band characterized by a spatial dimension of 512 by 217 pixels. The entire scene has been partitioned into 16 distinct classes, and the ground truth information is accessible for 54,129 individual pixels (refer to Figure \ref{fig:Sal_ds}, \cite{gualtieri1999support}).

\subsubsection{Pavia university dataset}
\begin{figure}[!htb]
    \centering
    \includegraphics[width=0.8\textwidth]{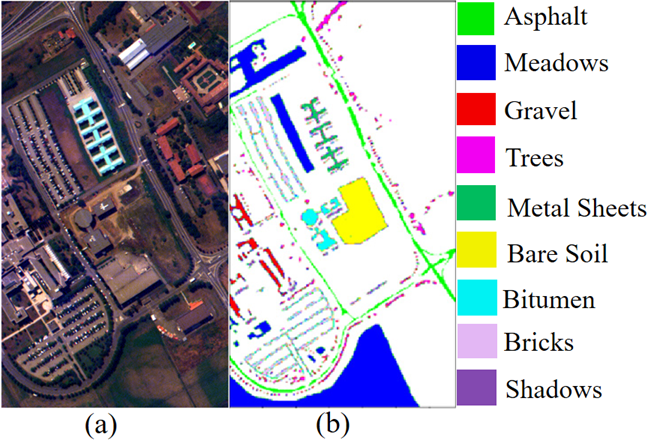}
    \caption{Pavia University hyperspectral dataset with (a) True colour composite (b) Groundtruth map}
    \label{fig:PavU_ds}
\end{figure}

The hyperspectral image is taken over Pavia city and its surroundings, utilizing the ROSIS sensor. The image spans spatial dimensions of 610 by 340 pixels and possesses 103 spectral dimensions. This image comprises 9 distinct land use and land cover classes, totaling 42,776 samples (see Figure \ref{fig:PavU_ds}, \cite{rasti2020feature}).

\subsubsection{Pavia centre dataset}
\begin{figure}[!htb]
    \centering
    \includegraphics[width=0.8\textwidth]{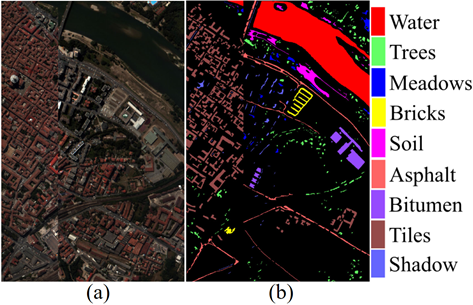}
    \caption{Pavia Centre hyperspectral dataset with (a) True colour composite (b) Groundtruth map}
    \label{fig:PavC_ds}
\end{figure}

Furthermore, the hyperspectral image is acquired over Pavia city, near Pavia centre, employing the ROSIS sensor. The image measures 1096 by 492 pixels and encompasses a total of 102 bands. It encompasses 9 land use and land cover classes, incorporating a collection of 39,332 annotated samples (Figure \ref{fig:PavC_ds}, \cite{rasti2020feature}).

\subsubsection{Botswana dataset}
\begin{figure}[!htb]
    \centering
    \includegraphics[width=\textwidth]{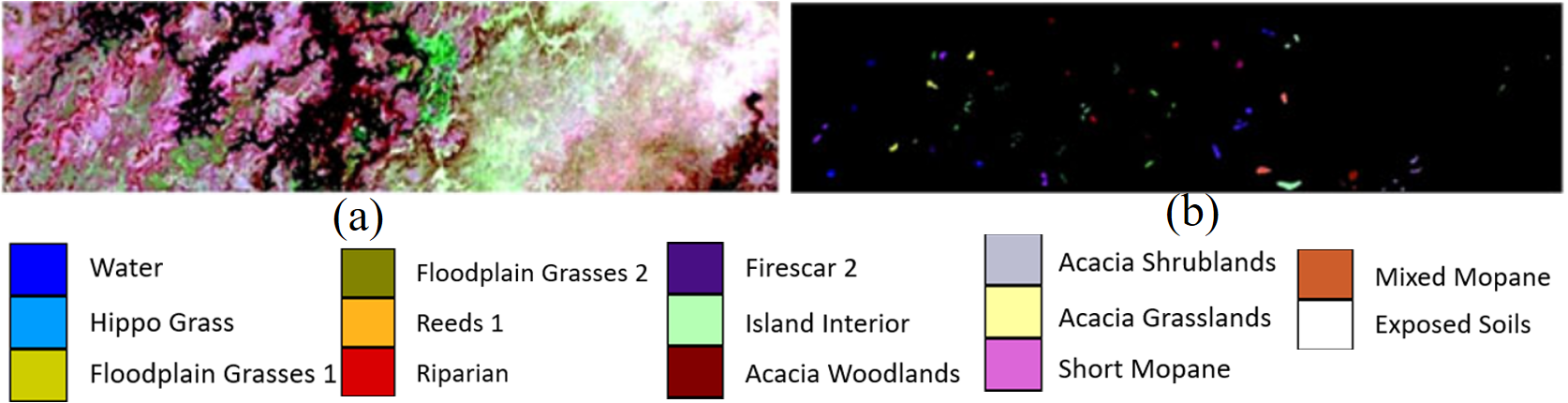}
    \caption{Botswana hyperspectral dataset with (a) True colour composite (b) Groundtruth map}
    \label{fig:Bot_ds}
\end{figure}

The initial dataset discussed is the Botswana hyperspectral imagery, as shown in Figure \ref{fig:Bot_ds}. This satellite imagery was captured by the NASA EO-1 satellite between 2001 and 2004. Utilizing the Hyperion sensor, the imagery boasts a spatial resolution of 30 meters and spans a strip of approximately 7.7 kilometers. Comprising a total of 242 bands, this imagery encompasses the spectral range spanning from 400 to 2500 nanometers. Notably, this dataset delineates fourteen distinct classes corresponding to various land cover features present on the ground \citep{ham2005investigation}.

\subsection{Hyperspectral-LiDAR datasets}

\subsubsection{University of Houston dataset, DFC-2013}
\begin{figure}[!htb]
    \centering
    \includegraphics[width=1.0\textwidth]{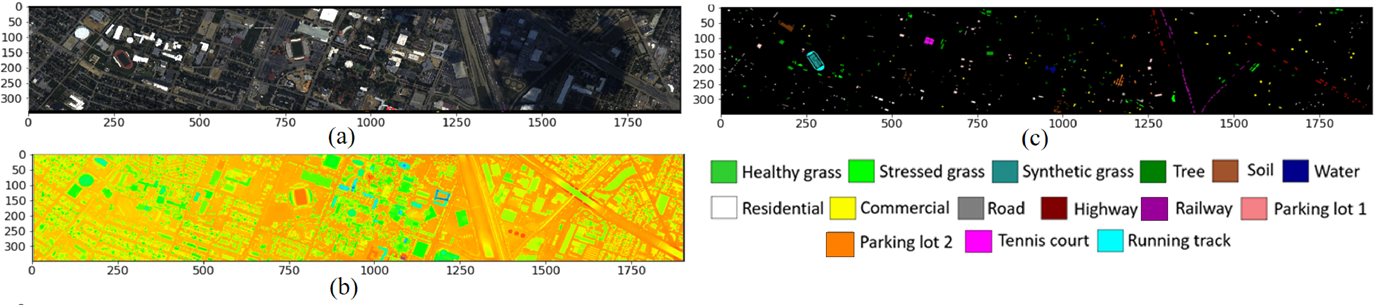}
    \caption{Houston 2013 hyperspectral-LiDAR dataset with (a) True colour composite (b) LiDAR based DSM (c) Groundtruth map}
    \label{fig:H13_ds}
\end{figure}

The dataset in question was made available for the IEEE Data Fusion Contest (DFC) in 2013. The hyperspectral image (HSI) was captured over the National Centre for Airborne Laser Mapping (NCALM). This image encompasses 144 bands and possesses a spatial dimension of 349 by 1905 pixels. Additionally, a LiDAR band is provided, offering a digital surface model (DSM) with identical spatial dimensions, wherein each cell contains depth information. Furthermore, this dataset includes two separate sets of masks, one for training and another for testing. These masks encompass 2,832 and 12,197 groundtruth samples respectively, pertaining to 15 distinct land use and land cover (LULC) classes. This dataset is introduced and explained in \citep{debes2014hyperspectral}. For the sake of simplicity in notation, this dataset will be referred to as ``Houston 13''. The HSI and LiDAR images, along with the corresponding ground truth, are displayed in Figure \ref{fig:H13_ds}.

\subsubsection{University of Houston dataset, DFC-2018}
\begin{figure}[!htb]
    \centering
    \includegraphics[width=1.0\textwidth]{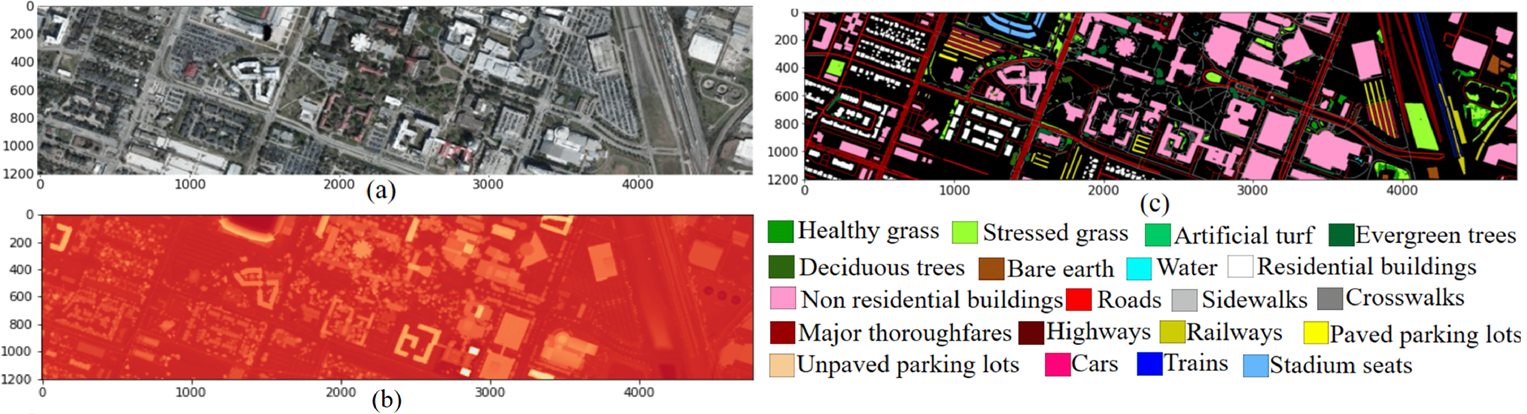}
    \caption{Houston 2018 hyperspectral-LiDAR dataset with (a) True colour composite (b) LiDAR based DSM (c) Groundtruth map}
    \label{fig:H18_ds}
\end{figure}

The dataset was released for the DFC-2018 (Data Fusion Contest 2018). The hyperspectral imagery (HSI) in the dataset includes 48 bands, each with dimensions of 601 by 2384 pixels. Additionally, a LiDAR band is present, presented as a digital surface model (DSM), with dimensions of 1202 by 4768 pixels, covering the same geographical area. The ground truth image, available at a higher spatial resolution, encompasses a total of 2,018,910 samples. This dataset, as outlined by \cite{le20182018}, features training and testing samples that span across 20 distinct classes. Due to the substantial number of annotated samples, the dataset proves highly suitable for the evaluation of deep learning models aimed at multimodal fusion techniques (Figure \ref{fig:H18_ds}).

\subsubsection{Trento dataset}
\begin{figure}[!htb]
    \centering
    \includegraphics[width=1.0\textwidth]{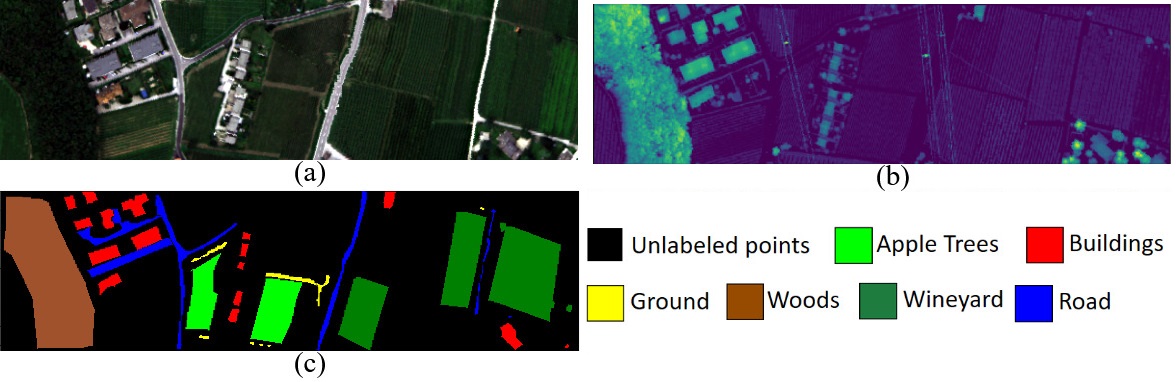}
    \caption{Trento hyperspectral-LiDAR dataset with (a) True colour composite (b) LiDAR based DSM (c) Groundtruth map}
    \label{fig:trt_ds}
\end{figure}

The dataset was gathered using the AISA Eagle sensor, situated above rural regions within Trento, Italy. The hyperspectral image (HSI) encompasses 63 bands, each with wavelengths ranging from 0.42 micrometers ($\mu$m) to 0.99 $\mu$m. Additionally, the LiDAR data contains two rasters depicting elevation information. Each band within the HSI has dimensions of 166 by 600 pixels, while both spectral and spatial resolutions are respectively set at 9.2 nm and 1.0 m. This dataset, detailed in \cite{rasti2017fusion}, features a total of 6 classes within the imagery. Ground truth information is accessible for 30,214 pixels, which are further divided into 819 training pixels and 29,395 test pixels. The dataset is visually represented in Figure \ref{fig:trt_ds}.

\subsubsection{MUUFL Gulfport dataset}
\begin{figure}[!htb]
    \centering
    \includegraphics[width=1.0\textwidth]{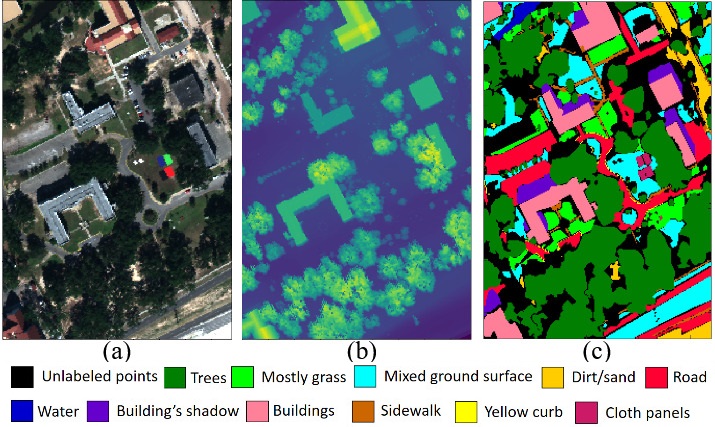}
    \caption{MUUFL Gulfport hyperspectral-LiDAR dataset with (a) True colour composite (b) LiDAR based DSM (c) Groundtruth map}
    \label{fig:muufl_ds}
\end{figure}

The dataset was captured above the University of Southern Mississippi Gulf Park campus in Long Beach, Mississippi, during November 2010. The hyperspectral imagery (HSI) in its original state included 72 bands, but due to noise, the initial and final four bands were excluded, resulting in a total of 64 bands. The LiDAR component comprises two elevation rasters. All these bands and rasters are co-registered, forming a dataset with dimensions 325 by 220 pixels. This dataset, outlined in  \cite{gader_muufl_2013} and \cite{du_technical_2017}, comprises a total of 53,687 ground truth pixels distributed across 11 classes. For training purposes, 100 pixels are chosen per class, leading to a total of 52,587 pixels for testing. The HSI and LiDAR images, along with the corresponding ground truth pixels, are visualized in Figure \ref{fig:muufl_ds}.

\subsection{Hyperspectral-SAR datasets}

\subsubsection{TU Berlin dataset}
\begin{figure}[!htb]
    \centering
    \includegraphics[width=1.0\textwidth]{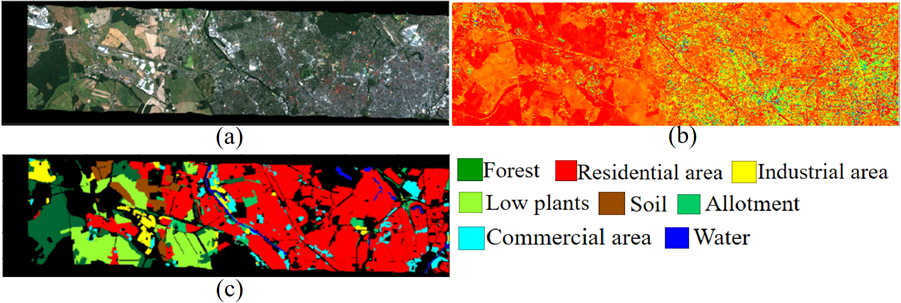}
    \caption{TU Berlin hyperspectral SAR dataset with (a) True colour composite (b) SAR band (VH+VV), (c) Groundtruth}
    \label{fig:tub_ds}
\end{figure}

The dataset comprises an urban area and its surrounding rural regions located in Berlin. It features an Environmental Mapping and Analysis Program (EnMAP) hyperspectral image (HSI) as well as a corresponding synthetic aperture radar (SAR) image from the Sentinel-1 satellite. Both the HSI and SAR images share spatial dimensions of 797 by 220 pixels. The HSI incorporates 244 spectral bands, whereas the SAR image is made up of 4 bands obtained from a dual-PolSAR setup. As depicted in Figure \ref{fig:tub_ds}, which is sourced from \cite{okujeni2016berlin}, the dataset includes ground truth samples. To maintain consistency with other datasets, only the first SAR band is employed. Two separate masks are provided for training and testing. The training set encompasses 2,820 samples, while the testing set is comprised of a total of 461,851 samples. Across both sets, the samples are distributed among 8 distinct land use and land cover (LULC) classes.

\subsection{Other datasets}

\subsubsection{Dual-source remote sensing image data set dataset (DSRSID)}
The dataset contains pairs of multispectral (MS) imagery, each consisting of 4 bands, and panchromatic (PAN) imagery, all covering a specific geographical region. In total, there are 80,000 pairs of images, representing eight different land-cover classes. These images are obtained from the multispectral and panchromatic sensors of the GF-1 satellite, as documented by \cite{chaudhuri2020cmir}. Each multispectral image has a size of 64$\times$64 pixels with 4 bands, and a spatial resolution of 2 meters. The corresponding panchromatic image has a size of 256$\times$256 pixels and a spatial resolution of 8 meters. An example of this imagery can be seen in Figure \ref{fig:mspan}, which displays a colour composite of one of the multispectral samples alongside its panchromatic counterpart. The dataset spans over $8$ landuse/land cover classes.
\begin{figure}[!htb]
    \centering
    \includegraphics[width=0.6\textwidth]{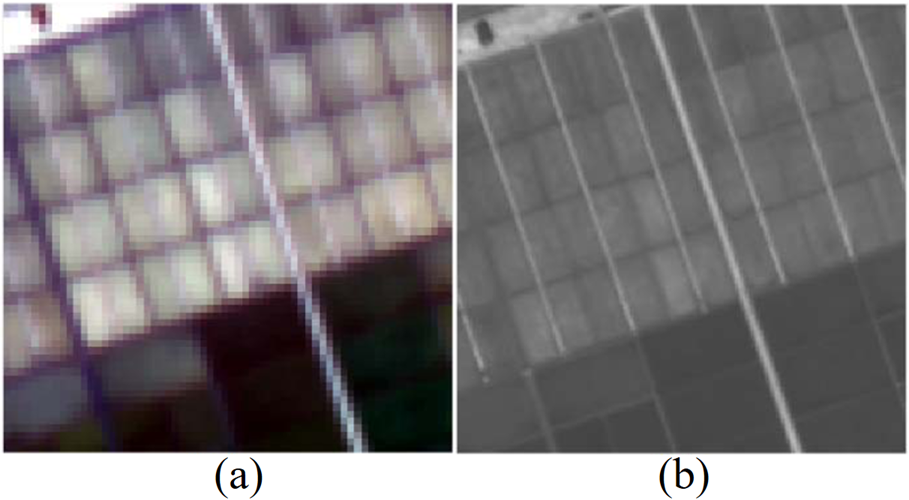}
    \caption{DSRSID dataset with (a) Multispectral image (b) Panchromatic image}
    \label{fig:MSP_ds}
\end{figure}

\section{Structure of thesis}

\noindent The thesis covers the works from the aforementioned domains in the following chapters: 

\begin{enumerate}
    \item \textbf{Chapter 1: Introduction}\\
    This chapter introduces the role of hyperspectral images in remote sensing domain. In this chapter, we see the the advantages associated with the usage of HSI in land use/land cover classification and the challenges associated with their processing using deep learning techniques. Additionally, we also discuss the role of the complementary modalities that could be used along side HSIs for better performance of the tasks. Following this, a thorough literature review is provided detailing the existing work that address the HSIs in single-modal and multimodal settings. Furthermore, a review of self-supervised techniques for HSI classification is provided as well. 
    
    \item \textbf{Chapter 2: Hyperspectral image classification in single-modal setting}\\
    In this chapter, various techniques and challenges with respect to LULC classification using HSIs in single-modal scenario are discussed. We start with the problem of unsupervised domain adaptation in HSIs. Then, we introduce the notion of attention mechanism in spectral and spatial domains to refine the feature extraction capabilities of deep learning models from hyperspectral images. Leading to this, we introduce the notion of feedback connections in HSIs to conserve the number of parameters for faster training. 
    
    \item \textbf{Chapter 3: Hyperspectral image classification in multimodal setting}\\
    This chapter focusses on processing and extracting information from HSIs in multimodal setting. We start with the problem of modality/band generation in HSIs, where it is assumed that a subset of bands are absent during the deployment phase. We then train a generative adversarial network to generate the missing bands/modalities on the fly. Secondly, we work on the problem of joint classification of HIS-LiDAR modalities. Here, we selectively extract out the spectral and spatial features using attention mechanism in order to obtain an synergistic and task oriented joint representation. Finally, we leverage the notion of self-supervised learning in classification of HSI in conjunction with LiDAR and SAR data. Here, we leverage the idea of shared cross-modal connection in the feature extractor alongwith the notion of feedback connections. Self-supervision is injected in the foray by forcing the features of one modality to construct those of another modality, without the annotation based supervision involved.
    
    \item \textbf{Chapter 4: Self-supervised learning for hyperspectral image analysis}\\
    This chapter focusses on self-supervised learning for HSI processing. We start with the problem of dimensionality reduction in HSI using deep learning based architectures (autoencoders), by leveraging different kinds of feature extractors to address different aspects of the HSI. Finally, we also introduce a contrastive learning based SSL work, where the feature extractors are trained with minimal supervision from the annotated samples and a hoard of unlabeled samples. The trained feature extractor is then used for LULC classification.  
    
    \item \textbf{Chapter 5: Conclusions and future scope}\\
    This chapter presents the relevant inferences and insights from all the experiments pertaining to the research work presented in the aforementioned chapters. In addition, the chapter extends to the possible future directions, where HSIs processing can be extended, with focuss on minimal supervision, minimal preprocessing, and faster computation and deployment.
\end{enumerate}
\chapter{Hyperspectral image classification in single-modal setting}

\section{Introduction}

The current era is witnessing a remarkable surge in the utilization of remote sensing due to advancements in imaging technology. Traditional imaging methods like RGB \citep{pohl1998review} and panchromatic \citep{zhang2014regions} are now accompanied by various other modalities/data sources, including synthetic aperture radar (SAR) \citep{zhang2011crop, wang2018polarimetric}, light detection and ranging (LiDAR) \citep{dong2017lidar}, and hyperspectral images (HSI) \citep{ghamisi2017advanced}. The field of hyperspectral imaging, in particular, has seen rapid growth as a result of the high spectral resolution of these images, enabling accurate classification based on reflectance. This research has been further accelerated by advancements in computer vision, machine learning, and deep learning. Learning-based algorithms are extensively used for information processing in the HSI domain due to their ability to capture the intrinsic characteristics of the data for specific applications \citep{zhang2018survey}. Numerous works have addressed HSI processing problems using traditional machine learning algorithms such as random forests, support vector machines, kNN, and others \citep{tarabalka2010svm, xia2017random, gewali2018machine}. Deep learning techniques, including convolutional neural networks (CNNs) \citep{luo2018hsi, roy2019hybridsn}, sequential models like recurrent neural networks (RNN) and long short-term memory (LSTM) \citep{hang2019cascaded}, as well as advanced algorithms such as generative adversarial networks (GANs) \citep{zhu2018generative} and graph neural networks (GNNs) \citep{hong2020graph}, have also been explored in this research domain. Although HSIs play a crucial role in land use/land cover (LULC) identification, challenges related to handling massive data remain. The ``curse of dimensionality'' \citep{ozdemir2020deep} poses a significant hurdle as HSIs exhibit a large number of features, demanding an extensive amount of training samples for deep learning classification models to perform well. The number of features can further increase when employing broader models such as inception networks \citep{szegedy2017inception}, which incorporate multiscale filter banks to cover larger spatial extents.

In the field of remote sensing, classification tasks face the challenge of domain gaps between training and test sets due to geographical variations, differences in reflectance characteristics, and changes in land terrain. Recent advancements in domain adaptation from the computer vision domain have also been applied to remote sensing, particularly in the context of unsupervised domain adaptation (UDA) \citep{liu2022deep, xu2022eyes}. UDA involves transductive learning, utilizing unannotated samples from the target domain during training with the source domain. For example, \citep{benjdira2019unsupervised} proposed a UDA method using a generative adversarial network (GAN) framework for aerial images, and \citep{yu2021unsupervised} introduced a self-training strategy for domain alignment in aerial image segmentation. In the domain of hyperspectral remote sensing, \citep{peng2019discriminative} proposed the discriminative transfer joint matching (DTJM) method for unsupervised land-cover classification, which matches features, performs instance reweighting, and preserves local manifold structure while maximizing the dependence between embedding and labels. Furthermore, \citep{li2020aligning} presented a UDA approach for damaged building detection by leveraging the maximum mean discrepancy loss between low-dimensional representations of source and target encodings.

Another issue in convolutional operations is that they often assign equal importance to all pixels regardless of their informational value, which may not be advantageous for classification tasks and can even hinder performance \citep{haut2019visual}. In order to address this limitation, techniques have been developed in the field of computer vision that focus on paying more \textit{attention} to relevant features while disregarding irrelevant ones \citep{jetley2018learn}. These algorithms, known for their success in the vision community, have started to be implemented in remote sensing applications, showing promising results. For example, in the context of hyperspectral imaging, \citep{haut2019visual} proposes an attention module based on ResNet within the CNN architecture to enhance spectral-spatial learning of features. Following this trend, attention modules have been introduced by \citep{dong2019band} and \citep{mou2019learning} to selectively emphasize the most informative bands for HSI classification.

Apart from the challenges associated with hyperspectral images, deep learning models also exhibit certain limitations. One prominent limitation is the forward-only nature of most deep learning models, especially convolutional neural networks (CNNs) such as AlexNet and GoogleNet \citep{ballester2016performance}. These models are designed in a feedforward manner, allowing information flow in a single direction. However, various studies have demonstrated that incorporating backward or feedback connections in CNNs can improve classification performance \citep{yang2018convolutional}. This improvement is attributed to the ability of feedback-based networks to utilize features multiple times for self-refinement, thereby establishing a training hierarchy characteristic of \textit{curriculum-based learning} \citep{zamir2017feedback}. Some research works have attempted to incorporate feedback mechanisms in CNNs. For instance, \citep{belagiannis2017recurrent} combined a recurrent neural network (RNN) module with CNNs for human pose estimation, and \citep{shi2015convolutional} employed a feedback mechanism in a joint CNN-LSTM model for short-term rainfall prediction.

In CNNs, additional challenges arise in relation to gradient propagation and information flow within the model. As the number of layers increases, the issue of \textit{vanishing gradient} becomes prominent, where the backpropagated gradients tend to diminish in the initial layers, resulting in minimal impact on the model weights \citep{huang2016deep}. To address this phenomenon, various approaches have been proposed in CNNs. For instance, ResNet incorporates residual or identity connections in CNNs \citep{he2016deep} to ensure smooth gradient flow throughout the network. Building upon the success of ResNets, several variants have been developed, such as ResNext \citep{xie2017aggregated}, which introduces residual connections in an inception-like framework, and PolyNet \citep{zhang2017polynet}, which explores polynomial combinations of different layers. Another advancement in gradient flow is observed in the DenseNet architecture \citep{huang2017densely}, where skip connections are established from all preceding layers to the current layer, facilitating enhanced information flow. Similarly, \citep{chen2017dual} extends the concepts of ResNet and DenseNet to a higher-order RNN, resulting in a model called Dual Path Network that combines the benefits of both ResNet and DenseNet architectures.

Building upon the aforementioned research, a CNN-based approach called CliqueNet was proposed \citep{yang2018convolutional}, integrating the advantages of DenseNets with feedback connections. While DenseNets ensure smooth gradient flow across the network, feedback connections maximize information sharing among network layers. In CliqueNet, feedback connections are established across all convolution layers, enabling each layer to serve as input to every other layer. Each such module within the network is referred to as a \textit{clique}, and the number of cliques can vary. To control the number of parameters, the authors introduce \textit{weights recycling}, allowing layers to recurrently update the weights. Another study proposes a looping pattern in CNNs, incorporating shared weights and skip connections \citep{caswell2016loopy}. While the former work adopts a denser architecture, the latter follows a relatively linear pattern.

When addressing image classification, it is essential to consider both a robust feature extractor and effective loss functions to enhance the discriminativeness among samples of known classes \citep{tian2022recent}. While cross-entropy loss \citep{zhang2018generalized} is conventionally used for classification tasks, incorporating metric learning techniques can further increase inter-class distances. For instance, contrastive loss \citep{wang2021understanding} and triplet loss \citep{dong2018triplet} can be utilized, which leverage negative sampling to deliberately amplify inter-class differences. Additionally, auxiliary losses such as center loss \citep{wen2016discriminative} or center invariant loss \citep{wu2017deep} can be combined with cross-entropy or triplet loss. The former minimizes the distance between clusters of the same class and their corresponding cluster centers, while the latter penalizes the distance between cluster centers themselves. Optimal transport approaches \citep{villani2009optimal} have also been explored to match probability distributions with target variables. Notably, Wasserstein distance \citep{frogner2015learning} has been used as a metric to quantify the effort required to transform one distribution into another.

Drawing from the existing literature on classification, we propose three classification paradigms specifically tailored for hyperspectral image (HSI) classification. The first paradigm focuses on domain adaptation between two geographically distinct HSI datasets that share common classes. Our approach incorporates adversarial learning by introducing a domain classifier alongside a source classifier. The objective of training is to confuse the domain classifier in distinguishing samples from the two domains, thereby creating a shared space for both domains. To enhance class discrimination, we introduce cross-sample reconstruction from the two classes. Simultaneously, we enforce an orthogonality constraint to promote non-redundancy within the extracted features.

In our second approach, we propose a novel hybrid network for HSI classification that incorporates attention mechanisms. Our architecture includes two attention modules: a spectral attention module based on 1D CNN and a spatial attention module based on 2D CNN. The spectral attention module leverages the spectral information present in the hyperspectral bands to generate an attention vector that highlights the relative importance of the bands in the input patches. Similarly, the spatial attention module generates an attention mask that focuses on the areas in the input patches contributing more efficiently to the classification task. The spectral-spatial features derived from attention are then combined with the original input using learnable weights, which we refer to as ``adaptive attention.'' The resulting combination is fed into a 3D CNN classification framework for pixel-based image classification. To enhance the learning framework, we incorporate residual connections in the spatial attention module. Additionally, to improve the model's class discrimination capacity, we introduce an additional metric learning-based loss called Wasserstein loss. This loss is jointly used with the cross-entropy loss to achieve robust and efficient classification.

In our third approach, we propose a method called HyperLoopNet. This method consists of three parallel ``self-looping'' blocks, which benefit from dense feedback connections across all the convolution layers. These connections ensure that each layer in the block is connected to every other layer, both as input and output. The self-looping structure is achieved by sharing weights for the same layers as the network unrolls. Unlike the CliqueNet, our network maintains a more compact architecture by using addition operators instead of concatenation operators. Moreover, we deviate from the sequential structure and adopt a parallel architecture. This allows our network to simultaneously pass information to different self-looping blocks that operate at different spatial scales. The integration of shared weights within the network induces a feedback effect, where the same layer in the previous instance is updated based on itself from the next instance after unrolling the blocks. The shallow structure of HyperLoopNet enables us to incorporate multiscale feature learning while controlling the number of parameters. This is achieved by using kernels of different sizes in different self-looping blocks. The inclusion of multiscale kernels ensures a more comprehensive representation of spectral-spatial features for HSI classification.

Based on the existing literature, we have made the following contributions in the research work:
\begin{enumerate}
\item We investigate the application of domain adaptation in hyperspectral image classification. Our approach incorporates adversarial learning by introducing a domain classifier into the classification framework. To further enhance the performance, we introduce classwise cross-sample reconstruction and an orthogonality constraint on the extracted features. This has been presented in section \ref{sec:DA}, published as \cite{pande2019class}.
\item We introduce the concept of spectral and spatial attention mechanisms in HSI classification. These attention modules enhance the features by selectively highlighting the relevant information. The attention-based features are then adaptively combined with the original features and subjected to the Wasserstein loss to improve discriminativeness. We call our proposed model `HybAtNet' and the research work is shown in section \ref{sec:AttHS}, published as \cite{pande2021adaptive}.
\item We propose a novel hyperspectral image classification model called `HyperLoopNet'. This model consists of multiscale self-looping blocks, where each layer is connected to all other layers through both forward and feedback connections. This design ensures maximum information flow and interconnection among all layers. The feedback loops are implemented using shared connections, reducing the number of parameters while maintaining effective information propagation. We incorporate three multiscale filter-banks in the model to capture features at different spatial scales, enabling efficient classification. We have discussed the research in section \ref{sec:FBHS}, published as \cite{pande2022hyperloopnet}.
\end{enumerate}  

\section{Domain Adaptation in hyperspectral images}
\label{sec:DA}
In this research, our main focus lies on the domain classifier (DC) based adversarial approach for unsupervised domain adaptation (UDA). Specifically, DC based UDA approaches involve training the domain classifier and a source-specific classifier simultaneously using a feature generator-discriminator framework. The domain classifier aims to make the domains overlap, while the source classifier prevents trivial mappings. However, we have identified the following limitations in standard DC based approaches: i) the learned space does not promote discriminativeness, failing to consider intra-class compactness and leading to sample overlap within fine-grained categories; ii) the learned space is unbounded and lacks meaningful interpretation, potentially containing redundant information.

To address these issues, we propose an advanced autoencoder-based approach that extends the conventional DC based UDA framework. In addition to jointly training the binary domain classifier and the source-specific multiclass classifier, we introduce two additional constraints on the learned latent space for source-specific samples. The first constraint is the reconstructive constraint, which aims to reconstruct one sample from another sample in $\mathcal{S}$ that share the same class label. This constraint helps capture classwise abstract attributes more effectively than a typical autoencoder setup. Moreover, it aids in focusing the samples from $\mathcal{S}$ at the category level. The second constraint is the orthogonality constraint, which ensures non-redundancy of the reconstructed features in the source domain. By optimizing all four loss measures together, we empirically observe a better alignment between $\mathcal{S}$ and $\mathcal{T}$ \footnote{Published as: Shivam Pande, Biplab Banerjee, and Aleksandra Pi{\v{z}}urica (2019).\href{https://link.springer.com/chapter/10.1007/978-3-030-31332-6_41}{Class reconstruction driven adversarial domain adaptation for hyperspectral image classification}. In: Iberian Conference on Pattern Recognition and Image Analysis. Springer, pp. 472–484.}. The details are presented below.

\subsection{Methodology}

In this section, we provide a detailed explanation of the Unsupervised Domain Adaptation (UDA) problem and present our proposed solution.

\textbf{Preliminaries}: We begin with the source domain training samples denoted as $\mathcal{X}_S = {(\mathbf{x}i^s, y_i^s)}^{N_S}_{i=1} \in X_S \otimes Y_S$. Here, $\textbf{x}_i^s \in \mathbb{R}^d$ represents the input features, and $y_i^s \in \{1,2,\ldots,C\}$ represents the corresponding labels. Similarly, we have the target domain samples denoted as $\mathcal{X}_T = {(\textbf{x}j^t)}^{N_T}_{j=1} \in X_T$. These target domain samples belong to the same categories as the source domain samples, but the probability distributions $P_S(X_S)$ and $P_T(X_T)$ are not equal. In this setup, the goal of the UDA problem is to learn a function $f_S: X_S \rightarrow Y_S$ that generalizes well for the target domain $\mathcal{X}_T$.

To effectively learn $f_S$, we propose an end-to-end encoder-decoder based neural network architecture consisting of the following components:
\begin{enumerate}
\item  A feature encoder $f_E$
\item  A domain classifier $f_D$
\item  A source-specific classifier $f_S$
\item  A reconstructive class-specific decoder $f_{DE}$
\end{enumerate}

\begin{figure}[h]
    \centering
    \includegraphics[width=14cm]{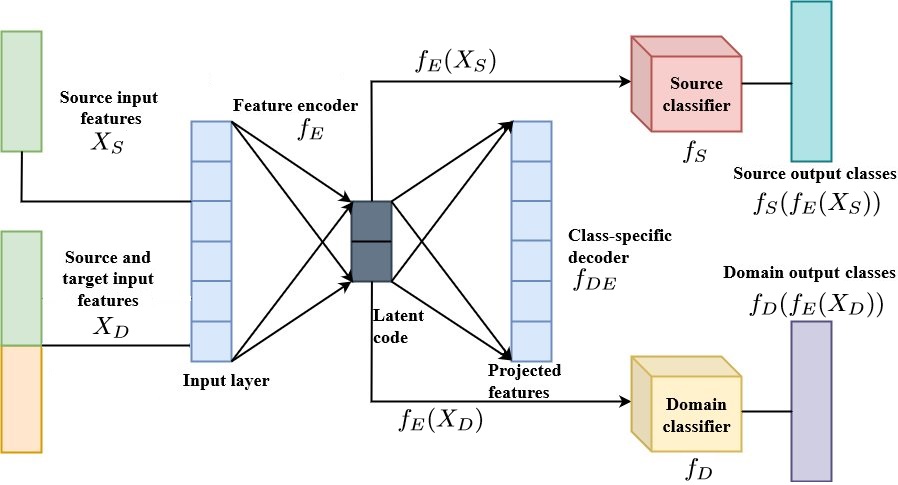}
    \caption{Schematic flow of the proposed UDA model.}
    \label{fig:Diagram_UDA}
\end{figure}

Note that the feature encoder is typically implemented using fully-connected (fc) layers with non-linearity. We denote the encoded feature representation for an input $\textbf{x}$ as $f_E(\textbf{x})$ for convenience. We provide further details on the proposed training and inference stages in the following sections. Figure \ref{fig:Diagram_UDA} illustrates a visualization of our model.

\subsubsection{Training}

The proposed loss measure in this work utilizes encoded feature representations and incorporates different components in the decoder to calculate the losses. These components include the source classifier $f_S$, the class-specific source reconstruction $f_{DE}$, the domain classifier $f_D$, and an orthogonality constraint. For the source classifier $f_S$, a multiclass softmax classifier is trained solely on $\mathcal{X}_S$. The corresponding loss is expressed using cross-entropy, which measures the log-likelihood between the training data and the model distribution. The empirical categorical cross-entropy based loss is defined as:

\begin{equation}
\mathcal{L}_S = - \mathbb{E}_{(\textbf{x}^s_i,y_i^s) \in \mathcal{X}_S} [y_i^s \log{f_S(f_E(\textbf{x}^s_i))}]
\end{equation}

The class-specific source reconstruction $f_{DE}$ aims to ensure both inter-class separation and intra-class compactness. It reconstructs the encoded source data $\hat{X}_S$ in the decoder based on $f_E(X_S)$. The loss for this reconstruction is formulated as:

\begin{equation}
\mathcal{L}_R = \sum_{i=1}^{N_S}{\|\tilde{X}_S-\hat{X}_S}\|_F^2
\end{equation}

To ensure within-class compactness and meaningfulness of the learned space, the joint minimization of $\mathcal{L}_S$ and $\mathcal{L}_R$ is performed.

The domain classifier $f_D$ projects the samples from both the source ($\mathcal{S}$) and target ($\mathcal{T}$) domains onto the shared space modeled by $f_E$. The loss for $f_D$ is defined as a typical binary cross-entropy based classification error:

\begin{equation}
\mathcal{L}_D = - \mathbb{E}_{(\textbf{x}^D_k, y^D_k) \in (X_D, Y_D)} [{y}^D_k\log{f_D(f_E(\textbf{x}^D_k))}]
\end{equation}

An orthogonality constraint is added to the total loss to ensure non-redundancy of the reconstructed source domain features. The constraint is given as:

\begin{equation}
{f_{DE}(X_S)}^Tf_{DE}(X_S) = I
\end{equation}

However, to avoid a hard constraint, a softer version is minimized:

\begin{equation}
\mathcal{L}_O = {f_{DE}(X_S)}^Tf_{DE}(X_S) - I
\end{equation}

where $I$ denotes the identity matrix.

The overall loss function $\mathcal{L}$ is optimized in two stages. In the first stage, $\mathcal{L}_1$ is minimized with respect to $f_S$ and $f_E$, incorporating $\mathcal{L}_S$, $\mathcal{L}_{R}$, and $\mathcal{L}_O$. In the second stage, $\mathcal{L}$ is further minimized with respect to $f_S$ and $f_E$, while simultaneously maximizing $\mathcal{L}_D$. The weight of the regularizer $\mathcal{R}$ on the learnable parameters is denoted by $\lambda$.

During optimization, a stochastic mini-batch gradient descent approach is followed. The order of optimization for individual terms does not affect the results. During testing, the labels for target samples are assigned using $f_S(f_E(X_T))$.

\subsubsection{Optimization and inference}

Stage 1:
The first stage involves minimizing $\mathcal{L}_1$ with respect to $f_S$ and $f_E$. This objective function consists of the sum of $\mathcal{L}_S$, $\mathcal{L}_{R}$, and $\mathcal{L}_O$:

\begin{equation}
\mathcal{L}_1 = \min\limits_{f_S, f_E}{{{\mathcal{L}_S + \mathcal{L}_{R}}}} + \mathcal{L}_O
\end{equation}

Stage 2:
In the second stage, the overall loss function $\mathcal{L}$ is optimized with respect to $f_S$ and $f_E$, while simultaneously maximizing with respect to $f_D$. This is achieved by subtracting $\mathcal{L}_D$ from $\mathcal{L}_1$, and including a regularization term $\lambda \mathcal{R}$:

\begin{equation}
\mathcal{L} = \min\limits_{f_S, f_E}\max\limits_{f_D}{{{\mathcal{L}_1 - \mathcal{L}_D}}} + \lambda \mathcal{R}
\end{equation}

Here, $\lambda$ represents the weight assigned to the regularizer $\mathcal{R}$ on the learnable parameters. The optimization process follows the standard alternate stochastic mini-batch gradient descent approach. The order of optimization for individual terms does not affect the results. During testing, labels are assigned to target samples using $f_S(f_E(X_T))$.

\subsection{Datasets and Experiments}

\subsubsection{Datasets}

Our approach has been evaluated using two standard hyperspectral datasets to verify its effectiveness.

\begin{figure}[!htb]
    \centering
    \includegraphics[width=\textwidth]{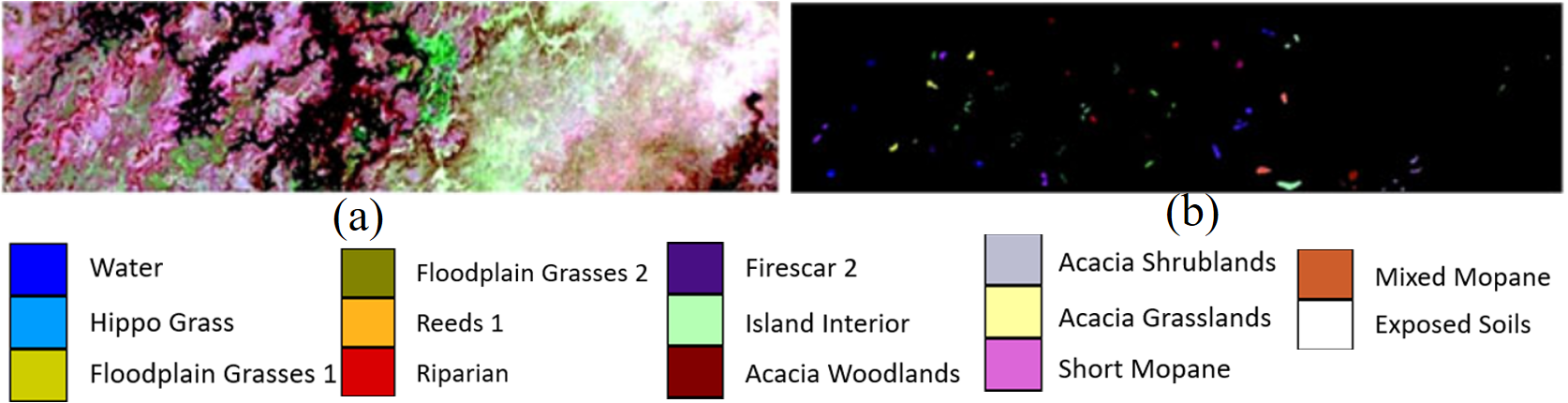}
    \caption{Botswana hyperspectral dataset with (a) True colour composite (b) Groundtruth map}
    \label{fig:botswana_dataset}
\end{figure}

\noindent\textbf{Botswana HSI dataset}: The first dataset used in our study is the Botswana hyperspectral imagery, as shown in Figure \ref{fig:botswana_dataset} \citep{neuenschwander2005results}. This dataset was collected by the NASA EO-1 satellite between 2001 and 2004, using the Hyperion sensor with a spatial resolution of 30 m. The imagery covers a strip spanning 7.7 km. It consists of 242 bands, capturing the spectral range from 400 nm to 2500 nm. However, for our study, we utilized a preprocessed version of the dataset that includes only 10 selected bands obtained through a feature selection strategy.

In this dataset, we have identified fourteen classes that correspond to different land cover features on the ground. Many of these classes exhibit fine-grained characteristics and have partially overlapping spectral signatures, making the adaptation task particularly challenging. To create the source dataset (SD) and the target dataset (TD), we selected spatially disjoint regions within the study area. The SD contains 2621 pixels, while the TD contains 1252 pixels. These spatially disjoint regions introduce subtle differences between $\mathcal{S}$ and $\mathcal{T}$, respectively.

\begin{figure}
  \centering
  \includegraphics[width=12cm]{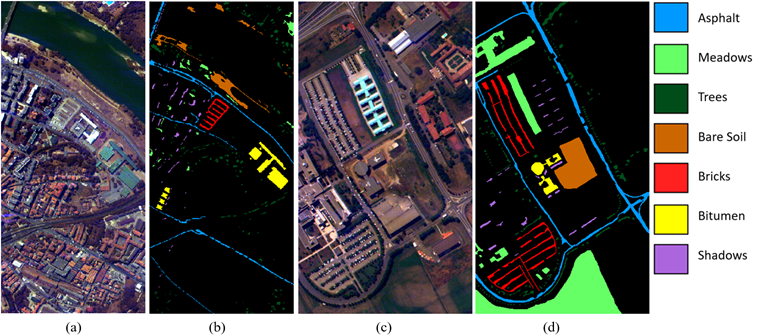}
  \caption[Pavia University and Centre dataset]{{Pavia University and Centre dataset. (a) False colour composite of Pavia Centre image (b) Groundtruth for Pavia Centre image (c) False colour composite of Pavia University image  (d) Groundtruth for Pavia University image}}
  \label{fig:paviauc_dataset}
\end{figure}

\noindent\textbf{Pavia University and Centre dataset}: The second dataset utilized in our study includes two hyperspectral images: one captured over the Pavia City Center and the other over the University of Pavia, as shown in Figure \ref{fig:paviauc_dataset} \citep{qin2018tensor}. These images were obtained using the Reflective Optics Spectrographic Image System (ROSIS). The Pavia City Center image comprises 1096 rows, 492 columns, and 102 bands, while the University of Pavia image consists of 610 rows, 340 columns, and 103 bands. Within both images, seven common classes have been identified. However, some of these classes exhibit similar spectral properties, posing challenges for their accurate classification. For our study, we designate the Pavia University image as the source dataset, and the Pavia City Center image as the target dataset. Since the Pavia City Center image has 102 bands, we ensure consistency by using the same number of bands for the Pavia University image, but excluding the last band. 

\subsubsection{Training Protocol}

The input for every model was the pixel vector of size 1$\times$1$\times$ $B$, where $B$ is the total number of hyperspectral bands. All the models were trained using Adam optimizer \citep{kingma2014adam} with learning rate of $0.0005$ for a total number of $2000$ epochs.  

\subsection{Results and Discussion}

\begin{table}[hbt!]
\centering \scriptsize
 \caption{\label{tab:bots_comp} Performance evaluation on the Botswana dataset (in \%).}
\begin{tabular}{|c|l|p{1 cm}|p{1 cm}|c|c|c|c|p{1.5 cm}|}
 \hline
 &Land-cover classes & Source Pixels & Target Pixels& TCA  & SA & GFK  & RevGrad  & Proposed Method\\
 \hline
 1&Water   & 213    &57&  60.0 & 46.0&  43.0 &\textbf{75}.0 &61.0\\
 2&Hippo grass   & 83    &81&  \textbf{100.0}  & \textbf{100.0} &  75.0 &97.0 &92.0\\
 3&Floodplain grasses 1   &199    &75 &56.0 &59.0&  69.0 &67.0 & \textbf{74.0}\\
 4&Floodplain grasses 2  & 169    &91&  75.0 &80.0&  \textbf{88.0}  &79.0 &76.0\\
 5&Reeds  & 219    &88&  78.0 & \textbf{83.0} &  81.0 &67.0 &75.0\\
 6&Riparian  & 221    &109&  58.0    &72.0&  \textbf{84.0} &65.0 &70.0\\
 7&Firescar 2   & 215    &83&  98.0 & \textbf{100.0} &  \textbf{100.0} &97.0 &\textbf{100.0}\\
 8&Island interior  & 166    &77&  62.0  &48.0&  60.0 &66.0 & \textbf{81.0}\\
 9&Acacia woodlands   & 253    &67&  27.0  &40.0&  44.0 &47.0 & \textbf{50.0}\\
 10&Acacia shrublands   & 202    &89&  40.0 &50.0&  62.0 &48.0 & \textbf{71.0}\\
 11&Acacia grasslands  & 243    &174&  79.0 &  \textbf{92.0} &  \textbf{92.0} &73.0 &74.0\\
 12&Short mopane   & 154    &85&  89.0 &\textbf{93.0}&  91.0 &73.0 &79.0\\
 13&Mixed mopane   & 203    &128&  48.0   &61.0&  65.0 &\textbf{77.0} &73.0\\
 14&Exposed soil   & 81    &48&  85.0  & \textbf{100.0} & \textbf{100.0} &79.0 &77.0\\
 \hline
 &Overall Accuracy  & -    & - &  61.0  &65.0&  70.0 &69.0 & \textbf{74.5}\\
 \hline
\end{tabular}
\end{table}

Tables \ref{tab:bots_comp} and \ref{tab:PUC_da} present the quantitative evaluation and comparison of performance between different approaches for the Botswana and Pavia datasets respectively. The highest accuracy achieved by a classifier for each class is indicated in bold.

\begin{table}[hbt!]
\centering \scriptsize
 \caption{\label{tab:PUC_da} Performance evaluation on the Pavia dataset (in \%). }
\begin{tabular}{|p{0.5cm} |p{2.8cm}| p{1.2cm} |p{1.2cm}  |p{1.0cm} |p{1.3cm} |p{1.3cm}|}
 \hline
 &Land-cover classes & Source Pixels & Target Pixels  & GFK & RevGrad & Proposed Method\\
 \hline
 1&Asphalt   & 6631    &7585&  50.0 &64.0 &\textbf{86.0}\\
 2&Meadows   & 18649    &2905 &  47.0 &61.5 &\textbf{92.0}\\
 3&Trees   &3064    &6508 &  92.0 &\textbf{94.0} & 84.0\\
 4&Baresoil   & 5029    &6549&  \textbf{97.0}  &72.5 &53.0\\
 5&Bricks   & 3682    &2140 &  62.0 &\textbf{67.0} &58.0\\
 6&Bitumen   & 1330    &7287  &  41.0  & 51.0 &\textbf{57.0}\\
 7&Shadows  & 947    &2165 &  \textbf{97.0} &83.5 & 95.0\\
 \hline
 &Overall Accuracy & -    &- &  66.0 &70.5 & \textbf{74.0}\\
 \hline
\end{tabular}
\end{table}

\begin{figure*}[hbt!]
    \centering
    \includegraphics[width=13cm]{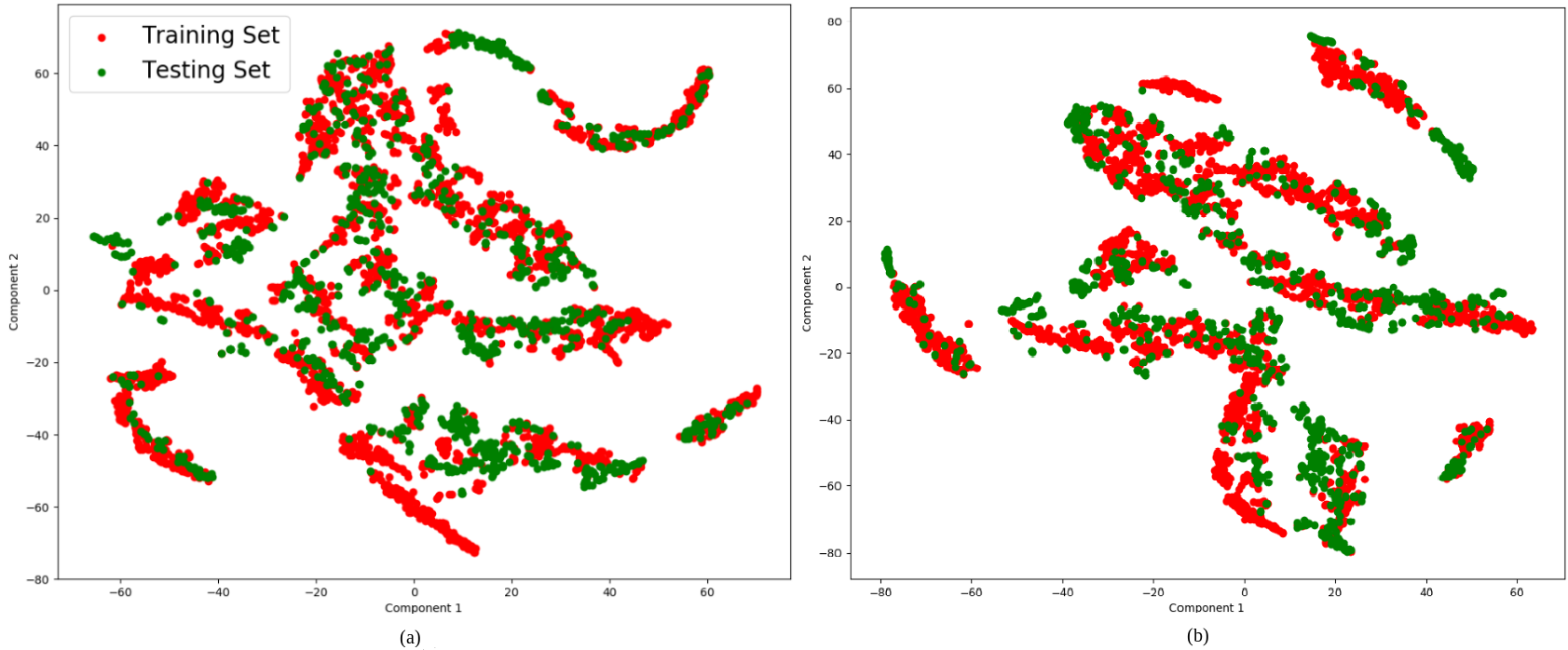}
    \caption[Two dimensional t-SNE of Botswana's source and target datasets]{Two dimensional t-SNE of Botswana's source and target datasets (a) before domain adaptation (b) after domain adaptation.}
    \label{fig:tSNE_bots}
\end{figure*}

Regarding the Botswana dataset, it can be concluded that the proposed approach outperforms the other methods, achieving an overall classification accuracy of 74.5\%. Conversely, the RevGrad technique yields an overall performance of 69\%, suggesting that aligning the domains without considering class information is not suitable for this dataset. The proposed method significantly improves the identification of the island interior (OA = 81\%), acacia woodlands (OA = 50\%), and acacia shrublands (OA = 71\%), which are challenging classes due to their similar spectral properties with other classes. The alternative approaches used for comparison mostly failed to identify these classes. The results for the remaining classes are comparable to those obtained by other techniques. Figure \ref{fig:tSNE_bots} illustrates the 2-D t-SNE plot comparing the source and target features before and after training.

 \begin{figure*}[hbt!]
    \centering
    \includegraphics[width=13cm]{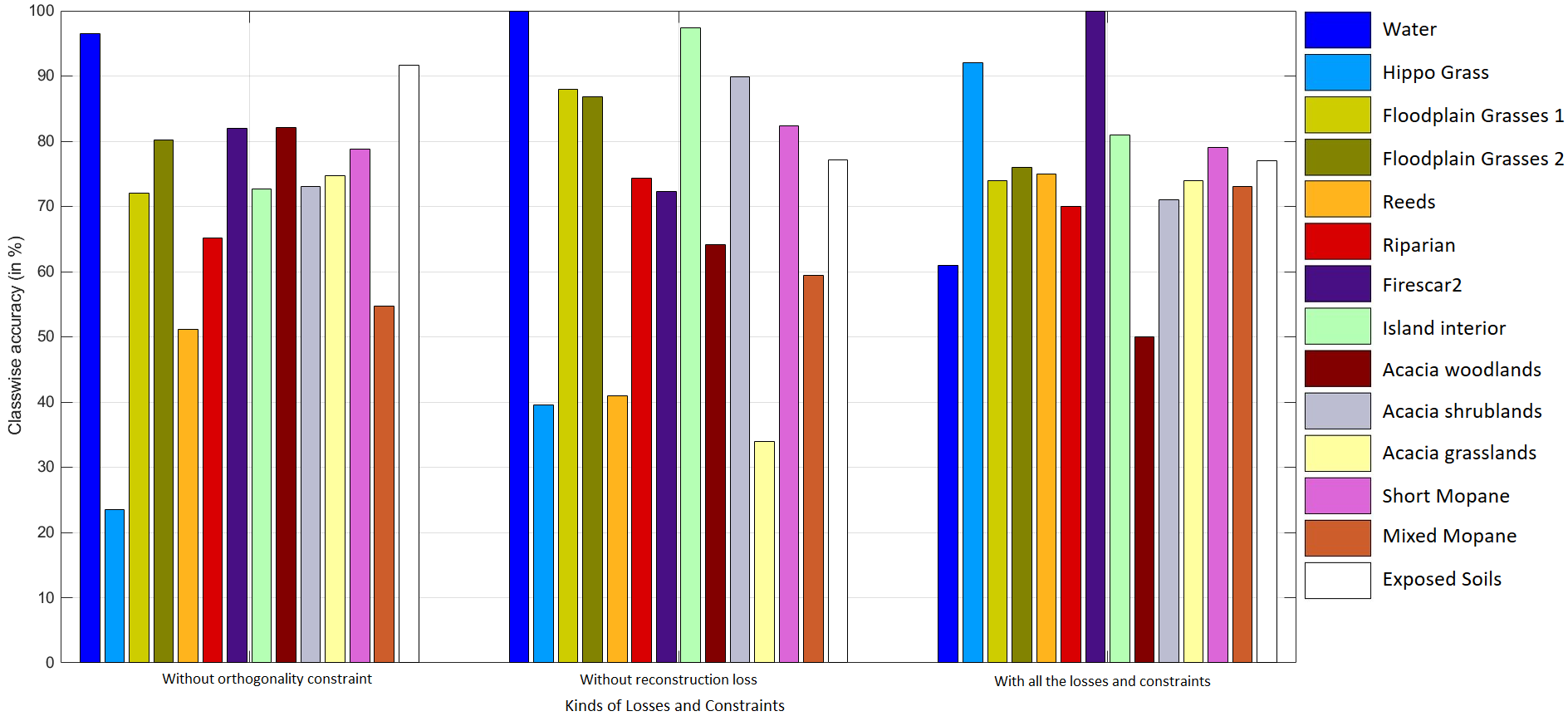}
    \caption{Bar chart for ablation study comparing effects of exclusion of different losses on the classes of Botswana dataset.}
    \label{fig:bots_ablation}
\end{figure*}

An ablation study conducted on the Botswana dataset demonstrated that an OA of 65\% is achieved when the reconstruction loss is not considered, while an OA of 64\% is recorded in the absence of the orthogonality constraint. Figure \ref{fig:bots_ablation} presents the accuracies obtained during the ablation study, showing a significant improvement in the accuracy of the hippo grass, reeds, and firescar 2 classes when all the losses and constraints are considered. However, it is observed that the accuracy of the water class considerably decreases in the same case, while the accuracies for the other classes remain more or less the same.

\begin{figure*}[hbt!]
    \centering
    \includegraphics[width=13cm]{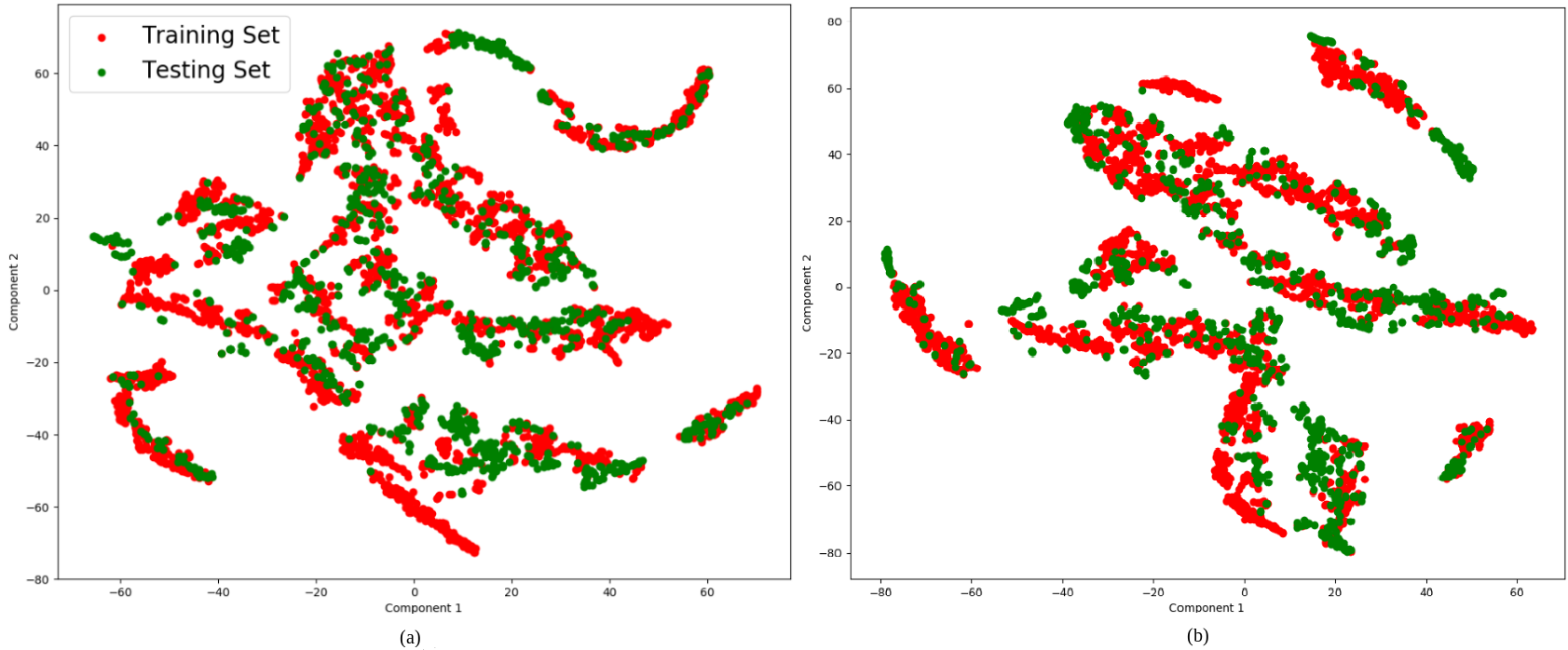}
    \caption[Two dimensional t-SNE of Pavia's source and target datasets]{Two dimensional t-SNE of Pavia's source and target datasets (a) before domain adaptation (b) after domain adaptation.}
    \label{fig:tSNE_pavia}
\end{figure*}

Similar trends are observed for the Pavia dataset, where our method surpasses the other classifiers with an overall accuracy (OA) of 74\%, while the benchmark RevGrad classifier achieves an OA of 70.5\%. This highlights the inefficiency of domain alignment without considering class information on the Pavia dataset as well. Additionally, there is a significant improvement in the classification of the asphalt (OA = 86\%) and meadows (OA = 92\%) classes. Although the spectral signatures of meadows and trees overlap (as they are both subsets of vegetation), our classifier performs better in identifying meadows compared to the other classifiers. The classification accuracies for the remaining classes are similar to those obtained by other classifiers. Figure \ref{fig:tSNE_pavia} presents the t-SNE plots of the source and target features before and after training.

 \begin{figure*}[hbt!]
    \centering
    \includegraphics[width=13cm]{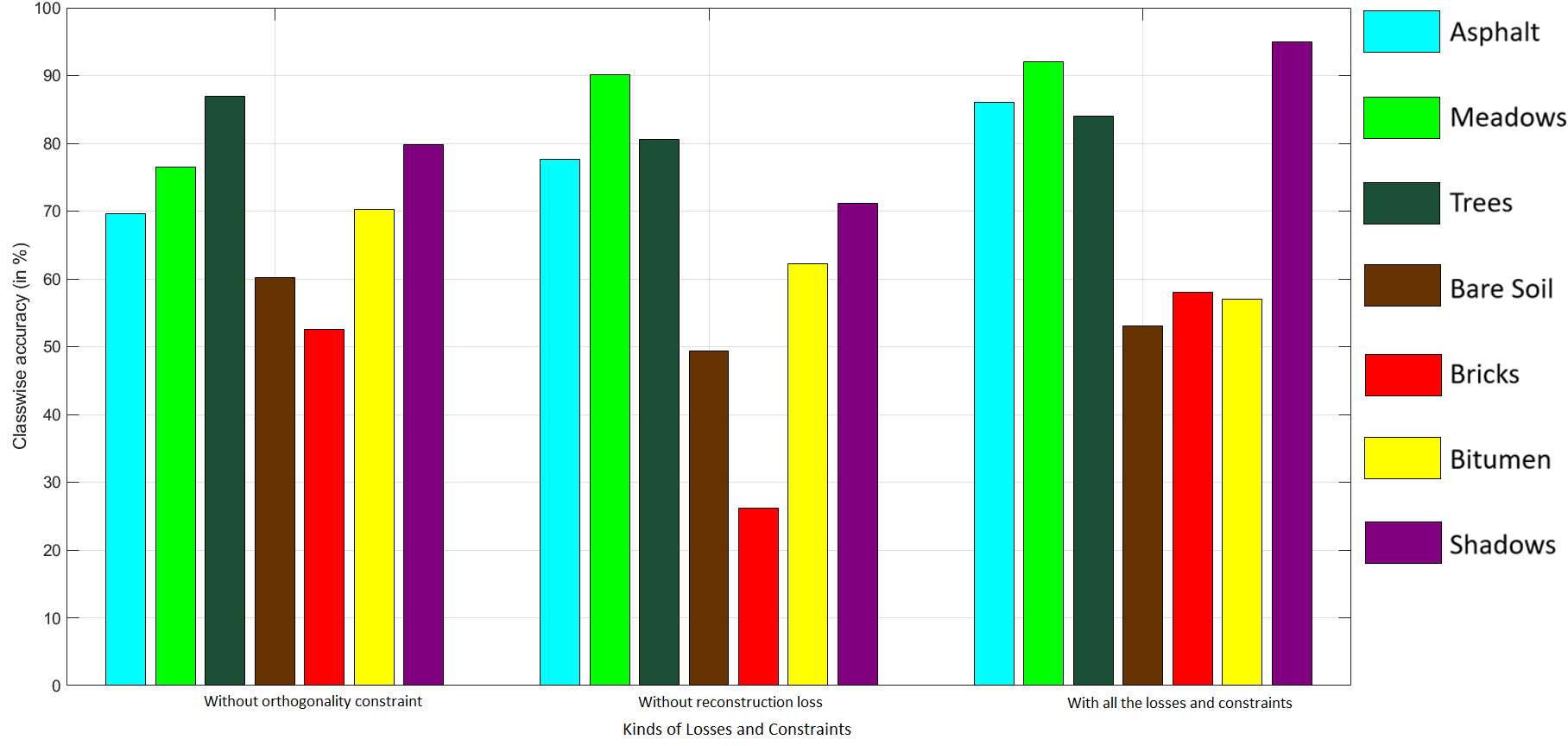}
    \caption{Bar chart for ablation study comparing effects of exclusion of different losses on the classes of Pavia dataset.}
    \label{fig:pavia_ablation}
\end{figure*}
The ablation study on the Pavia dataset reveals an overall accuracy of 65\% when the classifier is trained without the orthogonality constraint, while training without the reconstruction loss results in an OA of 71\%. Figure \ref{fig:pavia_ablation} compares the classwise accuracies for the Pavia dataset using different losses considered in our ablation study. The results demonstrate a significant improvement in the identification of shadows (OA = 95\%) and asphalt (OA = 86\%) when all the losses are considered.

\section{Attention based hyperspectral image classification}
\label{sec:AttHS}
In this section, we see a research paradigm on the attention based classification of hyperspectral images. Here, an end-to-end architecture is proposed that simultaneously extracts out the spectral and spatial attention masks from the input HSI patch. These attention masks exclusively highlight the spectral and spatial features of the HSI patch. These attention enhanced features are then combined with the original HSI features adaptively, to highlight the more contributing features for the classification task. Additionally, to enforce the notion of intraclass compactness, Wasserstein loss is added with the conventional cross-entropy loss during classification. The proposed method is discussed in subsequent sections\footnote{Published as: Shivam Pande, Biplab Banerjee, \href{https://www.sciencedirect.com/science/article/pii/S0167865521000283}{Adaptive hybrid attention network for hyperspectral image classification}, Pattern Recognition Letters 144 (2021) 6–12.}. 

\subsection{Methodology}

The objective of this research is to perform pixel classification on HSI imagery by dynamically utilizing both the spatial and spectral information through attention masks.

For this purpose, we consider an input $\mathcal{X} = \{\textbf{x}^i_P, \textbf{x}^i_V\}_{i=1}^n$, where $\textbf{x}^i_P \in \mathbb{R}^{M\times N \times B}$ represents a hyperspectral patch centered around the groundtruth pixel $\mathcal{Y} = \{y^i\}_{i=1}^n$, and $\textbf{x}^i_V \in \mathbb{R}^{1 \times B}$ denotes the hyperspectral pixel vector for the corresponding sample. The number of bands in the HSI imagery is denoted by $B$, while $M$ and $N$ respectively indicate the patch's rows and columns. Additionally, $n$ represents the total number of available groundtruth pixels. Moreover, the groundtruth labels, which belong to $K$ classes, are denoted as $\{y^i\}^n_{i=1} \in \{1,2,...,K\}$.

\subsubsection{Architecture}

\begin{figure*}[hbt!]
    \centering
    \includegraphics[width=14cm]{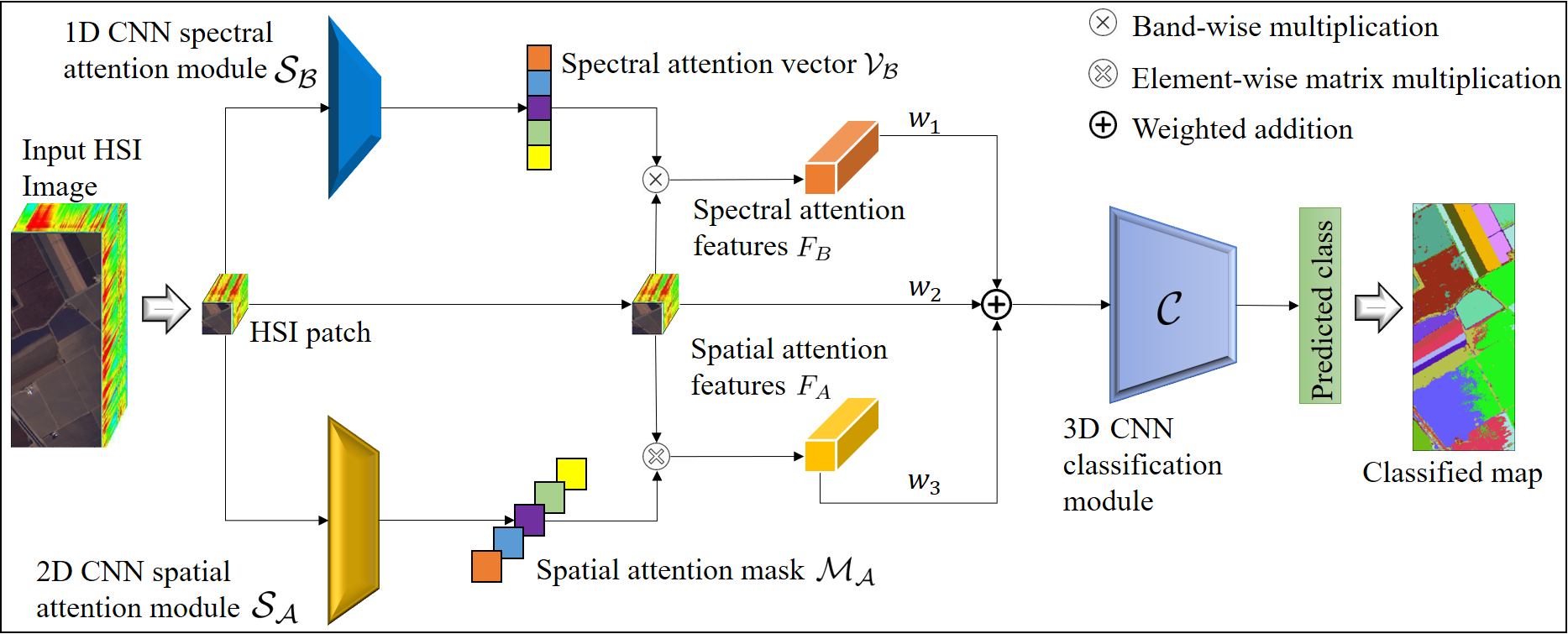}
    \caption{Schematic of hybrid attention based classification of hyperspectral images. From the figure, we can see that the hyperspectral image patch is simultaneously sent to the spectral and spatial attention modules to obtain the spectral attention vector and the 2D spatial attention mask. These attention masks selectively highlight the relevant information in the HSI patch using weighted addition (with learnable weights). The attention enhanced patch is the sent to a 3D CNN module for classification.}
    \label{fig:HyBAT}
\end{figure*}

The research aims to utilize the spectral and spatial characteristics of HSI for land use/land cover classification while investigating their combined contribution through an adaptive approach. The proposed model consists of three modules: (i) the spectral attention module $\mathcal{S_B}$ based on 1D CNN, (ii) the spatial attention module $\mathcal{S_A}$ based on 2D CNN, and (iii) the classification module $\mathcal{C}$ based on 3D CNN. Both $\mathcal{S_B}$ and $\mathcal{S_A}$ are employed simultaneously to generate the spectral attention vector $\mathcal{V_B}$ and the spatial attention mask $\mathcal{M_A}$, both having the same dimension as the input patch. The input patch is multiplied with $\mathcal{V_B}$ to highlight its spectral features, while it is multiplied with $\mathcal{M_A}$ to emphasize its spatial features. The spectrally and spatially enhanced features, along with the original input patch, are adaptively combined (using learnable weights) and passed to the classification module $\mathcal{C}$. To further improve the classification performance, the model is jointly trained using classwise Wasserstein loss and cross-entropy loss. The proposed architecture is illustrated in Figure \ref{fig:HyBAT}, and detailed explanations of the modules are provided in the subsequent sections.

\noindent\textbf{Spectral attention module $\mathcal{S_B}$}: The module $\mathcal{S_B}$ employs a 1D CNN structure, comprising of five convolution layers. It takes the hyperspectral pixel vector $\textbf{x}^i_V$ as input and produces an attention vector $\mathcal{V_B}$ of the same size, i.e., $1 \times B$. The kernel specifications for the spectral attention module are presented in Table \ref{tab:spectral}. Each layer, denoted as \textit{Conv1}, \textit{Conv2}, and so on, is activated using the rectified linear unit (ReLU) activation function and does not employ padding. Batch normalization is applied to all layers except \textit{Conv5}. The module can be represented as $\mathcal{S_B}(\theta_{SB}, \textbf{x}^i_V)$, and the spectrally enhanced features are expressed in Equation (\ref{eqn:Spec}).

\begin{equation}
F_B(\textbf{x}^i_V, \textbf{x}^i_P) = \mathcal{S_B}(\theta_{{SB}}, \textbf{x}_V^i) \otimes \textbf{x}_P^i
\label{eqn:Spec}
\end{equation}

In the equation, $F_B(\textbf{x}^i_V, \textbf{x}^i_P)$ represents the spectrally enhanced features, $\theta_{SB}$ denotes the parameters of the spectral attention module, and $\otimes$ signifies the elementwise multiplication operator applied element-wise between the attention vector and the input patch.

\begin{table}[!h]
\centering{\scriptsize
 \caption{\label{tab:spectral} Structure of $\mathcal{S_B}$ for Houston 13 dataset.}
\begin{tabular}{|p{1.0cm}|p{1.4cm}|p{1.3cm}|p{1.0cm}|}
 \hline
Layer &  Kernel size & Channels & Stride\\
 \hline
Conv1 & 20 & 18 & 1 \\
Conv2 & 20 & 36 & 1\\
Conv3 & 20 & 72 & 2\\
Conv4 & 20 & 108 & 2\\
Conv5 & 13 & 144 & 1\\
\hline
\end{tabular}}
\end{table}

\noindent\textbf{Spatial attention module $\mathcal{S_A}$}: The module $\mathcal{S_A}$ is a 2D CNN-based component consisting of three convolution layers, one residual layer (including the first and third convolutions), and three deconvolution layers. It takes the hyperspectral patch $\textbf{x}^i_P$ as input and produces an attention mask $\mathcal{M_A}$ of the same size, specifically $M \times N \times B$. The architecture specifications for $\mathcal{S_A}$ can be found in Table \ref{tab:spatial}. In the table, \textit{Conv1}, \textit{Conv2}, and \textit{Conv3} represent the spatial convolution kernels, \textit{Res1} and \textit{MP1} denote the residual and maxpooling layers respectively, while \textit{Deconv1}, \textit{Deconv2}, and \textit{Deconv3} correspond to the deconvolution layers. All layers employ the rectified linear unit (ReLU) activation function, and all layers, except \textit{Deconv3}, are followed by a batch normalization layer. The spatially enhanced features are denoted by $F_A(\textbf{x}^i_P)$, and they are obtained by element-wise multiplication between $\mathcal{S_A}(\theta_{SA}, \textbf{x}_P^i)$, which represents the output of $\mathcal{S_A}$ with its parameters $\theta_{SA}$, and the input patch $\textbf{x}_P^i$. Equation (\ref{eqn:Spat}) demonstrates this relationship, where $\theta_{SA}$ refers to the parameters of the spatial attention module.
\begin{equation}
F_A(\textbf{x}^i_P) =  \mathcal{S_A}(\theta_{SA}, \textbf{x}_P^i) \otimes  \textbf{x}_P^i
\label{eqn:Spat}
\end{equation}

\begin{table}[!h]
\centering{\scriptsize
\caption{\label{tab:spatial}Structure of $\mathcal{S_A}$ for Houston 13 dataset.}
\begin{tabular}{|p{1.0cm}|p{1.4cm}|p{1.3cm}|p{1.0cm}|p{1.0cm}|}
 \hline
Layer &  Kernel size & Channels & Stride & Padding\\
 \hline
Conv1 & 3$\times$3 & 256 & 1 & No\\
Conv2 & 3$\times$3 & 256 & 1 & Yes\\
Conv3 & 3$\times$3 & 512 & 1 & Yes\\
Res1 & - & 768 & - & -\\
MP1 & - & 768 & 2 & No\\
Deconv1 & 4$\times$4 & 256 & 1 & No\\
Deconv2 & 5$\times$5 & 256 & 1 & No\\
Deconv3 & 1$\times$1 & 144 & 1 & No\\
\hline
\end{tabular}}
\end{table}

\begin{figure*}[t!]
  \centering
  \centerline{\includegraphics[width=14.5cm]{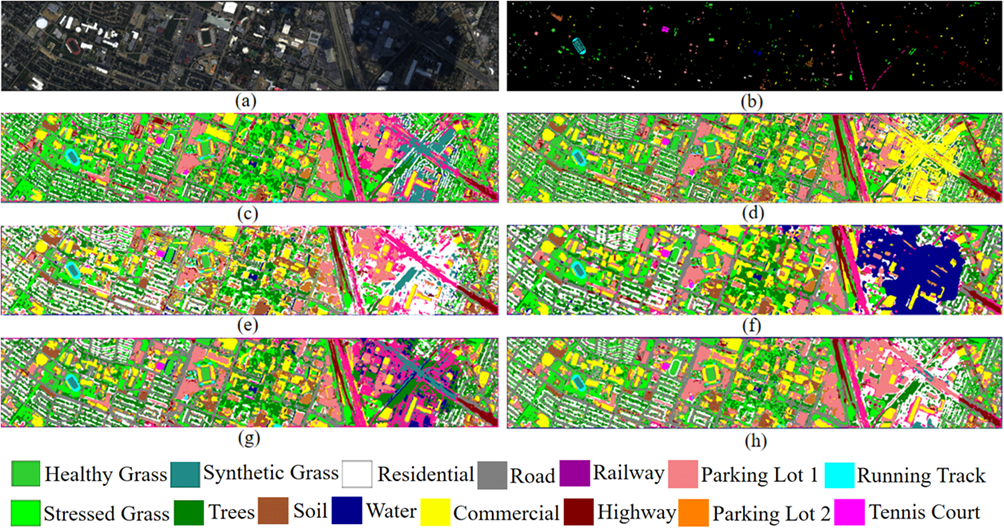}}
  \caption[Houston 13 HSI dataset with RGB image, groundtruth and classification maps]{Houston 13 hyperspectral dataset. (a) Colour composite of 3 bands from the image (b) Groundtruth with classes. Classification maps generated using (c) 2D CNN (d) Two branch CNN (e) FusAtNet (f) SSRN (g) 3D CNN and (h) Proposed.}\medskip
  \vspace{-0.5cm}
  \label{fig:Houston13}
\end{figure*}

\noindent\textbf{Classification module $\mathcal{C}$}: The classification module $\mathcal{C}(\theta_C, \textbf{x}^i_V, \textbf{x}^i_P)$, with $\theta_C$ representing the module's parameters, is built on a 3D CNN. It consists of four 3D convolution layers (\textit{Conv1}, \textit{Conv2}, \textit{Conv3}, and \textit{Conv4}) and two 3D maxpooling layers (\textit{MP1} and \textit{MP2}). The architecture of $\mathcal{C}$ can be found in Table \ref{tab:classifier}. All convolution layers, except the last one, employ ReLU activation and batch normalization. The final layer, \textit{Conv4}, uses a \textit{softmax} activation function. The input to $\mathcal{C}$ is obtained by weighted summation of spectrally and spatially enhanced features with the original input, as described in Eq. (\ref{eqn:WtAdd}). The output is a vector of size $1\times K$ ($K$ being the number of classes).
\begin{table}[!h]
\centering{\scriptsize
 \caption{\label{tab:classifier}Structure of $\mathcal{C}$ for Houston 13 dataset.}
\begin{tabular}{|p{1.0cm}|p{1.4cm}|p{1.3cm}|p{1.0cm}|p{1.0cm}|}
 \hline
Layer &  Kernel size & Channels & Stride & Padding\\
 \hline
Conv1 & 3$\times$3$\times$36 & 32 & 1 & Yes\\
MP1 & - & 32 & 2 & No\\
Conv2 & 3$\times$3$\times$36 & 64 & 1 & Yes\\
MP2 & - & 64 & 2 & No\\
Conv3 & 3$\times$3$\times$36 & 128 & 1 & Yes\\
Conv4 & 2$\times$2$\times$36 & 15 & 1 & No\\
\hline
\end{tabular}}
\end{table}

\begin{equation}
\mathcal{I}_\mathcal{C}(\textbf{x}^i_V, \textbf{x}^i_P) = w_{1}F_B(\textbf{x}^i_V, \textbf{x}^i_P) \oplus w_{2}F_A(\textbf{x}^i_P) \oplus w_{3}\textbf{x}_P^i 
\label{eqn:WtAdd}
\end{equation}

subject to the constraint
\begin{equation}
w_{1} + w_{2} + w_{3} = 1
\label{eqn:WtAdd2}
\end{equation}

In the above equations, $\mathcal{I}_\mathcal{C}(\textbf{x}^i_V, \textbf{x}^i_P)$ represents the input to the classifier, $\oplus$ denotes elementwise addition, and $w_1$, $w_2$, and $w_3$ are trainable weights.

\subsubsection{Training and inference}

The output of $\mathcal{C}$ is jointly trained using categorical cross-entropy loss and Wasserstein loss, as presented in Eq. (\ref{eqn:TLoss}). The Wasserstein loss is employed to enhance intraclass compactness.

\begin{equation}
\mathcal{L}_C = \textrm{min}\big(\mathcal{L}_{CE} - \mathcal{L}_W\big)
\label{eqn:TLoss}
\end{equation}

The first term in Eq. (\ref{eqn:TLoss}) corresponds to categorical cross-entropy loss, as expressed in Eq. (\ref{eqn:CELoss_att}).

\begin{equation}
\mathcal{L}_{CE} =  -\sum_{i=1}^{n}  y^i\ln{\hat{y}^i}
\label{eqn:CELoss_att}
\end{equation}

Here, $\hat{y}^i = \mathcal{C}(\theta_{C},\mathcal{I}_\mathcal{C}(\textbf{x}_V^i,\textbf{x}_P^i))$.

The second term in Eq. (\ref{eqn:TLoss}) represents the Wasserstein loss, as defined in Eq. (\ref{eqn:WLoss}). In this equation, $\mathbb{E}$ denotes the expectation operator.

\begin{equation}
\mathcal{L}_W = \mathbb{E}_{(y^i)} [y^i] - \mathbb{E}_{(\textbf{x}_V^i,\textbf{x}_P^i)} [\mathcal{C}(\theta_{C},\mathcal{I}_\mathcal{C}(\textbf{x}_V^i,\textbf{x}_P^i))]
\label{eqn:WLoss}
\end{equation}

Determining the Wasserstein distance is essentially a maximization problem \citep{weng2019gan} related to the term in Eq. (\ref{eqn:WLoss}). However, since the overall objective is to minimize the loss function, we subtract the Wasserstein loss $\mathcal{L}_W$ in Eq. (\ref{eqn:TLoss}).

\subsection{Experimental setup}

This section provides an overview of the datasets used for evaluation and presents the experimental details.

\subsubsection{Datasets}

To demonstrate the effectiveness of the proposed approach, three hyperspectral datasets were employed.

\textbf{Houston 2013}: This dataset was captured over the University of Houston campus by the National Centre for Airborne Laser Mapping (NCALM) and made available for the GRSS data fusion contest in 2013. It comprises a hyperspectral imagery with a total of 144 bands, where each band has dimensions of 349$\times$1905. The ground truth consists of 15 land use/land cover classes, containing a total of 15,029 samples. The dataset is already partitioned into a training set (with 2,832 samples) and a test set (with 12,197 samples) \citep{mohla2020fusatnet} (Figure \ref{fig:Houston13}). For reference, this dataset will be referred to as Houston 13.

\begin{figure}[htb]
\begin{minipage}[b]{1.0\linewidth}
  \centering
  \centerline{\includegraphics[width=\textwidth]{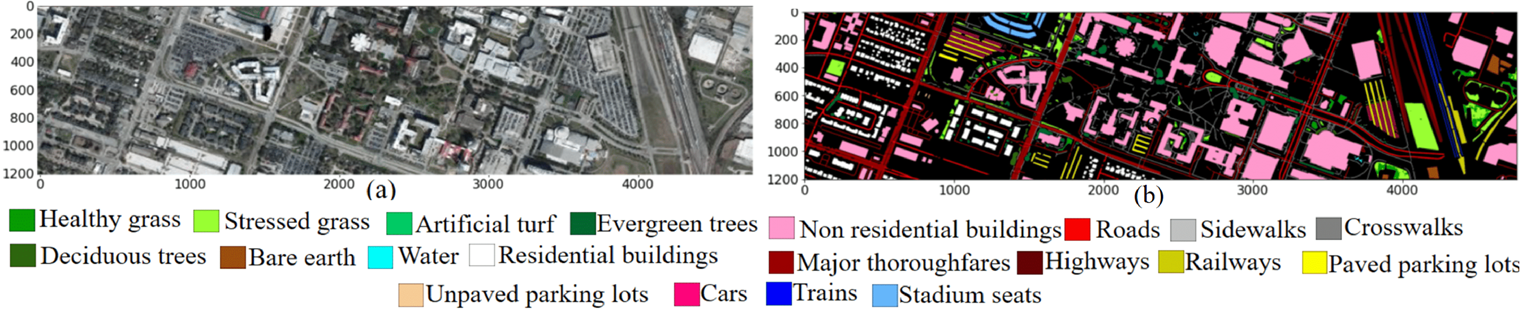}}
\end{minipage}
\caption[Houston 18 HSI dataset with RGB image and groundtruth.]{Houston 18 hyperspectral dataset. (a) Colour composite of 3 bands from the image (b) Groundtruth with classes.}
\label{fig:h18_data}
\end{figure}

\textbf{Houston 2018}: This dataset was specifically prepared for the 2018 data fusion contest. It comprises an image with 48 bands and a spatial size of 601$\times$2384. The ground truth was obtained from a high-resolution image measuring 1202$\times$4768. To align the hyperspectral image with the ground truth map, the HSI image was resized accordingly, and ground truth samples were extracted. The dataset includes 20 land use/land cover classes, totaling 2,018,910 samples. Due to the large number of samples, a random selection was made, with 100 samples per class used for training and 2000 samples per class used for testing. However, for classes with fewer than 2000 samples, the remaining samples were allocated for testing. As a result, the training set consists of 2000 samples, while the testing set comprises 37,451 samples \citep{xu2018multi} (Figure \ref{fig:h18_data}). This dataset will be referred to as Houston 18 for future reference.

\begin{figure}
\begin{minipage}[b]{1.0\linewidth}
  \centering
  \centerline{\includegraphics[width=0.8\textwidth]{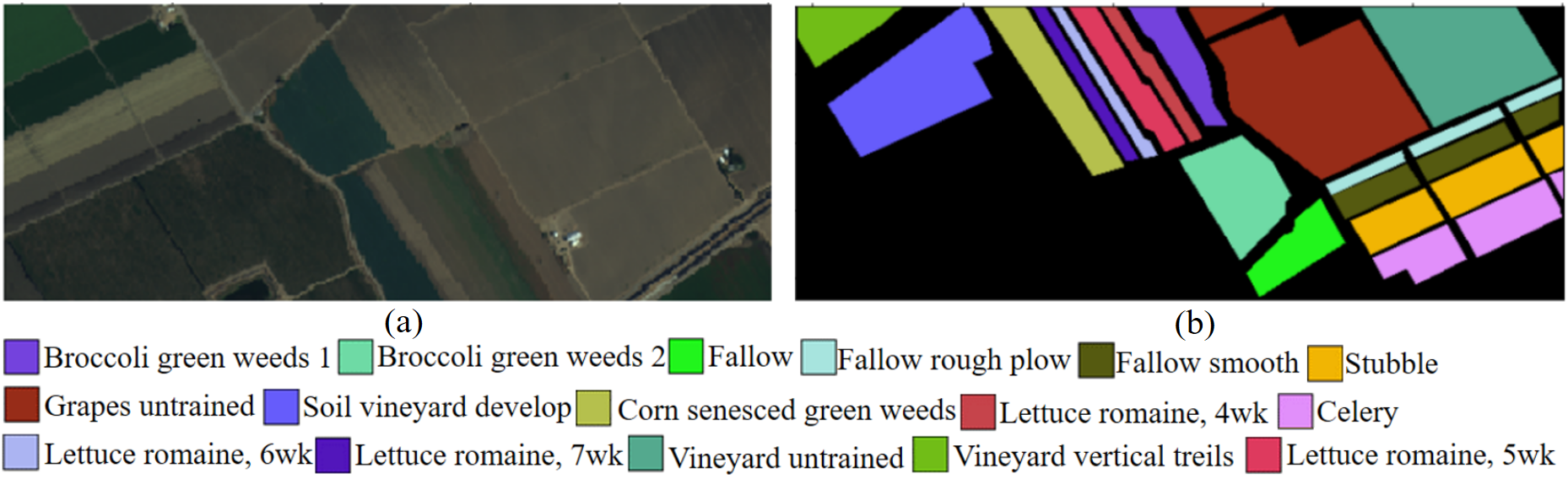}}
\end{minipage}
\caption[Salinas Valley HSI dataset with RGB image and groundtruth.]{Salinas Valley hyperspectral dataset. (a) Colour composite of 3 bands from the image (b) Groundtruth with classes.}
\label{fig:sal_data}
\end{figure}

\textbf{Salinas Valley}: The Salinas Valley hyperspectral imagery was captured by the AVIRIS sensor over the Salinas Valley in California. It comprises 204 bands, with each band having a spatial size of 512$\times$217. The scene is divided into 16 classes, and ground truth information is available for 54,129 pixels. For training purposes, a total of 1,600 samples were selected, with 100 samples per class, while the remaining samples were used for testing \citep{mou2019learning} (Figure \ref{fig:sal_data}).

For all the datasets, the background in the ground truth is represented by the colour black in the corresponding figures, and the class names are specified in Tables \ref{tab:hous13_perf_att}, \ref{tab:hous18_perf_att}, and \ref{tab:sal_perf_att}.

\subsubsection{Training protocols}

Our network is trained using hyperspectral patches of size 11$\times$11$\times B$ and the corresponding hyperspectral vectors of size 1$\times B$ for all the datasets. We employed the Nadam optimizer \citep{dozat2016incorporating}, with fixed learning rates of 0.00002, 0.00001, and 0.000015 for the Houston 13, Houston 18, and Salinas Valley datasets, respectively. The number of epochs used for training were 150, 100, and 250 for the respective datasets. To augment the training data, we applied four-fold rotation to the patches, rotating them by 90, 180, and 270 degrees. All the hyperparameters were empirically determined using a Grid Search approach \citep{syarif2016svm}.

We compared the performance of our model against state-of-the-art benchmark models, including 1D CNN \citep{chen2016deep}, spectral-spatial cascaded recurrent neural networks (SSCasRNN) \citep{hang2019cascaded}, 2D CNN \citep{mou2019learning}, spectral attention network (SpecAttenNet) \citep{mou2019learning}, spectral spatial residual network (SSRN) \citep{zhong2017spectral}, FusAtNet \citep{mohla2020fusatnet}, HybridSN \citep{roy2019hybridsn}, two-branch CNN \citep{xu2017multisource}, and 3D CNN \citep{chen2016deep}. To ensure a fair evaluation, we used the same patch size, number of epochs, and augmented data for all the benchmark models. Performance metrics such as overall accuracy (OA), class-wise accuracy, average accuracy (AA), and kappa coefficient ($\kappa$) \citep{mohla2020fusatnet} were employed for comparison. All experiments were conducted using \textit{Tensorflow 2.0} on \textit{Google Colaboratory}.

\subsection{Results and Discussion}

The proposed method is evaluated on three hyperspectral datasets: Houston 13, Houston 18, and Salinas Valley, and their results are presented in Tables \ref{tab:hous13_perf_att}, \ref{tab:hous18_perf_att}, and \ref{tab:sal_perf_att}, respectively. Across all three datasets, our proposed approach has outperformed existing benchmarks in terms of overall accuracy, with accuracies of 87.99\%, 86.24\%, and 97.16\%, respectively. Similarly, our approach achieves higher average accuracy (90.07\%, 87.11\%, 98.69\%) and $\kappa$ scores (0.8696, 0.8549, 9683) compared to the benchmarks. Moreover, our approach demonstrates consistent performance across different classes, surpassing benchmark methods in several classes such as \textit{stressed grass}, \textit{synthetic grass}, \textit{soil}, \textit{residential}, \textit{road}, and \textit{parking lot 1} for Houston 13; \textit{artificial turf}, \textit{residential buildings}, \textit{unpaved parking lots}, and \textit{trains} for Houston 18; and \textit{Lettuce romaine, 5wk}, \textit{Lettuce romaine, 6wk}, and \textit{Vineyard untrained} for Salinas Valley. Classification maps for the Houston 13 dataset are shown in Figure \ref{fig:Houston13}, where it can be observed that our maps exhibit precise and noise-free class delineation. Additionally, in the right portion of the image, our method correctly identifies the shadowy region as commercial, whereas other methods mislabel it as water or railway track. The improved performance of our model can be attributed to several factors. Firstly, the use of separate spectral and spatial attention modules enables the model to extract distinct features for each class, enhancing classification accuracy. Secondly, the adaptive combination of spectral, spatial, and original features ensures their optimal contributions to the classification process. Lastly, the incorporation of class-wise Wasserstein loss promotes compactness of classes in the feature space, resulting in improved classification performance.

\begin{sidewaystable*}[htbp]
  \centering{\scriptsize
  \caption{\label{tab:hous13_perf_att} Accuracy analysis on the Houston 13 dataset (in \%)}
    \begin{tabular}
    {|p{0.5cm}|p{3.2cm}|p{0.9cm}|p{0.9cm}|p{1.0cm}|p{1.0cm}| p{1.6cm}|p{1.0cm}|p{0.9cm}|p{1.0cm}|p{0.9cm}|p{0.9cm}|}
 \hline
 S.No.&Classes & 1D CNN & 2D CNN & Spec Atten Net & SS Cas RNN  & Two Branch CNN & FusAt Net & SSRN & Hybrid SN & 3D CNN & HybAt Net\\
 \hline
    1& Healthy grass &  83.00&	80.34&	81.01&	83.00&	82.34&	81.86&	81.96&	81.77&	83.10&	82.34\\
    2& Stressed grass&82.80  & \textbf{85.15} & 83.27 & 82.71 & 84.49 & 84.96 & 85.06 & 81.30  & \textbf{85.15} & \textbf{85.15} \\
    3& Synthetic grass&\textbf{100.00} & 99.60  & \textbf{100.00} & 98.81 & \textbf{100.00} & 99.80  & 99.80  & 97.23 & \textbf{100.00} & \textbf{100.00} \\
    4& Trees&91.67 & 99.81 & 93.18 & 88.64 & 99.34 & 93.09 & 91.19 & 91.10  & \textbf{99.91} & 99.43 \\
    5& Soil &98.20  & 93.18 & 97.92 & 95.45 & 99.34 & \textbf{100.00} & \textbf{100.00} & 99.91 & 96.78 & \textbf{100.00} \\
    6& Water &95.10  & \textbf{100.00} & \textbf{100.00} & 95.80  & \textbf{100.00} & \textbf{100.00}   & 97.20  & 94.41 & \textbf{100.00} & 99.30 \\
    7& Residential &81.44 & 91.51 & 91.98 & 85.73 & 90.11 & 92.16 & 83.21 & 86.75 & 90.76 & \textbf{97.67} \\
    8& Commercial&37.61 & 66.00 & 63.72 & 49.95 & 84.90  & \textbf{75.59} & 72.65 & 67.33 & 65.62 & 68.19 \\
    9& Road &75.64 & 81.96 & 74.50  & 81.59 & 76.30  & 84.32 & 78.85 & 80.08 & 87.16 & \textbf{88.67} \\
    10&Highway&53.96 & 52.41 & 63.22 & 89.96 & 62.36 & 63.80  & 48.55 & \textbf{65.15} & 60.71 & 64.77 \\
    11& Railway &76.47 & 77.04 & 48.01 & 78.37 & 73.34 & 66.51 & 79.98 & 70.68 & \textbf{80.46} & 78.56 \\
    12& Parking lot 1&76.95 & 92.22 & 91.16 & 87.42 & 92.51 & 89.72 & 89.05 & 94.91 & 97.69 & \textbf{99.14} \\
    13& Parking lot 2&69.12 & 90.18 & 83.51 & 81.75 & \textbf{92.98} & 92.63 & 91.23 & 84.91 & 90.53 & 91.23 \\
    14& Tennis court &99.60  & 98.79 & \textbf{100.00} & 95.55 & 97.98 & 99.60  & \textbf{100.00} & 97.98 & 99.60  & 96.76 \\
    15& Running track &98.31 & \textbf{100.00} & \textbf{100.00} & 98.94 & 98.52 & \textbf{100.00} & \textbf{100.00} & 96.19 & 99.37 & 99.79 \\
     \hline
     & Overall Accuracy (OA) &78.26 & 84.19 & 81.32 & 84.02 & 86.37 & 85.33 & 83.42 & 83.65 & 86.59 & \textbf{87.99} \\
     & Average Accuracy (AA) &81.32 & 87.21 & 84.77 & 86.24 & 88.97 & 88.27 & 86.58 & 85.98 & 89.12 & \textbf{90.07} \\
     & Cohen's Kappa $\kappa (\%)$ &0.7659 & 0.8287 & 0.7981 & 0.8270  & 0.8521 & 0.8410 & 0.8214 & 0.8235 & 0.8547 & \textbf{0.8696} \\
     \hline
    \end{tabular}}
\end{sidewaystable*}

\begin{figure}[htb]
\begin{minipage}[b]{1.0\linewidth}
  \centering
  \centerline{\includegraphics[width=8.5cm]{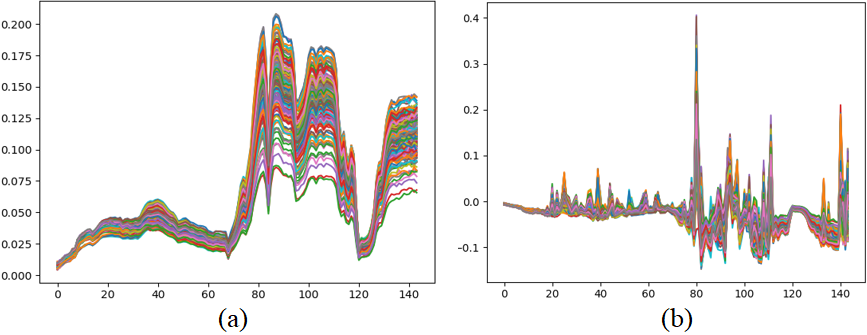}}
\end{minipage}
\caption{Comparison of spectral signatures for ``Trees'' class of Houston 13 dataset (a) before attention and (b) after attention.}
\label{fig:spectra}
\end{figure}

\begin{sidewaystable*}[htbp]
  \centering{\scriptsize
  \caption{\label{tab:hous18_perf_att}Accuracy analysis on the Houston 18 dataset (in \%)}
    \begin{tabular}{|p{0.5cm}|p{3.3cm}|p{0.9cm}|p{0.9cm}|p{1.0cm}|p{1.0cm}|p{1.6cm}|p{1.0cm}|p{0.9cm}|p{1.0cm}|p{0.9cm}|p{0.9cm}|}
 \hline
 S.No.&Classes & 1D CNN & 2D CNN & Spec Atten Net & SS Cas RNN  & Two Branch CNN & FusAt Net & SSRN & Hybrid SN & 3D CNN & HybAt Net\\
 \hline
    1     & Healthy grass     & 91.85 & 82.45 & 97.25 & \textbf{97.50} & 96.00 & 93.60  & 75.75 & 92.65 & 92.30  & 97.20 \\
    2     & Stressed grass     & 86.45 & 84.40  & 82.90  & 76.75 & 83.45 & 87.60  & 93.60  & 81.20  & \textbf{92.05} & 88.30 \\
    3     & Artificial turf     & \textbf{100.00} & 99.70  & 99.95 & 99.35 & \textbf{100.00} & 99.95 & \textbf{100.00} & 99.90  & \textbf{100.00} & \textbf{100.00} \\
    4     & Evergreen trees     & 92.35 & 95.45 & 96.25 & 95.65 & \textbf{97.40} & 95.40  & 96.40  & 97.05 & 96.70  & 95.85 \\
    5     & Deciduous trees     & 81.50  & 91.75 & 88.20  & 75.00  & 88.60  & 91.35 & 93.40  & 90.75 & \textbf{94.20} & 91.95 \\
    6     & Bare earth     & 92.65 & 71.20  & 96.70  & 94.90  & 98.70  & 97.10  & \textbf{100.00} & 98.30  & 95.50  & 97.60 \\
    7     & Water     & 96.78 & 99.59 & \textbf{100.00}   & 99.38 & 99.90  & 99.59 & \textbf{100.00} & 99.38 & \textbf{100.00} & 99.79 \\
    8     & Residential buildings & 72.80  & 80.65 & 77.25 & 75.65 & 82.65 & 78.30  & 77.30  & 76.80  & 77.10  & \textbf{83.80} \\
    9     & Non-residential buildings & 39.20  & 70.10  & 62.65 & 56.75 & \textbf{81.40} & 64.15 & 67.35 & 78.80  & 71.55 & 74.10 \\
    10    & Roads    & 31.65 & 33.85 & 22.80  & 16.05 & 42.80  & 48.70  & 40.75 & 41.95 & \textbf{70.95} & 54.50 \\
    11    & Sidewalks    & 34.75 & 56.25 & 52.40  & 56.00    & 59.30  & 53.15 & \textbf{68.40} & 40.30  & 51.10  & 49.30 \\
    12    & Crosswalks    & 39.60  & 66.30  & 75.50  & 38.90  & \textbf{80.25} & 64.45 & 60.75 & 79.40  & 31.30  & 75.75 \\
    13    & Major thoroughfares    & 33.85 & 49.65 & 44.40  & 41.45 & 57.40  & 55.20  & \textbf{67.45} & 50.85 & 19.70  & 60.85 \\
    14    & Highways    & 69.85 & 90.55 & 93.05 & 69.35 & 80.40  & 77.55 & 83.20  & 87.25 & \textbf{91.80} & 88.45 \\
    15    & Railways    & 97.15 & 96.10  & 97.55 & 92.35 & 96.90  & 97.45 & \textbf{99.85} & 98.75 & 94.90  & 98.45 \\
    16    & Paved parking lots    & 85.45 & 89.25 & 81.25 & 86.90  & 91.15 & \textbf{91.95} & 89.45 & 91.55 & 85.30  & 92.85 \\
    17    & Unpaved parking lots    & \textbf{100.00} & 98.56 & 99.79 & 99.79 & \textbf{100.00} & 99.59 & \textbf{100.00} & \textbf{100.00} & \textbf{100.00} & \textbf{100.00} \\
    18    & Cars    & 62.75 & 91.35 & 88.75 & 94.20  & 94.10  & 91.05 & 97.65 & 88.45 & \textbf{97.80} & 97.00 \\
    19    & Trains    & 74.55 & 96.40  & 94.05 & 84.95 & 95.15 & 97.50  & 94.60  & 91.75 & 98.25 & \textbf{98.35} \\
    20    & Stadium seats    & 82.95 & 99.80  & 94.70  & 93.35 & 98.00    & 97.50  & 96.60  & 97.80  & \textbf{98.75} & 98.20 \\
    \hline
        & Overall Accuracy (OA)    & 71.58 & 81.02 & 81.07 & 75.69 & 85.24 & 83.00   & 84.11 & 83.08 & 81.80  & \textbf{86.24} \\
        & Average Accuracy (AA)    & 73.31 & 82.17 & 82.27 & 77.21 & 86.18 & 84.06 & 85.13 & 84.14 & 82.96 & \textbf{87.11} \\
        & Cohen's Kappa $\kappa$ & 0.7002 & 0.7998 & 0.8003 & 0.7435 & 0.8443 & 0.8206 & 0.8324 & 0.8215 & 0.8080  & \textbf{0.8549} \\
     \hline
    \end{tabular}}
\end{sidewaystable*}

Furthermore, we conducted three ablation studies to assess the performance of our method under different conditions. Firstly, we evaluated the model's performance without the Wasserstein loss ($\mathcal{L_W}$) and adaptive learning. Secondly, we examined the contribution of attention branches in the classification. Lastly, we observed the model's performance without data augmentation. The results of the first ablation study are presented in Table \ref{tab:ablation}, where ``WtAdd'' represents weighted addition or adaptive learning. It is evident that the models perform less effectively in the absence of these modules. As expected, the lowest overall accuracy is observed when both modules are removed (86.45\% for Houston 13, 85.21\% for Houston 18, and 96.41\% for Salinas Valley). It is further observed that in the Houston 18 dataset, the effect of adaptive learning is dominant, as it outperforms $\mathcal{L_W}$ individually. On the other hand, in the Houston 13 and Salinas datasets, the model with only $\mathcal{L_W}$ achieves better accuracy than the one with only adaptive learning.

\begin{sidewaystable*}[htbp]
  \centering{\scriptsize
  \caption{\label{tab:sal_perf_att}Accuracy analysis on the Salinas dataset (in \%)}
    \begin{tabular}{|p{0.5cm}|p{3.5cm}|p{0.9cm}|p{0.9cm}|p{1.3cm}|p{1.2cm}|p{1.6cm}|p{1.0cm}|p{0.9cm}|p{0.9cm}|}
 \hline
 S.No.&Classes & 1D CNN & 2D CNN & Spec Atten Net & SS Cas RNN  & Two Branch CNN & FusAt Net & 3D CNN & HybAt Net\\
 \hline
    1& Broccoli green weeds 1 & 99.42 & 81.98 & 98.59 & \textbf{100.00} & \textbf{100.00} & 95.65 & \textbf{100.00} & 99.63 \\
    2& Broccoli green weeds 2 & 99.67 & \textbf{100.00} & \textbf{100.00} & 94.93 & \textbf{100.00} & 99.97 & 99.72 & 99.94 \\
    3& Fallow & 98.93 & \textbf{100.00} & 88.33 & 84.75 & 99.79 & 99.52 & 95.04 & 99.79 \\
    4& Fallow rough plow  & 99.54 & 99.85 & \textbf{100.00} & 99.38 & \textbf{100.00} & 99.61 & 96.06 & 99.61 \\
    5& Fallow smooth & 98.56 & \textbf{100.00} & 99.57 & 95.85 & 99.69 & 97.94 & \textbf{100.00} & 99.30 \\
    6& Stubble & 99.74 & 98.73 & 99.95 & 98.86 & \textbf{99.97} & 98.68 & 99.87 & 99.95 \\
    7& Celery & 99.43 & \textbf{100.00} & \textbf{100.00} & 99.54 & 99.51 & 99.63 & 99.17 & 99.20 \\
    8& Grapes untrained & 80.85 & \textbf{90.96} & 69.98 & 87.34 & 88.35 & 89.56 & 87.96 & 90.88 \\
    9&Soil vineyard develop & 99.84 & \textbf{100.00} & 98.89 & 99.02 & 99.97 & 99.34 & 99.33 & 99.92 \\
    10& Corn senesced green weeds   & 94.52 & 92.70  & 97.07 & 96.85 & \textbf{97.39} & 93.77 & \textbf{97.39} & 96.95 \\
    11& Lettuce romaine, 4wk & 99.28 & 32.54 & \textbf{100.00} & 99.28 & 99.79 & 99.79 & 98.97 & 99.79 \\
    12& Lettuce romaine, 5wk & 99.89 & 99.95 & 88.34 & \textbf{100.00} & \textbf{100.00} & 99.18 & 99.73 & \textbf{100.00} \\
    13& Lettuce romaine, 6wk & 97.79 & 98.90  & \textbf{100.00} & 98.16 & \textbf{100.00} & 99.88 & 99.51 & \textbf{100.00} \\
    14& Lettuce romaine, 7wk & 97.63 & 99.18 & \textbf{100.00} & 98.35 & 98.97 & 98.25 & 98.87 & 98.76 \\
    15& Vineyard untrained & 66.82 & 83.76 & 94.71 & 81.42 & 88.50  & 85.30  & 73.73 & \textbf{96.14} \\
    16& Vineyard vertical treils & 98.59 & 99.59 & \textbf{100.00} & 98.89 & 99.06 & 98.71 & 99.88 & 99.12 \\
    \hline
    &Overall Accuracy (OA) & 90.69 & 93.38 & 91.69 & 93.13 & 95.68 & 94.80  & 93.20 & \textbf{97.16} \\
    &Average Accuracy (AA) & 95.66 & 92.38 & 95.96 & 95.79 & 98.19 & 97.17 & 96.58 & \textbf{98.69} \\
    &Cohen's Kappa$\kappa$ (\%) & 0.8961 & 0.9260 & 0.9078 & 0.9233 & 0.9518 & 0.9420 & 0.9240 & \textbf{0.9683}\\
    \hline
    \end{tabular}}%
\end{sidewaystable*}%

The results of the ablation study on the contribution of attention branches are presented in Table \ref{tab:ablation_att}. It is evident from the table that the model's performance decreases in the absence of attention branches for all three datasets (87.55\% for Houston 13, 83.29\% for Houston 18, and 94.03\% for Salinas), as expected. Additionally, in Figure \ref{fig:spectra}, we visualize the effect of attention branches on the spectral reflectance for the ``Tree'' class in the Houston 13 dataset. Figure 5(a) displays the original reflectance for all wavelengths, while Figure 5(b) highlights the reflectance peaks that contribute more to the classification, while suppressing those that have less impact.

\begin{table}[!h]
\centering{\scriptsize
 \caption{\label{tab:ablation}Ablation study for Wasserstein loss and weighted summation (accuracy in \%)}
\begin{tabular}{ |p{1.5cm}|p{1.7cm}|p{1.7cm}|p{1.7cm}|p{1.7cm}|}
 \hline
Dataset & No $\mathcal{L_W}$ and WtAdd & Only $\mathcal{L_W} $ & Only WtAdd & Both $\mathcal{L_W}$ and WtAdd\\
 \hline
Houston 13 & 86.45 & 86.82 & 86.66 & \textbf{87.99} \\
Houston 18 & 85.21 & 85.50 & 85.86 & \textbf{86.24} \\
Salinas &  96.41 &  96.81 & 96.65 & \textbf{97.16}\\
\hline
\end{tabular}}
\end{table}

\begin{table}[!h]
\centering{\scriptsize
 \caption{\label{tab:ablation_att} Ablation study with and without attention module (accuracy in \%)}
\begin{tabular}{ |p{2.0cm}|p{1.5cm}|p{1.5cm}|p{1.5cm}|}
 \hline
Dataset & Houston 13 & Houston 18 & Salinas\\
 \hline
No attention & 87.55 & 83.29 & 94.03 \\
With attention & \textbf{87.99} & \textbf{86.24} & \textbf{97.16} \\
\hline
\end{tabular}}
\end{table}

\begin{table}[!h]
\centering{\scriptsize
 \caption{\label{tab:ablation_aug} Ablation study for data augmentation (accuracy in \%)}
\begin{tabular}{ |p{3.2cm}|p{1.5cm}|p{1.5cm}|p{1.5cm}|}
 \hline
& Houston 13 &Houston 18 & Salinas\\
 \hline
No augmentation & 82.82 & 82.49 & 95.11 \\
Partial augmentation & 86.71 & 84.49 & 96.26 \\
Full augmentation & \textbf{87.99} &  \textbf{86.24} &  \textbf{97.16} \\
\hline
\end{tabular}}
\end{table}

The performance of the model with different levels of data augmentation is presented in Table \ref{tab:ablation_aug}. We considered three cases: no data augmentation, partial augmentation with samples rotated by 180 degrees, and full data augmentation. As expected, as the level of augmentation decreases, the model's performance starts to decline for all three datasets. This indicates that, similar to many other deep learning models, the performance of our model is constrained by the amount of available data. Furthermore, using a large amount of data in conjunction with training the attention branches results in longer training times for the model.

\section{Self-looping convolutions for hyperspectral image classification}
\label{sec:FBHS}
In this section, we propose an approach that incorporates feedback connections throughout all convolution layers using shared parameters. These shared connections create a self-updating loop within the network, where each layer is updated based on its values from previous steps. Additionally, the shared weights not only facilitate maximum information flow but also greatly reduce the number of parameters within a block of shared layers. By leveraging parameter reduction in each self-looping block, we can introduce multiple blocks that operate with various spatial window sizes. Consequently, our network, HyperLoopNet, enables the learning of distinct features across different spatial extents using different spatial kernels and self-looping blocks. This approach significantly enhances the efficiency of hyperspectral image classification\footnote{Published as: Shivam Pande, Biplab Banerjee, \href{https://www.sciencedirect.com/science/article/pii/S0924271621003191}{HyperLoopNet: Hyperspectral image classification using multiscale self-looping convolutional networks}, ISPRS Journal of Photogrammetry and Remote Sensing 183 (2022) 422–438.}.

\subsection{Methodology}

In this section, we provide a mathematical explanation of HyperLoopNet. To enhance readability, we present a list of the notations in Table \ref{tab:not_hsi}. Let's consider a hyperspectral image $\mathcal{I} \in \mathbb{R}^{H \times W \times B}$, where $H$ and $W$ represent the image's height and width, respectively, and $B$ denotes the number of hyperspectral bands. From this image, we extract a hyperspectral patch of size $s \times s \times B$ centered around a ground truth pixel. Here, $s$ represents the spatial dimension of the patch. Thus, the input to the model can be denoted as $\mathcal{X} = \{\textbf{x}^i\}_{i=1}^n$ for every $\mathcal{Y} = \{y^i\}_{i=1}^n$, where $\textbf{x}^i \in \mathbb{R}^{s \times s \times B}$ and $y^i \in \{1, 2,..., C\}$. In this notation, $i$ is the index of the sample, $n$ is the total number of samples, and $C$ is the number of ground truth classes.

\begin{table}[ht]
\centering{\scriptsize
 \caption{\label{tab:not_hsi} List of notations.}
\begin{tabular}{|l|l|}
 \hline
Notation & Explanation\\
\hline
$\mathcal{I} \in R^{H \times W \times B}$ & HSI of spatial size $H$ and $W$, and spectral dimension $B$\\
$\textbf{x}^i \in \mathbb{R}^{s \times s \times B}$ & $i^{th}$ input HSI patch of spatial dimension $s \times s$\\
$\sigma(.)$  & ReLU activation function\\
$w$  & Weights for the corresponding layers\\
$b$  &Bias for the corresponding layers\\
$y^{i}$ & Groundtruth label for the $i^{th}$ sample\\
$\hat{y}^{i}$ & Softmax probability of the $i^{th}$ sample\\
$-\sum_{i=1}^{n}  y^i\ln{\hat{y}^i}$ & Categorical cross entropy loss\\
$dw$  & Gradient of the weight $w$\\
$||$ & Concatenation operator\\
$\odot$ & Convolution operator\\
$\beta$ & Batch normalization\\
 \hline
\end{tabular}}
\end{table}

Before passing through a self-looping block, the input patch undergoes a convolution layer to reduce the feature dimensions to 32 (selected empirically). This enables efficient parameter sharing without any conflicting dimensions. This operation can be represented by Equation \ref{equation:ini}.

\begin{equation}
\{c^{m}_0\}^i = \sigma(w^{m}_0 \odot \{x_0\}^i + b^{m}_0)
\label{equation:ini}
\end{equation}

In Equation \ref{equation:ini}, $x_0$ denotes the input HSI patch, $\odot$ represents the convolution operator, $w_0$ and $b_0$ represent the convolution weights and biases, respectively, and $\sigma$ represents the ReLU activation function. The superscript $m$ denotes the index of the self-looping block, which can be one of the three blocks with convolution kernel sizes of 3$\times$3, 5$\times$5, and 7$\times$7. $c^{m}_0$ represents the output of the convolution for the $m^{th}$ block.

\begin{figure}[t!]
  \centering
  \centerline{\includegraphics[width=8cm]{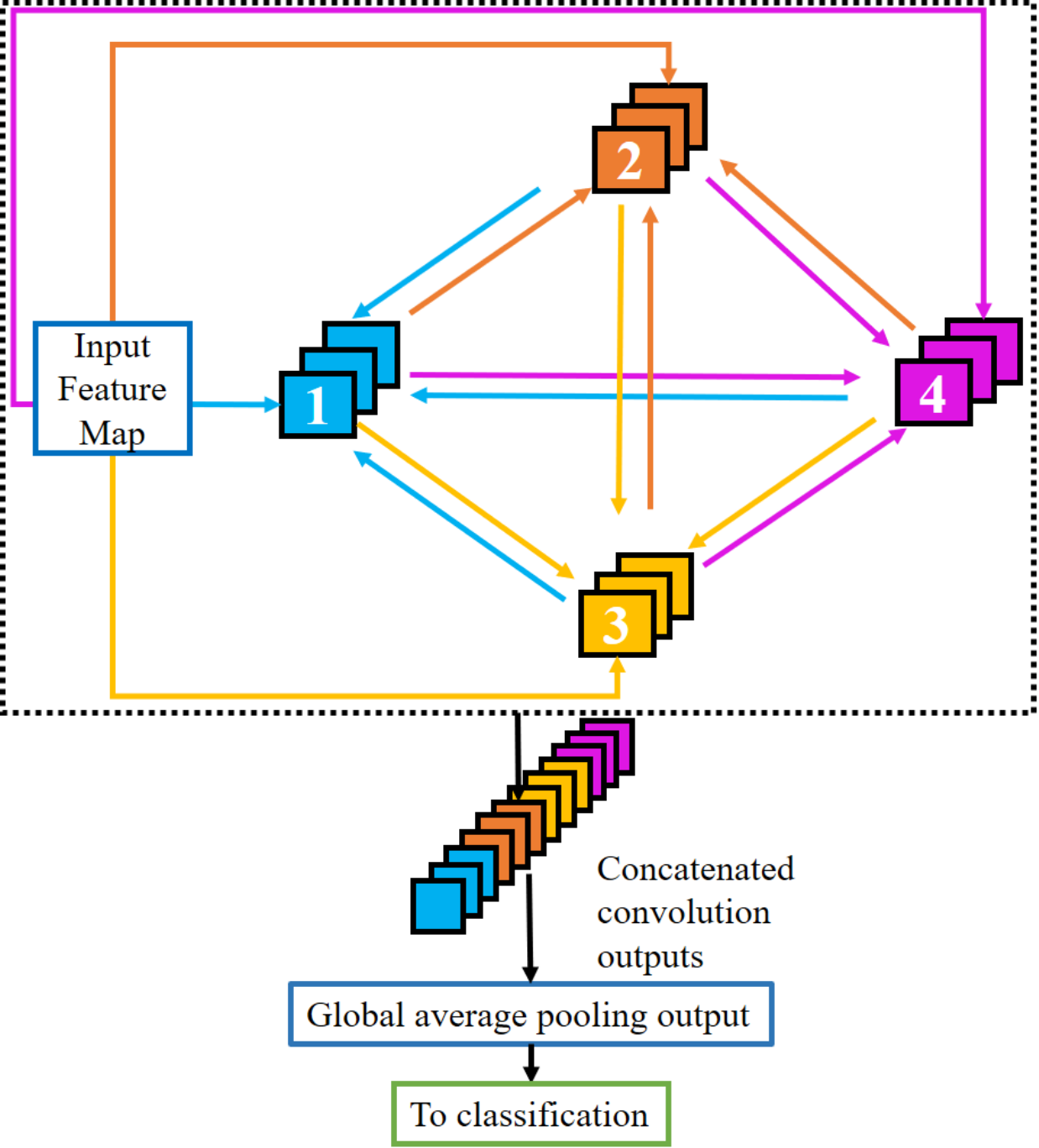}}
  \caption[Schematic of rolled representation of a self-looping block of HyperLoopNet]{The diagram illustrates a self-looping block in HyperLoopNet, where the squares labeled 1, 2, 3, and 4 depict the convolution blocks that follow a structure of convolution-ReLU-batch normalization. Input patches are simultaneously passed to all the convolution blocks. Each block is connected to all the other blocks through both forward and backward connections, creating a multi-loop structure in the network that facilitates optimal information flow. To aid visual comprehension, the arrows transferring information to a specific block are colour-coded based on the corresponding block's colour.}\medskip
  \vspace{-0.5cm}
  \label{fig:Diagram1}
\end{figure}

\begin{figure}[t!]
  \centering
  \centerline{\includegraphics[width=8cm]{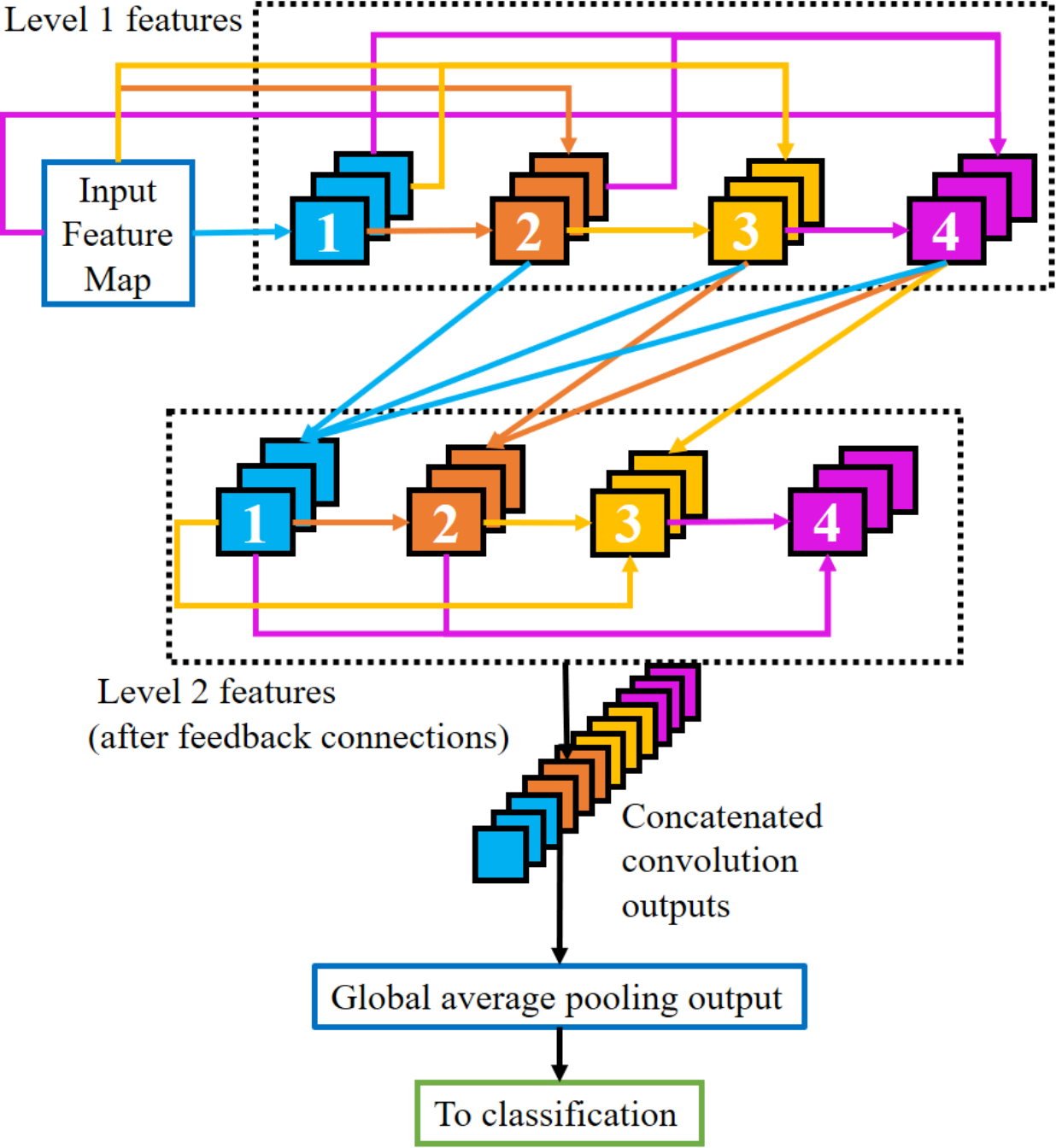}}
  \caption[Schematic of unrolled representation of a self-looping block of HyperLoopNet]{The diagram illustrates an unrolled representation of a self-looping block in HyperLoopNet. The self-looping block is divided into two levels. In the first level, a structure similar to DenseNet is followed, where features are generated through conventional operations. Moving to the second level, feedback connections are introduced, allowing input from all the remaining blocks. The features obtained in the second level are concatenated and utilized for the classification process.}\medskip
  \vspace{-0.5cm}
  \label{fig:Diagram2}
\end{figure}

A self-looping block in HyperLoopNet consists of two levels: \textit{level 1} and \textit{level 2}. The first level resembles a DenseNet-like architecture, where the connections are in the forward direction, and each layer takes the outputs of all previous layers as input. To control the number of parameters and facilitate parameter sharing in \textit{level 2}, the output feature maps are added instead of concatenated. The structure of a self-looping block is illustrated in Figure \ref{fig:Diagram1} and \ref{fig:Diagram2} for the rolled and unrolled versions, respectively. The detailed architecture of this block can be found in Table \ref{tab:archi_hln}. It should be noted that in Figure \ref{fig:Diagram2}, the self-looping block is unrolled only once for simplicity and fewer operations. However, the network can accommodate multiple unrolls without adding additional parameters. Equations \ref{equation:L1_input} and \ref{equation:L1_output} represent the inputs and outputs in \textit{level 1}, respectively.

\begin{table}[htbp]
  \centering{\scriptsize
  \caption[Structure of a self-looping block in HyperLoopNet.]{\label{tab:archi_hln} Structure of a self-looping block in HyperLoopNet. \textcolor{black}{(In the last row, GAP stands for global average pooling).}}
    \begin{tabular}{|c|c|c|c|c|}
     \hline
    Level & Input layers & Layer weights and bias & Output layer & Output size \\
   \hline
    Level I & $c_0$    & $w_1$, $b_1$   & $c_1$    & 11$\times$11$\times$32 \\
          & $c_0+c_1$ & $w_2$, $b_2$    & $c_2$    & for \\
          & $c_0+c_1+c_2$ & $w_3$, $b_3$    & $c_3$    & all the \\
          & $c_0+c_1+c_2+c_3$ & $w_4$, $b_4$    & $c_4$    & layers \\
     \hline
    Level II & $c_2+c_3+c_4$ & $w_1$, $b_1$ & $c_1$  & 11$\times$11$\times$32 \\
          & $c_1+c_3+c_4$ & $w_2$, $b_2$    & $c_2$    & for \\
          & $c_1+c_2+c_4$ & $w_3$, $b_3$    & $c_3$    & all the \\
          & $c_1+c_2+c_3$ & $w_4$, $b_4$    & $c_4$    & layers \\
     \hline
    GAP   & [$c_1,c_2,c_3,c_4$] & - & $g_1$ & 1$\times$128 \\
     \hline
    \end{tabular}}
\end{table}

\begin{figure}[t!]

  \centering
  \centerline{\includegraphics[width=10cm]{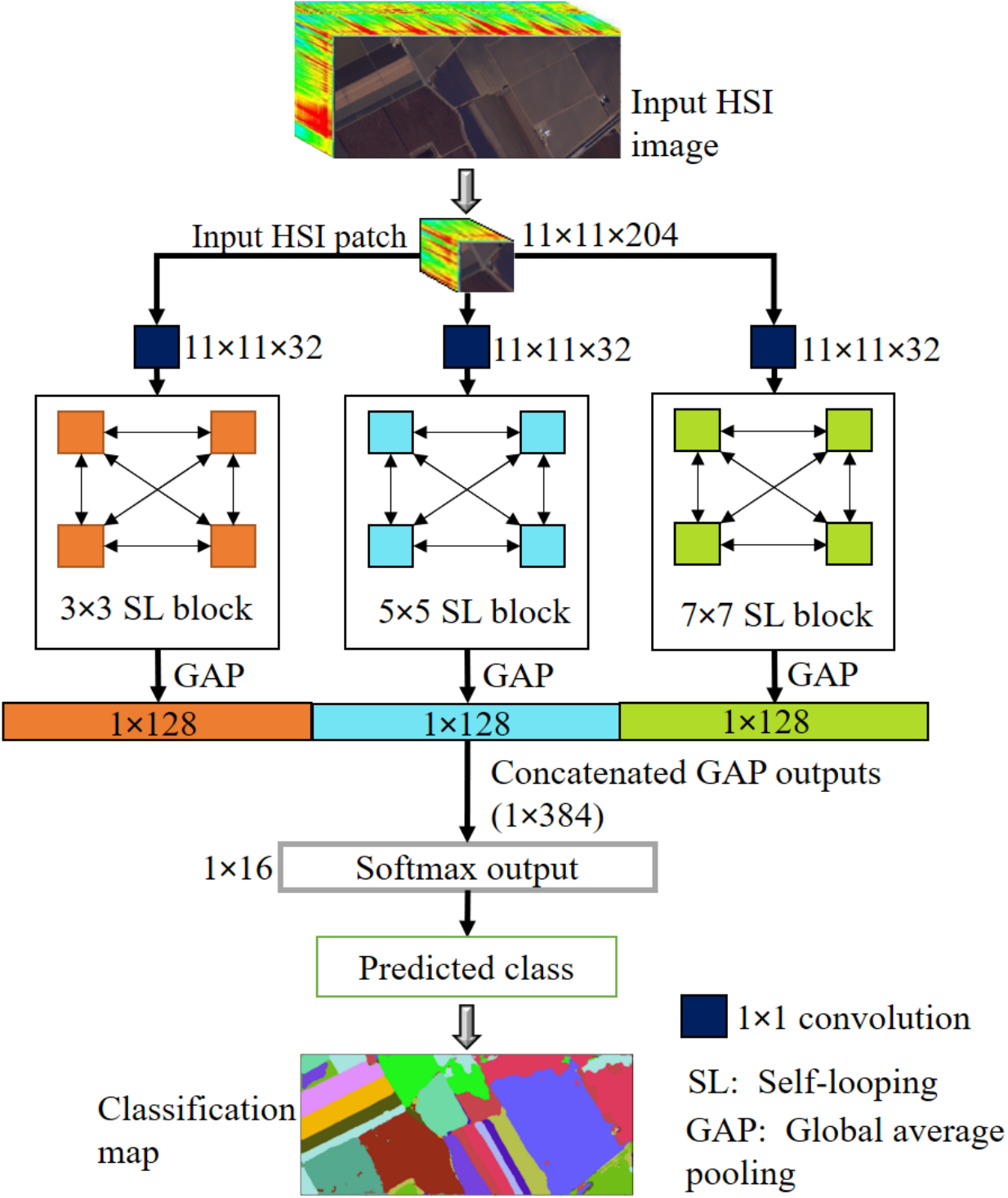}}

  \caption[Schematic of HyperLoopNet on Salinas valley dataset]{The schematic diagram showcases the proposed HyperLoopNet architecture applied to the Salinas Valley Dataset. HyperLoopNet is designed as a multiscale framework comprising three self-looping blocks, operating at distinct spatial scales of 3$\times$3, 5$\times$5, and 7$\times$7 using convolutional operations. Initially, the hyperspectral image (HSI) patch with dimensions 11$\times$11$\times$B (where B represents the number of bands) undergoes a 1$\times$1 convolution to generate initial feature maps sized 11$\times$11$\times$32. These feature maps are then fed into the self-looping blocks corresponding to different spatial scales. The outputs from these blocks are subsequently subjected to global average pooling (GAP) to obtain a one-dimensional representation. The GAP outputs are concatenated and forwarded for the classification process.}\medskip
  \vspace{-0.5cm}
  \label{fig:Diagram_full}
\end{figure}

\begin{equation}
\{x^{m}_k\}^i = \{c^{m}_0\}^i + \sum_{j=1}^{k-1}\{c^{m}_j\}^i
\label{equation:L1_input}
\end{equation}
\begin{table}[ht]
\centering{\scriptsize
 \caption{\label{tab:H13_data_isprs} \textcolor{black}{Houston 2013 hyperspectral dataset with number of training and test samples.}}
\begin{tabular}{|p{0.3cm}|p{2.0cm}|p{1.7cm}|p{1.5cm}|}
 \hline
&Class & Train samples & Test samples \\
\hline
1  &Healthy grass& 198 &1053\\
2  &Stressed grass& 190 &1064\\
3  &Synthetic grass& 192 &505\\
4  &Trees& 188 &1056\\
5  &Soil& 186 &1056\\
6  &Water& 182 &143\\
7  &Residential& 196 &1072\\
8  &Commercial& 191 &1053\\
9  &Road& 193 &1059\\
10 &Highway& 191 &1036\\
11 &Railway& 181 &1054\\
12 &Parking lot 1& 192 &1041\\
13 &Parking lot 2& 184 &285\\
14 &Tennis court& 181 &247\\
15 &Running track& 187 &473 \\
\hline
&Total& 2832 & 12197\\
&Percentage& 18.84 & 81.16\\
 \hline
\end{tabular}}
\end{table}
\begin{equation}
\{c^{m}_k\}^i = \beta\sigma(\{w^{m}_k\}^I \odot \{x_k\}^i + \{b^{m}_k\}^I)
\label{equation:L1_output}
\end{equation}

\begin{figure}[ht]
\begin{center}
\includegraphics[width=1.0\textwidth]{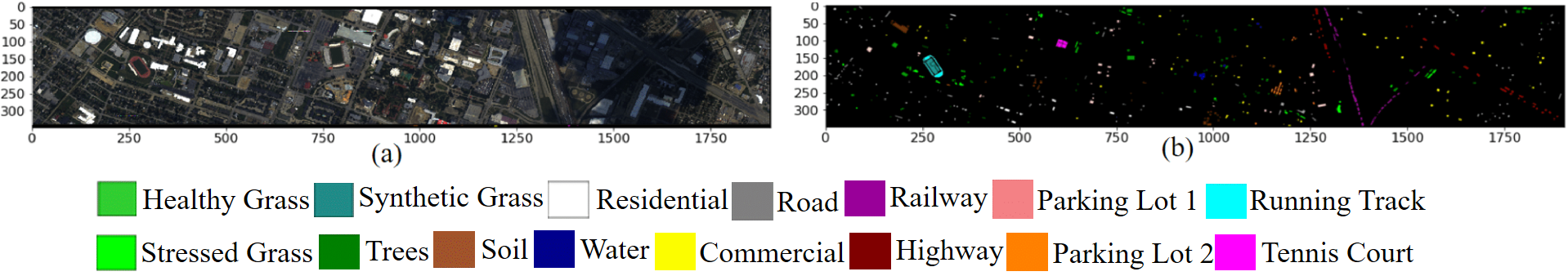}
\end{center}
\caption{Houston HSI dataset (DFC-2013): (a) RGB Colour composite. (b) Groundtruth with classes.}
\label{fig:h13_isprs_ds}
\end{figure}

\begin{figure}[ht]
\begin{center}
\includegraphics[width=1.0\textwidth]{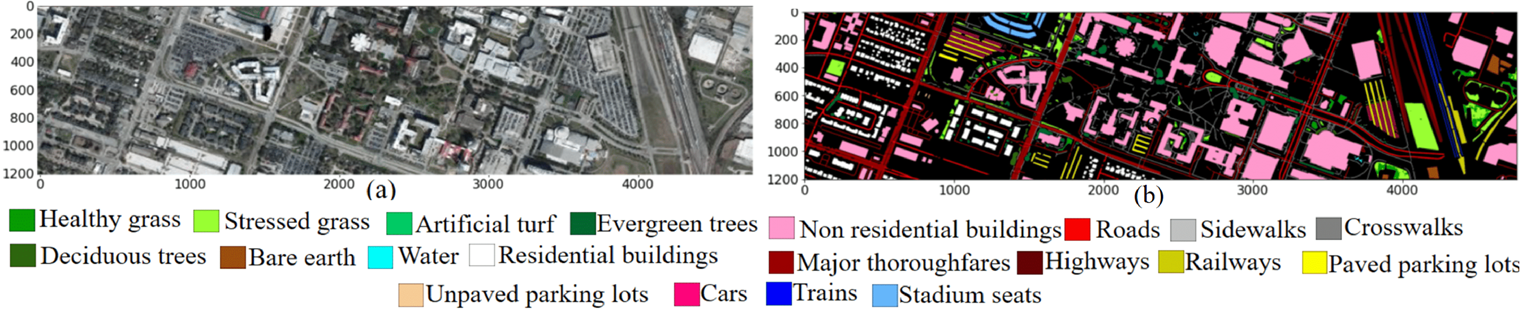}
\end{center}
\caption{Houston HSI dataset (DFC-2018): (a) Colour composite of three bands from red, green and blue wavelengths. (b) Groundtruth with classes.}
\label{fig:h18_isprs_ds}
\end{figure}

\begin{figure}[ht]
\begin{center}
\includegraphics[width=1.0\textwidth]{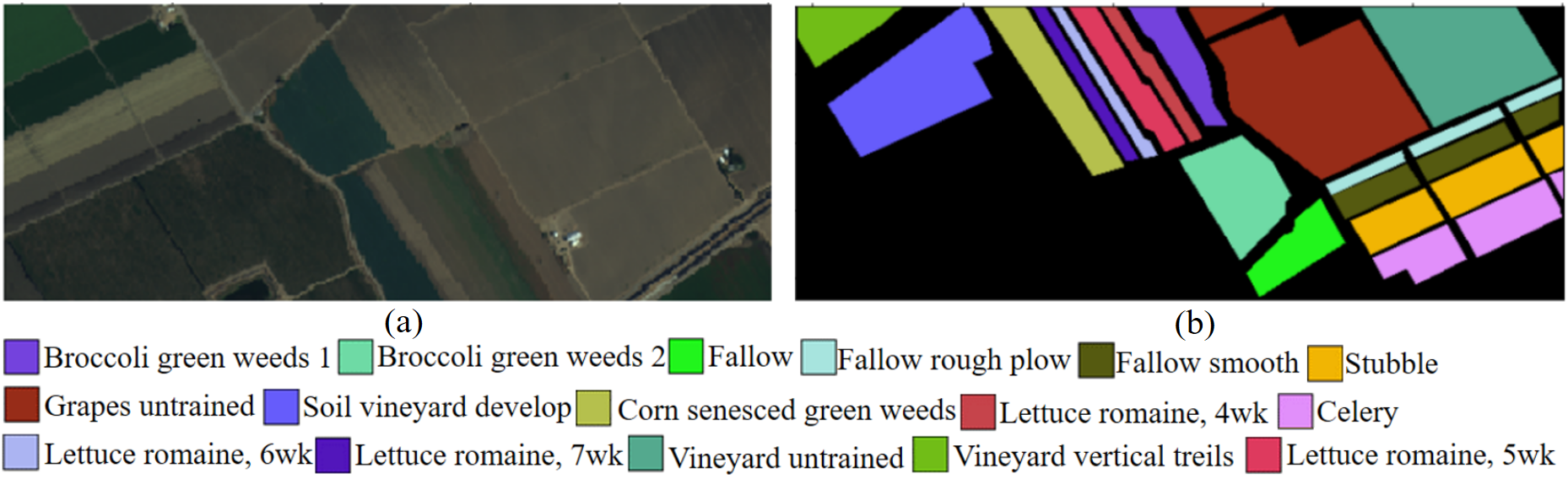}
\end{center}
\caption{Salinas Valley HSI dataset: (a) Colour composite of three bands from red, green and blue wavelengths. (b) Groundtruth with classes.}
\label{fig:sal_isprs_ds}
\end{figure}

\begin{figure}[t!]
  \centering
  \centerline{\includegraphics[width=8.5cm]{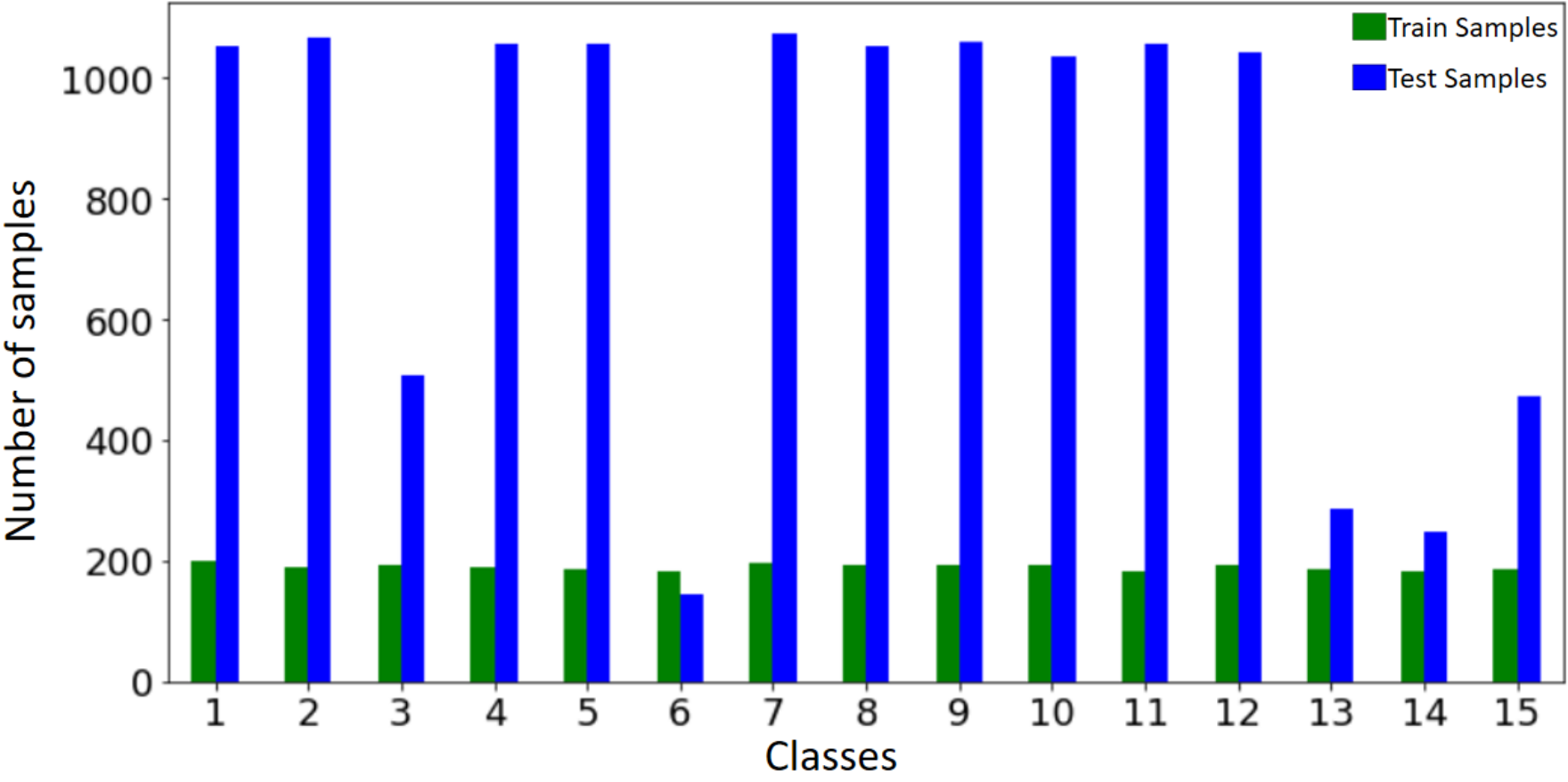}}
  \caption{\textcolor{black}{Bar chart representing train and test samples for Houston 2013 dataset.}}\medskip
  \vspace{-0.5cm}
  \label{fig:H13bar_isprs}
\end{figure}

Here, $c^{m}_0$ is the output from 1$\times$1 convolution (see Eqn. \ref{equation:ini}), $k$ is the $k^{th}$ layer in \textit{level 1}, $c_k$ is the output of the $k^{th}$ layer, while $w_k$ and $b_k$ respectively represents convolution weights and bias for the $k^{th}$ layer. $\beta$ represents batch normalization layer. The superscript $I$ stands for \textit{level 1} to denote the level in which the weights and bias are used. $\sigma$ is ReLU activation as before. 
\begin{table}[ht]
\centering{\scriptsize
 \caption{\label{tab:H18_data_isprs}\textcolor{black}{Houston 2018 hyperspectral dataset with number of training and test samples.}}
\begin{tabular}{|p{0.3cm}|p{3.4cm}| p{2.0cm} |p{2.0cm}|}
 \hline
&Class & Train samples & Test samples \\
\hline
1  &Healthy grass& 100 &39096\\
2  &Stressed grass& 100 &129908\\
3  &Artificial turf& 100 &2636\\
4  &Evergreen trees& 100 &54222\\
5  &Deciduous trees& 100 &20072\\
6  &Bare earth& 100 &17964\\
7  &Water& 100 &964\\
8  &Residential buildings& 100 &158895\\
9  &Non-residential buildings& 100 &894669\\
10 &Roads& 100 &183183\\
11 &Sidewalks& 100 &135935\\
12 &Cross walks& 100 &5959\\
13 &Major thoroughfares& 100 &185338\\
14 &Highways& 100 &39338\\
15 &Railways& 100 &27648 \\
16 &Paved parking lots& 100 &45832\\
17 &Unpaved parking lots& 100 &487\\
18 &Cars& 100 &26189\\
19 &Trains& 100 &21739\\
20 &Stadium seats& 100 &27196\\
\hline
&Total& 2000 & 2016910\\
&Percentage& 0.099 & 99.901\\
 \hline
\end{tabular}}
\end{table}
Moving onto the second level, the features from all the layers are circulated with the help of feedback/shared connections. Thus, now each layer sequentially gets the output features from all the other layers as its input. Eqn. \ref{equation:L2_input} and \ref{equation:L2_output} represent the formulation of the input and output  features respectively at \textit{level 2}. Here, $K$ stands for the total number of layers in a self-looping block and $II$ stands for \textit{level 2}. It goes without saying that $k\leq K$.

\begin{figure}[t!]
  \centering
  \centerline{\includegraphics[width=8.5cm]{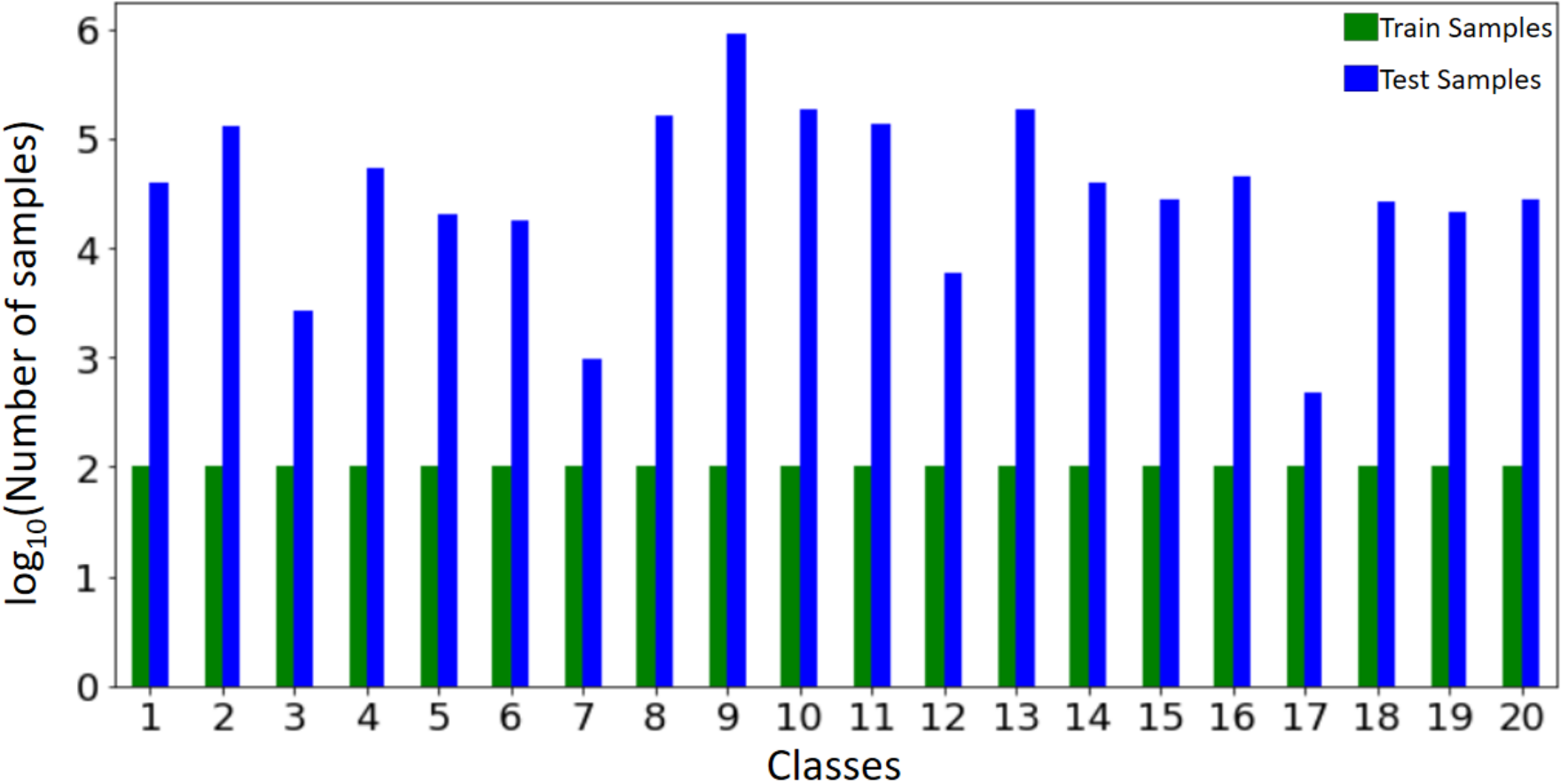}}
  \caption{\textcolor{black}{Logarithmic bar chart representing train and test samples for Houston 2018 dataset.}}\medskip
  \vspace{-0.5cm}
  \label{fig:H18bar_isprs}
\end{figure}

\begin{table}[ht]
\centering{\scriptsize
 \caption{\label{tab:Sal_data}\textcolor{black}{Salinas hyperspectral dataset with number of training and test samples.}}
\begin{tabular}{|p{0.5cm}|p{3.4cm}| p{2.0cm} |p{2.0cm}|}
 \hline
&Class & Train samples & Test samples \\
\hline
1  &Broccoli green weeds 1& 100 &1909\\
2  &Broccoli green weeds 2& 100 &3626\\
3  &Fallow& 100 &1876\\
4  &Fallow rough plow& 100 &1294\\
5  &Fallow smooth& 100 &2578\\
6  &Stubble& 100 &3859\\
7  &Celery& 100 &3479\\
8  &Grapes untrained& 100 &11171\\
9  &Soil vineyard develop& 100 &6103\\
10 &Corn senesced green weeds& 100 &3178\\
11 &Lettuce romaine, 4wk& 100 &968\\
12 &Lettuce romaine, 5wk& 100 &1827\\
13 &Lettuce romaine, 6wk& 100 &816\\
14 &Lettuce romaine, 7wk& 100 &970\\
15 &Vineyard untrained & 100 &7168\\
16 &Vineyard vertical treils & 100 &1707\\
\hline
&Total& 1600 & 52529\\
&Percentage& 2.96 & 97.04\\
 \hline
\end{tabular}}
\end{table}

\begin{figure}[t!]
  \centering
  \centerline{\includegraphics[width=8.5cm]{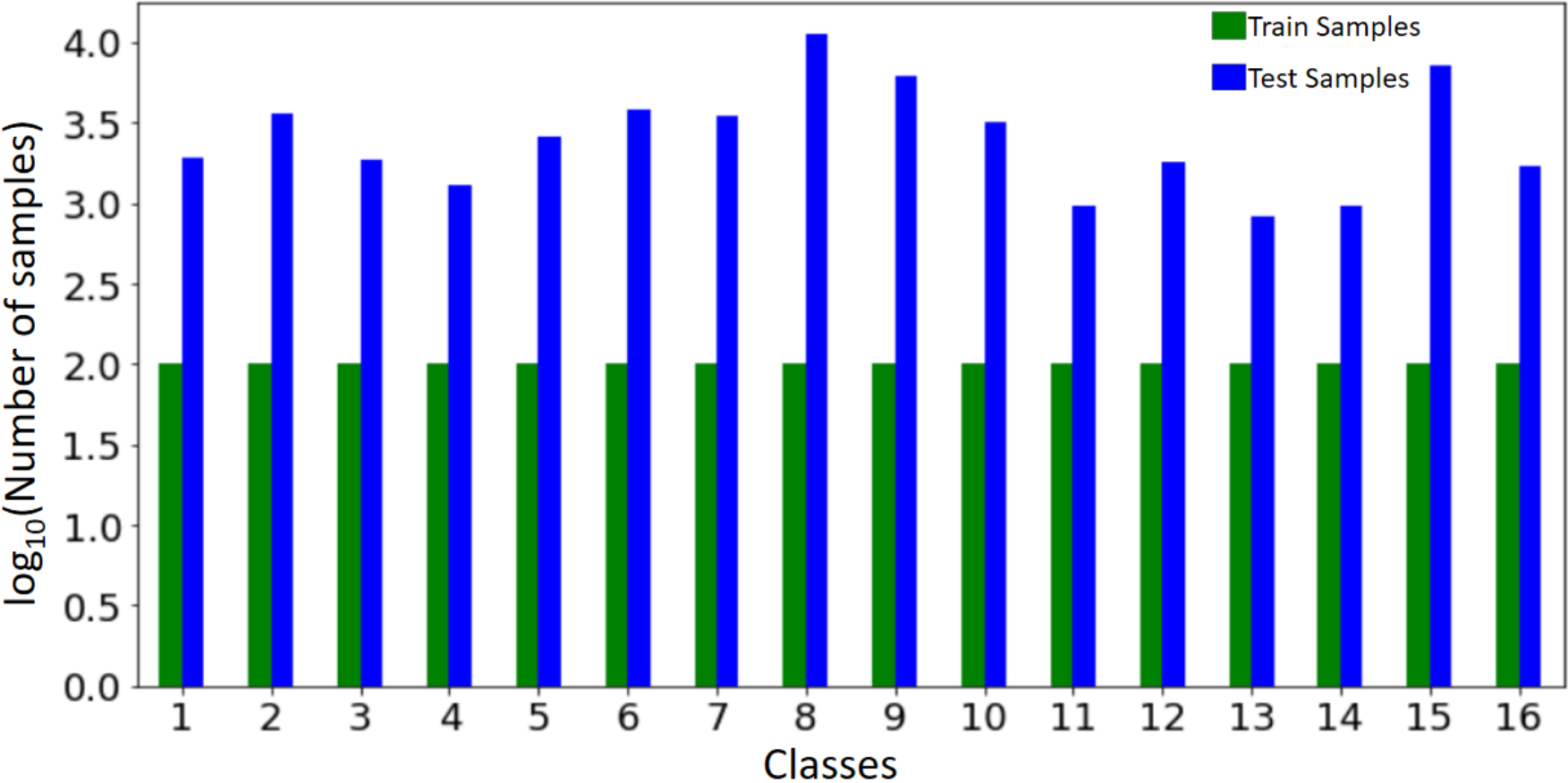}}
  \caption{\textcolor{black}{Logarithmic bar chart representing train and test samples for Salinas Valley dataset.}}\medskip
  \vspace{-0.5cm}
  \label{fig:Salbar}
\end{figure}

Here, $c^{m}_0$ is the output from 1$\times$1 convolution (see Eqn. \ref{equation:ini}), $k$ is the $k^{th}$ layer in \textit{level 1}, $c_k$ is the output of the $k^{th}$ layer, while $w_k$ and $b_k$ respectively represents convolution weights and bias for the $k^{th}$ layer. $\beta$ represents batch normalization layer. The superscript $I$ stands for \textit{level 1} to denote the level in which the weights and bias are used. $\sigma$ is ReLU activation as before. 

Moving onto the second level, the features from all the layers are circulated with the help of feedback/shared connections. Thus, now each layer sequentially gets the output features from all the other layers as its input. Eqn. \ref{equation:L2_input} and \ref{equation:L2_output} represent the formulation of the input and output  features respectively at \textit{level 2}. Here, $K$ stands for the total number of layers in a self-looping block and $II$ stands for \textit{level 2}. It goes without saying that $k\leq K$.
\begin{figure}[t!]
  \centering
  \centerline{\includegraphics[width=8.5cm]{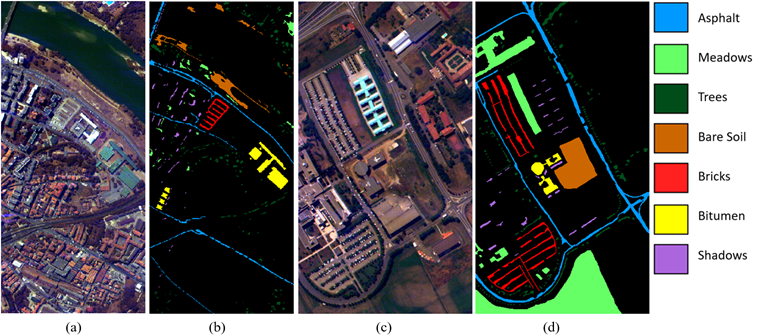}}
  \caption[RGB colour composites and groundtruth for Pavia Center and University datasets]{{Pavia University and Pavia Centre hyperspectral dataset (PaviaUC) with (a) RGB colour composite of the three bands for Pavia University (b) Groundtruth map for Pavia University (c) RGB colour composite of the three bands for Pavia Centre and (d) Groundtruth map for Pavia Centre}}\medskip
  \vspace{-0.5cm}
  \label{fig:dpuc}
\end{figure}

\begin{table}[ht]
\centering{\scriptsize
 \caption{\label{tab:puc_data}\textcolor{black}{Pavia University (PU) and Pavia Centre (PC) hyperspectral dataset with number of training and test samples.}}
\begin{tabular}{|p{0.2cm}|p{1.5cm}| p{2.3cm} |p{2.3cm}|}
 \hline
&Class & Train samples (PU)& Test samples (PC)\\
\hline
1  &Asphalt 1& 400 & 6631\\
2  &Meadows& 400 & 18649\\
3  &Trees& 400 & 3064\\
4  &Baresoil& 400 & 5029\\
5  &Bricks& 400 & 3682\\
6  &Bitumen& 400 & 1330\\
7  &Shadows& 400 & 947\\
\hline
&Total& 2800 & 39332\\
&Percentage& 6.64 & 93.36\\
 \hline
\end{tabular}}
\end{table}

\begin{figure}[t!]
  \centering
  \centerline{\includegraphics[width=8.5cm]{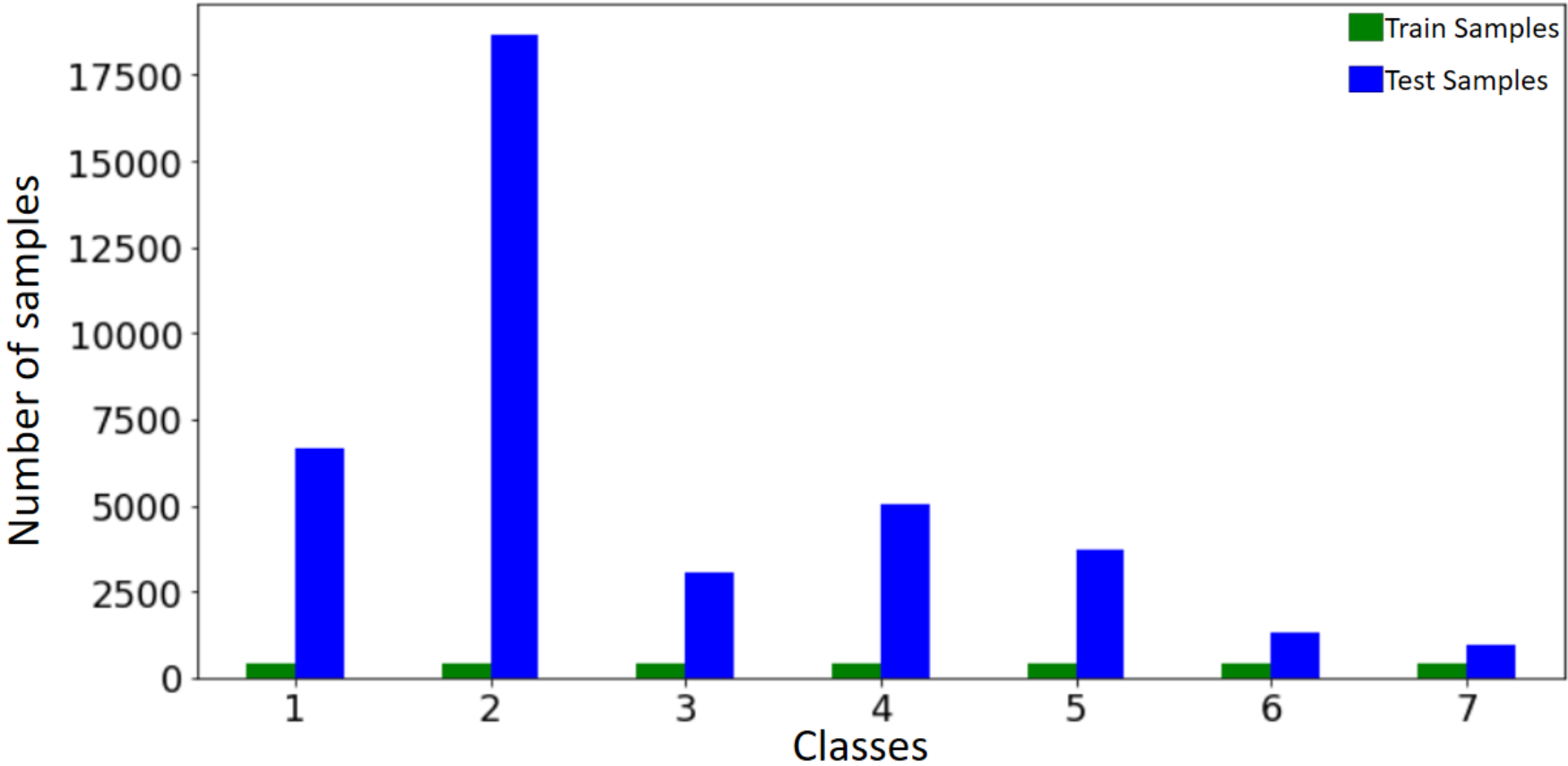}}
  \caption{{Bar chart representing train and test samples for PaviaUC dataset.}}\medskip
  \vspace{-0.5cm}
  \label{fig:ttpuc}
\end{figure}

\begin{equation}
\{x^{m}_k\}^i = \sum_{j=1, j \neq k}^{K}\{c^{m}_j\}^i
\label{equation:L2_input}
\end{equation}

\begin{equation}
\{c^{m}_k\}^i = \beta\sigma(\{w^{m}_k\}^{II} \odot \{x_k\}^i + \{b^{m}_k\}^{II})
\label{equation:L2_output}
\end{equation}

The convolution operations in Eqn. \ref{equation:L1_output} and \ref{equation:L2_output} are essentially same since the weights are being shared i.e. $\{w^{m}_k\}^{I} = \{w^{m}_k\}^{II}$ and $\{b^{m}_k\}^{I} = \{b^{m}_k\}^{II}$. The outputs from all the layers in \textit{level 2} are concatenated and subjected to global average pooling (GAP) for spatial averaging of the features (see Eqn. \ref{equation:GAP}). 

\begin{equation}
\{f^{m}_{gap}\}^i = \Gamma({\mathbin\Vert}^{k}_{j=1} \{c^{m}_j\}^i)
\label{equation:GAP}
\end{equation}

Here, $\mathbin\Vert$ represents concatenation operator, while $\Gamma$ represents global average pooling.

The GAP outputs from the three mutiscale blocks are then concatenated (Eqn. \ref{equation:F_GAP}) and sent to the softmax layer for classification (Eqn. \ref{equation:SM}).
\begin{sidewaystable*}[htbp]
  \centering{\scriptsize
  \caption{\label{tab:hous13_perf}Accuracy analysis on the Houston 13 dataset (in \%).}
    \begin{tabular}{|p{2.5cm}|p{0.9cm}|p{0.9cm}|p{1.3cm}|p{0.9cm} |p{0.9cm}| p{0.9cm}|p{1.7cm}|p{0.9cm}|p{0.9cm}|p{0.9cm}|p{1.0cm}|p{1.3cm}|}
 \hline
 Class name & SS Cas RNN & 2D CNN & Spec Atten Net & 3D CNN & Fus At Net & Hy- brid SN & Two Branch CNN & \textcolor{black}{2D Res Net} & \textcolor{black}{SAR- FN} & \textcolor{black}{LWt CNN} & \textcolor{black}{Octave CNN} & Hyper Loop Net\\
 \hline
    Healthy grass & 81.92 & 80.87 & 82.09 & \textbf{82.72} & 81.84 & 82.36 & 82.11 & 78.52 & 79.54 & 80.27 & 82.41 & 82.05 \\
    Stressed grass & 81.32 & \textbf{85.15} & 84.25 & 85.06 & 84.34 & 83.29 & 83.50 & 84.81 & 85.02 & 82.99 & 85.11 & 85.13 \\
    Synthetic grass & 94.53 & 99.05 & 98.06 & 97.39 & 95.92 & 97.86 & 95.29 & 99.25 & \textbf{99.72} & 98.97 & 98.34 & 98.30 \\
    Trees & 92.44 & 92.27 & 92.54 & 98.37 & 92.69 & 88.20 & \textbf{99.51} & 93.09 & 91.21 & 89.92 & 93.28 & 92.86 \\
    Soil  & 98.54 & 98.37 & 97.41 & 98.18 & 98.96 & \textbf{99.66} & 99.20 & 96.52 & 98.84 & 98.69 & 99.39 & 99.45 \\
    Water & 95.94 & 97.76 & 98.74 & 89.23 & 97.76 & 92.31 & 99.16 & \textbf{99.44} & 90.77 & 95.66 & 97.34 & 98.88 \\
    Residential & 83.68 & 86.47 & 87.85 & 86.34 & 88.34 & 85.62 & 91.44 & 88.34 & 84.53 & 82.93 & 88.08 & \textbf{92.61} \\
    Commercial  & 68.74 & 70.33 & 70.10 & 69.86 & 77.45 & 70.37 & 68.91 & 72.69 & 72.02 & 78.02 & 79.64 & \textbf{85.94} \\
    Road  & 77.24 & 79.41 & 80.25 & \textbf{87.55} & 82.97 & 81.98 & 79.06 & 83.89 & 84.23 & 79.85 & 81.45 & 84.21 \\
    Highway & 58.38 & 59.59 & 59.69 & 61.95 & 60.04 & 56.54 & 59.21 & 60.42 & \textbf{65.75} & 56.97 & 62.51 & 62.10 \\
    Railway & 78.71 & 83.36 & 78.69 & \textbf{95.18} & 84.76 & 71.80 & 88.27 & 87.99 & 77.76 & 84.91 & 92.16 & 94.27 \\
    Parking lot 1 & 84.94 & 90.22 & 87.53 & 92.76 & 86.82 & 90.53 & 91.89 & 90.01 & 94.49 & \textbf{95.93} & 94.29 & 91.28 \\
    Parking lot 2 & 86.74 & 91.02 & 91.09 & 86.95 & 90.60 & 81.89 & \textbf{91.79} & 88.91 & 86.81 & 86.53 & 85.82 & 84.84 \\
    Tennis court & 90.93 & 96.76 & 84.86 & 94.82 & 94.33 & \textbf{100.00} & 83.40 & 86.64 & 97.65 & 96.11 & 96.84 & 96.19 \\
    Running track & 95.94 & 96.96 & 99.75 & 99.41 & 99.70 & 96.96 & 99.20 & 99.07 & 97.59 & 96.03 & \textbf{99.92} & 99.62 \\
    \hline
    Overall accuracy & 82.33 & 84.54 & 83.90 & 87.09 & 85.52 & 82.93 & 85.71 & 85.28 & 85.05 & 84.73 & 87.29 & \textbf{88.27} \\
    Average accuracy & 84.67 & 87.17 & 86.19 & 88.38 & 87.77 & 85.29 & 87.46 & 87.31 & 87.06 & 86.92 & 89.11 & \textbf{89.85} \\
    Kappa ($\kappa$ \%) & 0.8086 & 0.8328 & 0.8261 & 0.8598 & 0.8430 & 0.8157 & 0.8452 & 0.8408 & 0.8384 & 0.8346 & 0.8623 & \textbf{0.8728} \\
     \hline
    \end{tabular}}
\end{sidewaystable*}

\begin{equation}
f^i_{gap} = {\mathbin\Vert}^{m}_{j=1} \{f^{j}_{gap}\}^i
\label{equation:F_GAP}
\end{equation}

\begin{equation}
\hat{y}^i = softmax(FC(f^{i}_{gap}))
\label{equation:SM}
\end{equation}

Here, $FC$ represents a fully connected layer and $\hat{y}^i$ are the class probabilities. For training the model, categorical cross entropy is used as the loss function. The cross entropy loss is presented for the $i^{th}$ sample in Eqn. \ref{eqn:CELoss_hln}. 

\begin{equation}
\mathcal{L}_{CE} =  -\sum_{i=1}^{n}  y^i\ln{\hat{y}^i}
\label{eqn:CELoss_hln}
\end{equation}

Here, $dw$ denotes the gradient, $p$ represents the layer index, and $q$ represents the specific instance at which the gradient is being calculated. $r$ indicates the number of possible instances obtained through multiple unrolls. In our study, we have set $r=1$ (since increasing the value of $r$ does not affect the number of parameters but can lead to increased computational operations).

The training and testing processes in HyperLoopNet follow the conventional approach, where training patches are fed to the model in batches during training. Once the training is completed, the performance of the model is evaluated using test patches.

\subsection{Datasets and Experiments}

This section provides a description of the datasets used, the training protocols employed, the evaluation metrics, and the methods used for comparison. In general, our approach is evaluated using four hyperspectral datasets, namely Houston 2013, Houston 2018, Salinas Valley, and Pavia University and Pavia Centre. For the first three datasets, both the train and test patches are extracted from the same image. However, to address the potential issue of overfitting due to using the same image and limited sample size, we also include the Pavia dataset. In this dataset, the training and testing patches are generated from two non-overlapping (disjoint) images. This ensures that the models can be evaluated without any bias towards specific datasets. Additionally, for the first three datasets, only a small number of pixels are selected for training and testing purposes (details can be found in section \ref{sec:datasets} below) to ensure the proposed model's robust performance.

\subsubsection{Datasets}
\label{sec:datasets_hln}
\noindent \textbf{Houston 2013}: This dataset was publicly released for the IEEE Data Fusion Contest 2013. It consists of hyperspectral imagery captured over the National Centre for Airborne Laser Mapping (NCALM), containing 144 bands with each band sized at 349 $\times$ 1905. There are a total of 15,029 groundtruth samples available, distributed across 15 land-use/land-cover (LULC) classes. These samples are pre-divided into 2,832 training samples and 12,197 testing samples. The sample distribution is presented in Table \ref{tab:H13_data_isprs} and Figure \ref{fig:H13bar_isprs}, and the spatial extent of the samples can be seen in Figure \ref{fig:h13_isprs_ds} (a). We refer to this dataset as Houston 13 \citep{debes2014hyperspectral}.

\noindent \textbf{Houston 2018}: This dataset is sourced from the IEEE Data Fusion Contest 2018. It consists of 48 hyperspectral bands, with each band sized at 601 $\times$ 2384. The groundtruth samples are collected on a higher-resolution grid of size 1202 $\times$ 4768. There are a total of 20 LULC classes represented by 2,018,910 groundtruth samples. From these samples, 2,000 samples (100 samples per class) are randomly selected for training, while the remaining 2,018,910 samples are reserved for testing. The overlap between the training and testing samples is negligible, as the number of training samples is extremely small ($<$ 0.1\%) compared to the test samples. The class information and spatial distribution are shown in Table \ref{tab:H18_data_isprs} and Figure \ref{fig:h18_isprs_ds} (b), respectively. The visual idee of the number of samples is presented in logarathmic bar graph in Figure \ref{fig:H18bar_isprs} For simplicity, we denote this dataset as Houston 18 \citep{pande2021adaptive}.

\noindent \textbf{Salinas Valley}: The Salinas Valley dataset comprises imagery captured over Salinas Valley, California, using the AVIRIS sensor. It consists of a total of 204 bands, and each band has a spatial dimension of 512 $\times$ 217. The dataset is categorized into 16 classes, with a total of 54,129 groundtruth samples. For training purposes, 1,600 samples (100 per class) are selected, while the remaining 52,529 samples are used for testing. The training samples account for less than 3\% of the total number of samples. Figure \ref{fig:sal_isprs_ds} and Table \ref{tab:Sal_data} provide information about the groundtruth, which is also visualised in Figure \ref{fig:Salbar}. We refer to this dataset as the SV dataset \citep{pande2021adaptive}.

\noindent\textbf{Pavia University and Pavia Centre Dataset}: The Pavia University and Pavia Centre datasets consist of hyperspectral images captured over the Pavia city using the ROSIS sensor. These datasets comprise two separate images: Pavia University and Pavia Centre. The former has a spatial size of 610$\times$340$\times$102, while the latter has a size of 1096$\times$492$\times$102 (where 102 represents the number of hyperspectral bands). Both images share the same land-use/land-cover (LULC) classes \citep{qin2019tensor}. For training the models, the Pavia University image is chosen, and 400 samples are randomly selected from each class. For testing, the Pavia Centre image is selected, which contains a total of 39,332 groundtruth samples (see Figure \ref{fig:dpuc} for the images and Table \ref{tab:puc_data} for groundtruth information. Figure \ref{fig:ttpuc} shows the visual distribution of train and test sample for this dataset. This dataset differs from the previous three datasets in that it includes images from different and non-overlapping domains to evaluate the robustness of the models. We refer to this dataset as PaviaUC.

To enhance the training sets for spatial methods, augmentation is applied through clockwise rotations of 90, 180, and 270 degrees. However, for the Houston 13 dataset, only the original samples and those rotated by 180 degrees are considered due to the dataset containing approximately 200 samples per class. On the other hand, for the Houston 18 and Salinas Valley datasets, all rotation angles are considered as they consist of only 100 samples per class. For the PaviaUC dataset, augmentation is not performed since the number of training samples is sufficient.

\subsubsection{Training protocols}

All models were trained using hyperspectral patches with a fixed size of 11 $\times$ 11 $\times$ $B$. The training process consisted of 500 epochs. We employed the Adam optimizer with Nesterov momentum \citep{dozat2016incorporating}, setting the initial learning rate to 0.00002. \textcolor{black}{The batch size for all models was fixed at 64}. The training was conducted on the TensorFlow 2.0 framework using Google Colaboratory. To evaluate the performance of our model, we compared it against several state-of-the-art algorithms from the machine learning and deep learning domains, which are discussed below:

\textbf{Spectral Spatial Cascaded Recurrent Neural Network (SSCasRNN)}: Introduced by \citep{hang2019cascaded}, this method incorporates convolution layers followed by a sequence of recurrent layers. The convolution layers initially extract sequential patterns from the images, resulting in a multi-dimensional vector. This vector is then divided into sub-vectors and fed into individual gated recurrent units (GRU) modules. The outputs of these modules are concatenated and passed through a single GRU layer before classification.

\textbf{2D CNN}: This model follows a VGG Net-like architecture, as presented in \citep{mou2019learning}. It consists of three convolution blocks, each comprising two convolution layers and a max-pooling layer. The kernel size is fixed at 3 $\times$ 3, while the number of channels varies as 32, 64, and 128 for each block. The last block is followed by a global average pooling (GAP) layer, and the output is then processed for softmax classification.

\textbf{SpecAttenNet}: Similar to the 2D CNN, this model has a comparable structure but reinforces the input features using a spectral attention mask. The spectral attention mask is obtained by globally convolving the input patch with a sigmoid activation function. The resulting mask is multiplied element-wise with the original features to highlight the bands that contribute more to the classification \citep{mou2019learning}.

\textbf{Residual CNNs}: This model follows a structure similar to that of a 2D CNN, comprising three convolution blocks, each consisting of two convolution layers. In each convolution block, the output of the first layer is connected to the final output through a concatenation-based skip connection. The output from the final block is passed through a global average pooling (GAP) layer, and the resulting output is used for softmax classification \citep{he2016deep}.

\textbf{3D CNN}: Inspired by the model proposed by \citep{chen2016deep}, this model consists of four 3D convolution layers and two 3D pooling layers after the first two convolutions. The model jointly extracts spectral-spatial information from hyperspectral image patches and utilizes the extracted features for classification.

\textbf{FusAtNet}: This method is derived from \citep{mohla2020fusatnet}, which was primarily designed for fusing HSI and LiDAR data. In our study, we have excluded the LiDAR component and utilized the method solely for HSI classification. A key feature of this method is the use of attention mechanisms at different levels to capture spatial and spectral features from the HSI data, which are then jointly utilized for classification.

\begin{sidewaystable*}[htbp]
  \centering{\scriptsize
  \caption{\label{tab:hous18_perf}Accuracy analysis on the Houston 18 dataset (in \%).}
    \begin{tabular}{|p{2.9cm}|p{0.9cm}|p{0.9cm}|p{1.3cm}|p{0.9cm} |p{0.9cm}| p{0.9cm}|p{1.7cm}|p{0.9cm}|p{0.9cm}|p{0.9cm}|p{1.0cm}|p{1.3cm}|}
 \hline
 Class name & SS Cas RNN & 2D CNN & Spec Atten Net & 3D CNN & Fus At Net & Hy- brid SN & Two Branch CNN & \textcolor{black}{2D Res Net} & \textcolor{black}{SAR- FN} & \textcolor{black}{LWt CNN} & \textcolor{black}{Octave CNN} & Hyper Loop Net\\
 \hline
    Healthy grass & 92.50 & 94.21 & 93.95 & 91.63 & 90.90 & 94.65 & 94.70 & 94.82 & 93.46 & 90.49 & \textbf{96.34} & 94.02 \\
    Stressed grass & 87.70 & 88.91 & 86.05 & 88.93 & 87.69 & \textbf{89.45} & 85.55 & 86.80 & 83.47 & 84.61 & 85.54 & 87.34 \\
    Artificial turf & 99.57 & 99.66 & 98.59 & 99.69 & 99.03 & 99.86 & \textbf{99.98} & 99.60 & 99.92 & 99.63 & 99.80 & 99.81 \\
    Evergreen trees & 95.83 & 93.82 & 94.89 & 94.98 & 96.16 & 96.67 & 96.32 & 96.63 & 96.94 & 96.27 & 96.89 & \textbf{97.43} \\
    Deciduous trees & 89.96 & 85.32 & 89.84 & \textbf{96.56} & 90.67 & 90.47 & 93.19 & 91.19 & 92.79 & 89.26 & 94.25 & 93.67 \\
    Bare earth  & 97.86 & 99.10 & 94.79 & 99.37 & 98.59 & 99.21 & 99.63 & 98.42 & 98.59 & \textbf{99.86} & 99.67 & 99.83 \\
    Water & 99.71 & 99.81 & 99.90 & 99.71 & 99.61 & \textbf{100.00} & \textbf{100.00} & 98.67 & 99.90 & \textbf{100.00} & 99.90 & \textbf{100.00} \\
    Residential buildings & 75.41 & 78.09 & 76.58 & 81.05 & 78.40 & 76.16 & 79.43 & 80.52 & 78.90 & 80.97 & 81.38 & \textbf{82.61} \\
    Non-residential buildings & 65.17 & 69.82 & 64.49 & 74.01 & 70.72 & 73.42 & 76.17 & 72.96 & 76.68 & 75.96 & 73.67 & \textbf{77.92} \\
    Roads & 45.07 & 36.86 & 43.37 & 62.24 & 44.43 & 45.68 & 46.93 & 47.15 & 45.82 & 48.39 & 48.83 & \textbf{50.86} \\
    Sidewalks & 48.87 & 52.47 & 48.62 & 46.08 & 47.00 & 51.76 & 51.82 & \textbf{56.60} & 55.10 & 51.61 & 51.03 & 55.83 \\
    Cross walks & 59.79 & 70.72 & 57.72 & 64.34 & 63.50 & 67.82 & 67.64 & 66.36 & 72.01 & 71.28 & 73.67 & \textbf{75.91} \\
    Major thoroughfares & 46.89 & 55.58 & 50.49 & 49.98 & 64.89 & 53.23 & 55.87 & 55.22 & 51.45 & 61.64 & 63.49 & \textbf{65.06} \\
    Highways & 85.87 & 85.65 & 90.40 & \textbf{94.66} & 86.86 & 89.91 & 92.38 & 94.42 & 90.56 & 91.24 & 94.03 & 93.04 \\
    Railways & 97.81 & 96.64 & 95.94 & 98.86 & 97.17 & 99.15 & 98.86 & 98.66 & 98.35 & 98.55 & 99.24 & \textbf{99.43} \\
    Paved parking lots & 91.01 & 88.84 & 87.58 & 88.32 & 92.20 & 90.35 & \textbf{92.27} & 89.42 & 89.85 & 87.50 & 91.83 & 92.06 \\
    Unpaved parking lots & 98.23 & 99.88 & 97.58 & \textbf{100.00} & 99.96 & 100.00 & \textbf{100.00} & 99.79 & 99.71 & 99.96 & \textbf{100.00} & \textbf{100.00} \\
    Cars  & 85.66 & 93.82 & 92.42 & 94.06 & 87.51 & 90.51 & 94.22 & 92.88 & 91.07 & 91.89 & \textbf{95.34} & 94.65 \\
    Trains & 92.44 & 96.85 & 95.20 & 96.19 & 94.65 & 95.49 & 96.57 & 97.81 & 97.16 & 98.74 & 98.66 & \textbf{99.03} \\
    Stadium seats & 97.88 & 97.86 & 98.31 & 98.88 & 98.85 & 97.90 & 99.46 & 99.29 & 99.30 & 98.45 & 99.27 & \textbf{99.77} \\
    \hline
    Overall accuracy & 67.20 & 69.91 & 67.12 & 73.65 & 71.47 & 72.15 & 73.93 & 72.89 & 73.55 & 74.28 & 73.93 & \textbf{76.64} \\
    Average accuracy & 82.66 & 84.20 & 82.84 & 85.98 & 84.44 & 85.08 & 86.05 & 85.86 & 85.55 & 85.81 & 87.14 & \textbf{87.91} \\
    Kappa ($\kappa$ \%) & 0.6055 & 0.6351 & 0.6059 & 0.6766 & 0.6525 & 0.6591 & 0.6788 & 0.6688 & 0.6741 & 0.6830 & 0.6808 & \textbf{0.7108} \\
     \hline
    \end{tabular}}
\end{sidewaystable*}%

\textbf{HybridSN}: This method combines the use of 3D CNN and 2D CNN for feature extraction and refinement in HSI. The 3D CNN is employed to extract spectral-spatial features, which are subsequently passed to a 2D CNN to further refine the spatial information within the extracted features. The output is then used for classification \citep{roy2019hybridsn}.

\textbf{Two Branch CNN}: This model consists of two parallel layers, one processing the spatial information using 2D CNNs, while the other focuses on the spectral information using 1D CNNs. The outputs from these two layers are concatenated and forwarded for classification \citep{xu2017multisource}.

\textbf{Light Weight CNN}: This network architecture comprises six lightweight blocks organized in pairs. Each lightweight block consists of a convolution layer followed by a separable convolution layer. The outputs from each pair of blocks are combined using an addition-based skip connection, and the result is passed to the next lightweight module. The final residual layer is followed by a global average pooling (GAP) layer and a softmax classification layer \citep{meng2021lightweight}. This method will be referred to as LWt CNN.

\textbf{Feedback Attention CNN}: This approach introduces feedback connections within an attention-based framework. The network is composed of a convolution layer (with Batch Normalization and ReLU activation) followed by two consecutive feedback attention blocks. Each feedback attention block consists of a two-branch network with an attention layer. The output from the block is connected to the input through a feedback connection. The final output from the second feedback attention block is then passed through another convolution layer, and the resulting output is used for softmax classification \citep{li2021recurrent}. This method will be abbreviated as SARFN (spatial attention recurrent feedback network).

\textbf{Octave CNN}: This method draws inspiration from the works of \citep{chen2019drop} and \citep{xu2020csa}. The network comprises six octave convolution layers designed specifically for hyperspectral image classification. Octave convolution is an advancement over traditional convolution, allowing for similar or improved results with fewer parameters. The approach involves dividing the input channels into higher and lower frequencies (with reduced height and width), and subsequently adaptively combining them using a multiplier $\alpha$ (where $0 \leq \alpha \leq 1$) after processing \citep{chen2019drop, xu2020csa}.

For evaluation purposes, several metrics are utilized, including overall accuracy (OA), average accuracy (AA), producer's accuracy (class-wise accuracy, PA), and Cohen's kappa coefficient ($\kappa$) \citep{rossiter2004statistical}. In order to mitigate the impact of differences in initialization, all models are trained five times, and the mean values of the metrics are recorded. Moreover, McNemar's test is employed to assess the statistical significance of performance differences among the models \citep{mou2019learning}. The expression to calculate the \textit{z-statistic} (referred to as $z_{12}$, where 1 and 2 represent the models being compared) is provided in Equation \ref{equation:MN}.

\begin{sidewaystable*}[htbp]
  \centering{\scriptsize
  \caption{\label{tab:sal_perf}Accuracy analysis on the Salinas Valley dataset (in \%).}
    \begin{tabular}{|p{3.5cm}|p{0.9cm}|p{0.9cm}|p{1.3cm}|p{0.9cm} |p{0.9cm}| p{0.9cm}|p{1.7cm}|p{0.9cm}|p{0.9cm}|p{0.9cm}|p{1.0cm}|p{1.3cm}|}
 \hline
 Class name & SS Cas RNN & 2D CNN & Spec Atten Net & 3D CNN & Fus At Net & Hy- brid SN & Two Branch CNN & \textcolor{black}{2D Res Net} & \textcolor{black}{SAR- FN} & \textcolor{black}{LWt CNN} & \textcolor{black}{Octave CNN} & Hyper Loop Net\\
    \hline
    Broccoli green weeds1 & 99.20 & 99.97 & 99.46 & 99.33 & 96.45 & 98.92 & \textbf{100.00} & 99.90 & \textbf{100.00} & 99.80 & \textbf{100.00} & \textbf{100.00} \\
    Broccoli green weeds 2 & 99.17 & 99.92 & 99.64 & 99.94 & 98.49 & 99.71 & 99.75 & 99.79 & 99.97 & 99.96 & \textbf{100.00} & \textbf{100.00} \\
    Fallow & 98.59 & 99.73 & 99.09 & 98.28 & 92.27 & 99.91 & 99.77 & 99.55 & 99.99 & 99.99 & \textbf{100.00} & \textbf{100.00} \\
    Fallow rough plow & 99.55 & 99.61 & 99.78 & 97.30 & 99.37 & 99.35 & 99.63 & 99.91 & 99.75 & 99.41 & \textbf{99.98} & 99.88 \\
    Fallow smooth & 97.66 & 99.44 & 99.56 & 99.91 & 93.07 & 98.83 & 99.49 & 99.01 & 99.17 & 99.05 & \textbf{99.70} & 99.37 \\
    Stubble & 99.39 & 99.62 & 99.74 & 99.99 & 98.92 & 100.00 & 99.92 & 99.88 & 99.99 & 99.73 & \textbf{100.00} & \textbf{100.00} \\
    Celery & 98.69 & 99.94 & 99.72 & 99.30 & 99.28 & 99.65 & 99.52 & 99.80 & 99.94 & 99.94 & \textbf{99.99} & 99.92 \\
    Grapes untrained & 86.42 & 88.28 & 85.30 & 89.01 & 86.28 & 91.44 & 87.44 & 88.84 & 95.99 & 91.43 & 96.96 & \textbf{98.55} \\
    Soil vineyard develop & 99.42 & 99.73 & 99.61 & 99.88 & 98.93 & 99.80 & 99.77 & 99.79 & 99.92 & \textbf{100.00} & 99.98 & 99.89 \\
    Corn senesced green weeds & 88.68 & 97.09 & 91.91 & 96.66 & 94.21 & 96.07 & 95.92 & 98.04 & 98.99 & 99.23 & 99.49 & \textbf{99.60} \\
    Lettuce romaine, 4wk & 97.36 & 99.19 & 98.49 & 99.17 & 98.04 & 98.68 & 99.81 & 99.30 & 99.98 & 99.92 & 99.98 & \textbf{100.00} \\
    Lettuce romaine, 5wk & 93.10 & 99.93 & 99.53 & 99.95 & 97.01 & 99.93 & 99.98 & 99.97 & \textbf{99.99} & 99.97 & 99.93 & 99.92 \\
    Lettuce romaine, 6wk & 98.80 & 99.63 & 99.58 & 99.68 & 98.01 & 99.98 & 99.98 & 99.85 & 99.88 & 99.68 & 99.95 & \textbf{100.00} \\
    Lettuce romaine, 7wk & 96.14 & 99.03 & 97.32 & 99.05 & 96.97 & 98.25 & 98.72 & 98.43 & 99.42 & 99.53 & 99.86 & \textbf{99.98} \\
    Vineyards untrained & 77.36 & 87.82 & 81.14 & 79.09 & 90.11 & 92.42 & 83.30 & 88.17 & 96.02 & 94.03 & 97.13 & \textbf{97.89} \\
    Vineyard vertical treils & 99.05 & 99.14 & 99.23 & 99.34 & 99.52 & 99.21 & 99.21 & 99.52 & 99.50 & 99.75 & 99.65 & \textbf{99.94} \\
    \hline
    Overall accuracy & 92.47 & 95.48 & 93.50 & 94.33 & 94.02 & 96.64 & 94.63 & 95.70 & 98.45 & 97.20 & 98.90 & \textbf{99.32} \\
    Average accuracy & 95.54 & 98.01 & 96.82 & 97.24 & 96.06 & 98.26 & 97.64 & 98.11 & 99.28 & 98.84 & 99.54 & \textbf{99.68} \\
    Kappa ($\kappa$ \%)  & 0.9159 & 0.9496 & 0.9274 & 0.9366 & 0.9334 & 0.9625 & 0.9400 & 0.9520 & 0.9827 & 0.9687 & 0.9877 & \textbf{0.9924} \\
    \hline
    \end{tabular}}
\end{sidewaystable*}

\begin{equation}
z_{12} = \frac{|f_{12}-f_{21}|}{\sqrt{f_{12}+f_{21}}}
\label{equation:MN}
\end{equation}

Here, $f_{12}$ represents the number of samples correctly classified by model 1 and incorrectly by model 2, while $f_{21}$ represents the number of samples correctly classified by model 2 and incorrectly by model 1. Initially, the null hypothesis assumes that the two models do not exhibit different performance. If the value of $z_{12}$ exceeds $1.96$, the null hypothesis can be rejected with a $95$\% confidence level.

\subsection{Results and Discussion}

The performance metrics, including OA, AA, PA, and $\kappa$ (\%), for all models, including HyperLoopNet, are presented in Tables \ref{tab:hous13_perf}, \ref{tab:hous18_perf}, \ref{tab:sal_perf}, \textcolor{black}{and \ref{tab:puc_perf}} for the Houston 13, Houston 18, Salinas Valley, \textcolor{black}{and PaviaUC datasets}, respectively. The highest accuracies are highlighted in bold. Across all datasets, it is evident that our HyperLoopNet model outperforms all other state-of-the-art models in terms of OA (88.27\%, 76.64\%, and 99.32\% for Houston 13, Houston 18, and SV datasets, respectively), $\kappa$ (0.8728\%, 0.7108\%, and 0.9924\% for the three datasets), and AA (89.85\%, 87.91\%, and 99.67\%). For the PaviaUC dataset, HyperLoopNet achieves the highest $\kappa$ coefficient of 0.6844\% and the second highest overall accuracy of 77.76\%. Notably, our model consistently demonstrates superior performance in class-wise accuracies across all datasets, often surpassing other methods. In the Houston 13 dataset, the highest performance is observed in the \textit{residential} and \textit{commercial} classes. Similarly, for the Houston 18 dataset, our model outperforms other models in 11 classes (\textit{evergreen trees}, \textit{water}, \textit{residential buildings}, \textit{non-residential buildings}, \textit{roads}, \textit{crosswalks}, \textit{major thoroughfares}, \textit{railways}, \textit{unpaved parking lots}, \textit{trains}, and \textit{stadium seats}). Likewise, for the SV dataset, the highest accuracies are observed in 11 classes: \textit{broccoli green weeds 1}, \textit{broccoli green weeds 2}, \textit{fallow}, \textit{stubble}, \textit{grapes untrained}, \textit{corn senesced weeds}, \textit{lettuce romaine 4 wk}, \textit{lettuce romaine 6 wk}, \textit{lettuce romaine 7 wk}, \textit{vineyard untrained}, and \textit{vineyard vertical trails}. \textcolor{black}{Regarding the PaviaUC dataset, although HyperLoopNet does not achieve the highest class-wise accuracy, the accuracies are consistent, and no single class appears to be classified better at the expense of other classes compared to other models.} Furthermore, to establish the statistical significance of the proposed method over other methods, we present the z-values from McNemar's test between HyperLoopNet and other models in Table \ref{tab:tab_mcnmr}. A z-statistic value greater than 1.96 indicates the superior performance of our model, which is clearly observed.
\begin{table*}[htbp]
  \centering{\tiny
  \caption{\label{tab:puc_perf}{Accuracy analysis on the PaviaUC dataset (in \%).}}
    \begin{tabular}{|p{2.0cm}|p{0.7cm}|p{0.6cm}|p{0.8cm}|p{0.7cm}|p{0.6cm}| p{0.8cm}|p{0.8cm}|p{0.8cm}|p{0.6cm}|p{0.8cm}|}
 \hline
 Class name & SS Cas RNN & 2D CNN & Spec Atten Net & 3D CNN & FusAt Net & Hybrid SN & Two Branch CNN & 2D Res Net & LWt CNN & Hyper Loop Net\\
\hline
    Asphalt & 86.84 & 83.19 & 79.38 & 90.23 & 85.40 & 85.22 & 84.66 & \textbf{92.46} & 87.10 & 85.91 \\
    Meadows & 83.69 & 85.05 & 89.16 & 84.31 & 78.45 & 75.77 & 86.57 & 86.28 & \textbf{92.94} & 87.00 \\
    Trees & 87.11 & 87.36 & 86.84 & 84.73 & \textbf{89.11} & 81.63 & 85.68 & 87.06 & 75.10 & 80.09 \\
    Baresoil & 35.41 & 30.98 & 29.23 & 36.76 & 40.16 & \textbf{44.28} & 30.94 & 37.61 & 21.70 & 41.03 \\
    Bricks & 63.45 & 73.47 & 64.63 & 54.32 & 71.59 & \textbf{83.62} & 70.56 & 66.90 & 75.58 & 73.82 \\
    Bitumen & 32.35 & 65.62 & 63.10 & 36.32 & \textbf{72.60} & 68.38 & 37.38 & 38.06 & 45.68 & 46.95 \\
    Shadows & 82.32 & 92.33 & 88.03 & 88.19 & 84.86 & 89.57 & \textbf{96.87} & 83.61 & 75.29 & 84.73 \\
\hline
    Overall accuracy & 74.65 & 76.44 & 76.46 & 74.92 & 74.87 & 74.61 & 76.15 & 77.65 & \textbf{77.81} & 77.76 \\
    Average accuracy & 67.31 & 74.00 & 71.48 & 67.84 & 74.59 & \textbf{75.50} & 70.38 & 70.28 & 67.63 & 71.36 \\
    Kappa ($\kappa$ \%) & 0.6431 & 0.6679 & 0.6639 & 0.6478 & 0.6530 & 0.6501 & 0.6598 & 0.6814 & 0.6766 & \textbf{0.6844} \\
\hline
    \end{tabular}}
\end{table*}
To visualize the training and validation patterns of the models for each epoch, we present the loss and accuracy curves of the two best-performing models (HyperLoopNet and Octave CNN) for the Houston 13, Houston 18, and Salinas Valley datasets in Figure \ref{fig:la}. It can be observed in all scenarios that the training loss steadily decreases (with minor fluctuations) until it reaches a saturation point after 500 epochs. Simultaneously, the training accuracy increases and exhibits mild fluctuations, reaching around 99-100\%. On the other hand, for the test set, the loss initially decreases until a certain threshold and then gradually starts to increase. Similarly, the test accuracy steadily increases with each epoch and eventually reaches a fixed value, showing minor fluctuations. Considering this behavior, all models are trained for the full 500 epochs, and only the average of the maximum overall accuracy for the test set (across the 5 runs) is recorded. \textcolor{black}{Furthermore, to mitigate overfitting, we employ early stopping during model training, where the models are evaluated based on the point of achieving maximum test accuracy.} \textcolor{black}{Moreover, we address the issue of potential overfitting that can occur when training and testing are performed on patches from the same dataset. To alleviate this, we utilize the Pavia dataset, where two separate images are used for training and testing: the HSI image of Pavia University for extracting training patches and the HSI image of Pavia Centre for testing. From Table \ref{tab:puc_perf}, it is observed that even in this scenario, our model surpasses other models in terms of kappa ($\kappa$) and closely matches other models in overall and average accuracy. However, it is important to note that due to domain differences between the two images, the training process exhibited fluctuations in loss values for all models, and some models such as Octave CNN and SARFN failed to effectively converge, resulting in their exclusion from the table.}
\begin{table}[ht]
\centering{\scriptsize
 \caption{\label{tab:tab_mcnmr}Z - values of McNemar's test between HyperLoopNet and other models.}
\begin{tabular}{|l|r|r|r|}
 \hline
Method& Houston 13 & Houston 18 & Salinas Valley\\
\hline
    SSCasRNN & 20.34 & 284.21 & 59.97 \\
    2D CNN & 15.62 & 223.59 & 42.34 \\
    SpecAttenNet & 17.89 & 296.97 & 53.27 \\
    3D CNN & 4.80  & 105.86 & 49.02 \\
    FusAtNet & 11.29 & 172.23 & 50.71 \\
    HybridSN & 20.17 & 155.61 & 34.16 \\
    Two Branch CNN & 10.22 & 95.34 & 47.36 \\
    \textcolor{black}{2D ResNet} & 13.06 & 129.97 & 40.98 \\
    \textcolor{black}{SARFN} & 13.08 & 106.66 & 16.19 \\
    \textcolor{black}{LWt CNN} & 13.43 & 80.43 & 39.39 \\
    \textcolor{black}{Octave CNN} & 5.18  & 100.96 & 9.48 \\
 \hline
\end{tabular}}
\end{table}

\begin{figure}[t!]
  \centering
  \centerline{\includegraphics[width=8.5cm]{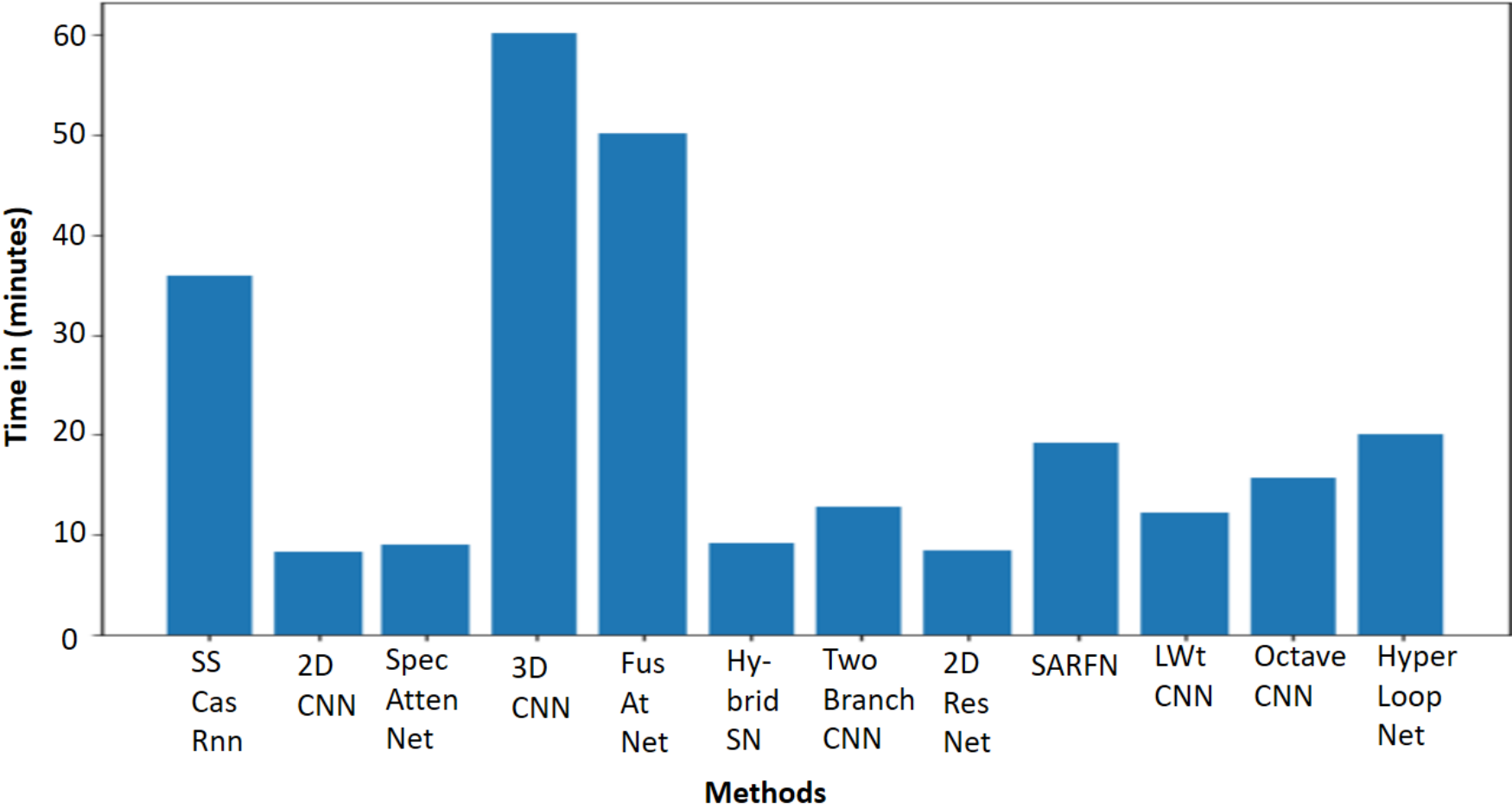}}
  \caption{\textcolor{black}{Bar chart representing approximate training time (in minutes) taken by the models.}}\medskip
  \vspace{-0.5cm}
  \label{fig:time_plot}
\end{figure}

\begin{figure}[t!]
  \centering
  \centerline{\includegraphics[width=14cm]{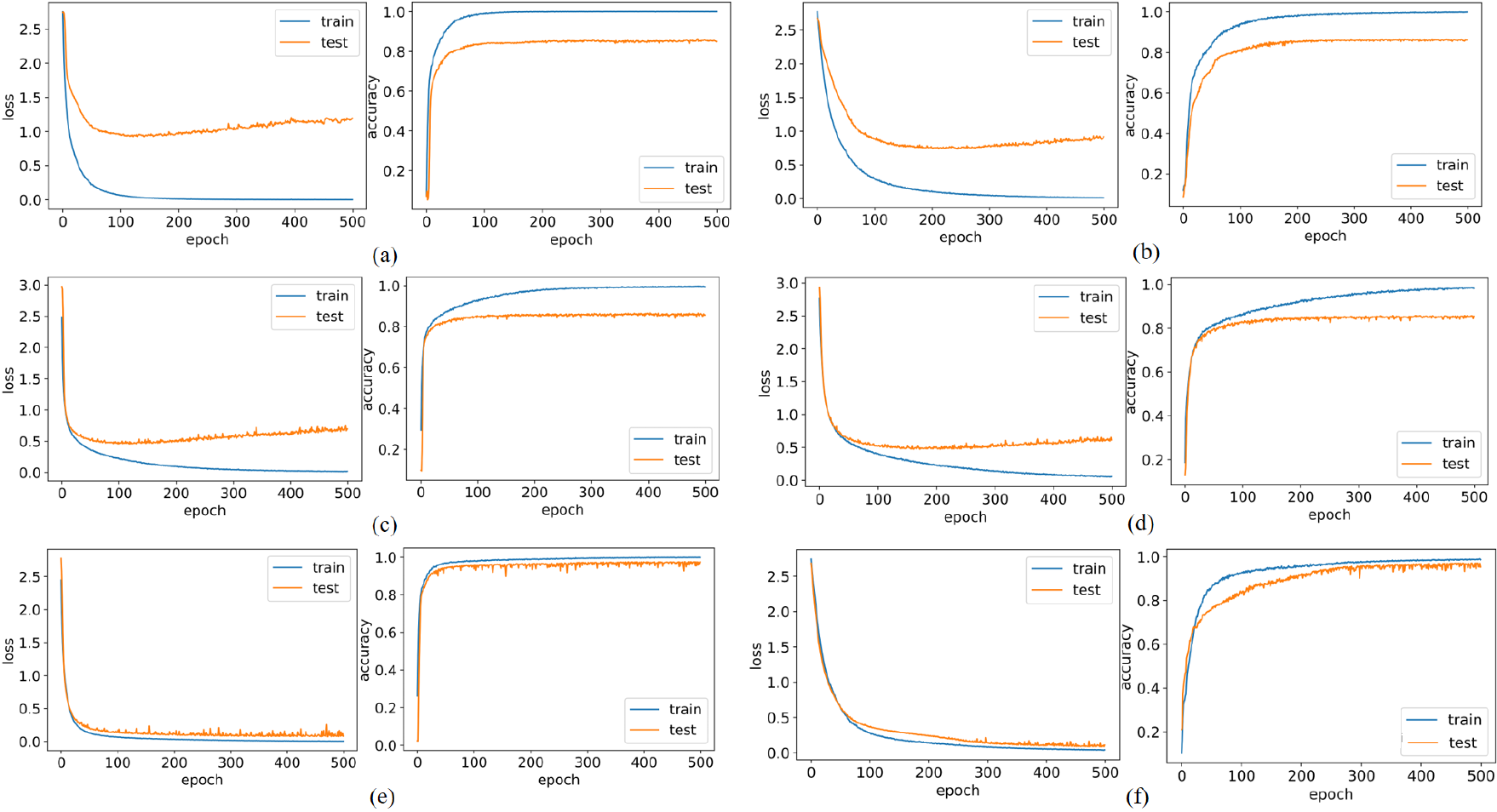}}
  \caption[Loss and accuracy curves for HyperLoopNet and Octave CNNs]{(a) and (b) Loss and accuracy curves with HyperLoopNet and Octave CNN for Houston 13 dataset. (c) and (d) Loss and accuracy curves with HyperLoopNet and Octave CNN for Houston 2018 dataset. (e) and (f) Loss and accuracy curves with HyperLoopNet and Octave CNN for Salinas dataset. \textcolor{black}{Here, we can observe that the loss continuously decreases upto a certain point and then starts to increase. Similarly, the test accuracy increases to a certain point and then starts to saturate Hence, we take the model corresponding to the best test accuracy.}}\medskip
  \vspace{-0.5cm}
  \label{fig:la}
\end{figure}

The approximate number of parameters required by all the deep learning models is provided in Table \ref{tab:para}. It is evident that our model achieves significantly better performance with a much lower number of parameters compared to several other models. Additionally, the training time for all the models is displayed in Figure \ref{fig:time_plot}. It can be observed that despite having fewer parameters, our model HyperLoopNet does not exhibit a proportionally reduced training time. This can be mainly attributed to the recursive or loopy nature of the model. Although the number of parameters is significantly lower, the number of operations with the same number of parameters increases. Consequently, the overall training time of the model increases. However, our model still requires less time than some of the other models (as evident in Figure \ref{fig:time_plot}).

\begin{table}[ht]
\centering{\scriptsize
 \caption{\label{tab:para}\textcolor{black}{Approximate number of parameters for the models.}}
\begin{tabular}{|p{2.5cm}|p{2.8cm}|}
 \hline
    Method & Approximate  number of parameters (in 1000s)\\
    \hline
    SS CasRNN & 1000 \\
    2D CNN & 500 \\
    SpecAttenNet & 3000 \\
    3D CNN & 3500 \\
    FusAtNet & 30000 \\
    HybridSN & 1500 \\
    Two Branch CNN & 15000 \\
    2D ResNet & 1000 \\
    SARFN & 150 \\
    LWt CNN & 10 \\
    Octave CNN & 550 \\
    HyperLoopNet & 350 \\
 \hline
\end{tabular}}
\end{table}

\begin{figure*}[t!]
  \centering
  \centerline{\includegraphics[width=14cm]{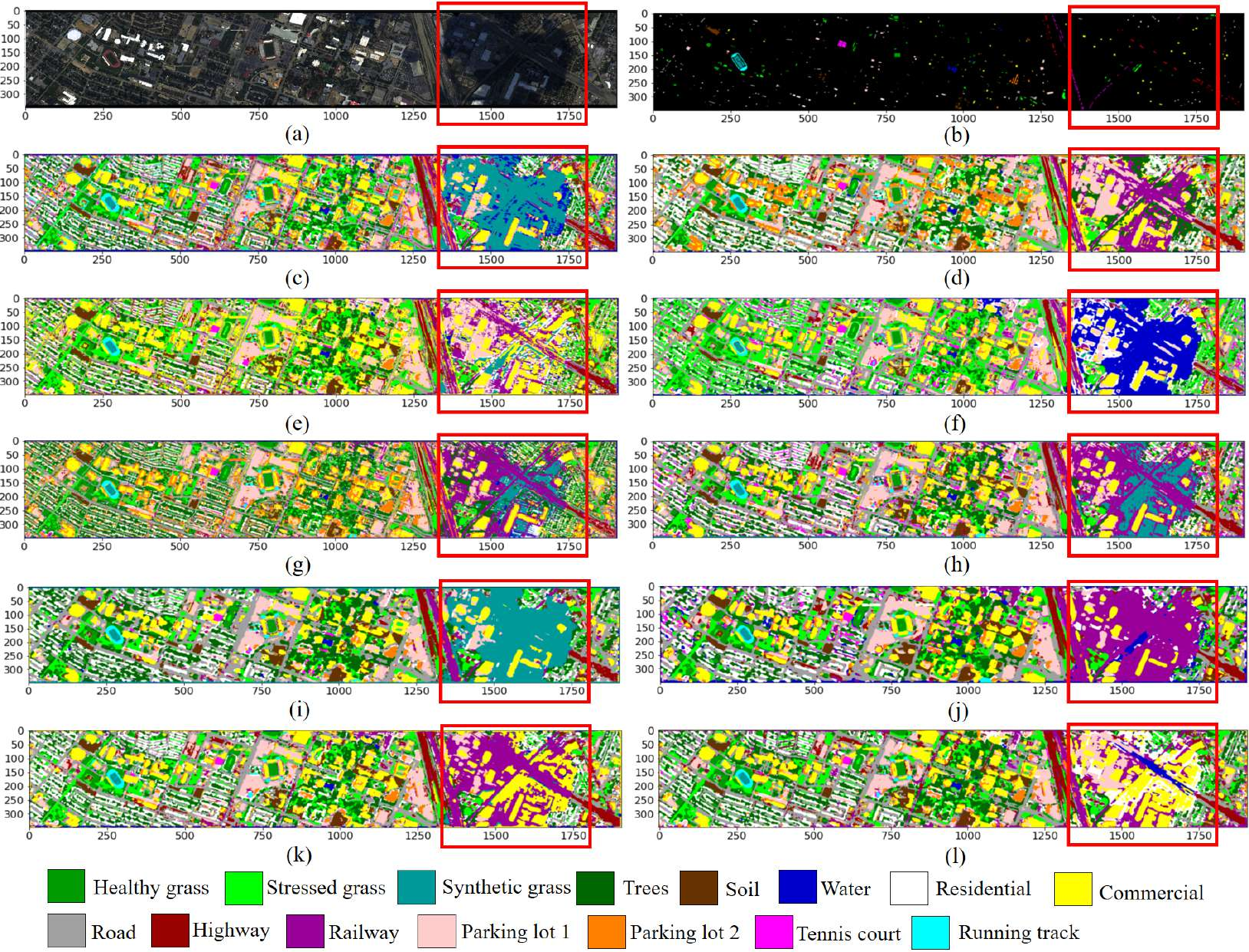}}
  \caption[Classification maps for Hosuton 13 dataset with proposed method and SOTA methods]{\textcolor{black}{(a) Three band true colour composite of Houston 13 HSI (b) Groundtruth of Houston 13 dataset. Classification maps using (c) SpecAttenNet (d) 3D CNN (e) FusAtNet (f) HybridSN (g) Two Branch CNN (h) Residual CNN (i) SARFN (j) LWt CNN (k) Octave CNN (l) HyperLoopNet}}\medskip
  \vspace{-0.5cm}
  \label{fig:IH13}
\end{figure*}

\begin{figure*}[t!]
  \centering
  \centerline{\includegraphics[width=14cm]{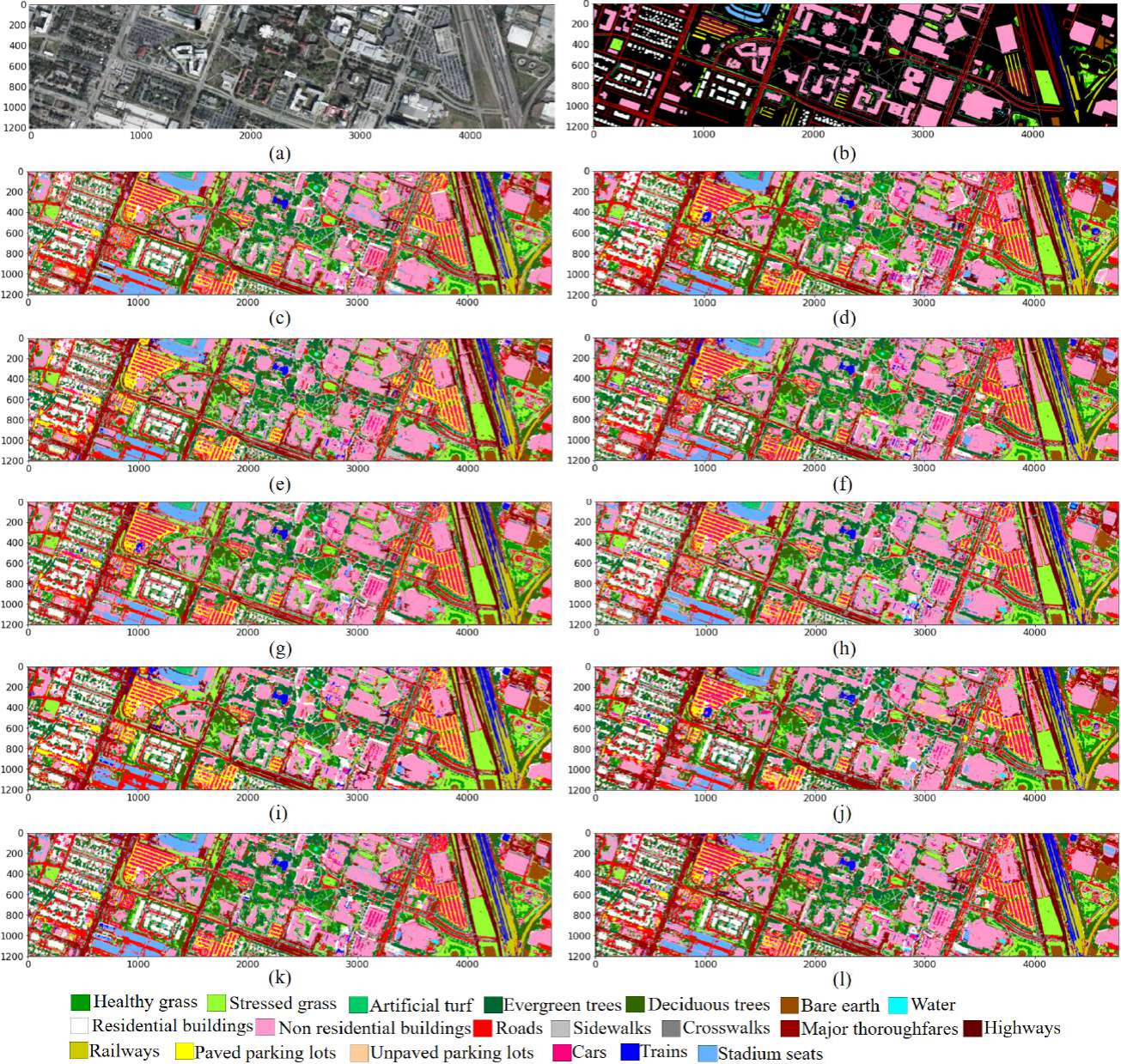}}
  \caption[Classification maps for Hosuton 18 dataset with proposed method and SOTA methods]{\textcolor{black}{(a) Three band true colour composite of Houston 18 HSI (b) Groundtruth of Houston 18 dataset. Classification maps using (c) SpecAttenNet (d) 3D CNN (e) FusAtNet (f) HybridSN (g) Two Branch CNN (h) Residual CNN (i) SARFN (j) LWt CNN (k) Octave CNN (l) HyperLoopNet}}\medskip
  \vspace{-0.5cm}
  \label{fig:IH18}
\end{figure*}

\begin{figure*}[t!]
  \centering
  \centerline{\includegraphics[width=14cm]{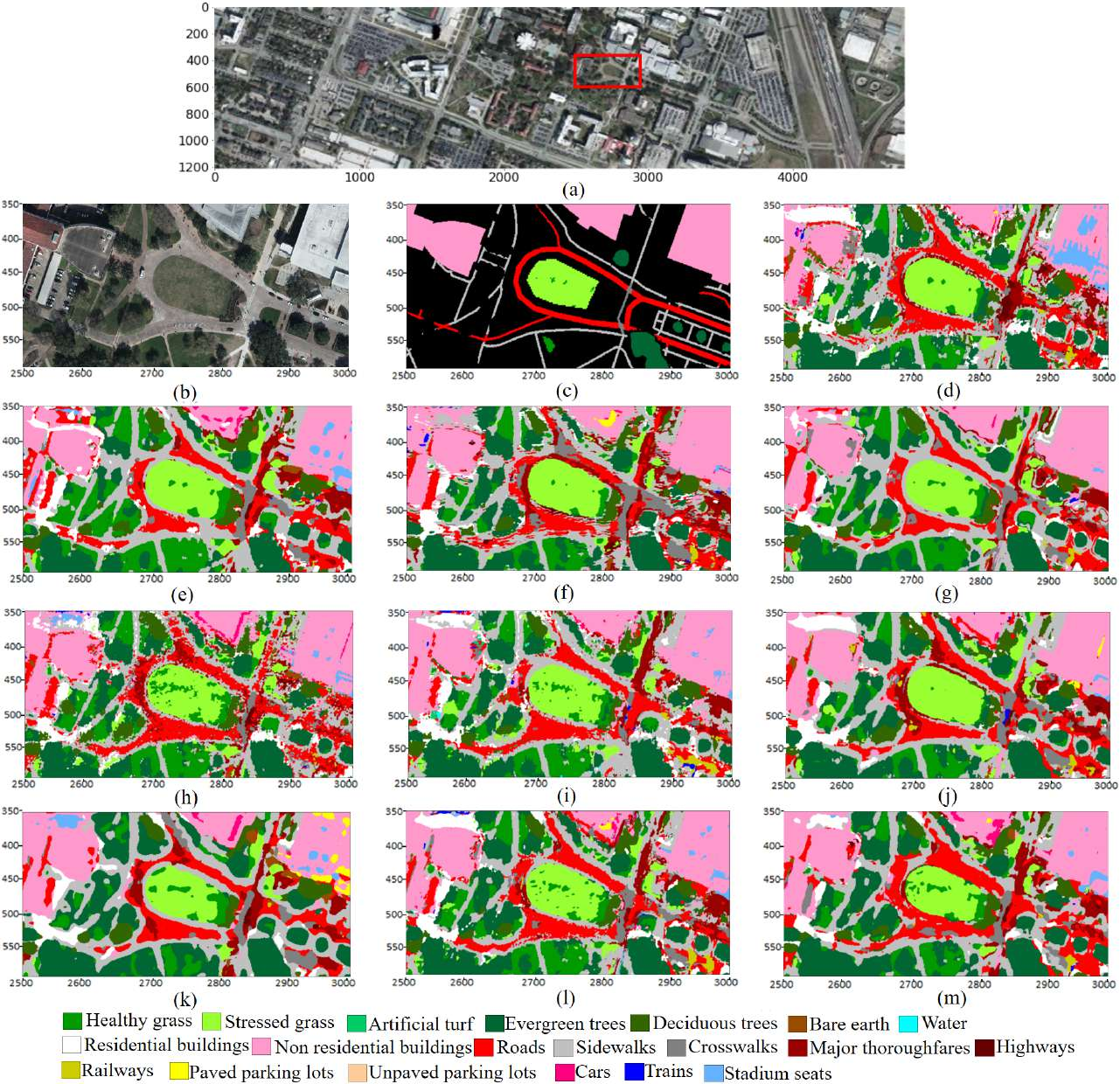}}
  \caption[Classification maps for subset of Hosuton 18 dataset with proposed method and SOTA methods]{\textcolor{black}{(a) Three band true colour composite of Houston 18 HSI (b) Three band true colour composite of a subset of Houston 18 HSI (c) Groundtruth of a subset of Houston 18 dataset. Classification maps using (d) SpecAttenNet (e) 3D CNN (f) FusAtNet (g) HybridSN (h) Two Branch CNN (i) Residual CNN (j) SARFN (k) LWt CNN (l) Octave CNN (m) HyperLoopNet}}\medskip
  \vspace{-0.5cm}
  \label{fig:IH18_sub}
\end{figure*}

\begin{figure*}[t!]
  \centering
  \centerline{\includegraphics[width=14cm]{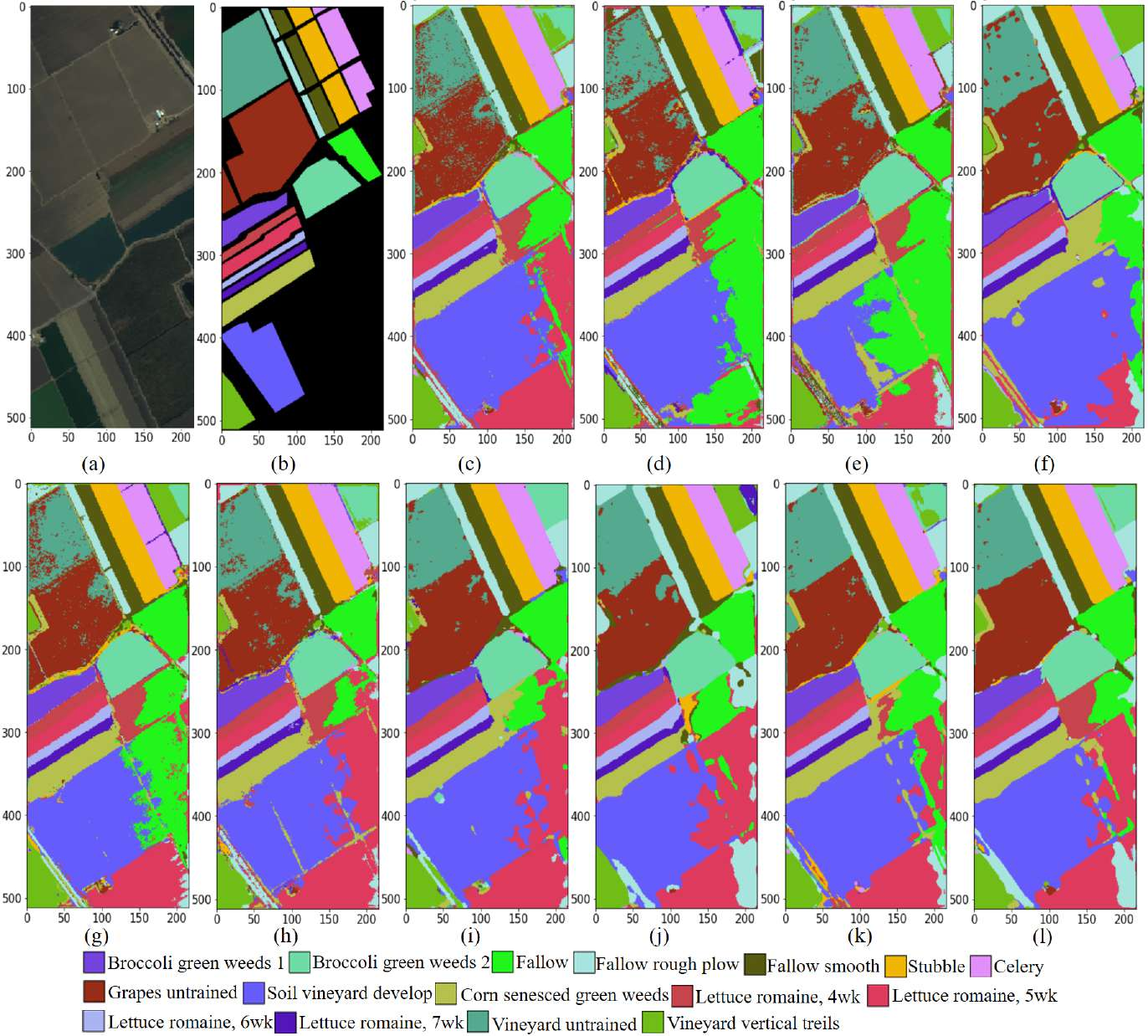}}
  \caption[Classification maps for Salinas Valley dataset with proposed method and SOTA methods]{\textcolor{black}{(a) Three band true colour composite of Salinas Valley HSI (b) Groundtruth of Salinas Valleys dataset. Classification maps using (c) SpecAttenNet (d) 3D CNN (e) FusAtNet (f) HybridSN (g) Two Branch CNN (h) Residual CNN (i) SARFN (j) LWt CNN (k) Octave CNN (l) HyperLoopNet}}\medskip
  \vspace{-0.5cm}
  \label{fig:Isal}
\end{figure*}

Furthermore, the classified maps of the Houston 13, Houston 18, and Salinas datasets are presented in Figure \ref{fig:IH13}, \ref{fig:IH18}, and \ref{fig:Isal}, respectively, for visualization purposes. In the case of Houston 13 (Figure \ref{fig:IH13}), it is evident that our method produces smoother delineation of classes, resulting in a relatively noise-free classification map. \textcolor{black}{Moreover, in the rightmost section of the image, there is a darker area (predominantly urban classes) due to a cloud shadow (refer to Figure \ref{fig:IH13} (a) for the original image, \textcolor{black}{outlined with a red rectangle}). This shadow might cause certain models (like HybridSN) to mistakenly interpret it as `water'. However, HyperLoopNet accurately identifies the urban area as primarily commercial or residential, showcasing its superior performance.} For the Houston 18 dataset (Figure \ref{fig:IH18}), the classification map obtained from HyperLoopNet demonstrates enhanced homogeneity among classes, avoiding speckling segments. \textcolor{black}{Additionally, to emphasize the spatial information and delineation in the classification maps of the Houston 18 dataset, we present a subset classification extracted from the main image (see Figure \ref{fig:IH18_sub}). The chosen segment size is 250$\times$500, which is referenced in Figure \ref{fig:IH18_sub} (a). In Figure \ref{fig:IH18_sub}, it is evident that our model achieves smoother boundary delineation. Furthermore, it can be observed that some models incorrectly classify the `Roads' class as `Highways' due to the similarity in reflectance patterns between the two classes. However, HyperLoopNet successfully avoids this issue and strives for more precise pixel classification.} Similarly, for the Salinas dataset (Figure \ref{fig:Isal}), the classification map generated by HyperLoopNet clearly and neatly generalizes classes into specific regions.
\begin{table}[ht]
\centering{\scriptsize
 \caption[Ablation study with different number of channels (all accuracies in \%).]{\label{tab:feat_abl}Ablation study with different number of channels (all accuracies in \%). \textcolor{black}{For \textit{HyperLoopNet}, with decrease in the number of channels, accuracy decreases for all the datasets (specially for Houston 18).}}
\begin{tabular}{|c|c|c|c|}
 \hline
& Houston 13 & Houston 18 & Salinas \\
\hline
8 channels per layer   &86.33 & 71.51 & 98.41\\
16 channels per layer  &87.54 & 74.13 & 99.00\\
32 channels per layer  &\textbf{88.27} & \textbf{76.64} & \textbf{99.32}\\
 \hline
\end{tabular}}
\end{table}

\begin{table}[ht]
\centering{\scriptsize
 \caption[Ablation study without feedback connections (all accuracies in \%).]{\label{tab:nofb}Ablation study without feedback connections (all accuracies in \%). \textcolor{black}{It can be observed that in the absence of feedback connection, the accuracy has decreased consistently for all the datasets.}}
\begin{tabular}{|c|c|c|c|}
 \hline
& Houston 13 & Houston 18 & Salinas \\
\hline
No feedback connections  & 87.74 & 76.55 & 99.27\\
With feedback connections &\textbf{88.27} & \textbf{76.64} & \textbf{99.32}\\
 \hline
\end{tabular}}
\end{table}
To further emphasize the superior performance of our model, we conducted several ablation studies on HyperLoopNet. The impact of varying the number of channels in each convolutional layer is presented in Table \ref{tab:feat_abl}. Reducing the number of channels has a more noticeable effect on the Houston 18 dataset, but for the other two datasets, the decrease in accuracy is not substantial. Even with only 16 channels, our model still outperforms state-of-the-art methods in the respective tables. 
Another ablation study (Table \ref{tab:nofb}) was performed to assess the influence of weight sharing and feedback connections on the network. It is evident that models with feedback connections achieve higher performance for all three datasets compared to models without feedback connections.
The third ablation study examines the effect of data augmentation on HyperLoopNet for the three datasets (Table \ref{tab:feat_aug}). There is a slight drop in accuracy for the Houston 13 dataset, but the difference is not significant for the other datasets.

Furthermore, we present an ablation study by considering only one of the three multiscale blocks (Table \ref{tab:feat_block}). As expected, using only one block results in lower accuracy. However, the accuracy trend for single blocks is not consistent across all three datasets. For example, in the case of the Salinas dataset, the highest accuracy is achieved with a spatial block of window size 3$\times$3, indicating the dominance of spectral features. Conversely, for Houston 13 and Houston 18, the highest accuracies are obtained with a window size of 5$\times$5, highlighting the contribution of spatial information. Notably, a window size of 7$\times$7 does not yield the best performance in any of the datasets, indicating significant information loss and relatively lower accuracy.

The final ablation study involves reducing the number of convolution blocks from 4 to 3 in each of the three self-looping blocks (Table \ref{tab:num_b}). Despite the decrease in the number of convolution blocks, the model's performance is not significantly affected for any of the datasets, demonstrating the effectiveness of HyperLoopNet even with fewer parameters.

\begin{table}[!htb]
\centering{\scriptsize
 \caption[Ablation study with and without data augmentation (all accuracies in \%).]{\label{tab:feat_aug}Ablation study with and without data augmentation (all accuracies in \%). \textcolor{black}{As the augmentation was removed from the training dataset, the decrement in the accuracy is observed for all the datasets.}}
\begin{tabular}{|c|c|c|c|}
 \hline
& Houston 13 & Houston 18 & Salinas \\
\hline
No augmentation  & 84.67 & 75.39 & 97.45\\
With augmentation  &\textbf{88.27} & \textbf{76.64} & \textbf{99.32}\\
 \hline
\end{tabular}}
\end{table}

\begin{table}[!htb]
\centering{\scriptsize
 \caption[Ablation study for individual spatial blocks (all accuracies in \%).]{\label{tab:feat_block}Ablation study for individual spatial blocks (all accuracies in \%). \textcolor{black}{We observe that when filters of only one scale are used, the accuracy decreases for all the datasets. But, accuracy for 5$\times$5 and 3$\times$3 filter are higher than 7$\times$7 for Houston 18 and Salinas datasets. This tells us about their higher contribution in feature extraction.}}
\begin{tabular}{|c|c|c|c|}
 \hline
& Houston 13 & Houston 18 & Salinas \\
\hline
3$\times$3  & 86.68 & 74.31 & 99.22\\
5$\times$5  & 87.37 & 74.53 & 98.97\\
7$\times$7  & 87.05 & 73.34 & 98.66\\
All  &\textbf{88.27} & \textbf{76.64} & \textbf{99.32}\\
 \hline
\end{tabular}}
\end{table}

\begin{table}[!htb]
\centering{\scriptsize
 \caption[Ablation study for number of convolution blocks per self-looping block (all accuracies in \%).]{\label{tab:num_b}Ablation study for number of convolution blocks per self-looping block (all accuracies in \%). \textcolor{black}{We observe that as we decrease the number of convolution blocks, the accuracy decreases, but not so significantly.}}
\begin{tabular}{|c|c|c|c|}
 \hline
& Houston 13 & Houston 18 & Salinas \\
\hline
3 Conv blocks  & 87.97 & 76.49 & 99.30\\
4 Conv blocks  &\textbf{88.27} & \textbf{76.64}& \textbf{99.32}\\
 \hline
\end{tabular}}
\end{table}

\section{Summary}

This chapter describes three research works focusing on the challenges of hyperspectral image classification in single-modal setting. The first research work focusses on the problem of domain adaptation in hyperspectral images. In the research work,we present a cross-domain classification algorithm for HSI using adversarial learning. The algorithm incorporates an additional loss for class-level cross-sample reconstruction in the source domain S within the framework of a standard domain classifier (DC). This enhances the meaningfulness and compactness of the learned space while imposing an orthogonality constraint to avoid redundant reconstructed features. To evaluate the effectiveness of the proposed technique, we conduct experiments on the Botswana and Pavia datasets. The results demonstrate the superiority of my approach over several existing ad hoc and neural network-based methods. Currently, the proposed method solely relies on spectral information, but we have plans to introduce spatial considerations, leveraging the advantages of CNNs, to achieve improved semantic segmentation of the scene.

The second work deals with the feature extraction aspect of the CNNs from spatial and spectral perspectives using attention mechanism. Here, we present a novel hybrid attention-based approach for hyperspectral image classification. The approach capitalizes on the salient features of 1D CNN-based spectral attention and 2D CNN-based spatial attention to enhance the spectral-spatial characteristics of HSI patches. By adaptively combining these features with the input using learnable weights, and employing a 3D CNN classifier, our method achieves improved classification performance. Additionally, to reduce intraclass variance, the classifier is jointly trained using both cross-entropy and Wasserstein losses. The effectiveness of our approach is evident from surpassing the accuracy of all benchmark methods in the experimental results, when presented on Houston 2013, Houston 2018 and Salinas Valley datasets. Moving forward, we aim to enhance the efficiency of our model by incorporating sparsity and constraining the number of training samples. 

The final research work also addresses the feature extraction capabilities of CNNs using a self-refining loop in the CNNs. Here, we introduce a novel approach known as HyperLoopNet for the classification of hyperspectral images. What sets our approach apart is the incorporation of feedback connections within a densely connected network. This design ensures that each convolution layer is connected to every other layer, creating a self-reinforcing structure. Due to these shared connections, our model exhibits a highly efficient information flow, allowing us to work with a relatively small number of model parameters. This design also facilitates the development of a multiscale structure, with convolution kernels of varying sizes within each self-reinforcing block. Consequently, our model enriches feature representations with information from different spatial scales. We evaluate our model using four established hyperspectral datasets: Houston 2013, Houston 2018, Salinas Valley, and Pavia Centre and University datasets. In all cases, HyperLoopNet outperforms other existing models. Notably, our model excels in achieving superior performance, even when equipped with a limited number of parameters compared to state-of-the-art models. For future research, we plan to incorporate attention mechanisms into our model to explicitly emphasize more discriminative features for classification. Additionally, we aim to explore the application of our model in the open-set scenario, encompassing classes that were not part of the model's training data.

\section{Publications}

\begin{enumerate}
    \item Shivam Pande, Biplab Banerjee, and Aleksandra Pi\v{z}urica (2019).Class reconstruction driven adversarial domain adaptation for hyperspectral image classification. In: Iberian Conference on Pattern Recognition and Image Analysis. Springer, pp. 472–484.
    \item  Shivam Pande, Biplab Banerjee, Adaptive hybrid attention network for hyperspectral image classification, Pattern Recognition Letters 144 (2021) 6–12.
    \item Shivam Pande, Biplab Banerjee, HyperLoopNet: Hyperspectral image classification using multiscale self-looping convolutional networks, ISPRS Journal of Photogrammetry and Remote Sensing 183 (2022) 422–438. 
\end{enumerate}
\chapter{Hyperspectral image classification in multimodal setting}

\section{Introduction}

As advancements in remote sensing research continue, an increasing amount of data for earth observation is being gathered from various sources such as multispectral images (MSI), hyperspectral images (HSI), synthetic aperture radar (SAR), light detection and ranging (LiDAR), and more. Each of these modalities plays a unique role in earth observation. For instance, HSI offers superior spectral resolution and color discrimination capabilities, which are crucial for distinguishing elements of similar color, like crops \citep{belwalkar2022evaluation}. Similarly, MSI provides higher spatial resolution and decent spectral resolution, making it useful for differentiating urban features such as roads and buildings, where shape information is as important as color information \citep{liu2018roadnet}. To identify classes with variations in height, such as trees and shrubs, or road and roofs, the height information derived from the LiDAR point cloud's digital surface model (DSM) can be utilized \citep{shinde2021lidarcsnet}. Additionally, SAR sensors are employed for sub-surface monitoring applications due to their high penetrative power and reliance on active sensing \citep{tsai2019remote}. Several research works have analyzed these modalities separately in various applications. Initially, numerous machine learning algorithms were employed for remote sensing image analysis. For instance, crop identification from SAR images was performed using entropy decomposition and support vector machines (SVM) \citep{tan2011agricultural}. Hyperspectral image classification was accomplished using canonical correlation forests \citep{xia2016hyperspectral}. In their work, \citep{jafarzadeh2021bagging} utilized bagging and boosting ensembles of tree-based classifiers for the classification of SAR, hyperspectral, and multispectral images (individually taken).

However, real-world scenarios often present challenges where multiple classes coexist within images, making it difficult for a single data source to accurately identify these classes in a given geographical area. Therefore, instead of solely focusing on the capabilities of individual modalities, it becomes crucial to judiciously combine multiple modalities to achieve improved and accurate land use/land cover (LULC) classification. To address these needs, numerous studies have employed a joint utilization of multiple data sources for the aforementioned task. An integrated approach based on wavelets and intensity-hue-saturation (IHS) transform was proposed by \citep{hong2009wavelet} to fuse SAR images with medium-resolution MSI. \citep{iervolino2019novel} presented an algorithm for fusing multispectral, panchromatic, and SAR images, utilizing generalized IHS transform and Á-trous wavelet transform for land cover classification. For monitoring mangrove species, \citep{cao2021combining} introduced a method that fuses HSI and LiDAR images using rotation forest. Object-level feature extraction was employed by \citep{hu2016object} to fuse PolSAR and HSI data, followed by fusion and utilization for land cover classification. Deep learning techniques have also been applied to address multimodal fusion for various modalities. For example, \citep{hu2017fusionet} presented a two-stream convolutional neural network (CNN) for fusing PolSAR and HSI modalities. In the context of combining HSI and SAR modalities, \citep{li2022asymmetric} introduced the concept of asymmetric feature fusion, where scaling factors in batch normalization layers are utilized to identify redundant features and eliminate them using sparse constraint. Fusion techniques have also been extended to hyperspectral and LiDAR imagery. \citep{xu2017multisource} proposed a method for fusing HSI and LiDAR data using a trainable two-branch architecture, where each branch independently extracted features from the two modalities. The resulting features were then concatenated and used for classification.

The challenge of multimodal learning in the field of remote sensing can be approached from various perspectives, particularly for classification tasks. One perspective involves predicting the class label of a geographical area based on all available modalities. Another approach is to train a model with multiple modalities but test it with only a subset of modalities. In our research work, we have addressed this problem in both scenarios.

\subsection{Missing Modality Generation}

In remote sensing image classification models, it is not always feasible to obtain all cross-modal information simultaneously, especially in time-critical situations such as disaster management where different sensors may have varying temporal resolutions. This creates a scenario in developing machine learning systems: \textit{while training data can be captured offline with multimodal information, test samples may not always have access to all modalities in real-time}.

Nevertheless, it is advantageous to train prediction models with additional multimodal information, a concept known as \textit{learning with privileged information} \citep{lopez2015unifying}. However, if the test samples do not have consistent feature dimensions compared to the training data, the trained model cannot be effectively evaluated on the test set. Two potential solutions exist in this context: i) utilize the information available during both training and testing, even if it compromises the model's performance to some extent, or ii) train the model with privileged side information and devise a method to approximate the missing information of the test data using the available information, commonly referred to as \textit{modality distillation through hallucination} \citep{garcia2018learning}.

In our research, we propose a solution to the distillation problem in RS image classification by employing a deep generative model-driven teacher-student architecture. The teacher model is trained with all modalities, and subsequently, we train a student model that operates on both the available features and hallucinated features.

We consider two experimental scenarios: i) RS scene classification using multispectral (modality-1) and panchromatic (modality-2) data, and ii) HSI classification where two non-overlapping subsets of spectral bands represent the respective modalities. It should be noted that discriminative feature descriptors are initially learned specifically for each modality, and the distillation module focuses on learning the feature mapping from modality-1 to modality-2.

In the field of computer vision, there has been a recent increase in knowledge transfer approaches, particularly within the teacher-student distillation framework \citep{Garcia_2018_ECCV, hoffman2016learning}. However, despite its significant importance in remote sensing (RS), this problem has received limited attention. To the best of our knowledge, the notable endeavor by \citep{kampffmeyer2018urban} is the only work that directly extends the approach of \citep{hoffman2016learning} to support RS images. Nonetheless, the reconstruction loss-based method of \citep{hoffman2016learning} does not consistently capture the overall data distributions effectively. Another crucial aspect is the discriminativeness of modality-specific feature descriptors. A conditional generative adversarial network (C-GAN) has recently been applied to address this concern \citep{roheda2018cross}. However, vanilla GAN models with binary discriminators often encounter the issue of mode collapse during the hallucination learning phase. Therefore, based on our review, we extend the idea of modality generation in remote sensing using conditional GAN and following that, transfer the information from a network with all the modalities to that with limited number of modalities.

\subsection{Multimodal Fusion}

When dealing with hyperspectral images (HSIs), the challenge of high dimensionality, often referred to as the \textit{curse of dimensionality} \citep{koppen2000curse, rasti2020feature}, arises, requiring algorithms to have a larger amount of data for satisfactory performance. Currently, deep learning algorithms are widely used for processing HSIs in various applications such as classification, object detection, and unmixing. However, this issue is further compounded with deeper and wider architectures that require a greater number of parameters. Moreover, deep learning models primarily follow a forward direction, resulting in predominantly unidirectional information flow and potentially limiting the generation of robust representations \citep{yang2018convolutional}. Effective feature extraction in multimodal scenarios is being addressed through various approaches.
\begin{figure}[t] 
\centering
  \centerline{\includegraphics[width=8cm]{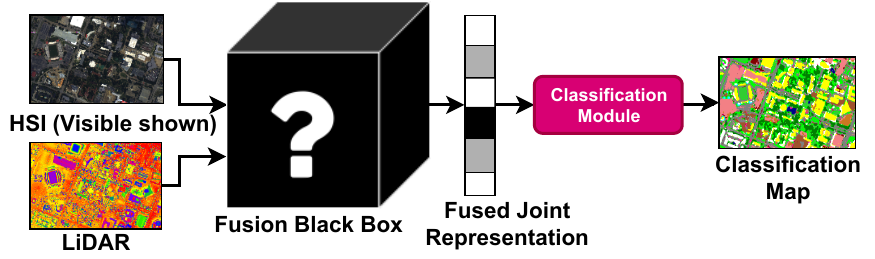}}
\caption[Schematic of general fusion framework for HSI-LiDAR fusion.]{A general diagram illustrating a classification task based on multimodal fusion. The goal is to combine two modalities (in this case, HSI and LiDAR) in a manner that produces a fused representation with informative and resilient features suitable for precise classification.}\medskip
  \vspace{-0.5cm}
\label{fig:IntroDiagram}
\end{figure}
The usage of attention learning mechanisms has recently demonstrated remarkable performance gains in different visual inference tasks \citep{jetley2018learn, mou2019learning, haut2019visual}. Ideally, attention modules highlight salient features while suppressing irrelevant ones through self-supervised learning. However, most of these research works focus on attention-based learning within a single modality, resulting in the highlighting of similar features. Therefore, the challenge lies in designing a network that takes attention masks from one modality and uses them to enhance the representations of other modalities (Figure \ref{fig:IntroDiagram}). Motivated by this, the concept of multimodal attention is envisioned, where a complementary modality not only adds relevant information synergistically to the existing modality but also highlights features that may have gone unnoticed by the attention map derived from the existing modality. Based on these discussions, we propose FusAtNet, an attention-based multimodal fusion network for land-cover classification using an HSI-LiDAR pair as input, as illustrated in Figure \ref{fig:Diagram}. Our approach involves extracting spectral features using ``self-attention'' in HSI and incorporating multimodal attention through the proposed ``cross-attention'' mechanism, which utilizes the LiDAR modality to derive an attention mask highlighting the spatial features of the HSI (Figure \ref{fig:self_cross}). This interaction between spectral and spatial features leads to an intermediate representation that is further refined through self-attention-based learning. Finally, this rich representation is utilized for classification.

Another approach to address this problem is the incorporation of feedback connections in convolutional neural networks (CNNs), where information from future stages is propagated backwards multiple times, enabling derived features to become more informative. This concept also aligns with the idea of \textit{curriculum learning} \citep{wang2021survey}, where the model is initially trained on relatively easier samples and gradually progresses to harder ones \citep{yang2018convolutional, pande2022feedback}. Furthermore, the inclusion of feedback connections simulates an attention-like framework within the models, enhancing their ability to select informative features. For example, \citep{li2019feedback} utilizes convolutional long short-term memory (LSTM) for image super-resolution. The network incorporates future information through feedback connections to refine initial features and generate sharper images. Similarly, a feedback connection-based architecture is presented in \citep{cheng2019multi} for multiclass object detection, where future information from later layers serves as posterior probabilities to refine information from initial layers. The research on feedback connections is also extended to the remote sensing domain. \citep{fu2020two} introduces the concept of feedback connections in the pansharpening of multispectral images (MSI). The network consists of a two-path CNN architecture that takes MS-PAN pairs as input in each branch. After feature extraction, the features are passed through an RNN-like feedback block multiple times to refine low-level features, which are then sent to the decoder to obtain the sharpened image. \citep{pande2022hyperloopnet} proposes a method for HSI classification using densely connected feedback blocks. The network consists of convolutional self-looping (SL) feedback blocks with three different kernel sizes. Each block has convolutional layers connected to each other to facilitate maximum information flow. The outputs from the three blocks are concatenated and used for classification.

Furthermore, most existing fusion approaches have certain associated drawbacks. Figure \ref{fig:fa} illustrates the different fusion approaches found in the literature, along with the proposed model. The fusion paradigms mainly employ \textit{single-level} feature fusion, where fusion occurs either at the beginning of the deep learning model (\textit{early fusion}, see Figure \ref{fig:fa} (a)) or at later stages of the network (\textit{middle fusion} and \textit{late fusion}, see Figure \ref{fig:fa} (b) and \ref{fig:fa} (c)). However, in \textit{single-level} fusion, the heterogeneity between features may not be adequately considered. To address this issue, \citep{hang2020classification} proposed a method that introduced shared connections among the layers of HSI and LiDAR-based encoders. This ensured that the two encoders mutually guided each other while learning the hidden representations between the modalities, while also controlling the number of parameters within the model.
\begin{figure}[!t]
  \centering
  \centerline{\includegraphics[width=8cm]{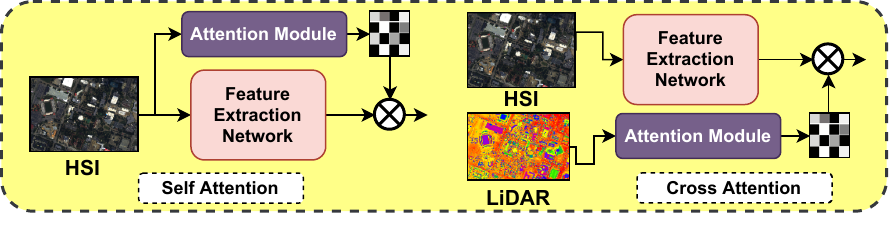}}
\caption[A visual depiction of self-attention and cross attention.]{Self-attention and cross-attention mechanisms for multimodal fusion are compared. The self-attention module (left) operates solely within a single modality, where both the hidden representations and the attention mask are derived from the same modality (HSIs). In contrast, the cross-attention module (right) utilizes an attention mask obtained from a different modality (LiDAR) to enhance the latent features of the first modality.}\medskip
  \vspace{-0.5cm}
\label{fig:self_cross}
\end{figure}
Additionally, another factor associated with deep fusion-based models is ensuring the robustness of features generated for classification with limited training samples. Several approaches in the deep learning domain have tackled similar issues by employing self-supervised pretraining of models \citep{ohri2021review}. Typically, these approaches involve training the model on a task (referred to as the \textit{pretext} task) that is independent of the ground truth information of image patches. This can include generative tasks such as reconstructing the original sample or discriminative tasks such as identifying similar and different samples based on their augmentations. Learning on such tasks allows the models to capture the data distribution, which can then be utilized for other downstream tasks like classification. Due to these characteristics, these self-supervision tasks are sometimes included as \textit{auxiliary tasks} in conjunction with the main supervised task. \citep{pande2019class} implemented a domain adaptation-based framework for HSI classification, where cross-sample-based reconstruction from the same class was used as an auxiliary task. \citep{zhang2018feature} proposed a method called patch-to-patch CNN, where a separate network was trained to construct LiDAR-based DSM from HSI imagery in an unsupervised manner. After training, features from intermediate layers were used for classification, assuming that the transition features from HSI to LiDAR would be more effective. \citep{hong2020deep} proposed a fully connected network for pixel-level fusion of HSI and LiDAR features, utilizing two separate branches to reconstruct the original features from the fused representation (Figure \ref{fig:fa} (d)). \citep{wu2021convolutional} presented a cross feature reconstruction framework, where features extracted from one modality were used to reconstruct the same features from another modality. However, unlike our proposed work, the aforementioned research solely focuses on feature-level reconstruction and does not involve the original modalities in the reconstruction process. Moreover, the reconstruction is performed on the features obtained after the transformation of the fused domain, which imposes a more relaxed constraint on the fused features.

\begin{enumerate}
    \item Since we are dealing with multiple modalities simultaneously, it is crucial to systematically emphasize the relevant information from each modality while disregarding unnecessary information for the task at hand. This can be achieved by selectively attending to specific channels in hyperspectral images and spatial locations in HSI and LiDAR/SAR images.
    \item Obtaining more robust feature representations from the modalities before fusion is necessary to achieve more accurate classification. One efficient approach is leveraging future information to refine past feature representations within the network.
    \item Due to the limited number of training samples, it is important to develop models with a controlled number of parameters. This can be accomplished by incorporating the idea of shared weights across modalities and across time in the networks.
    \item To achieve better fusion, it is essential for the pre-fusion features to effectively capture the inherent properties of all involved modalities, thereby enhancing the information content of the fused features.
\end{enumerate}

We propose a novel fusion paradigm for HSI-LiDAR/SAR modalities. Our method utilizes multiscale self-looping blocks that share the same set of parameters across time and modalities. Weight sharing across time creates a compact architecture where each convolution layer exchanges information with every other layer (both forward and backward). This ensures more robust feature extraction by refining past layers' features using future information. Weight sharing across modalities enables the network to learn correlated features of the two modalities through cross-modal communication. Moreover, sharing the same set of parameters across time and modalities helps control the number of parameters in the network. Additionally, to explicitly enforce the characteristics of one modality on the extracted features of another, we introduce a \textit{self-supervised auxiliary task} for cross-modal reconstruction (CMR) for both modalities involved. These cross-modal reconstruction modules take the pre-fusion extracted features of one modality as input and use them to reconstruct the input features of the other modality. This ensures that the pre-fusion features contain embedded information from both modalities, leading to more robust classification.

Based on the aforementioned studies, we present a research work for modality generation for HSI and two other research works on hyperspectral image fusion with a LiDAR-based DSM model and SAR imagery. Our contributions are summarised as follows:

\begin{enumerate}

    \item We propose a novel teacher-student framework for modality distillation in RS image classification, incorporating a new C-GAN-based cross-modality mapping module. Additionally, we employ the knowledge distillation technique to ensure that the student's classifier remains consistent with the teacher's classifier. We introduce data augmentation through noise perturbation on the teacher's training samples to train the hallucination and student models effectively. We conduct extensive experiments on HSI classification and RS scene classification using MS-PAN image pairs, demonstrating improved results. The methodology can be referred in section \ref{sec:HalNet} and is published as \cite{pande2019adversarial}.

    \item We propose a novel HSI-LiDAR fusion model called FusAtNet. Our approach is one of the pioneering attempts to introduce attention learning for HSI-LiDAR fusion in the context of land-cover classification. In this regard, we propose the concept of ``cross-attention'' based feature learning among the modalities, which is a novel and intuitive fusion method. It utilizes attention from one modality (LiDAR) to highlight features in the other modality (HSI). We demonstrate superior classification performance on three benchmark HSI-LiDAR datasets, surpassing all existing deep fusion strategies and conduct a robust comprehensive analysis, that gives detailed insights for our model. The details of the approach are presented section \ref{sec:FAN} (published as \cite{mohla2020fusatnet}).

    \item In the third research work, we introduce dense feedback connections-based convolutional feature extractors for remote sensing image fusion. Each feature extractor consists of ``coupled self-looping'' blocks, where weights are shared across time and simultaneously with the extractor of the other modality. To enhance spatial information, we employ a wider network that accounts for multiple scales of convolutional kernels. This represents the primary novelty of our approach, as it is the first time the same set of parameters is utilized across time and modalities to create a compact fused representation of multisource data. The sharing of the same parameters across time and modalities also ensures control over the number of parameters, making the model manageable and less prone to overfitting. Another contribution of our approach is the incorporation of self-supervised cross-modal reconstruction modules. These modules utilize features from one modality prior to fusion to reconstruct the input channels of the other modality. This ensures that even before concatenation-based fusion, the features are representative of both the modalities. We call our method `HyperFuseNet' and it can be referred in section \ref{sec:HFN} (published as \cite{pande2023self}).
    
\end{enumerate}

\section{Missing modality prediction from hyperspectral images}
\label{sec:HalNet}
In this research study, we aim to address the issues related to discriminative modality distillation using a teacher-student network approach. The teacher network comprises multiple streams and a multi-layer classifier, with each stream focused on learning discriminative feature representations specific to a particular modality{\footnote{Published as: Shivam Pande, Avinandan Banerjee, Biplab Banerjee, Subhasis Chaudhuri. \href{https://openaccess.thecvf.com/content_ICCVW_2019/papers/CROMOL/Pande_An_Adversarial_Approach_to_Discriminative_Modality_Distillation_for_Remote_Sensing_ICCVW_2019_paper.pdf}{An adversarial approach to discriminative modality distillation for remote sensing image classification}, Oct. 2019, In: Proceedings of the IEEE/CVF International Conference on Computer Vision Workshops, pp. 4571–4580.}}. We then employ a C-GAN based hallucination model to generate features for the missing modality, conditioned on the available modality's features. To ensure discriminativeness of the hallucinated features, we propose using $2C$ nodes for $C$ classes in the C-GAN discriminator and optimize the model using a min-max approach. To carry out the $C$-class classification task, we design the student network, which takes both the available and hallucinated modality features into account. This setup not only addresses the mode collapse problem but also ensures discriminativeness in the hallucinated features. Furthermore, we incorporate soft-target based knowledge distillation (KD) for training the student's classifier. The details are addressed in the subsequent sections. 

\subsection{Methodology}
We provide a detailed explanation of the algorithm proposed in this section. The training process consists of three main phases: i) training the teacher network, ii) training the hallucination module, and iii) training the student's classifier for practical use during inference. Each of these stages is described below.

\begin{figure}[t]
\begin{center}
\includegraphics[width=0.7\textwidth]{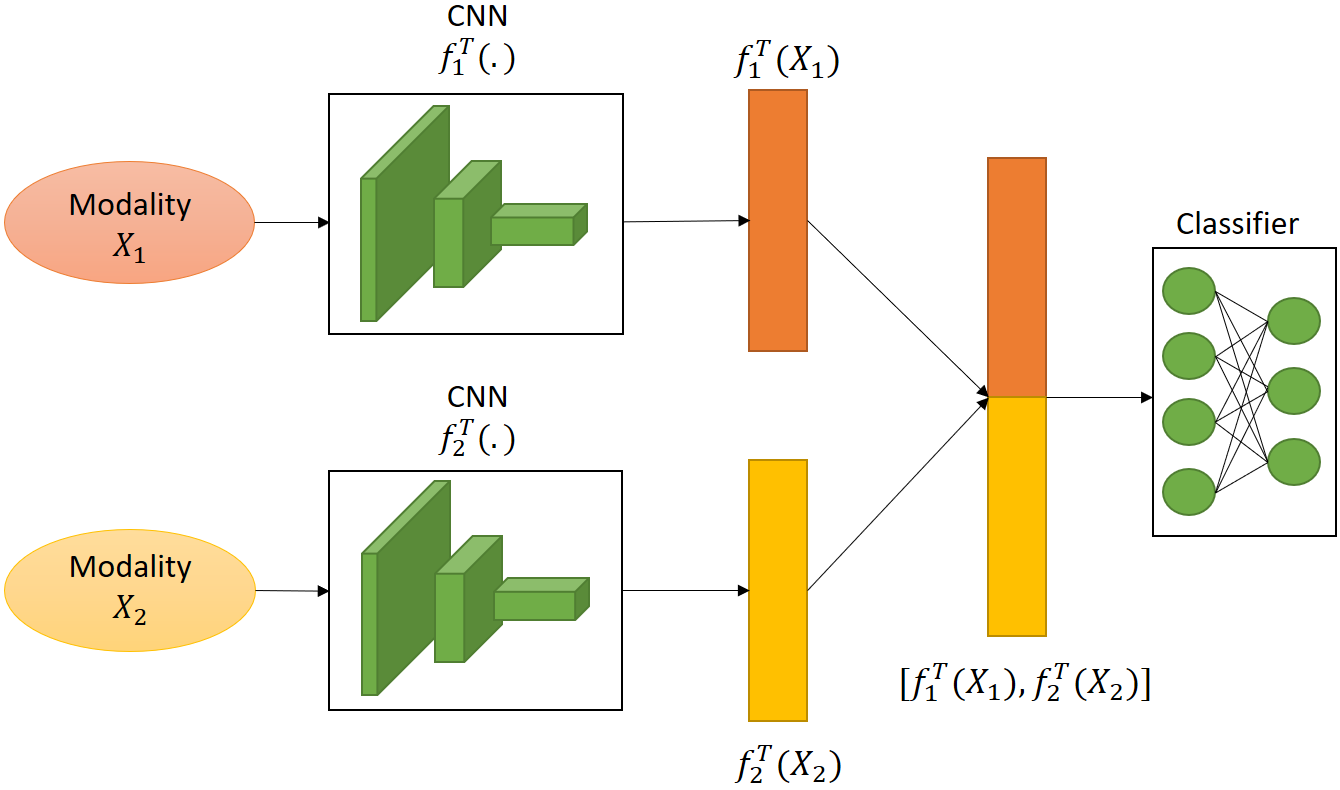}
\end{center}
\caption[Illustration of the teacher network]{The diagram illustrates the structure of the teacher network. Both modalities $X_1$ and $X_2$ are inputted into the respective feature extractors $f_1^T(.)$ and $f_2^T(.)$. The extracted features from both modalities are then concatenated and passed to the classifier $f_C^T$.}
\label{fig:TN}
\end{figure}

\begin{figure}[t]
\begin{center}
\includegraphics[width=10cm]{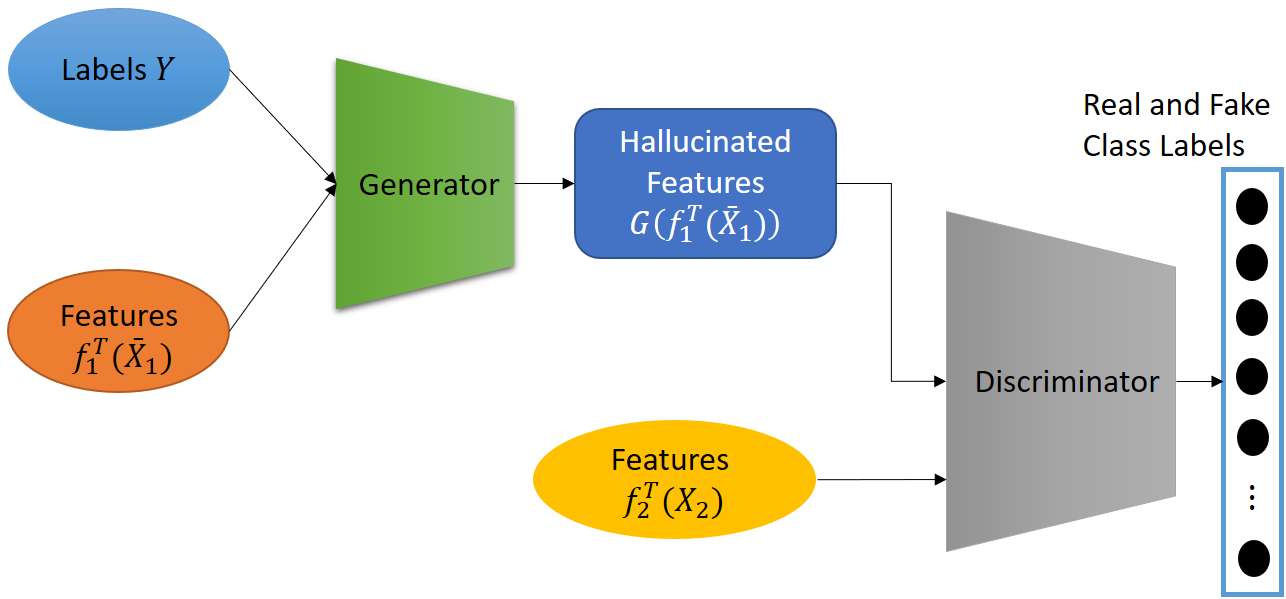}
\end{center}
\caption[Illustration of the modality hallucination network]{The schematic diagram illustrates the process of feature hallucination using a C-GAN. The extracted features $f_1^T(\Bar{\mathcal{X}}_1)$ and corresponding labels are inputted into the generator $G$, which generates hallucinated features $G(f_1^T(\Bar{\mathcal{X}}_1))$. These hallucinated features, along with the real features $f_2^T(\mathcal{X}_2)$, are then fed into the discriminator $D$. The discriminator is trained to distinguish between the real and fake sets of classes.}
\label{fig:Hal}
\end{figure}

\begin{figure}[t]
\begin{center}
\includegraphics[width=10cm]{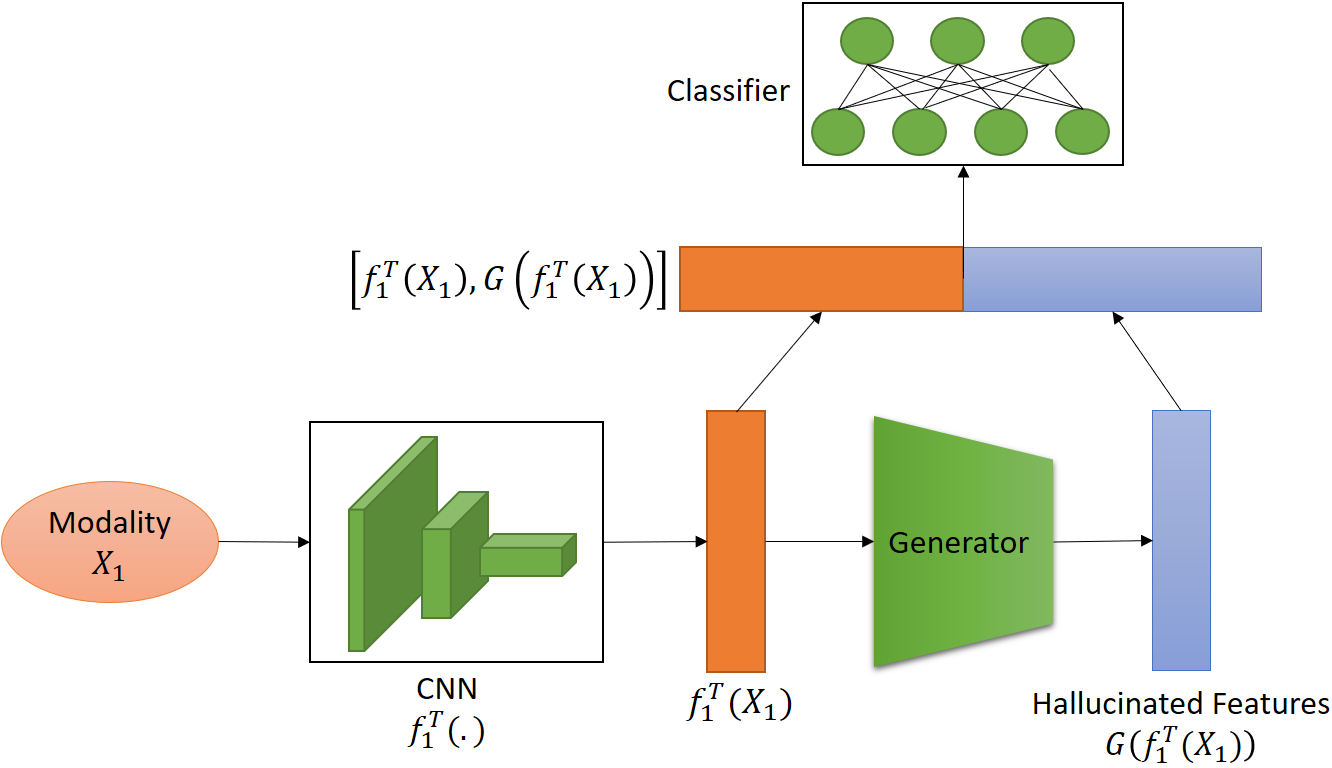}
\end{center}
\caption[Illustration of the student network]{The diagram represents the structure of the student network. Modality $X_1$ is inputted into the feature extractor $f_1^T(.)$, which extracts features $f_1^T(\mathcal{X}_1)$. These extracted features are then passed to the trained generator $G$, which hallucinates the missing modality by generating features $G(f_1^T(\mathcal{X}_1))$. Both the original features $f_1^T(\mathcal{X}_1)$ and the hallucinated features $G(f_1^T(\mathcal{X}_1))$ are concatenated and forwarded to the classifier $f_C^S$.}
\label{fig:SN}
\end{figure}

\subsubsection{Training the teacher network}

Let us consider a dataset $\mathcal{X} = \{x_1^i, x_2^i,y^i\}_{i=1}^N$, where $x_1^i \in \mathcal{X}_1$ and $x_2^i \in \mathcal{X}_2$ represent the inputs for two specific modalities, and $y^i \in \mathcal{Y}$ denotes the class labels from a predefined set of $C$ land-cover classes. In the case of MS-PAN data, we use images as inputs for both modalities, while for HSI datasets, local patches centered around pixel locations are considered to capture both spectral and spatial information jointly. The teacher model is a complex model trained with privileged information.

Specifically, the teacher network $\mathcal{T}$ comprises two modality-specific feature extractors $f_1^T(\cdot)$ and $f_2^T(\cdot)$ (referred to as featurenets), and a classification model $f_C^T$, which takes the concatenated feature representations ($\tilde{x}^i = [f_1^T(x_1^i), f_2^T(x_2^i)]$) obtained from the featurenets as inputs and maps them to the label space. Both feature extractors are implemented as convolutional networks, while the classification module is defined as a feed-forward neural network model. Classification is performed using softmax cross-entropy loss, given by:

\begin{equation}
\mathcal{L}_T = - \mathbb{E}_{(x_1^i, x_2^i, y^i) \in \mathcal{X}} [y^i \log{f_C^T(\tilde{x}^i})]
\end{equation}

The diagram representing the teacher network is depicted in Figure \ref{fig:TN}.

\subsubsection{Modality hallucination using C-GAN}

In our experimental setup, the modality $\mathcal{X}_2$ is assumed to be unavailable during inference, while $\mathcal{X}_1$ is considered the available modality. Hence, our hallucination model $\mathcal{H}$ is designed to generate the hallucinated features $f_2^T(x_2^i)$ given the input $f_1^T(x_1^i)$. It is important to note that our focus is on hallucinating the features learned by $\mathcal{T}$ rather than the actual data itself, considering the discriminative nature of the features learned by $f_2^T(\cdot)$. The C-GAN model consists of a feature generator ($G$) or the hallucination stream and a discriminator ($D$) network. The feature generator is conditioned on the samples from $f_1^T(\mathcal{X}_1)$ and the label vector $\mathcal{Y}$.

Furthermore, we introduce two intuitive modifications in the C-GAN architecture to avoid potential trivial solutions and ensure the generation of more discriminative hallucinated features that are comparable to the features learned by $\mathcal{T}$. These modifications are as follows:

\begin{itemize}
    \item We consider samples from $\mathcal{X}$ as well as augmented samples obtained by adding random Gaussian noise $z \in \mathcal{N}(0,1)$ to $f_1^T(\mathcal{X}_1)$. It is important to note that for a given $(x_1^i, x_2^i,y^i)$, while perturbing $f_1^T(x_1^i)$ to generate a cloud of feature points surrounding $f_1^T(x_1^i)$, we do not modify $f_2^T(x_2^i)$ during hallucination to ensure the robustness of the cross-modality mapping.
    \item The discriminator $D$ is designed to output $2C$ class scores, considering both \textit{real} and \textit{fake} samples on a per-class basis. This approach helps prevent the generation of potentially spurious samples for modality $f_2^T(\mathcal{X}_2)$. Specifically, the label outputs of $D$ (denoted as $\Bar{\mathcal{Y}}$) are encoded as vectors of length $2C$, where the $j^{th}$ and $(j+1)^{th}$ indices are labeled as $1$ each based on whether the input sample is the feature descriptor corresponding to a real sample from the $j^{th}$ class in $\mathcal{Y}$ category from $\mathcal{X}_2$ or it is a hallucinated (fake) sample.
\end{itemize}

Let $\mathcal{\Bar{X}} = (\Bar{x}_{1}^{j}, x_{2}^{j}, \Bar{y}^{j})_{j=1}^{M}$, where $\Bar{x}_1^j$ represents either a sample obtained from $\mathcal{X}$ (referred to as $x_1^j$) or a perturbed version of $x_1^j$ ($x_1^j + z$), and $\Bar{y}^j \in \Bar{\mathcal{Y}}$. These training samples are used to train $\mathcal{H}$. To this end, both $G$ and $D$ are trained using the adversarial min-max strategy, optimizing $\mathcal{L}_{hal}$ as stated in equation 2.

\begin{equation}
\min\limits_{G}\max\limits_{D}\mathcal{L}_{hal} = \mathbb{E}_{(\Bar{x}_{1}^{j}, \Bar{y}^{j}, y^{j})} [\log{D(G(f^{T}_1(\Bar{x}_{1}^{j}}|{y}^j, \Bar{y}^{j})))]\\ + \mathbb{E}_{(x_{2}^{j}, \Bar{y}^{j}, y^{j})}[\log{D(f^{T}_2(x_{2}^{j}), \Bar{y}^{j}})]
\end{equation}

where $|$ represents the concatenation operation. The schematic of modality hallucination using C-GAN is illustrated in Figure \ref{fig:Hal}.\subsubsection{Training the student network}

In the student model $\mathcal{S}$, the featurenet $f_1^T(\cdot)$ for the available modality $\mathcal{X}_1$ from the teacher $\mathcal{T}$ and the trained generator of the hallucination network $\mathcal{H}$ are kept unchanged, while only the student's classifier is trained. As the modality $\mathcal{X}_2$ is not available during testing, the student network $\mathcal{S}$ is trained using only the available modality $\mathcal{X}_1$. The process involves sending $\Bar{x}_1^j$ to featurenet $f_1^T(\cdot)$ to obtain the feature representation $f_1^T(\Bar{x}_1^j)$, which is then passed to the generator $G$ to generate the hallucinated features $G(f_1^T(\Bar{x}_1^j))$ for the missing modality. The concatenated representation $\hat{x}^j = [f_1^T(\Bar{x}_1^j), G(f_1^T(\Bar{x}_1^j))]$ is used to train the student's classifier $f_C^S$. To ensure that the predictions of $f_C^S$ align closely with the predictions of $f_C^T$ on similar samples, a cross-entropy based classification loss and a knowledge distillation loss between the teacher and the student are jointly optimized:

\begin{equation}
\mathcal{L}_S = - \mathbb{E}_{(\hat{x}^j,y^j)} [y^j \log{f_C^S({\hat{x}^j}})] + \lambda (||\textbf{q}_{T}^{j} - \textbf{q}_{S}^{j}||_{2})
\end{equation}

where $\textbf{q}_{T}^{j}$ and $\textbf{q}_{S}^{j}$ represent the softmax probability vectors of size $C \times 1$ (where $C$ is the number of classes) for the $j^{th}$ sample from the teacher and student networks, respectively. In the softmax formulation, a high temperature value is used to normalize the probabilities. Specifically, the temperature-based softmax normalization is defined as:

\begin{equation}
{q}_{c}^{j} = \frac{e^{z_{c}^{j}/T}}{\sum_{c=1}^{C}e^{z_{c}^{j}/T}}
\end{equation}

where ${q}_{c}^{j}$ represents the softened softmax probability for the $c^{th}$ class, and $z_{c}^{j}$ denotes the logit scores for the $c^{th}$ class given $\hat{x}^{j}$.

The hyperparameter $\lambda$ is determined empirically, as it determines the relative importance of the classification term, considering that the teacher network may also make misclassifications. The temperature value is initially set to a higher value ($>1$) during the training of both the teacher and student networks. However, during the testing phase of the student network, the temperature value is set back to $1$ (Figure \ref{fig:SN}).

\subsubsection{Inference}

During the inference phase, for a given test sample $x$ in modality-1, we utilize the featurenet $f_1^T$ and the hallucination generator $G$ to obtain the feature representations $f_1^T(x)$ and $G(f_1^T(x))$, respectively. These features are concatenated and fed into $f_C^S$ to obtain the predicted class label for the sample.

\subsection{Datasets and Experiments}

\subsubsection{Datasets}

In our study, we utilize two benchmark hyperspectral imaging (HSI) datasets and a multimodal remote sensing (RS) dataset that includes a large number of multispectral-panchromatic (MS-PAN) image pairs. The HSI datasets are employed for pixel classification tasks, leveraging both spatial and spectral information, while the MS-PAN dataset is used for scene recognition.

\begin{figure}[ht]
\begin{center}
\includegraphics[width=0.68\linewidth]{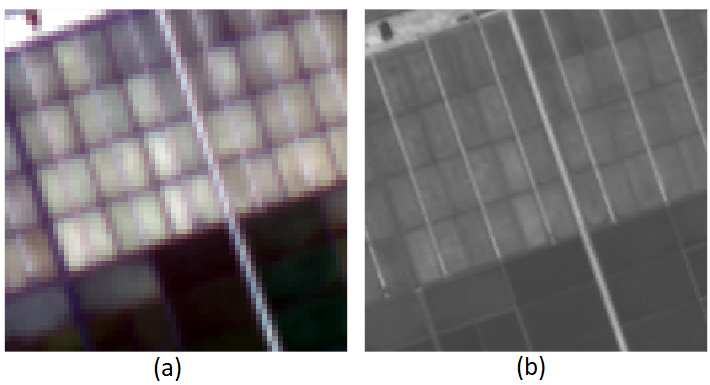}
\end{center}
\caption{MSPAN dataset: (a) Colour composite of three bands for MS image. (b) PAN image.}
\label{fig:mspan}
\end{figure}

\noindent\textbf{Multispectral-panchromatic dataset}: This dataset consists of multispectral (MS) imagery with 4 bands and panchromatic (PAN) imagery captured over a specific geographic area. It contains a total of 80,000 image pairs, representing eight land-cover classes. The MS imagery is collected from multispectral sensors, and the PAN imagery is obtained from a panchromatic sensor on the GF-1 satellite \citep{li2018learning}. Each MS image has a size of $64 \times 64 \times 4$ with a spatial resolution of 2 meters, while the corresponding PAN imagery has a size of $256 \times 256$ pixels with a spatial resolution of 8 meters. Figure \ref{fig:mspan} illustrates a color composite of an MS sample and its corresponding panchromatic image.

\begin{figure}[ht]
\begin{center}
\includegraphics[width=11cm]{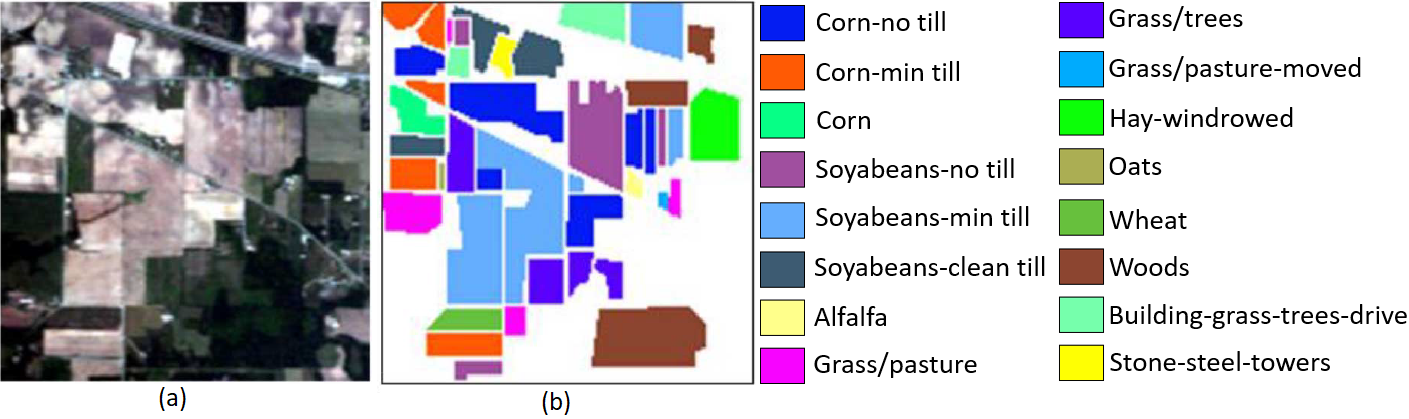}
\end{center}
\caption{Indian pines hyperspectral dataset: (a) Colour composite of three bands from red, green and blue wavelengths. (b) Groundtruth image.}
\label{fig:ip_dataset}
\end{figure}

\noindent\textbf{Indian Pines dataset}: The Indian Pines hyperspectral dataset \citep{zhang2015active} was acquired by the AVIRIS sensor over Northwestern Indiana. It consists of 200 bands, each with a size of $145 \times 145$ pixels and a spatial resolution of 20 meters. The dataset encompasses pixels from sixteen predefined land-cover classes. In total, there are 10,249 pixels in the scene, each with an associated ground-truth label (Figure \ref{fig:ip_dataset} \citep{zhang2015active}).

\begin{figure}[ht]
\begin{center}
\includegraphics[width=1.0\textwidth]{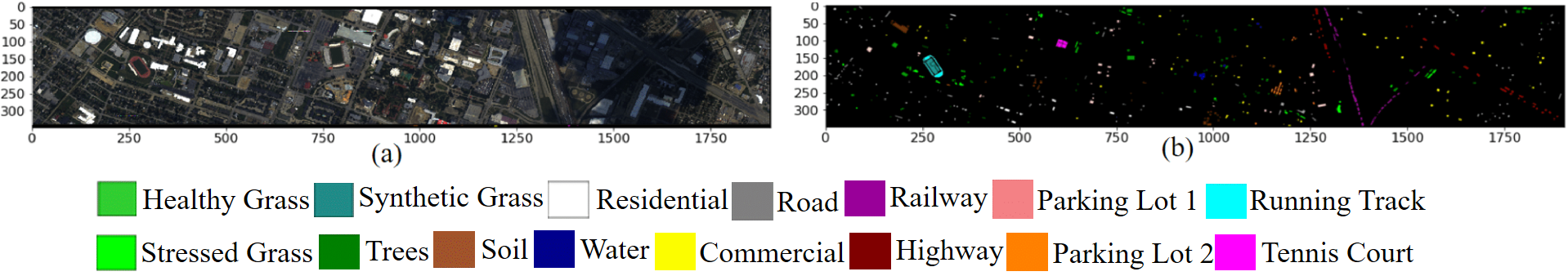}
\end{center}
\caption{Houston hyperspectral dataset: (a) Colour composite of three bands from red, green and blue wavelengths. (b) Groundtruth with classes.}
\label{fig:hous_dataset}
\end{figure}

\noindent\textbf{Houston 2013 dataset}: The Houston hyperspectral dataset \citep{debes2014hyperspectral} is a collection of imagery captured by the National Center for Airborne Laser Mapping over the University of Houston campus and its surrounding urban areas. The dataset consists of 144 bands, each with a size of $1905 \times 349$ pixels and a spatial resolution of 2.5 meters. There are a total of 15 land-use/land-cover classes, and the dataset contains 15,029 pixel vectors with associated ground truth classes (Figure \ref{fig:hous_dataset}).

\subsubsection{Model Architectures}

In this section, we provide detailed information about the experimental protocols and the model architectures for $\mathcal{T}$, $\mathcal{H}$, and $\mathcal{S}$, respectively, for each dataset.
\begin{table}
\begin{center}{\scriptsize
\begin{tabular}{|l|c|}
\hline
Network & Accuracy (in \%)\\
\hline\hline
Stream MS & 90.17 \\
Stream PAN & 92.58 \\
Teacher& 95.46\\
Teacher (avg) & 98.07\\

&\\
Student with MS absent \citep{garcia2018learning} & 81.06 \\
Student with PAN absent \citep{garcia2018learning} & 81.06 \\

&\\
Student with MS absent \citep{Garcia_2018_ECCV} & 37.24 \\
Student with PAN absent \citep{Garcia_2018_ECCV} & 37.35 \\

&\\
Student with MS absent (proposed) & 82.75 \\
Student with PAN absent (proposed) & 86.40 \\
\hline
\end{tabular}}
\end{center}
\caption{Results on MS-PAN dataset.}\label{tab:MSPAN}
\end{table}
\noindent\textbf{Teacher model}: The teacher network consists of two parallel featurenets implemented using CNN architectures, which include convolutional, pooling, and fully-connected layers. In the MS-PAN dataset, the MS images are input to featurenet-1, while the PAN images are input to featurenet-2. Each MS image has a size of $64 \times 64 \times 4$, and each PAN image has a size of $256 \times 256$. Therefore, the CNN architecture for the MS images consists of 3 layers of convolutions and pooling, while the architecture for the PAN images comprises 4 convolutional and pooling layers, as the latter requires further size reduction. For both CNN encoders, the first layer utilizes 128 filters of size $5 \times 5$, the second layer uses 128 filters of size $5 \times 5 \times 128$, and the third layer uses 64 filters of size $5 \times 5 \times 128$. The fourth layer of featurenet-2 consists of 64 filters with a size of $5 \times 5 \times 64$. Both layers employ ``same padding'' and ``max-pooling'', followed by batch normalization, dropout with a rate of $10\%$, and ReLU activation function. The strides in both convolutional filters and pooling kernels are set to 1.
\begin{table}
\begin{center}{\scriptsize
\begin{tabular}{|l|c|}
\hline
Network & Accuracy (in \%)\\
\hline\hline
Stream 1 & 73.53 \\
Stream 2 & 54.91 \\
Teacher & 70.28 \\
Student (proposed) & 80.57 \\
\hline
\end{tabular}}
\end{center}
\caption{Results on Indian Pines dataset.}\label{table:IP}
\end{table}
In the hyperspectral datasets, each pixel is represented as a $1 \times 1 \times B$ pixel vector, where $B$ is the number of bands. Consequently, a patch of size $17 \times 17$ is created centered around each pixel location. Then, patches from each modality are fed into their corresponding CNN encoders for training. The CNN encoders share the same architecture as those used for the MS-PAN dataset up to the third layer.

The output of each CNN feature extractor is flattened and passed through a two-layer fully connected (FC) encoder. The first layer has 512 nodes, and the second layer has 200 nodes, reducing the dimension of the features to 200. The outputs from the FC encoder are concatenated and fed into a three-layer FC neural network classifier $f_{C}^{T}(.)$, which consists of 400 and 200 nodes in the first and second layers, respectively, followed by a softmax layer with C nodes, where C is the number of classes. The classifier is trained using categorical cross-entropy loss.

\noindent\textbf{Hallucination model}: In the hallucination model based on C-GAN, both the generator and discriminator are fully connected neural networks with three layers, and the discriminator has a softmax layer at the end. However, the softmax layer of the discriminator has 2C nodes to account for both real and fake samples per class. Both networks utilize batch normalization, dropout with a rate of 10\%, and ReLU activation.

\noindent\textbf{Student model}: The student model comprises the featurenet corresponding to the available modality, the trained generator (with fixed weights), and an FC multilayer classifier. The classifier is a three-layer fully connected neural network $f_{C}^{S}(.)$ that replicates the architecture of $f_{C}^{T}(.)$ and is trained using the extracted features from the available modality and the generated features for the absent modality.

All models are trained using the Adam optimizer \citep{kingma2014adam} with a learning rate of 0.001.

\subsection{Results and Discussion}

We evaluate the performance of the student's classification module using two baselines. First, we train separate classification networks for each modality and report their classification performances. Second, we evaluate the performance of the teacher model where the modality-specific features are fused to train the teacher's classifier. For baseline comparison, we randomly select 25\% of the samples to train the student-teacher model, and the remaining 75\% of the samples are used for testing. In the case of hyperspectral imagery (HSI), we randomly select two sets of non-overlapping bands to constitute modality-1 and modality-2.
\begin{table}
\begin{center}{\scriptsize
\caption{Results on Houston dataset.}\label{table:Hous}
\begin{tabular}{|l|c|}
\hline
Network & Accuracy (in \%)\\
\hline\hline
Stream 1 & 95.28 \\
Stream 2 & 91.80 \\
Teacher & 98.17 \\
Student (proposed) & 97.96 \\
\hline
\end{tabular}}
\end{center}
\end{table}
In addition to the baselines, we compare our model to hallucination techniques inspired by \citep{garcia2018learning} (where hallucination is performed using an encoder) and \citep{Garcia_2018_ECCV} (where softmax probabilities from two streams are averaged before classification). These techniques have been state-of-the-art in the field of modality hallucination. The results of the baseline comparison for the MS-PAN dataset, Indian Pines dataset, and Houston dataset are presented in Tables \ref{tab:MSPAN}, \ref{table:IP}, and \ref{table:Hous}, respectively.

We also conduct sensitivity analysis on the Indian Pines dataset by varying the ratio of bands in each modality and the temperature. Additionally, we compare our C-GAN model, trained with a discriminator on 2C classes, against C-GANs with a binary discriminator and a discriminator with C+1 classes.

\textbf{MS-PAN dataset}: When considering the MS-PAN dataset, we observe that the accuracies of the MS stream (90.17\%) and PAN stream (92.58\%) are comparable, while the accuracy of the teacher model (95.46\%) surpasses both. This result was expected. However, for the student model, the accuracies obtained for hallucination of PAN imagery (86.40\%) and MS imagery (82.75\%) are lower than both streams and the teacher network. Similar trends are observed in \citep{garcia2018learning}, where hallucination of PAN imagery from MS and vice versa resulted in accuracies of 81.06\% each. These observations suggest that the MS and PAN bands do not share much correlated information, and the features from one modality do not efficiently generate features of the other modality. 

\begin{figure}[t]
\begin{center}
\includegraphics[width=0.77\linewidth]{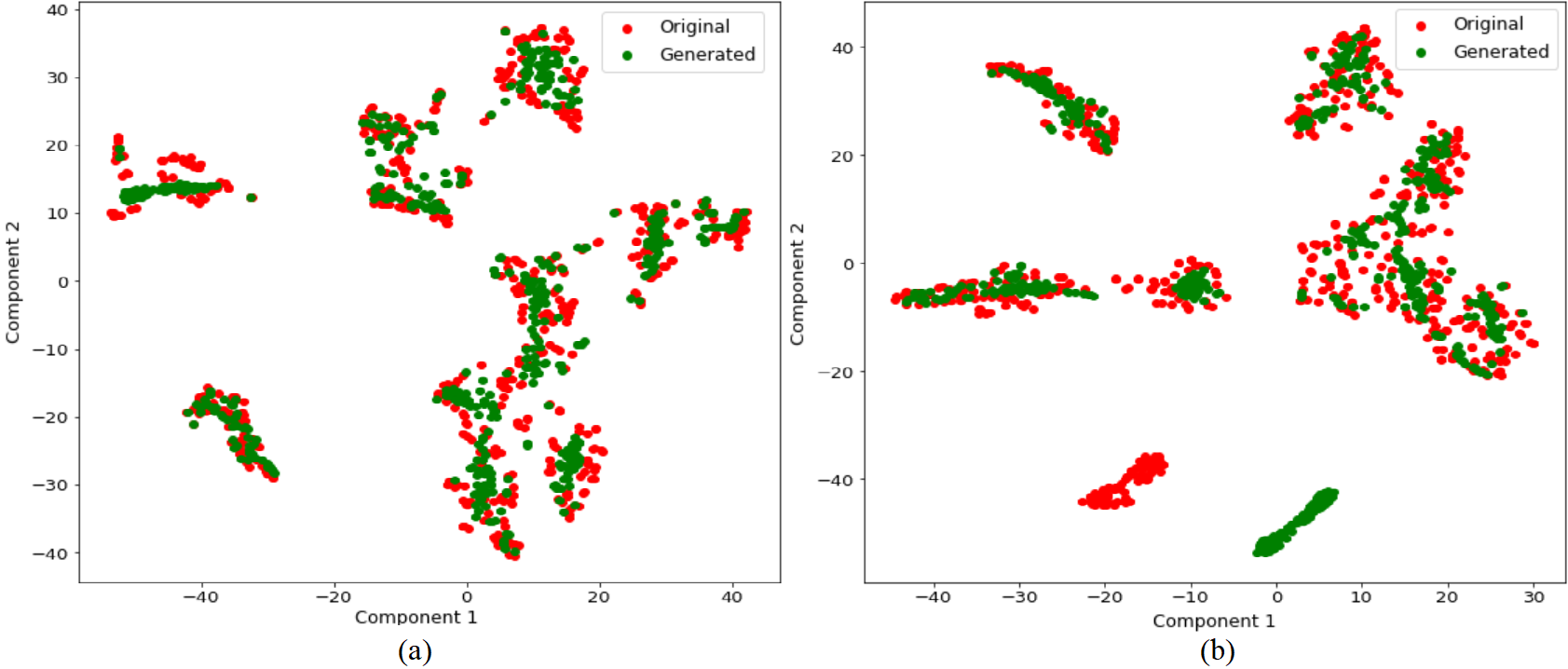}
\end{center}
\caption{t-SNE comparing the original and encoder generated features for (a) MS imagery (b) PAN imagery.}
\label{fig:tSNE_ms_pan}
\end{figure}

Furthermore, when using the technique based on \citep{Garcia_2018_ECCV}, we observe that although the teacher model achieves an accuracy as high as 98.07\%, the student networks achieve accuracies as low as 37.24\% (when hallucinating MS) and 37.35\% (when hallucinating PAN). This suggests that the CNN weights obtained through averaging do not extract informative features that can be used to generate features of the absent modality. Figure \ref{fig:tSNE_ms_pan} show t-SNE plots comparing generated PAN features from available MS features and vice versa through the hallucination framework based on \citep{garcia2018learning}. The t-SNE plots indicate that the generated features of the absent modality and the original features of the same modality effectively overlap, demonstrating efficient hallucination.

\textbf{Indian Pines dataset}: In the case of the Indian Pines dataset, the accuracy of the teacher model (70.28\%) is lower than that of stream 1 (73.53\%) by 3.25\%. Additionally, the accuracy for stream 2 (54.91\%) is much lower than stream 1. This suggests that the bands in modality 2 contain irrelevant information, which hinders the performance of the teacher model. On the other hand, the accuracy of the student model (80.57\%) surpasses both single streams and the teacher model. This indicates that the generated features from the hallucination model were able to fully capture the distribution of the original features and overcome the effects of band correlation, leading to better results.

Figure \ref{fig:BandSplit} presents the sensitivity analysis results for varying the percentage of bands included in the first and second modalities for the Indian Pines dataset. For every instance, except the 75\%-25\% split, the student model outperforms the teacher model. This suggests that even with a smaller number of available bands, the hallucination model is able to generate better features.

\begin{figure}[ht]
\begin{center}
\includegraphics[width=6cm]{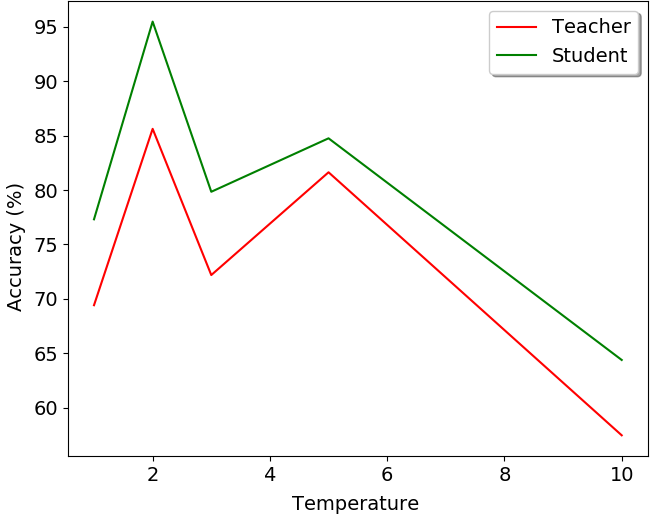}
\end{center}
\caption{Variation in test accuracy with change in temperature conducted on Indian pines dataset.}
\label{fig:tg}
\end{figure}

\begin{figure}[ht]
\begin{center}
\includegraphics[width=6cm]{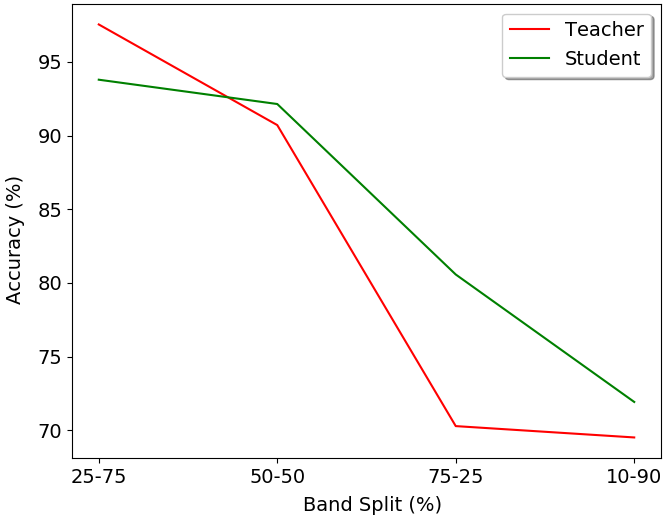}
\end{center}
\caption{Variation in test accuracy with change in modality split ratios conducted on Indian pines dataset.}
\label{fig:BandSplit}
\end{figure}

Figure \ref{fig:tg} depicts the variation in accuracies of $\mathcal{T}$ and $\mathcal{S}$ with respect to changes in temperature, while keeping the train-test split fixed at 25\%-75\% and the band split at 50\%-50\%. The maximum accuracy is achieved when the temperature is set to 2.

\textbf{Houston dataset}: In the case of the Houston dataset, we observe that the accuracy of $\mathcal{T}$ (98.17\%) outperforms both stream 1 (95.28\%) and stream 2 (91.80\%), which aligns with our expectations. The accuracy of $\mathcal{S}$ (97.96\%) also surpasses the accuracies obtained from both streams and only lags behind $\mathcal{T}$ by a difference of 0.21\%. This suggests that $\mathcal{H}$ is able to generate the absent modality with higher accuracy, allowing the student classifier to learn efficient representations.
We do not compare the performance on HSI datasets to \citep{garcia2018learning} and \citep{Garcia_2018_ECCV} because in those works, the student network either surpasses the teacher network or lags behind by a very small margin, demonstrating the effectiveness of hallucination of the absent modality.

\textbf{Choice of discriminator architecture}: We opt for a discriminator that is trained on two sets of classes (real and fake) based on the intuition that training on two sets would make the model more sensitive to identifying interclass and intraclass variances and reduce the misclassification rate among the generated features. This represents an improvement over using a binary discriminator, which fails to capture the intraclass variance. Furthermore, our discriminator outperforms the one trained on $C+1$ classes, as the latter is more prone to misclassifying the generated features. Models employing binary and $C+1$ class discriminators yielded classification accuracies as low as 30\%, further supporting our rationale.

\section{Cross-modal attention for hyperspectral LiDAR classification}

\label{sec:FAN}
In this section, we explore the technique of combining HSI and LiDAR data through an attention-based approach. We utilize the concept of ``self-attention'' to extract spectral features from the HSI data. Additionally, we propose a ``cross-attention'' mechanism that incorporates the LiDAR modality to create an attention mask, emphasizing the spatial characteristics of the HSI data (Figure \ref{fig:self_cross}). By integrating spectral and spatial features, we obtain an intermediate representation, which is further enhanced through self-attention learning. This refined representation serves as a comprehensive input for the classification process\footnote{Published as: Satyam Mohla, Shivam Pande, Biplab Banerjee, Subhasis Chaudhuri. \href{https://openaccess.thecvf.com/content_CVPRW_2020/papers/w6/Mohla_FusAtNet_Dual_Attention_Based_SpectroSpatial_Multimodal_Fusion_Network_for_Hyperspectral_CVPRW_2020_paper.pdf}{FusAtNet: Dual Attention based SpectroSpatial Multimodal Fusion Network for Hyperspectral and LiDAR Classification}, In: Proceedings of the IEEE/CVF Conference on Computer Vision and Pattern Recognition Workshops, Jun. 2020 pp. 92–93.}.

\subsection{Methodology}
The aim of this study is to perform classification at the pixel level by utilizing the combined spectral and spatial information present in hyperspectral images (HSIs) and the depth and intensity information encoded in LiDAR data.

To achieve this objective, we work with patches from HSI and LiDAR, denoted as $\mathcal{X} = \{\textbf{x}^i_H, \textbf{x}^i_L\}_{i=1}^n$, which are centered around the ground truth pixels $\mathcal{Y} = \{y^i\}_{i=1}^n$. Here, $\textbf{x}^i_H \in \mathbb{R}^{M\times N \times B_1}$ and $\textbf{x}^i_L \in \mathbb{R}^{M\times N \times B_2}$, where $B_1$ and $B_2$ represent the number of channels in the HSI and LiDAR modalities, respectively. The variable $n$ denotes the number of available ground truth samples. The ground truth labels are denoted as $y_i^n \in \{1, 2,..., K\}$, where $K$ represents the number of ground truth classes. These patches are fed into the proposed FusAtNet model, which processes them as they pass through various modules discussed in Section \ref{subsec:MA}.

\subsubsection{Overview of the model}
\begin{figure*}[!ht]
  \centering
  \centerline{\includegraphics[width=14cm]{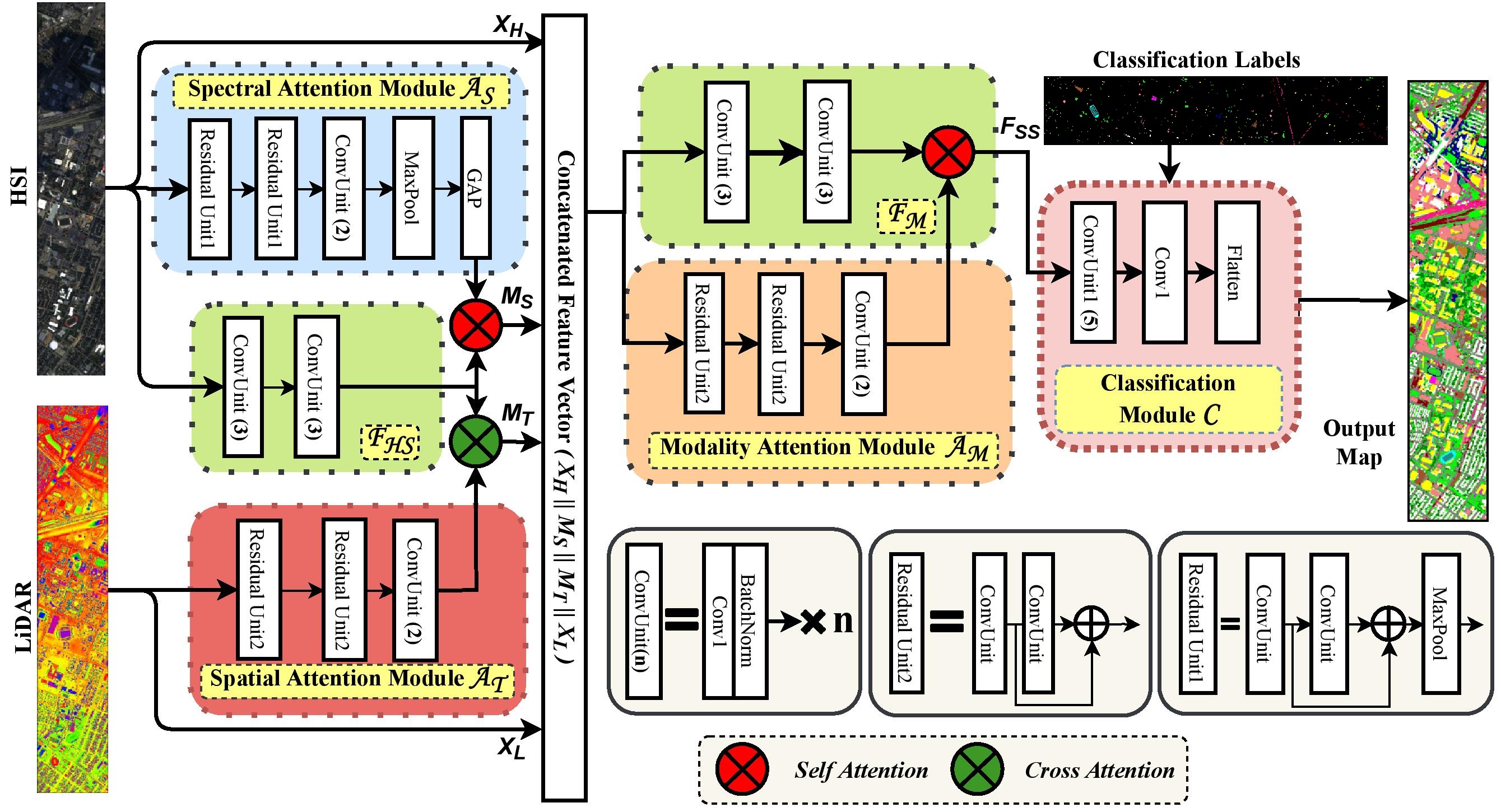}}
\caption[Schematic of FusAtNet method for fusion of HSI-LiDAR images]{Diagram illustrating the FusAtNet model (applied to the Houston dataset). Initially, the hyperspectral training samples $X_H$ are passed through the feature extractor $\mathcal{F_{HS}}$ to obtain latent representations, and the spectral attention module $\mathcal{A_S}$ generates a spectral attention mask. Concurrently, the corresponding LiDAR training samples $X_L$ are processed by the spatial attention module $\mathcal{A_T}$ to obtain the spatial attention mask. The attention masks are then applied individually to the latent HSI representations, resulting in $M_S$ and $M_T$. $M_S$ and $M_T$ are subsequently concatenated with $X_H$ and $X_L$, which are then fed into the modality feature extractor $\mathcal{F_M}$ and the modality attention module $\mathcal{A_M}$. The outputs from these modules are multiplied to produce $F_{SS}$, which is further passed to the classification module $\mathcal{C}$ for pixel-level classification.}\medskip
  \vspace{-0.5cm}

\label{fig:Diagram}
\end{figure*}

The purpose of this research is to synergistically investigate the spectral-spatial properties of hyperspectral imaging (HSI) and the spatial/elevation characteristics of LiDAR using the ``cross-attention'' framework. The attention modules play a role in selectively highlighting the significant areas in the extracted hyperspectral features to enhance the interclass variance and improve classification accuracy. This is accomplished in two steps: Firstly, the HSI features are processed through a feature extractor and spectral attention module, which combine to emphasize the spectral information in the HSI features. Simultaneously, the LiDAR features undergo spatial attention processing, resulting in a mask that accentuates the spatial characteristics of the HSI. Secondly, the highlighted features are reinforced with the original features and passed through modality extraction and modality attention modules. The outputs of these modules are combined to strategically emphasize the important sections of both modalities. The resulting features are then fed into the classification module.

\subsubsection{Network architecture}
\label{subsec:MA}
The complete architecture of FusAtNet is depicted in Figure \ref{fig:Diagram}, and all experiments follow this architecture. FusAtNet consists of six modules that are utilized in three phases. In the first phase, the hyperspectral feature extractor $\mathcal{F_{HS}}$, spectral attention module $\mathcal{A_S}$, and spatial attention module $\mathcal{A_T}$ are employed to jointly extract and emphasize the spatial-spectral features from the HSI. In the second phase, the modality feature extractor $\mathcal{F_M}$ and modality attention module $\mathcal{A_M}$ are employed to selectively highlight the modality-specific features. In the third phase, the modality-specific spectral-spatial features are fed into the classification module $\mathcal{C}$. All modules are CNN-based, with fixed kernel size of 3$\times$3 and ReLU activation function. The modules are discussed as follows:\\

\begin{figure*}[!t] 
  \centering
  \centerline{\includegraphics[width=14cm]{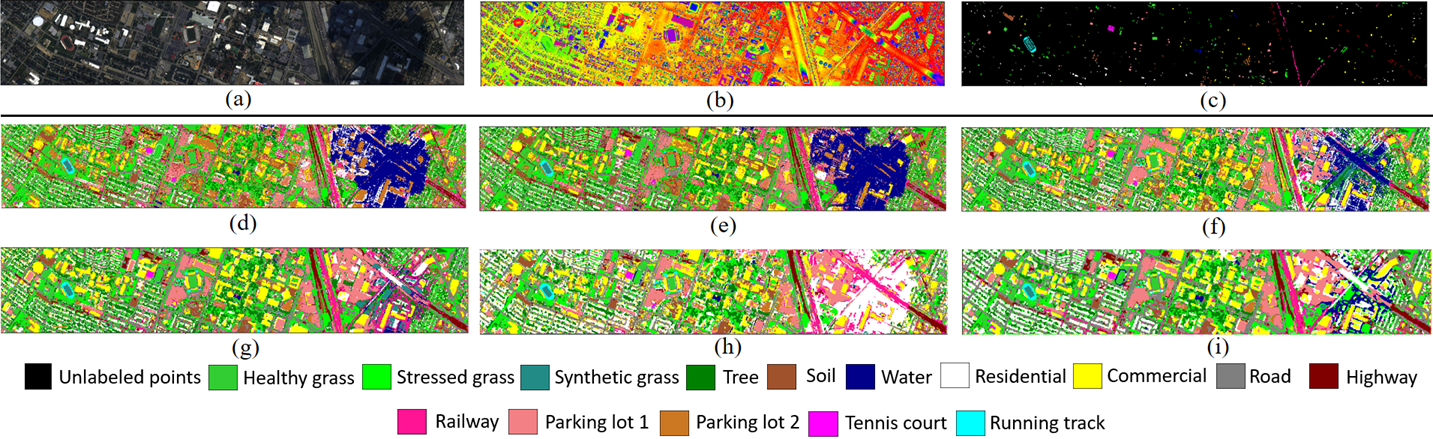}}
\caption[Houston 2013 HSI-LiDAR dataset with corresponding RGB image, groundtruth image and classification maps.]{Visualization of the Houston hyperspectral and LiDAR dataset with corresponding classification maps. (a) True color composite of HSI, (b) LiDAR image, (c) Ground truth. Classification maps from (d) SVM (H), (e) SVM (H+L), (f) Two-branch CNN (H), (g) Two-branch CNN (H+L), (h) FusAtNet (H), (i) FusAtNet (H+L).}\medskip
  \vspace{-0.5cm}
\label{fig:Houston}
\end{figure*}

\noindent\textbf{Hyperspectral feature extractor $\mathcal{F_{HS}}$}:
$\mathcal{F_{HS}}$ comprises a 6-layer CNN and is employed to extract spectral-spatial features from HSIs. The first five layers consist of 256 filters, while the sixth layer has 1024 filters. All convolution operations employ zero padding, and batch normalization is applied after each convolution operation. This module can be represented as $\mathcal{F_{HS}}(\theta_{F_{HS}}, \textbf{x}^i_H)$, where $\theta_{F}$ represents the module's weights. The output of $\mathcal{F_{HS}}$ is a patch of size 11$\times$11$\times$1024.\\

\noindent\textbf{Spectral attention module $\mathcal{A_S}$}:
$\mathcal{A_S}$ derives its attention mask from the HSI. This module is a CNN with 3 convolution blocks, each containing 2 convolution layers. Additionally, the first and second convolution blocks are followed by a residual block each. Maxpooling layers are applied after each residual block and the sixth convolution layer. The final layer of this module is a global average pooling (GAP) layer. The overall architecture of this module is inspired by \citep{mou2019learning}. The number of kernels in the first five convolution layers is 256, while the sixth layer has 1024 kernels, all utilizing zero padding. Batch normalization layers follow each convolution operation. The model is denoted as $\mathcal{A_{S}}(\theta_{A_{S}}, \textbf{x}^i_H)$, where $\theta_{A_{S}}$ represents the weights of this attention module. The output of this module is a vector of size 1$\times$1024, which is multiplied element-wise with the output of $\mathcal{F_{HS}}$ to obtain the highlighted spectral features as expressed in Eq. (\ref{eqn:SpecMask}).
\begin{equation}
M_S(\textbf{x}^i) =  \mathcal{F_{HS}}(\theta_{F_{HS}}, \textbf{x}_H^i) \otimes \mathcal{A_S}(\theta_{A_S}, \textbf{x}_H^i)
\label{eqn:SpecMask}
\end{equation}
Here, $M_S$ denotes the extracted features highlighted by the spectral attention mask, and $\otimes$ represents the broadcasted element-wise matrix multiplication operation, ensuring that the resulting product retains the size of the matrix with the higher dimension.
\begin{figure}[ht!]
  \centering
  \centerline{\includegraphics[width=14cm]{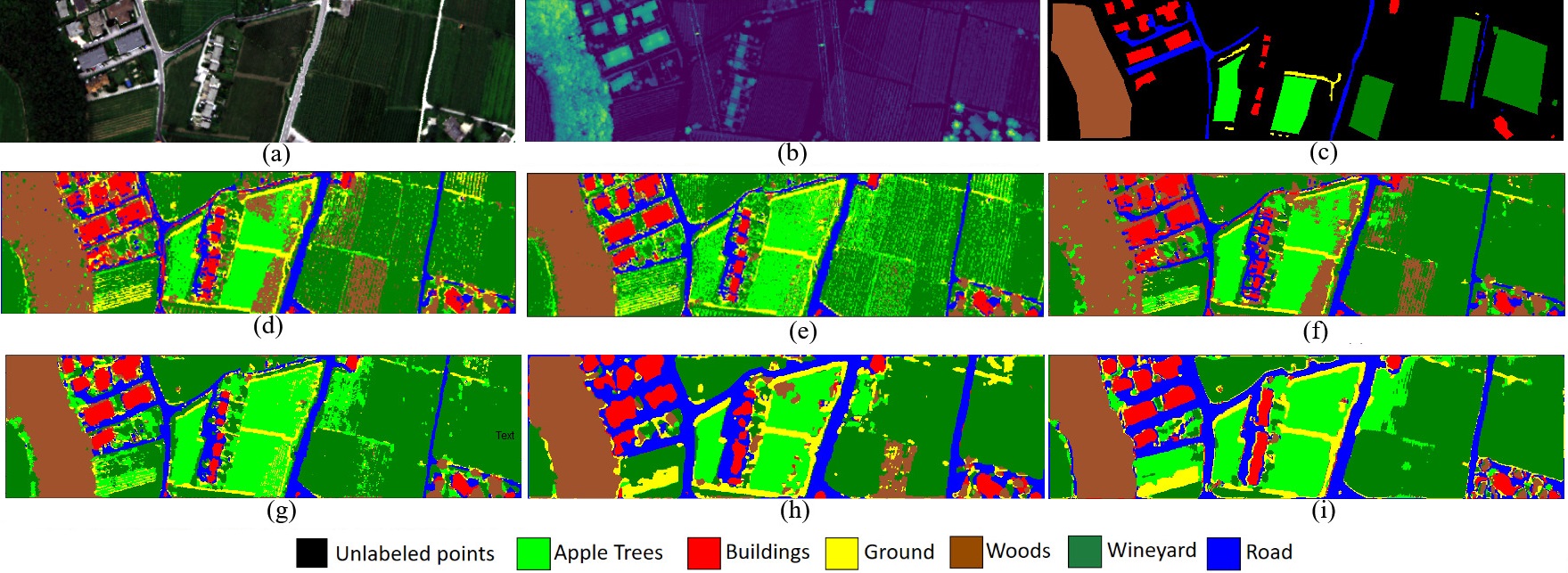}}
\caption[Trento HSI-LiDAR dataset with corresponding RGB image, groundtruth image and classification maps.]{Trento hyperspectral and lidar dataset with classification maps. (a) True colour composite of HSI, (b) LiDAR image, (c) Groundtruth. Classification maps from (d) SVM (H), (e) SVM (H+L), (f) Two-branch CNN (H), (g) Two-branch CNN (H+L), (h) FusAtNet (H), (i) FusAtNet (H+L).}\medskip
  \vspace{-0.5cm}
\label{fig:Trento}
\end{figure}

\noindent\textbf{Spatial attention module $\mathcal{A_T}$}:
The spatial attention module, denoted as $\mathcal{A_{T}}(\theta_{A_{T}}, \textbf{x}^i_L)$, generates an attention mask from the LiDAR modality. It consists of a 6-layer CNN with the first 3 layers containing 128 filters each, and the last three layers having 256 filters each. Two residual layers are present, one following the second convolution layer and the other following the fourth convolution layer. Batch normalization is applied after each convolution layer. The output of this module is a patch of size 11$\times$11$\times$1024, which is multiplied element-wise with the extracted features from $\mathcal{F_{HS}}$ to obtain spatially highlighted features $M_T$, as denoted in Eq. (\ref{eqn:SpatMask}):
\begin{equation}
M_T(\textbf{x}_H^i,\textbf{x}_L^i) =  \mathcal{F_{HS}}(\theta_{F_{HS}}, \textbf{x}_H^i) \otimes \mathcal{A_T}(\theta_{A_T}, \textbf{x}_L^i)
\label{eqn:SpatMask}
\end{equation}

\noindent\textbf{Modality feature extractor $\mathcal{F_M}$}:
The $\mathcal{F_M}$ module follows the same structure as $\mathcal{F_{HS}}$ and can be represented as $\mathcal{F_{M}}(\theta_{F_M}, \textbf{x}^i_H, \textbf{x}^i_L)$, where $\theta_{F_M}$ represents the module's weights. It takes as input the spectrally and spatially highlighted features $M_S$ and $M_T$, along with the original $\mathcal{X}$. The output is a patch of size 11$\times$11$\times$1024, represented as shown in Eq. (\ref{eqn:ModF}):
\begin{equation}
F_M(\textbf{x}_H^i,\textbf{x}_L^i) =  \mathcal{F_M}(\theta_{F_M}, \textbf{x}^i_H \oplus \textbf{x}^i_L \oplus M_S(\textbf{x}^i) \oplus
M_T(\textbf{x}_H^i,\textbf{x}_L^i))
\label{eqn:ModF}
\end{equation}
Here, $\oplus$ represents concatenation along the channel axis.
\begin{figure}[!h] 
  \centering
  \centerline{\includegraphics[width=14cm]{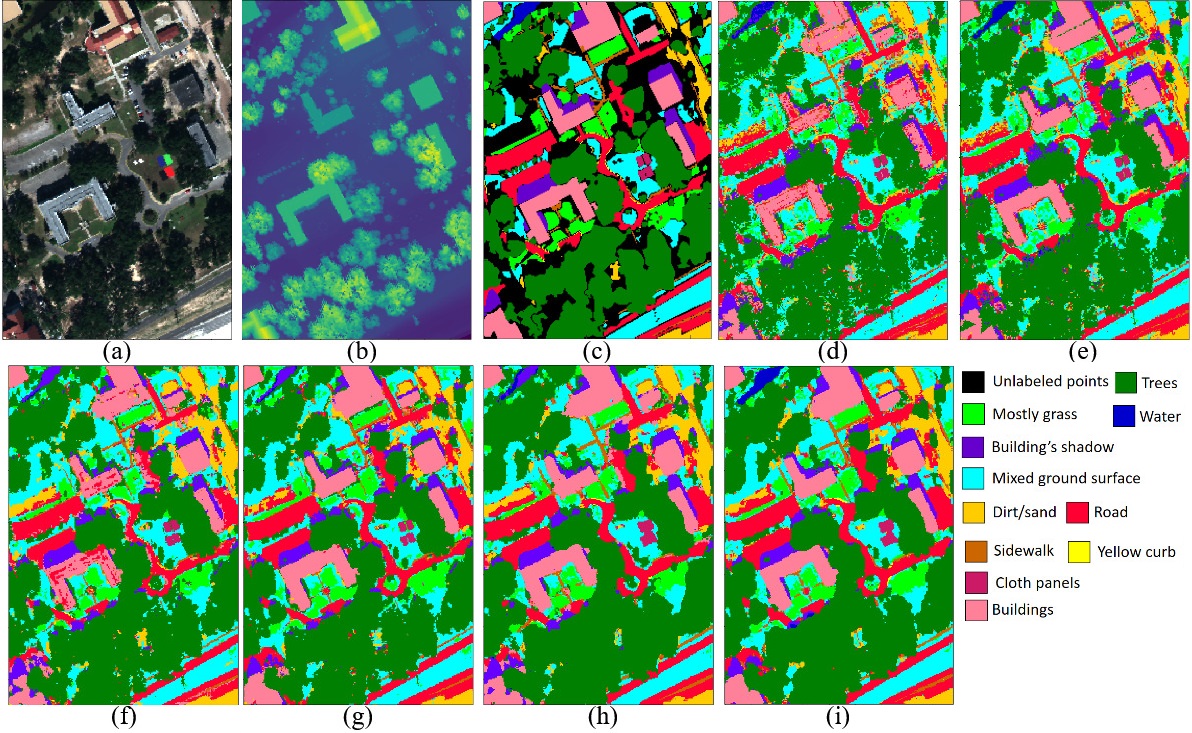}}
\caption[MUUFL Gulfport HSI-LiDAR dataset with corresponding RGB image, groundtruth image and classification maps.]{MUUFL hyperspectral and lidar dataset with classification maps.  (a) True colour composite of HSI, (b) LiDAR image, (c) Groundtruth. Classification maps from (d) SVM (H), (e) SVM (H+L), (f) Two-branch CNN (H), (g) Two-branch CNN (H+L), (h) FusAtNet (H), (i) FusAtNet (H+L).}\medskip
  \vspace{-0.7cm}
\label{fig:Muufl}
\end{figure}

\noindent\textbf{Modality attention module $\mathcal{A_M}$}:
The architecture of $\mathcal{A_M}$ is similar to that of $\mathcal{A_T}$ and is denoted as $\mathcal{A_{M}}(\theta_{A_M}, \textbf{x}^i_H, \textbf{x}^i_L)$, with $\theta_{A_M}$ representing the weights. This module generates an attention mask that focuses on specific traits of each modality. The input is the same as that of $\mathcal{F_M}$. This is represented in Eq. (\ref{eqn:ModA}):
\begin{equation}
A_M(\textbf{x}_H^i,\textbf{x}_L^i) = \mathcal{A_M}(\theta_{A_M}, \textbf{x}^i_H \oplus \textbf{x}^i_L \oplus M_S(\textbf{x}^i) \oplus \\
M_T(\textbf{x}_H^i,\textbf{x}_L^i))
\label{eqn:ModA}
\end{equation}
The output of the module is an 11$\times$11$\times$1024 patch that is element-wise multiplied with the output of $\mathcal{F_M}$, as shown in Eq. (\ref{eqn:FSS}), and the result is forwarded to the classification module.
\begin{equation}
F_{SS}(\textbf{x}_H^i,\textbf{x}_L^i) =  
F_M(\textbf{x}_H^i,\textbf{x}_L^i) \otimes A_M(\textbf{x}_H^i,\textbf{x}_L^i)
\label{eqn:FSS}
\end{equation}
Here, $F_{SS}$ represents the final spectral-spatial features.

\noindent\textbf{Classification module $\mathcal{C}$}:
The input to the $\mathcal{C}$ module consists of the final spectral-spatial features $F_{SS}(\textbf{x}_H^i,\textbf{x}_L^i)$. This module is a 6-layer fully convolutional neural network, where the first four layers have 256 filters each, while the fifth and sixth layers have 1024 and $K$ filters, respectively, where $K$ is the number of classes. The filter size for the last layer is set to 1$\times$1, and no padding is applied in any layer. ReLU activation function and batch normalization are applied to all layers except the last one, which is the softmax layer. The module can be defined as $\mathcal{C}(\theta_{C}, F_{SS}(\textbf{x}_H^i,\textbf{x}_L^i))$, where $\theta_{C}$ represents the classification weights. The output of $\mathcal{C}$ is a vector of size $1\times K$.
\begin{table}[!t]
\centering{\tiny
 \caption[Accuracy analysis on the Houston 2013 dataset (in \%).]{\label{tab:hous_perf} Accuracy analysis on the Houston 2013 dataset (in \%). `H' represents only HSI while `H+L' represents fused HSI and LiDAR.}
\begin{tabular}{|p{1.6cm}|p{0.7cm}|p{0.7cm}|p{0.8cm} |p{0.9cm}| p{0.8cm}|p{1.1cm} |p{1.1cm}| p{1.1cm}|p{0.9cm}|p{0.9cm}|}
 \hline
Classes &SVM (H) & SVM (H+L) & ELM (H) & ELM (H+L) & CNN-PPF (H) & CNN-PPF (H+L)  & Two Branch CNN (H) & Two Branch CNN (H+L) & FusAtNet (H) & FusAtNet (H+L)\\
 \hline
Healthy Grass &  81.86 &82.43& 82.91 &83.10 &82.24&\textbf{83.57} &83.38 &83.10 &83.00& 83.10\\
Stressed Grass &  82.61 &82.05& 83.93 &83.70 & \textbf{98.31}& 98.21 &84.21 & 84.10 &84.96& 96.05\\
Synthetic Grass &  99.80 & 99.80 & \textbf{100.00} & \textbf{100.00} & 70.69 & 98.42 & 99.60 & \textbf{100.00} &\textbf{100.00}& \textbf{100.00}\\
Trees & 92.50 & 92.80 & 91.76 & 91.86 & 94.98 & \textbf{97.73} & 93.18 & 93.09 &92.23& 93.09\\
Soil &  98.39 & 98.48 & 98.77 & 98.86 & 97.25 & 96.50 & 98.58 & \textbf{100.00} &97.06& 99.43\\
Water & 94.41 & 95.10 & 95.10 & 95.10 & 79.02 & 97.20 & 99.30 & 99.30 &\textbf{100.00}&\textbf{100.00}\\
Residential & 76.87 & 75.47 & 89.65 & 80.04 & 86.19 & 85.82 & 85.45 & 92.82 &\textbf{93.81}& 93.53\\
Commercial & 43.02 & 46.91 & 49.76 & 68.47 & 65.81 & 56.51 & 69.14 & 82.34 &76.35&\textbf{92.12}\\
Road & 79.04 & 77.53 & 81.11 & 84.80 & 72.11 & 71.20 & 78.66 & 84.70 &\textbf{85.15}& 83.63\\
Highway & 58.01 & 60.04 & 54.34 & 49.13 & 55.21 & 57.21 & 52.90 & \textbf{65.44} &62.64& 64.09\\
Railway & 81.59 & 81.02 & 74.67 & 80.27 & 85.01 & 80.55 & 82.16 & 88.24 &72.11&\textbf{ 90.13}\\
Parking Lot 1 & 72.91 & 85.49 & 69.07 & 79.06 & 60.23 & 62.82 & \textbf{92.51} & 89.53 &88.95& 91.93\\
Parking Lot 2 & 71.23 & 75.09 & 69.82 & 71.58 & 75.09 & 63.86 & 92.63 & 92.28 &\textbf{92.98}& 88.42\\
Tennis Court & 99.60 & \textbf{100.00} & 99.19 & 99.60 & 83.00 & \textbf{100.00} & 94.33 & 96.76 &\textbf{100.00}& \textbf{100.00}\\
Running Track & 97.67 & 98.31 & 98.52 & 98.52 & 52.64 & 98.10 & 99.79 & 99.79 &\textbf{100.00}& 99.15\\
 \hline
OA & 79.00 & 80.49 & 79.87 & 81.92 & 78.35 & 83.33 & 84.08 & 87.98 &85.72& \textbf{89.98}\\
AA & 81.94 & 83.37 & 82.57 & 84.27 & 77.19 & 83.21 & 86.98 & 90.11 &88.62& \textbf{94.65}\\
$\kappa$ & 0.7741 & 0.7898 & 0.7821 & 0.8045 & 0.7646 & 0.8188 & 0.8274 & 0.8698 &0.8450& \textbf{0.8913}\\
 \hline
\end{tabular}}
\end{table}
\subsubsection{Training and inference}
The output from $\mathcal{C}$ undergoes categorical cross-entropy loss, which is utilized for backpropagation to train the FusAtNet model in an end-to-end manner (see Eq. (\ref{eqn:clf})).
\begin{equation}
\mathcal{L}_C = - \mathbb{E}_{(\textbf{x}_H^i,\textbf{x}_L^i,y^i)} [y^i\log{\mathcal{C}(\theta_{C},F_{SS}(\textbf{x}_H^i,\textbf{x}_L^i}))]
\label{eqn:clf}
\end{equation}
Here, $\mathcal{L}_C$ represents the classification loss.

During the testing phase, the given test sample $(\textbf{x}^j_H, \textbf{x}^j_L)$ is passed through the fusion module and follows the same path as the training samples. The resulting output $F_{SS}(\textbf{x}^j_H, \textbf{x}^j_L)$ is forwarded to the classification module $\mathcal{C}$, where it is assigned the predicted class label.

\subsection{Experimental setup}

This section presents the details of the datasets used to evaluate FusAtNet and the protocols followed during the training process.

\subsubsection{Datasets}
To assess the effectiveness of our method, we utilized three HSI-LiDAR datasets.\\

\noindent\textbf{Houston dataset}: This dataset comprises a hyperspectral imagery and a LiDAR depth raster, which was initially introduced in the GRSS Data Fusion Contest 2013. The dataset was acquired over the Houston university campus and its surroundings by the National Airborne Centre for Laser Mapping (NCALM). The HSI consists of 144 hyperspectral bands with wavelengths ranging from 0.38 $\mu$m to 1.05 $\mu$m. Each raster has a size of 349$\times$1905 and a spatial resolution of 2.5 m. A total of 15,029 ground truth samples are available, distributed across 15 classes. The dataset is divided into training and testing sets, containing 2,832 and 12,197 pixels, respectively \citep{xu2017multisource}. However, for our experiments, we considered 12,189 pixels in the test set, as some pixels were excluded due to data preprocessing issues. The dataset can be visualized in Figure \ref{fig:Houston}.\\

\noindent\textbf{Trento dataset}: This dataset was collected using the AISA Eagle sensor over rural regions in Trento, Italy. The HSI image in the Trento dataset consists of 63 bands with wavelengths ranging from 0.42 $\mu$m to 0.99 $\mu$m, while the LiDAR data comprises 2 rasters representing elevation information. Each band has dimensions of 166 $\times$ 600, with a spatial resolution of 9.2 nm and a spectral resolution of 1.0 m. The dataset contains a total of 6 classes, with ground truth available for 30,214 pixels. These pixels are further divided into 819 training pixels and 29,395 test pixels \citep{xu2017multisource}. The dataset can be observed in Figure \ref{fig:Trento}.\\

\noindent\textbf{MUUFL Gulfport dataset}: This dataset was acquired over the campus of the University of Southern Mississippi Gulf Park in Long Beach, Mississippi in November 2010. The original HSI imagery consisted of 72 bands, but the first four and last four bands were excluded due to noise, resulting in a total of 64 bands. The LiDAR data consists of two elevation rasters. All the bands and rasters are coregistered, resulting in a total size of 325$\times$220. The dataset contains a total of 53,687 ground truth pixels, covering 11 classes \citep{gader_muufl_2013, du_technical_2017}. For training purposes, 100 pixels per class were selected, leaving a total of 52,587 pixels for testing. The HSI and LiDAR images, along with the ground truth pixels, can be visualized in Figure \ref{fig:Muufl}.

\subsection{Results and Discussion}

\begin{table}[!t]
\centering{\tiny
 \caption[Accuracy analysis on the Trento dataset (in \%).]{\label{tab:tren_perf} Accuracy analysis on the Trento dataset (in \%). `H' represents only HSI while `H+L' represents fused HSI and LiDAR.}
\begin{tabular}{|p{0.9cm}|p{0.7cm}|p{0.7cm}|p{0.8cm} |p{0.9cm}| p{0.8cm}|p{1.1cm} |p{1.1cm}| p{1.1cm}|p{0.9cm}|p{0.9cm}|}
 \hline
Classes &SVM (H) & SVM (H+L) & ELM (H) & ELM (H+L) & CNN-PPF (H) & CNN-PPF (H+L)  & Two Branch CNN (H) & Two Branch CNN (H+L) & FusAtNet (H) & FusAtNet (H+L)\\
 \hline
Apples & 90.80 & 85.49 & 91.32 & 95.81 & 92.22 & 95.88 & 98.04 & 98.07 &99.06& \textbf{99.54}\\
Buildings & 84.22 & 89.76 & 85.74 & 96.97 & 87.08 & \textbf{99.07} & 97.45 & 95.21 &97.05& 98.49\\
Ground & 98.12 & 59.56& 97.59 & 96.66 & 66.81 & 91.44 & 83.09 & 93.32 &\textbf{100.00}& 99.73\\
Woods & 97.01 & 97.42 & 88.44 & 99.39 & 65.24 & 99.79 & 98.29 & 99.93 &\textbf{100.00}& \textbf{100.00}\\
Vineyard & 79.02 & 93.85 & 86.39 & 82.24 & 98.98 & 98.56 & 98.29 & 98.78 &99.85& \textbf{99.90}\\
Roads & 66.92 & 89.96 & 64.06 & 86.52 & 73.19 & 88.72 & 68.21 & 89.98 &89.39& \textbf{93.32}\\
 \hline
OA & 85.56 & 92.30 & 85.43 & 91.32 & 83.52 & 97.48 & 95.35 & 97.92 &98.50& \textbf{99.06}\\
AA & 86.02 & 86.01 & 85.59 & 92.93 & 80.59 & 95.58 & 90.86 & 96.19 &97.56& \textbf{98.50}\\
$\kappa$ & 0.8102 & 0.8971 & 0.8065 & 0.9042 & 0.7843 & 0.9664 & 0.9379 & 0.9681 &0.9796& \textbf{0.9875}\\
 \hline
\end{tabular}}
\end{table}

\begin{table}[!ht]
\centering{\tiny
 \caption{\label{tab:muufl_perf} Accuracy analysis on the MUUFL dataset (in \%). `H' represents only HSI while `H+L' represents fused HSI and LiDAR.}
\begin{tabular}{|p{2.0cm}|p{0.7cm}|p{0.7cm}|p{0.8cm} |p{1.1cm}|p{1.1cm}| p{1.1cm}|p{0.9cm}|p{0.9cm}|}
 \hline
Classes &SVM (H) & SVM (H+L) & ELM (H) & ELM (H+L) & Two Branch CNN (H) & Two Branch CNN (H+L)& FusAtNet (H) & FusAtNet (H+L)\\
 \hline
Trees & 93.91 & 95.97 & 91.99 & 94.89 &97.07 &97.40 &97.74& \textbf{98.10}\\
Grass Pure &  59.54 & 62.71 & 39.44 & 62.23 &62.93 &\textbf{76.84} &63.71& 71.66\\
Grass Groundsurface &  82.72 & 83.60 & 76.87 & 83.15 & 87.44& 84.31 &86.48& \textbf{87.65}\\
Dirt and Sand & 79.11 & 78.60 & 74.51 & 57.88 & \textbf{90.74}& 84.93 & 87.34 & 86.42\\
Road Materials &  91.20 & 92.72 & 92.14 & 93.33 & 85.30 &93.41 &93.07& \textbf{95.09}\\
Water & 54.31 & 95.10 & 0.00 &68.32 & 5.39 &10.78 &24.78& \textbf{90.73}\\
Buildings' Shadow &  58.35 & 71.23 & 63.88 & 47.01 & 67.33 & 63.34&72.55&\textbf{74.27}\\
Buildings & 75.94 &  87.96 & 68.26 & 77.58 & 82.33 & 96.20 &96.38& \textbf{97.55}\\
Sidewalk & 43.85 & 41.11 & 24.22 & 32.15& 54.59& 54.30&56.07&\textbf{60.44}\\
Yellow Curb & 11.05 & 11.05 & 0.00 &0.00&\textbf{24.31}& 2.21 &7.73& 9.39\\
Cloth Panels & 88.37 & 88.76 & 89.92 & 78.29 & 90.31 & 87.21 &92.25&\textbf{93.02}\\
 \hline
OA & 83.39 & 86.90 & 78.49 & 83.10 & 86.30 & 89.38 &89.41& \textbf{91.48}\\
AA & 67.12 & 83.37& 56.48 & 63.17 & 67.97 & 68.26 &70.74& \textbf{78.58}\\
$\kappa$ & 0.7790 & 0.8255 & 0.7137 & 0.7742 & 0.8197 & 0.8583 &0.8581& \textbf{0.8865}\\
 \hline
\end{tabular}}
\end{table}

\subsubsection{Training protocols}

Our method is compared against various conventional and state-of-the-art multimodal learning methods from \citep{xu2017multisource} that fuse HSI and LiDAR modalities. These methods include SVM \citep{mercier2003support}, extreme learning machines \citep{li2015local}, CNN-PPF \citep{li2016hyperspectral}, and two-branch CNN \citep{xu2017multisource} with spectral and spatial feature extraction. For the Trento dataset, the SVM models (both hyperspectral and LiDAR) and the ELM model (only hyperspectral) have been retrained and re-evaluated, as the values reported in \citep{xu2017multisource} appeared to be incorrect. The analysis has been conducted on both HSI-only data (denoted as (H) in the results and classified maps) and fused HSI and LiDAR data (denoted as (H+L)) to demonstrate the effectiveness of multimodal learning compared to unimodal learning. The evaluation metrics used to assess the performance of the methods include overall accuracy (OA), producer's accuracy (PA), average accuracy (AA), and Cohen's kappa ($\kappa$). Both the HSI and LiDAR data are normalized using min-max normalization to scale the modalities and accelerate convergence.

The network adopts a fixed patch size of 11 $\times$ 11 for all datasets. These patches are created around pixels with known ground truth labels. Additionally, to enhance the performance of our model, we apply a data augmentation technique (as used in \citep{zhang2018feature}) by rotating the training patches by 90\textdegree, 180\textdegree, and 270 \textdegree in a clockwise direction. All the weights are initialized using the glorot initialization method \citep{glorot2010understanding}, and training is performed for 1000 epochs. A small initial learning rate of 0.000005 is chosen to prevent excessive fluctuations when using the Adam optimizer with Nesterov momentum \citep{dozat2016incorporating}.

Our proposed method is evaluated on the Houston, Trento, and MUUFL datasets, as shown in Tables \ref{tab:hous_perf}, \ref{tab:tren_perf}, and \ref{tab:muufl_perf}, respectively. It is evident that our method outperforms all state-of-the-art methods by a significant margin in terms of overall accuracy (OA), average accuracy (AA), and Cohen's kappa ($\kappa$) across all datasets. Specifically, the Houston, Trento, and MUUFL datasets achieve respective OA values of 89.98\%, 99.06\%, and 91.48\%, and AA values of 94.65\%, 98.50\%, and 78.58\% using our method. 

Moreover, our method demonstrates superior performance in class-wise accuracy (producer's accuracy) for most of the classes, only marginally surpassed by other methods for a few classes. Notably, for the Houston dataset, our method significantly improves the accuracy of the ``commercial'' class (92.12\%) compared to other methods. This improvement can be attributed to the fact that commercial regions often exhibit variable layouts with frequent elevation changes, which are effectively captured by the LiDAR-based attention maps. Similarly, for the Trento dataset, our method achieves a notable increase in accuracy for the ``road'' class (93.32\%), which can be attributed to the variation in road profiles with respect to elevation.

The classification maps generated by FusAtNet for the Houston, Trento, and MUUFL datasets are presented in Figures \ref{fig:Houston}, \ref{fig:Trento}, and \ref{fig:Muufl}, respectively. Visual inspection of these maps confirms that the classification maps obtained from FusAtNet exhibit reduced noise and smooth interclass transitions. Additionally, in Figure \ref{fig:Houston}, it can be observed that methods such as SVM and two-branch CNN tend to misclassify shadowy areas as water (in the right portion of the maps) due to their darker appearance. Our approach largely mitigates this issue as well.

\subsubsection{Ablation study}

To investigate the contribution of individual components of our model, we conducted several ablation studies. Table \ref{tab:ablation1} presents the performance evaluation of our model when each attention module is removed iteratively. The results clearly demonstrate that the absence of any attention module leads to a decline in model performance. Furthermore, the importance of the spatial characteristics of the LiDAR modality is highlighted, as the presence of the LiDAR-based spatial attention module alone achieves higher accuracy compared to the HSI-based spectral attention module for all three datasets.

\begin{table}[!h]
\centering{\scriptsize
 \caption{\label{tab:ablation1} Ablation study by changing attention layers on all the datasets (accuracy in \%).}
\begin{tabular}{|c|c|c|c|}
 \hline
 Attention layers & OA (Houston) & OA (Trento) & OA (MUUFL)\\
 \hline

Only \textbf{$\mathcal{A_S}$} & 86.48 & 98.04 & 90.09 \\
Only \textbf{$\mathcal{A_T}$} & 88.39 & 98.79 &91.13 \\
Only \textbf{$\mathcal{A_M}$} & 87.51 & 98.48 & 89.31\\
Only \textbf{$\mathcal{A_S}$} and \textbf{$\mathcal{A_T}$} & 89.04 & 98.77 & 90.69\\
Only \textbf{$\mathcal{A_T}$} and \textbf{$\mathcal{A_M}$} & 87.78 & 98.95 & 88.24\\
Only \textbf{$\mathcal{A_S}$} and \textbf{$\mathcal{A_M}$} & 86.90 & 98.24 & 89.14\\
All \textbf{$\mathcal{A_S}$}, \textbf{$\mathcal{A_T}$} and \textbf{$\mathcal{A_M}$} &  \textbf{89.98} &  \textbf{99.06} &  \textbf{91.48}\\
 \hline
\end{tabular}}
\end{table}

Table \ref{tab:ablation2} presents the results of our method when trained without data augmentation. Due to the depth of our model, there is a decrease in performance when training samples are not augmented. The magnitude of this decrease is most significant for the Houston dataset (4.76\%) as it has the highest number of features compared to the other datasets. Therefore, it requires more iterations to converge and achieve higher accuracy.

\begin{table}[!h]
\centering{\scriptsize
 \caption{\label{tab:ablation2} Ablation study for training with and without data augmentation (accuracy in \%).}
\begin{tabular}{|c|c|c|c|}
 \hline
 Data & OA (Houston) & OA (Trento) & OA (MUUFL)\\
 \hline
No augmentation & 85.22 & 98.32 &88.81 \\
With augmentation &  \textbf{89.98} &  \textbf{99.06} &  \textbf{91.48}\\
 \hline
\end{tabular}}
\end{table}

Additionally, we conducted an extra ablation study on all the datasets to examine the impact of reducing the training size and evaluate the performance of our model, as shown in Table \ref{tab:ablation3}. As anticipated, the accuracy gradually decreases as the number of training samples decreases, reaffirming the high data requirement of deep learning models.

\begin{table}[ht]
\centering{\scriptsize
 \caption{\label{tab:ablation3} Model performance by changing the fraction of training samples on all datasets (accuracy in \%).}
\begin{tabular}{|c|c|c|c|c|}
 \hline
 \centering
 &30\% data & 50\% data & 75\% data & 100\% data \\
  \hline
 \centering
OA (Houston) & 84.75 & 87.92 & 88.66 & \textbf{89.98} \\
 \centering
OA (Trento) & 97.78 & 98.48 & 98.93 & \textbf{99.06} \\
 \centering
OA (MUUFL) & 86.90 & 89.58 & 89.78 & \textbf{91.48} \\
 \hline
\end{tabular}}
\end{table}
\vspace{-0.5cm}

\section{Self-supervision assisted multimodal fusion using feedback networks and cross-modal network sharing}
\label{sec:HFN}
\begin{figure*}[t!]
  \centering
  \centerline{\includegraphics[width=14cm]{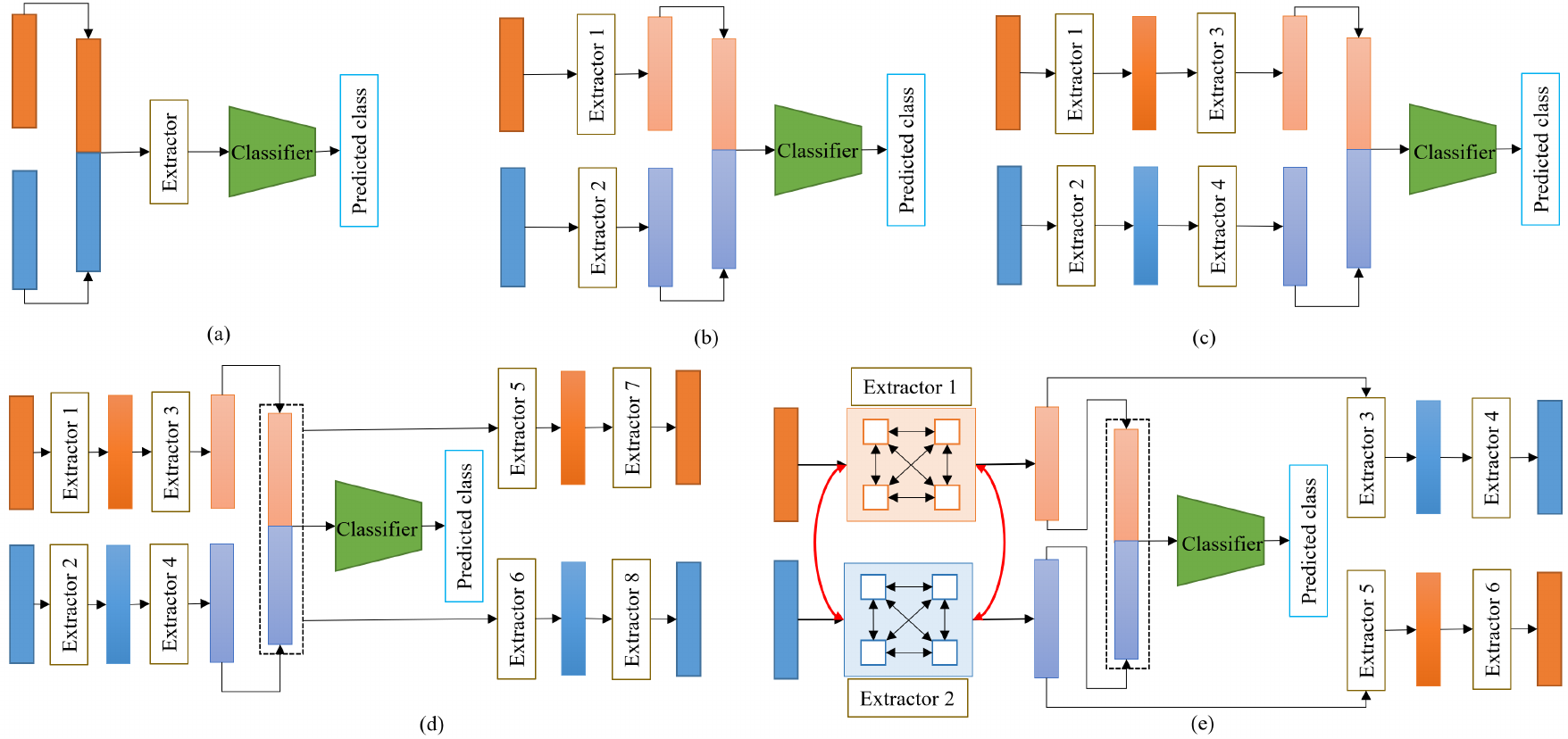}}
  \caption[Different types of fusion approaches.]{Figures (a) to (e) illustrate various fusion paradigms found in existing literature. These paradigms are represented as follows: (a) Early fusion, (b) Middle fusion, (c) Late fusion, (d) Reconstruction-based fusion, and (e) Proposed cross-modal reconstruction based fusion with shared weights across time within the same feature extractors (for the same modality) and across feature extractors (for different modalities). The red curved arrows indicate the shared weights between the feature extraction modules across the two modalities. In all figures, the orange and blue boxes represent the features of the two modalities considered for fusion.}\medskip
  \vspace{-0.5cm}
  \label{fig:fa}
\end{figure*}

We have proposed an innovative fusion approach for integrating HSI-LiDAR/SAR modalities. Our method leverages multiscale self-looping blocks that share the same parameters across time and modalities. By sharing weights across time, we achieve a compact architecture where each convolution layer exchanges information bidirectionally with all other convolution layers. This enhances feature extraction by refining past layers with future information. Weight sharing across modalities facilitates the network in learning correlated features of both modalities through cross-modal communication. Additionally, the shared parameters across time and modalities ensure controlled network parameters. To further enforce modality characteristics on extracted features, we incorporate a multi-task learning \citep{jha2020mt} based framework. We introduce a self-supervised auxiliary task for cross-modal reconstruction in both modalities. These reconstruction modules utilize extracted features from one modality (pre-fusion) as input to construct features for the other modality. This embedding of information from both modalities in pre-fusion features contributes to achieving more robust classification\footnote{Published as: Shivam Pande, Biplab Banerjee, \href{https://www.sciencedirect.com/science/article/pii/S0893608023002058}{Self-supervision assisted multimodal remote sensing image classification with coupled self-looping convolution networks}, Neural Networks 164 (2023) 1–20.}. Our proposed approach is visually compared against other methods in Figure \ref{fig:fa}.

\subsection{Methodology}
\begin{figure*}[t!]
  \centering
  \centerline{\includegraphics[width=14cm]{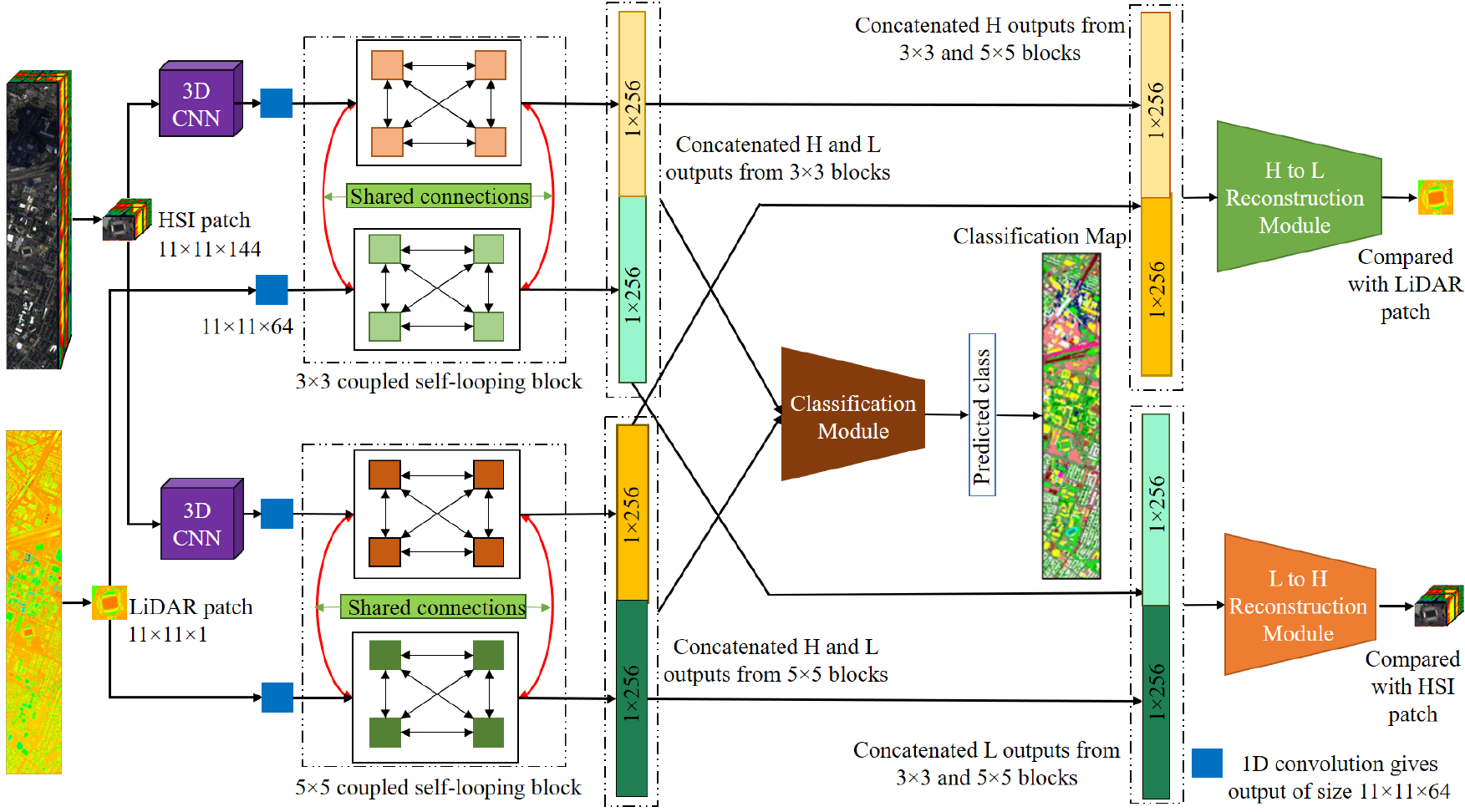}}
  \caption[Schematic of cross-modal selfsupervsion driven fusion model with feedback connections and coupling.]{The diagram represents the architecture of the proposed model. It accepts HSI patches and LiDAR/SAR patches as inputs. The HSI patches undergo a 3D CNN module to obtain refined features. These HSI features, along with the LiDAR/SAR features, are then fed into their respective multiscale coupled self-looping blocks. The outputs from these self-looping blocks are concatenated and passed to the classification module for final classification. At the same time, the extracted features from one modality at both scales are combined and used to reconstruct the other modality. This ensures that the pre-fusion features capture the characteristics of both modalities effectively.}\medskip
  \vspace{-0.5cm}
  \label{fig:diagram_nn}
\end{figure*}

This section will focus on the mathematical formulation and the different modules of the proposed model. Let us consider a dataset $\{\mathcal{H}$, $\mathcal{L}\}$, where $\mathcal{H} \in \mathbb{R}^{D \times W \times B_1}$ represents the HSI modality, and $\mathcal{L} \in \mathbb{R}^{D \times W \times B_2}$ represents the LiDAR/SAR modality. Here, $D$ and $W$ denote the height and width of the image, while $B_1$ and $B_2$ represent the number of bands in the HSI and LiDAR/SAR modalities. From these two images, we extract patches of size $p \times p \times B_1$ and $p \times p \times B_2$, respectively, centered on the groundtruth pixel. Thus, we obtain the input dataset $\mathcal{X} \in \{x_{H}, x_{L}\}^n_i$, where $x_{H} \in \mathbb{R}^{p \times p \times B_1}$ and $x_{L} \in \mathbb{R}^{p \times p \times B_2}$. The corresponding groundtruth labels are $\mathcal{Y} \in \{y\}^n_i$. Here, $i$ is the index of the sample, and $n$ is the total number of groundtruth samples. The overall schematic of the proposed model can be seen in Figure \ref{fig:diagram_nn}. To facilitate comprehension, Table \ref{tab:not} includes all the relevant notations with their corresponding descriptions.

\begin{table}[ht]
\centering{\scriptsize
 \caption{\label{tab:not}List of notations.}
\begin{tabular}{|l|r|}
 \hline
Notation & Description\\
\hline
$\textbf{x}_H^i \in \mathbb{R}^{p \times p \times B_1}$ & $i^{th}$ input HSI patch of spatial dimension $p \times p$ and channels $B_1$\\
$\textbf{x}_L^i \in \mathbb{R}^{p \times p \times B_2}$ & $i^{th}$ input LiDAR/SAR patch of spatial dimension $p \times p$ and channels $B_2$\\
$n$ & Number of training samples\\
$K$ & Total number of layers in a coupled self-looping block\\
$\sigma(.)$  & ReLU non-linearity\\
$w$  & Weights for the corresponding layers\\
$b$  &Bias for the corresponding layers\\
$y^{i}$ & Groundtruth for the $i^{th}$ sample\\
$\hat{y}^{i}$ & Softmax values of the $i^{th}$ sample\\
$-\sum_{i=1}^{n}  y^i\ln{\hat{y}^i}$ & Categorical cross entropy loss\\
$dw$  & Gradient of the weight $w$\\
$\bigcup$ & Concatenation \\
$\odot$ & 2D convolution \\
$\bigodot$ & 3D convolution \\
$\beta$ & Batch normalization\\
 \hline
\end{tabular}}
\end{table}

\begin{figure*}[t!]
  \centering
  \centerline{\includegraphics[width=14cm]{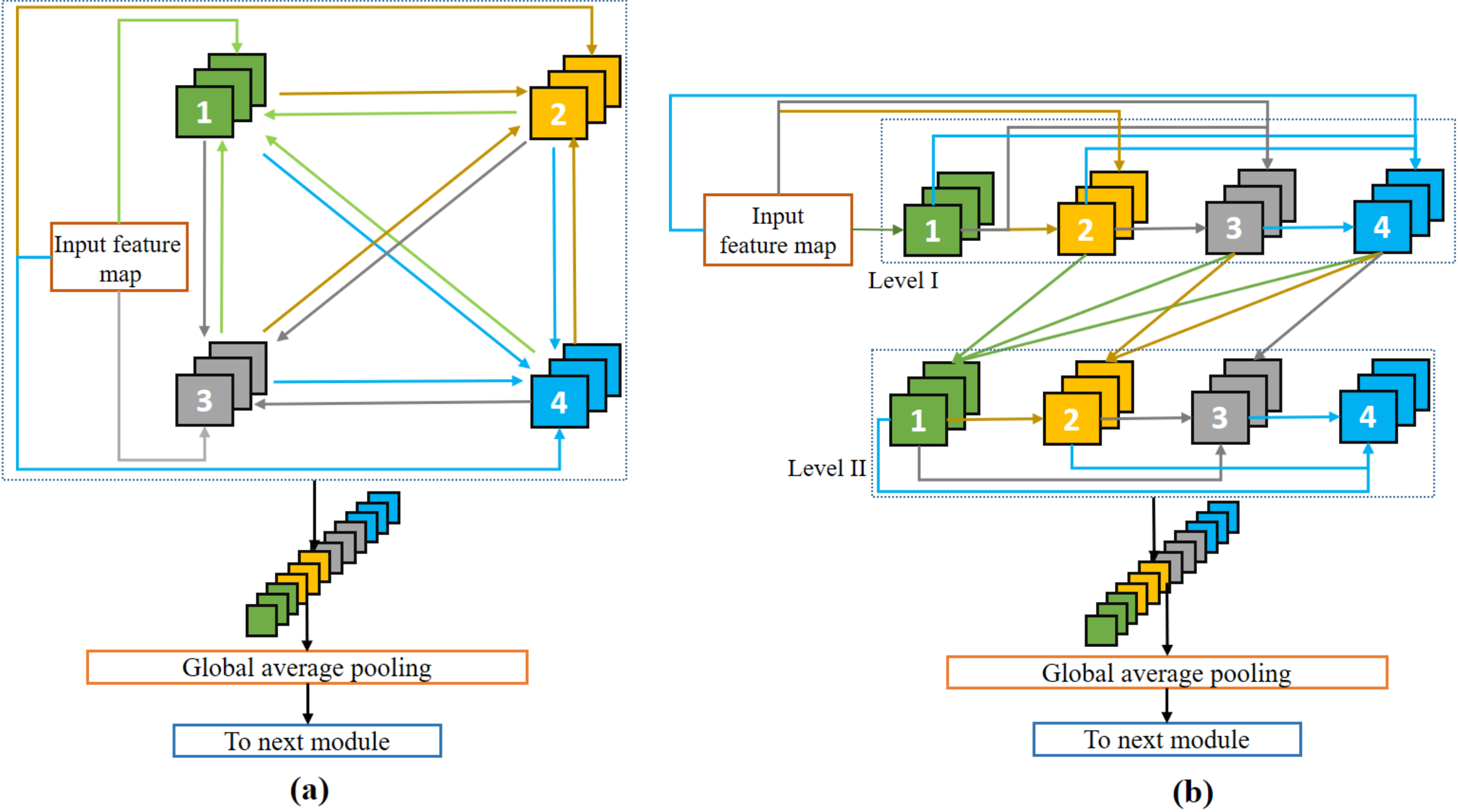}}
  \caption[Schematic of rolled and unrolled self-looping blocks of the proposed model.]{\textcolor{black}{The figure illustrates the self-looping blocks used for feature extraction. There are two versions of the block: the rolled version (a) and the unrolled version (b). Each block consists of a convolution layer, followed by ReLU and batch-normalization layers. The input data passes through two levels within the self-looping blocks. In the first level, the output from all previous layers is fed as input to the second layer. In the second level, the output from the last layer is circulated back to all previous layers using shared parameters across time.}}\medskip
  \vspace{-0.5cm}
  \label{fig:fbblock}
\end{figure*}

\subsubsection{Model Overview}
\label{sec:mo}

Our proposed model consists of four main components: the \textit{spectral-spatial feature extractor}, \textit{multiscale coupled self-looping (MCSL) blocks}, \textit{classification module}, and the \textit{self-supervised cross-modal reconstruction modules}. The \textit{spectral-spatial feature extractor} (refer to section \ref{sec:sfe}) is responsible for extracting refined features from HSI patches before fusion. The outputs from this module are then processed by the \textit{MCSL blocks} for joint feature extraction of the HSI patches and LiDAR/SAR patches (section \ref{sec:mcf}). The resulting features from the \textit{MCSL blocks} are passed to the \textit{classification module} (section \ref{sec:cm}) to obtain the final classification probabilities. Additionally, to ensure that the self-looping modules capture the characteristics of both modalities, they are simultaneously utilized in the \textit{self-supervised cross-modal reconstruction modules} (section \ref{sec:cmr}) to reconstruct inputs from the other modality. The details and mathematical background of each module are discussed in subsequent sections.

\subsubsection{Spectral-spatial feature extractor}
\label{sec:sfe}
HSIs contain abundant information in both the spectral and spatial domains. The initial layers of our module focus on jointly extracting spectral-spatial features. We utilize a 3D CNN for extracting spectral-spatial features from HSI patches within each multiscale block. The first two dimensions of the feature extractor capture spatial information, while the third dimension considers contiguous spectral information. The convolution operation is described in Eq. \ref{equation:3dini}.

\begin{equation}
\{c^{s}_{3d}\}^i = \sigma \left(w^{s}_{3d} \bigodot \{x_H\}^i + b^{s}_{3d} \right)
\label{equation:3dini}
\end{equation}

Here, $w^{s}_{3d}$ represents the 3D convolution kernel for the $s^{th}$ scale, such as 3$\times$3 or 5$\times$5. $b^{s}_{3d}$ denotes the corresponding bias term, and $\bigodot$ signifies the 3D convolution operation. Padding is employed in the convolutions to maintain the size of the input. It is important to note that the input $x_H$ needs to be reshaped accordingly to accommodate 3D convolutions and then reshaped back to the original dimensions for 2D CNNs. To avoid any dimension mismatch during parameter sharing across features and time, the extracted spectral features and the input LiDAR/SAR features are subsequently subjected to 2D convolutions, resulting in a feature dimension of 64 (chosen empirically). The corresponding equations can be found in Eq. \ref{equation:conv_hsi} and \ref{equation:conv_lidar}.

\begin{equation}
\{c^{s}_{H0}\}^i = \sigma \left(w^{s}_{H} \odot \{c^{s}_{3d}\}^i + b^{s}_{H} \right)
\label{equation:conv_hsi}
\end{equation}
\begin{equation}
\{c^{s}_{L0}\}^i = \sigma \left(w^{s}_{L} \odot \{x_L\}^i + b^{s}_{L} \right)
\label{equation:conv_lidar}
\end{equation}

Here, $\{c^{s}_{H0}\}^i$ and $\{c^{s}_{L0}\}^i$ represent the outputs of the 2D CNN-based feature extractors for HSI and LiDAR/SAR, respectively. $w^{s}_{H}$, $w^{s}_{L}$, $b^{s}_{H}$, and $b^{s}_{L}$ are the weights and biases associated with the two modalities.

\subsubsection{MCSL blocks for HSI and LiDAR fusion}
\label{sec:mcf}
The HSI and LiDAR features extracted previously are processed in the coupled feedback blocks for further refinement. In our proposed model, the feedback block is divided into two levels: \textit{level 1} and \textit{level 2}. In \textit{level 1}, we employ \textit{dense} connections \citep{huang2017densely} across the layers in a feedforward manner. Each layer receives input from all the previous layers using the addition operator. This process is represented by Eqn. \ref{equation:HL1_input} and \ref{equation:HL1_output} for the HSI modality and Eqn. \ref{equation:LL1_input} and \ref{equation:LL1_output} for the LiDAR/SAR modality. Here, $k$ represents the index of the layer in that level, $z^{s}_{Hk}$ and $z^{s}_{Lk}$ denote the HSI and LiDAR/SAR outputs obtained after combining all the inputs, respectively. $\beta$ represents batch normalization, $\sigma$ represents the ReLU activation function, while $w^{s}_{Hk}$, $w^{s}_{Lk}$, $b^{s}_{Hk}$, and $b^{s}_{Lk}$ are the weights and biases associated with the HSI and LiDAR/SAR modalities. The symbol $\odot$ represents the 2D convolution operator, and $c^{s}_{Hk}$ and $c^{s}_{Lk}$ are the convolution outputs for HSI and LiDAR/SAR, respectively. The index $i$ refers to the sample, and the superscript $I$ denotes \textit{level 1}, while $s$ represents the scale of the convolutional kernel. The diagram illustrating the self-looping block can be seen in Figure \ref{fig:fbblock}.

\begin{equation}
\{z^{s}_{Hk}\}^i = \{c^{s}_{H0}\}^i + \sum_{j=1}^{k-1}\{c^{s}_{Hj}\}^i
\label{equation:HL1_input}
\end{equation}

\begin{equation}
\{c^{s}_{Hk}\}^i = \beta\sigma \left(\{w^{s}_{Hk}\}^{I} \odot \{z^{s}_{Hk}\}^i + \{b^{s}_{Hk}\}^{I} \right)
\label{equation:HL1_output}
\end{equation}

\begin{equation}
\{z^{s}_{Lk}\}^i = \{c^{s}_{L0}\}^i + \sum_{j=1}^{k-1}\{c^{s}_{Lj}\}^i
\label{equation:LL1_input}
\end{equation}

\begin{equation}
\{c^{s}_{Lk}\}^i = \beta\sigma \left(\{w^{s}_{Lk}\}^{I} \odot \{z^{s}_{Lk}\}^i + \{b^{s}_{Lk}\}^{I} \right)
\label{equation:LL1_output}
\end{equation}

The processed HSI and LiDAR/SAR features obtained from \textit{level 1} are further refined in \textit{level 2} by sending these features backward using shared weights. In each layer of \textit{level 2}, the outputs of all the layers are used as inputs while keeping the weights of the layer the same as in the previous level. This process is described by Eqn. \ref{equation:HL2_input} and \ref{equation:HL2_output} for the HSI modality and Eqn. \ref{equation:LL2_input} and \ref{equation:LL2_output} for the LiDAR/SAR modality. Here, $K$ represents the total number of layers, and $II$ denotes \textit{level 2}.

\begin{equation}
\{z^{s}_{Hk}\}^i = \sum_{j=1, j \neq k}^{K}\{c^{s}_{Hj}\}^i
\label{equation:HL2_input}
\end{equation}

\begin{equation}
\{c^{s}_{Hk}\}^i = \beta\sigma \left(\{w^{s}_{Hk}\}^{II} \odot \{z^{s}_{Hk}\}^i + \{b^{s}_{Hk}\}^{II}\right)
\label{equation:HL2_output}
\end{equation}

\begin{equation}
\{z^{s}_{Lk}\}^i = \sum_{j=1, j \neq k}^{K}\{c^{s}_{Lj}\}^i
\label{equation:LL2_input}
\end{equation}

\begin{equation}
\{c^{s}_{Lk}\}^i = \beta\sigma \left(\{w^{s}_{Lk}\}^{II} \odot \{z^{s}_{Lk}\}^i + \{b^{s}_{Lk}\}^{II}\right)
\label{equation:LL2_output}
\end{equation}

\begin{table*}[htbp]
  \centering{\scriptsize
  \caption{\label{tab:archi}Structure of a coupled self-looping block in the proposed model.}
    \begin{tabular}{|p{1.7cm}|p{2.0cm}|p{1.9cm}|p{1.9cm}|p{1.5cm}|p{1.8cm}|p{1.4cm}|}
    \hline
    Level & HSI Input layers & LiDAR/SAR Input layers & Layer weights and bias & HSI Output layer & LiDAR/ SAR Output layer& Output size \\
   \hline
  Level I & $c_{H0}$ & $c_{L0}$    & $w_1$, $b_1$   & $c_{H1}$ & $ c_{L1}$    & 11$\times$11$\times$64 \\
          & $c_{H0}+c_{H1}$ & $ c_{L0}+c_{L1}$ & $w_2$, $b_2$    & $c_{H2}$ & $ c_{L2}$    & for \\
          & $c_{H0}+c_{H1}+c_{H2}$ & $ c_{L0}+c_{L1}+c_{L2}$ & $w_3$, $b_3$    & $c_{H3}$ & $ c_{L3}$    & all the \\
          & $c_{H0}+c_{H1}+c_{H2}+c_{H3}$ & $ c_{L0}+c_{L1}+c_{L2}+c_{L3}$ & $w_4$, $b_4$ & $c_{H4}$ & $ c_{L4}$    & layers \\
    \hline
 Level II & $c_{H2}+c_{H3}+c_{H4}$ &$c_{L2}+c_{L3}+c_{L4}$& $w_1$, $b_1$ & $c_{H1}$ & $ c_{L1}$  & 11$\times$11$\times$64 \\
          & $c_{H1}+c_{H3}+c_{H4}$ &$c_{L1}+c_{L3}+c_{L4}$& $w_2$, $b_2$ & $c_{H2}$ & $ c_{L2}$    & for \\
          & $c_{H1}+c_{H2}+c_{H4}$ &$c_{L1}+c_{L2}+c_{L4}$& $w_3$, $b_3$ & $c_{H3}$ & $ c_{L3}$    & all the \\
          & $c_{H1}+c_{H2}+c_{H3}$ &$c_{L1}+c_{L2}+c_{L3}$& $w_4$, $b_4$ & $c_{H4}$ & $ c_{L4}$    & layers \\
    \hline
    Global Aver- age Pooling  & [$c_{H1},c_{H2},$ $c_{H3},c_{H4}$] &[$c_{L1},c_{L2},$ $c_{L3},c_{L4}$]& - & $g_{H1}$ & $ g_{L1}$ & 1$\times$256 \\
    \hline
\end{tabular}}
\end{table*}

It is important to note that the weights and biases are shared across both levels, resulting in $\{w^{s}_{Hk}\}^{I} = \{w^{s}_{Hk}\}^{II}$ and $\{b^{s}_{Hk}\}^{I} = \{b^{s}_{Hk}\}^{II}$ for the HSI modality. Similarly, for the LiDAR/SAR modality, $\{w^{s}_{Lk}\}^{I} = \{w^{s}_{Lk}\}^{II}$ and $\{b^{s}_{Lk}\}^{I} = \{b^{s}_{Lk}\}^{II}$. Additionally, since coupling is applied between the HSI and LiDAR/SAR modalities, we have $\{w^{s}_{Hk}\}^{I} = \{w^{s}_{Lk}\}^{I}$, $\{b^{s}_{Hk}\}^{I} = \{b^{s}_{Lk}\}^{I}$, $\{w^{s}_{Hk}\}^{II} = \{w^{s}_{Lk}\}^{II}$, and $\{b^{s}_{Hk}\}^{II} = \{b^{s}_{Lk}\}^{II}$. These parameter equalities are summarized in Eqn. \ref{equation:weq} and \ref{equation:beq}. These equalities indicate that the same parameters are shared across both time and modalities.

\begin{equation}
\{w^{s}_{Hk}\}^{I} = \{w^{s}_{Hk}\}^{II} = \{w^{s}_{Lk}\}^{I} = \{w^{s}_{Lk}\}^{II}
\label{equation:weq}
\end{equation}
\begin{equation}
\{b^{s}_{Hk}\}^{I} = \{b^{s}_{Hk}\}^{II} = \{b^{s}_{Lk}\}^{I} = \{b^{s}_{Lk}\}^{II}
\label{equation:beq}
\end{equation}

Table \ref{tab:archi} illustrates the utilization of shared weights and biases with their respective HSI and LiDAR/SAR inputs. To enhance clarity, we have simplified the notation of input and output convolution layers. The subscript $H$ denotes HSI, while the subscript $L$ denotes LiDAR/SAR.

The outputs obtained from the coupled self-looping modules for both modalities undergo global average pooling (GAP) and are subsequently concatenated (Eqn. \ref{equation:GAPH} and \ref{equation:GAPL}). Here, $\Gamma$ represents the GAP operator, and $\bigcup$ denotes the concatenation operator. These concatenated outputs are then forwarded to the classification module $\mathcal{C}$.

\begin{equation}
\{f^{s}_{gapH}\}^i = \Gamma \left({\bigcup}^{K}_{j=1} \{c^{s}_{Hj}\}^i\right) 
\label{equation:GAPH}
\end{equation}

\begin{equation}
\{f^{s}_{gapL}\}^i = \Gamma \left({\bigcup}^{K}_{j=1} \{c^{s}_{Lj}\}^i\right) 
\label{equation:GAPL}
\end{equation}

\subsubsection{Classification module}
\label{sec:cm}
\begin{figure}[t!]
  \centering
  \centerline{\includegraphics[width=5.0cm]{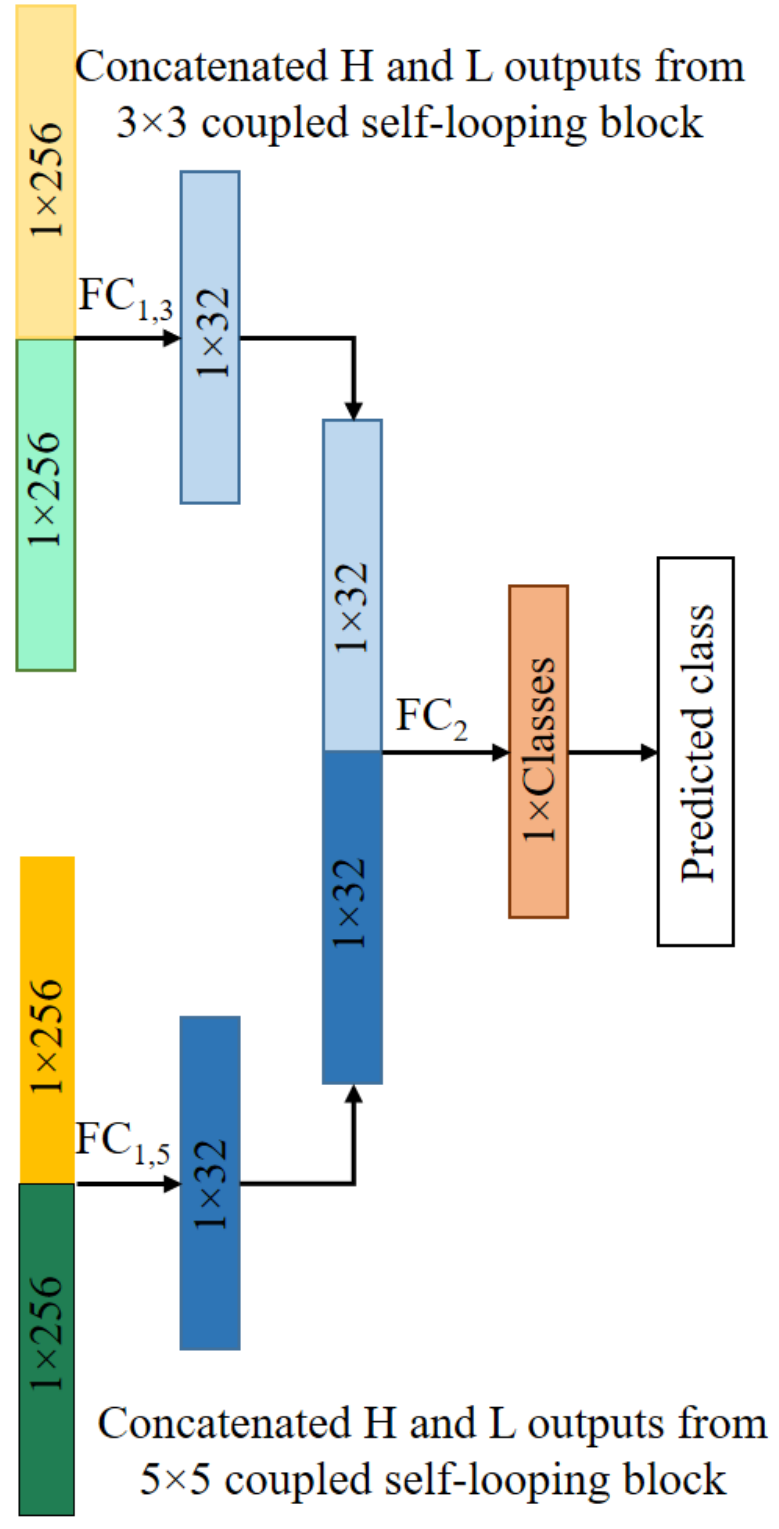}}
  \caption[Schematic of the classification module $\mathcal{C}$]{Schematic of the classification module $\mathcal{C}$ is presented in Figure \ref{fig:diagram_nn}, employing the same colour scheme and notations for ease of comprehension. In the figure, H denotes HSI and L denotes LiDAR/SAR.}\medskip
  \vspace{-0.5cm}
  \label{fig:cmodule}
\end{figure}
The classification module $\mathcal{C}$ is depicted in Figure \ref{fig:cmodule}, illustrating its structure. The outputs from the global average pooling (GAP) layers are concatenated and passed through the fully connected layer $FC_{1,s}$, where $s$ denotes the specified kernel size. ReLU activation is applied to adjust the dimensions to 32, as shown in Eqn. \ref{equation:FC1}. The concatenation of features from different modalities is denoted by $[.]$.

\begin{equation}
\{f^{s}_{dense}\}^i = FC_{1,s} \left(\left[ \{f^{s}_{gapH}\}^i, \{f^{s}_{gapL}\}^i \right]\right) 
\label{equation:FC1}
\end{equation}
The outputs from the self-looping blocks of various scales are combined by concatenation and fed into another fully connected layer, denoted as $FC_2$. Subsequently, a softmax activation function is applied for classification. This process is illustrated in Eqn. \ref{equation:tocm}.

\begin{equation}
\hat{y}^i = softmax\left(FC_2\left({\bigcup}^{S}_{s=1} \{f^{s}_{dense}\}^i \right)\right)
\label{equation:tocm}
\end{equation}

In this context, $\hat{y}^i$ denotes the softmax class probabilities, which are utilized in the categorical cross-entropy loss function as described in Eqn. \ref{equation:ce}. The loss function is computed for $n$ samples, where $y$ represents the original probabilities and $\hat{y}$ denotes the calculated probabilities. Additionally, $S \in \{3,5\}$ refers to the set of spatial sizes for the convolutional kernel employed.

\begin{equation}
\mathcal{L}_{CE} =  -\sum_{i=1}^{n}  y^i\ln{\hat{y}^i}
\label{equation:ce}
\end{equation}

\subsubsection{Self-supervised cross-modal reconstruction module}
\label{sec:cmr}
\begin{figure*}[t!]
  \centering
  \centerline{\includegraphics[width=14cm]{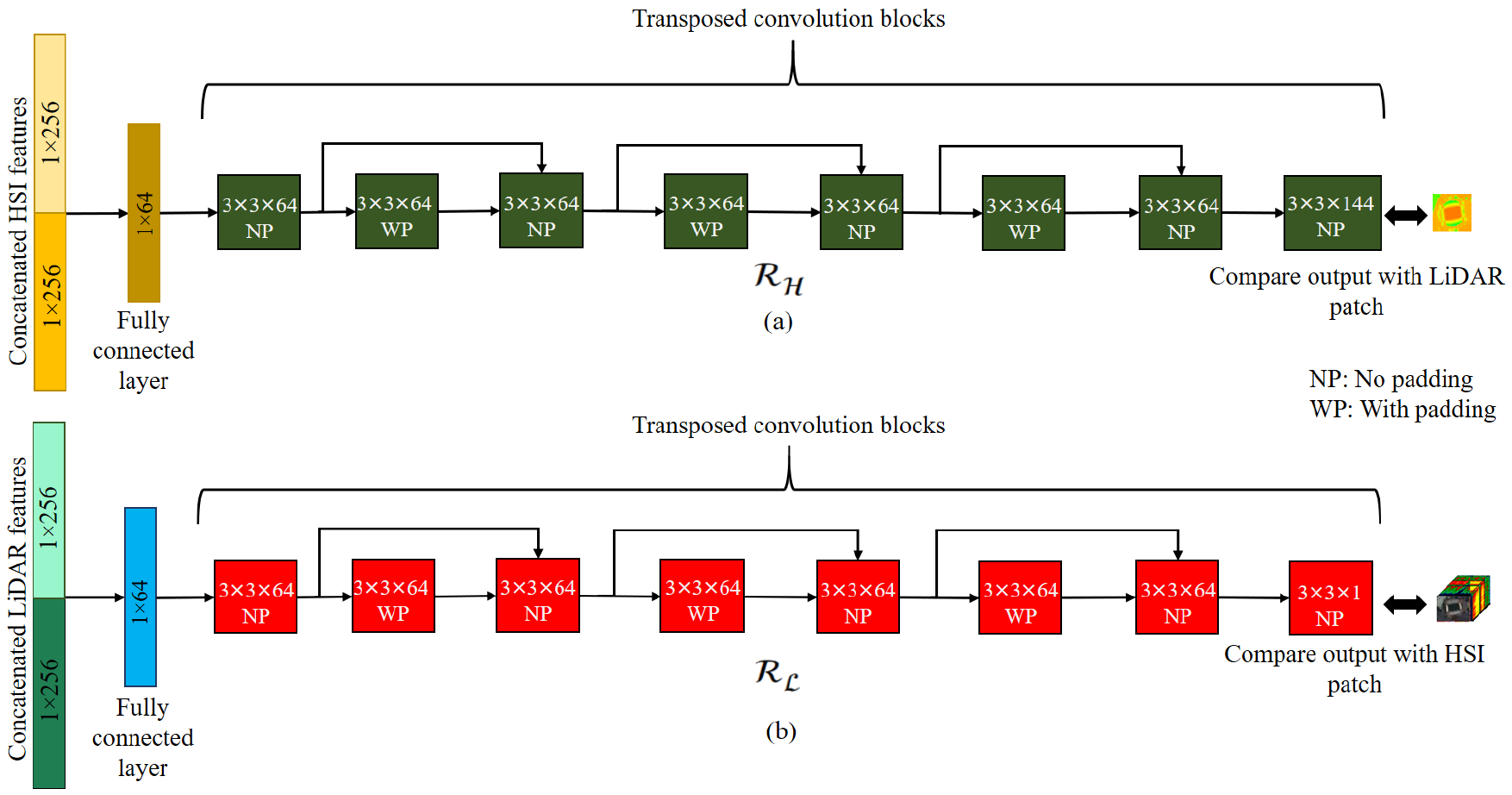}}
  \caption[Schematic of self-supervised cross-modal reconstruction modules.]{The diagram illustrates the self-supervised cross-modal reconstruction modules for reconstructing (a) LiDAR/SAR from HSI ($\mathcal{R_H}$) and (b) HSI from LiDAR/SAR ($\mathcal{R_L}$) using the Houston 13 dataset. These reconstruction modules utilize the extracted features of one modality from the MCSL blocks of both the 3$\times$3 and 5$\times$5 scales. The features are concatenated and passed through a fully connected layer, followed by a series of deconvolution layers, to generate the reconstructed patch of the other modality. Each deconvolution/transposed convolution kernel size is indicated in the respective blocks. ReLU activation and batch normalization are applied after each convolution operation.}\medskip
  \vspace{-0.5cm}
  \label{fig:rec}
\end{figure*}
In order to enhance the robustness of the extracted features, we leverage the features from the last layer of the MCSL modules to construct features from the complementary modality. This approach ensures that the extracted features reflect the joint characteristics of both modalities, leading to more reliable classification outcomes. The Global Average Pooling (GAP) outputs from the MCSL modules are concatenated for both the 3$\times$3 and 5$\times$5 scales of the corresponding modalities, and then passed to their respective reconstruction modules. Each reconstruction module consists of a fully connected layer followed by eight transposed convolution blocks, which include convolution layers with batch normalization and ReLU activation. To control the gradient flow, three skip connections are incorporated between the third, fifth, and seventh convolution blocks. The last convolution block employs a sigmoid activation function. The reconstructed output of the HSI module has a channel size of $B_1$, while the reconstructed output of the LiDAR/SAR module has a channel size of $B_2$. The HSI and LiDAR/SAR reconstruction modules, denoted as $\mathcal{R_H}$ and $\mathcal{R_L}$, are depicted in Figure \ref{fig:rec} (a) and \ref{fig:rec} (b), respectively. For clarity, the colour scheme of the output vectors from the MCSL blocks, which serve as inputs to the reconstruction modules, remains consistent between Figure \ref{fig:diagram_nn} and Figure \ref{fig:rec}. To facilitate effective reconstruction, we employ the mean squared error loss function to compare the reconstructed LiDAR/SAR features with the original HSI features, and vice versa. The loss functions are respectively represented in Eqn. \ref{equation:lossH} and \ref{equation:lossL}.
\begin{figure}[t!]
  \centering
  \centerline{\includegraphics[width=1.0\textwidth]{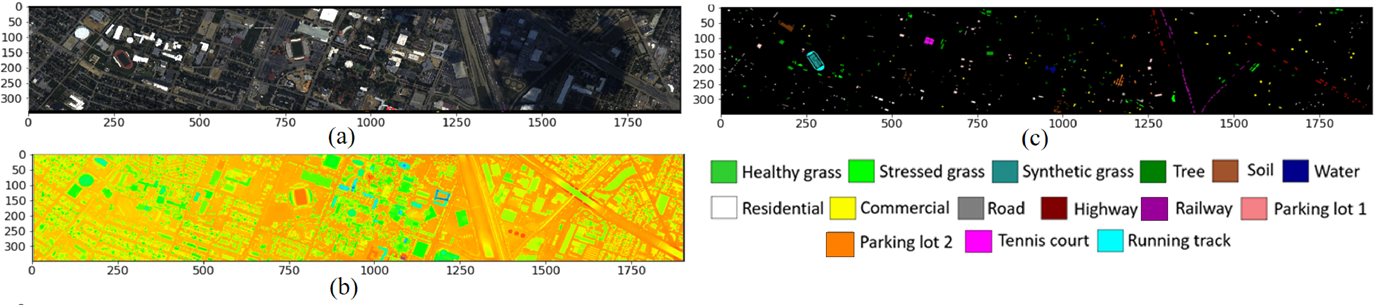}}
  \caption{\textcolor{black}{Houston 2013 dataset (a) Three band colour composite for HSI dataset. (b) DSM obtained from LiDAR (c) Grountruth map}}\medskip
  \vspace{-0.5cm}
  \label{fig:H13ds}
\end{figure}

\begin{equation}
\mathcal{L}_{H2L} = \left|\left|\mathcal{R_H}\left({\bigcup}^{S}_{j=1} \{f^{s}_{gapH}\}^i \right)-\{x_L\}^i\right|\right|^2_2 
\label{equation:lossH}
\end{equation}

\begin{equation}
\mathcal{L}_{L2H} = \left|\left|\mathcal{R_L}\left({\bigcup}^{S}_{j=1} \{f^{s}_{gapL}\}^i \right)-\{x_H\}^i\right|\right|^2_2 
\label{equation:lossL}
\end{equation}

The model is trained using a joint optimization approach that incorporates the cross-entropy loss and two reconstruction losses (Eqn. \ref{equation:Lossfinal}). The balancing weights for the reconstruction losses are denoted as $\lambda_1$ and $\lambda_2$. In our empirical analysis, we set both $\lambda_1$ and $\lambda_2$ to 1, as this configuration yielded the best performance. The impact of different values of $\lambda_1$ and $\lambda_2$ on the model's performance is evaluated in the ablation study presented in Table \ref{tab:aug_lam}.

\begin{equation}
\mathcal{L}_{final} = \mathcal{L}_{CE} +\lambda_1 \mathcal{L}_{L2H} + \lambda_2 \mathcal{L}_{H2L}
\label{equation:Lossfinal}
\end{equation}

\begin{table}[ht]
\centering{\scriptsize
 \caption{\label{tab:H13_data_nn} Houston 2013 hyperspectral dataset with number of training and test samples.}
\begin{tabular}{|p{0.3cm}|p{2.0cm}|p{2.0cm}|p{2.0cm}|}
 \hline
&Class & Train samples & Test samples \\
\hline
1  &Healthy grass& 198 &1053\\
2  &Stressed grass& 190 &1064\\
3  &Synthetic grass& 192 &505\\
4  &Trees& 188 &1056\\
5  &Soil& 186 &1056\\
6  &Water& 182 &143\\
7  &Residential& 196 &1072\\
8  &Commercial& 191 &1053\\
9  &Road& 193 &1059\\
10 &Highway& 191 &1036\\
11 &Railway& 181 &1054\\
12 &Parking lot 1& 192 &1041\\
13 &Parking lot 2& 184 &285\\
14 &Tennis court& 181 &247\\
15 &Running track& 187 &473 \\
\hline
&Total& 2832 & 12197\\
&Percentage& 18.84 & 81.16\\
 \hline
\end{tabular}}
\end{table}

\subsubsection{Gradient flow in the proposed model}
\label{sec:gfp}

The proposed model consists of shared parameters that are shared across both the temporal domain and the feature domain for the two modalities. Consequently, when updating the gradients during training, it is necessary to aggregate the gradients across all layers and time steps. This aggregation process is depicted in Eqn. \ref{equation:WU}. In the equation, $d$ represents the number of modalities, $c$ represents the number of time steps, and $a$ represents the layer index. In order to limit the computational complexity, we set the value of $c$ to 2, but it can be adjusted for experimental purposes to unroll the model multiple times.
\begin{equation}
dw_a =  \sum_{m=1}^{d} \sum_{b=1}^{c} dw_{a,b,m}
\label{equation:WU}
\end{equation}

During the training phase, the model is supplied with HSI and LiDAR/SAR patches extracted from the training dataset. On the other hand, the performance of the model is evaluated using the patches from the test dataset.

\begin{figure}[t!]
  \centering
  \centerline{\includegraphics[width=1.0\textwidth]{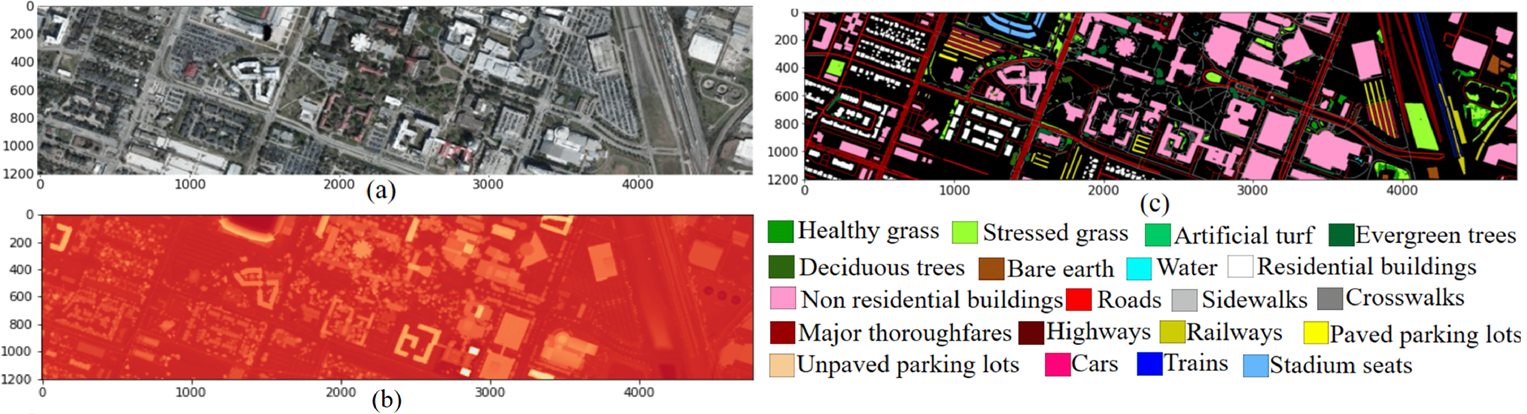}}
  \caption{\textcolor{black}{Houston 2018 dataset (a) Three band colour composite for HSI dataset. (b) DSM obtained from LiDAR (c) Grountruth map}}\medskip
  \vspace{-0.5cm}
  \label{fig:H18ds}
\end{figure}

\subsection{Datasets and experiments}

\begin{table}[ht]
\centering{\scriptsize
 \caption{\label{tab:H18_data_nn}\textcolor{black}{Houston 2018 hyperspectral dataset with number of training and test samples.}}
\begin{tabular}{|p{0.3cm}|p{3.4cm}| p{2.0cm} |p{2.0cm}|}
 \hline
&Class & Train samples & Test samples \\
\hline
1  &Healthy grass& 100 &39096\\
2  &Stressed grass& 100 &129908\\
3  &Artificial turf& 100 &2636\\
4  &Evergreen trees& 100 &54222\\
5  &Deciduous trees& 100 &20072\\
6  &Bare earth& 100 &17964\\
7  &Water& 100 &964\\
8  &Residential buildings& 100 &158895\\
9  &Non-residential buildings& 100 &894669\\
10 &Roads& 100 &183183\\
11 &Sidewalks& 100 &135935\\
12 &Cross walks& 100 &5959\\
13 &Major thoroughfares& 100 &185338\\
14 &Highways& 100 &39338\\
15 &Railways& 100 &27648 \\
16 &Paved parking lots& 100 &45832\\
17 &Unpaved parking lots& 100 &487\\
18 &Cars& 100 &26189\\
19 &Trains& 100 &21739\\
20 &Stadium seats& 100 &27196\\
\hline
&Total& 2000 & 2016910\\
&Percentage& 0.099 & 99.901\\
 \hline
\end{tabular}}
\end{table}

This section focuses on the multimodal datasets utilized to validate our approach. Specifically, three datasets were employed, including two HSI-LiDAR datasets (Houston 2013 and Houston 2018) and one HSI-SAR dataset (TU Berlin). Additionally, we will elaborate on the training protocols and the methods employed for comparison with our approach.

\begin{figure}[t!]
  \centering
  \centerline{\includegraphics[width=1.0\textwidth]{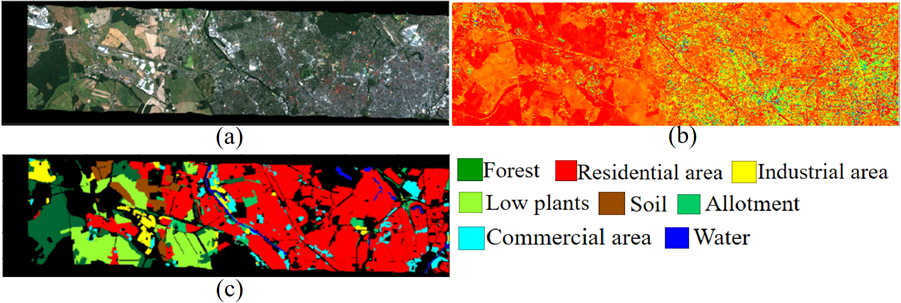}}
  \caption{TU Berlin dataset (a) Three band colour composite for HSI dataset. (b) DSM obtained from LiDAR (c) Grountruth map}\medskip
  \vspace{-0.5cm}
  \label{fig:TUBds}
\end{figure}

\begin{table}[ht]
\centering{\scriptsize
 \caption{\label{tab:TUB_data}{Houston 2018 hyperspectral dataset with number of training and test samples.}}
\begin{tabular}{|p{0.3cm}|p{2.9cm}| p{2.0cm} |p{2.0cm}|}
 \hline
&Class & Train samples & Test samples \\
\hline
1  &Healthy grass& 443 &54511\\
2  &Stressed grass& 423 &268219\\
3  &Artificial turf& 499 &19067\\
4  &Evergreen trees& 376 &58906\\
5  &Deciduous trees& 331 &17095\\
6  &Bare earth& 280 &13025\\
7  &Water& 298 &24526\\
8  &Residential buildings& 170 &6502\\
\hline
&Total& 2820 & 461851\\
&Percentage& 0.61 & 99.39\\
 \hline
\end{tabular}}
\end{table}

\subsubsection{Datasets}
\begin{figure}[t!]
  \centering
  \centerline{\includegraphics[width=8.5cm]{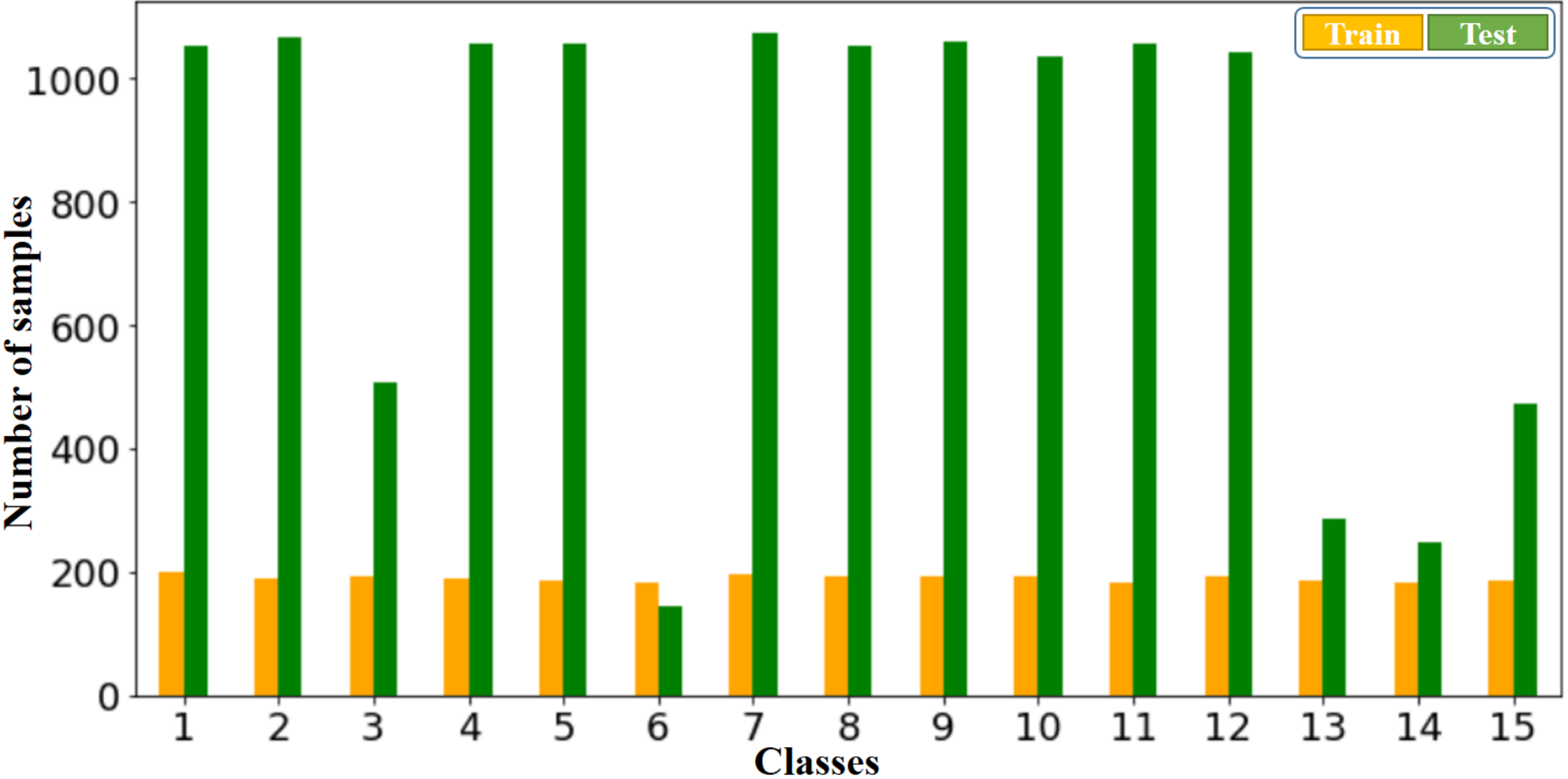}}
  \caption{\textcolor{black}{Bar chart representing train and test samples for Houston 2013 dataset.}}\medskip
  \vspace{-0.5cm}
  \label{fig:H13bar}
\end{figure}

\noindent \textbf{Houston 2013}: The Houston 2013 dataset was provided for the IEEE Data Fusion Contest (DFC) 2013. The hyperspectral image (HSI) was acquired over the National Centre for Airborne Laser Mapping (NCALM) and consists of 144 bands with a spatial dimension of 349$\times$1905. The LiDAR data includes a digital surface model (DSM) with the same spatial dimensions, providing depth information for each cell. The dataset is accompanied by two separate train and test masks, containing 2832 and 12197 ground truth samples, respectively, representing 15 land use and land cover (LULC) classes \citep{debes2014hyperspectral}. For simplicity, we will refer to this dataset as Houston 13. The HSI and LiDAR images, along with the ground truth, are presented in Figure \ref{fig:H13ds}, and the details of the train and test samples are provided in Table \ref{tab:H13_data_nn}. Additionally, the frequencies of the train and test samples are plotted in Figure \ref{fig:H13bar} to visually illustrate their contrast.

\begin{figure}[t!]
  \centering
  \centerline{\includegraphics[width=8.5cm]{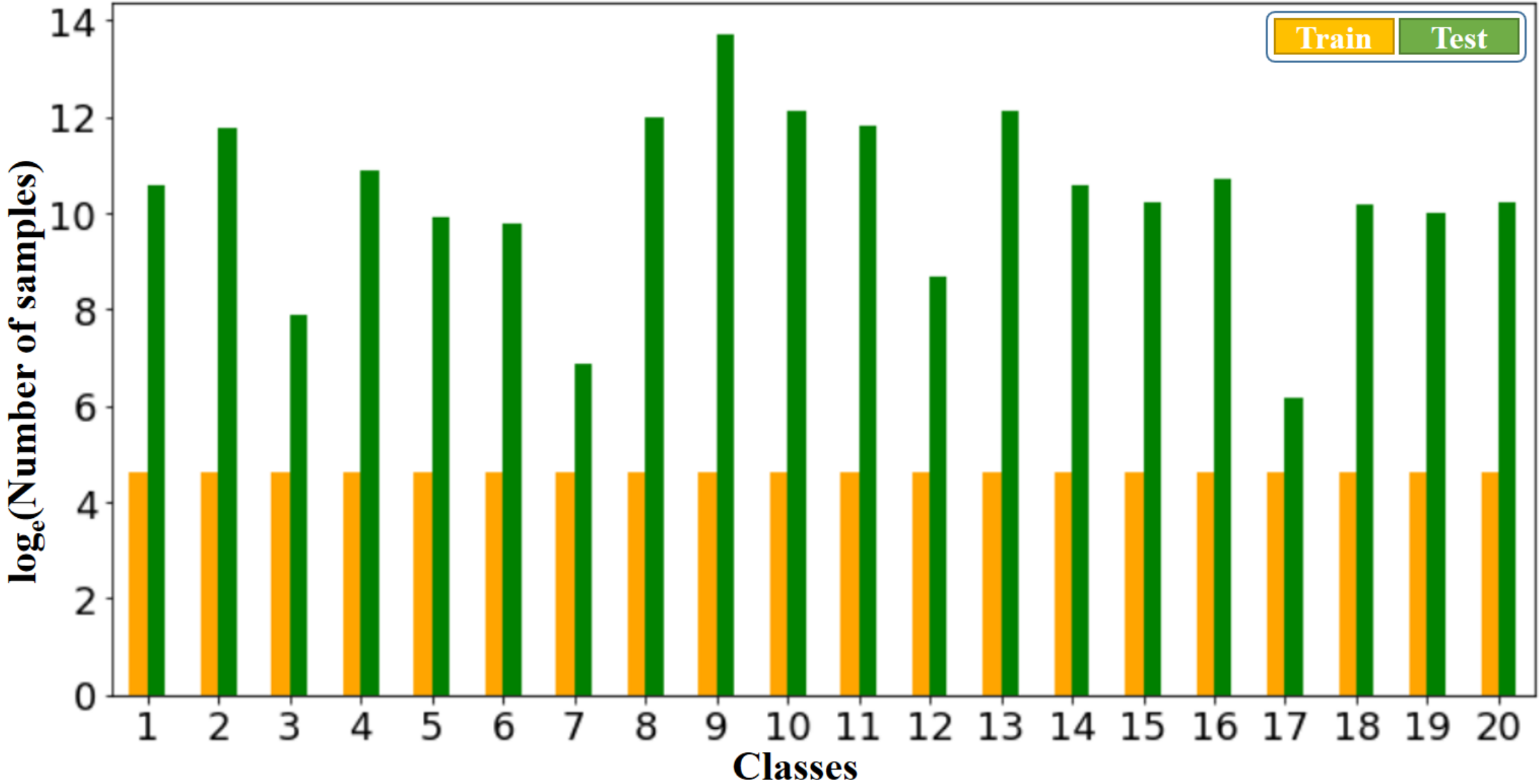}}
  \caption{\textcolor{black}{Logarithmic bar chart representing train and test samples for Houston 2018 dataset.}}\medskip
  \vspace{-0.5cm}
  \label{fig:H18bar}
\end{figure}

\noindent \textbf{Houston 2018}: The Houston 2018 dataset was provided for the DFC-2018. The dataset includes hyperspectral imagery (HSI) consisting of 48 bands with a size of 601$\times$2384, and a LiDAR digital surface model (DSM) with a size of 1202$\times$4768. The ground truth image is provided at a higher spatial resolution and contains a total of 2,018,910 samples \citep{xu2019advanced}. For training, we have selected 2,000 samples, with 100 samples per class, ensuring minimal overlap. The remaining 2,016,910 samples are reserved for testing, and their details can be found in Table \ref{tab:H18_data_nn}. To visualize the distribution of train and test samples, we have plotted the logarithm of the values in Figure \ref{fig:H18bar}. The dataset consists of a total of 20 classes. Since the number of training samples is significantly smaller (<0.01\%) than the test samples, the model avoids overfitting due to minimal overlap between the two sets. We will refer to this dataset as Houston 18. The corresponding HSI, LiDAR, and ground truth images can be seen in Figure \ref{fig:H18ds}.

\begin{figure}[t!]
  \centering
  \centerline{\includegraphics[width=8.5cm]{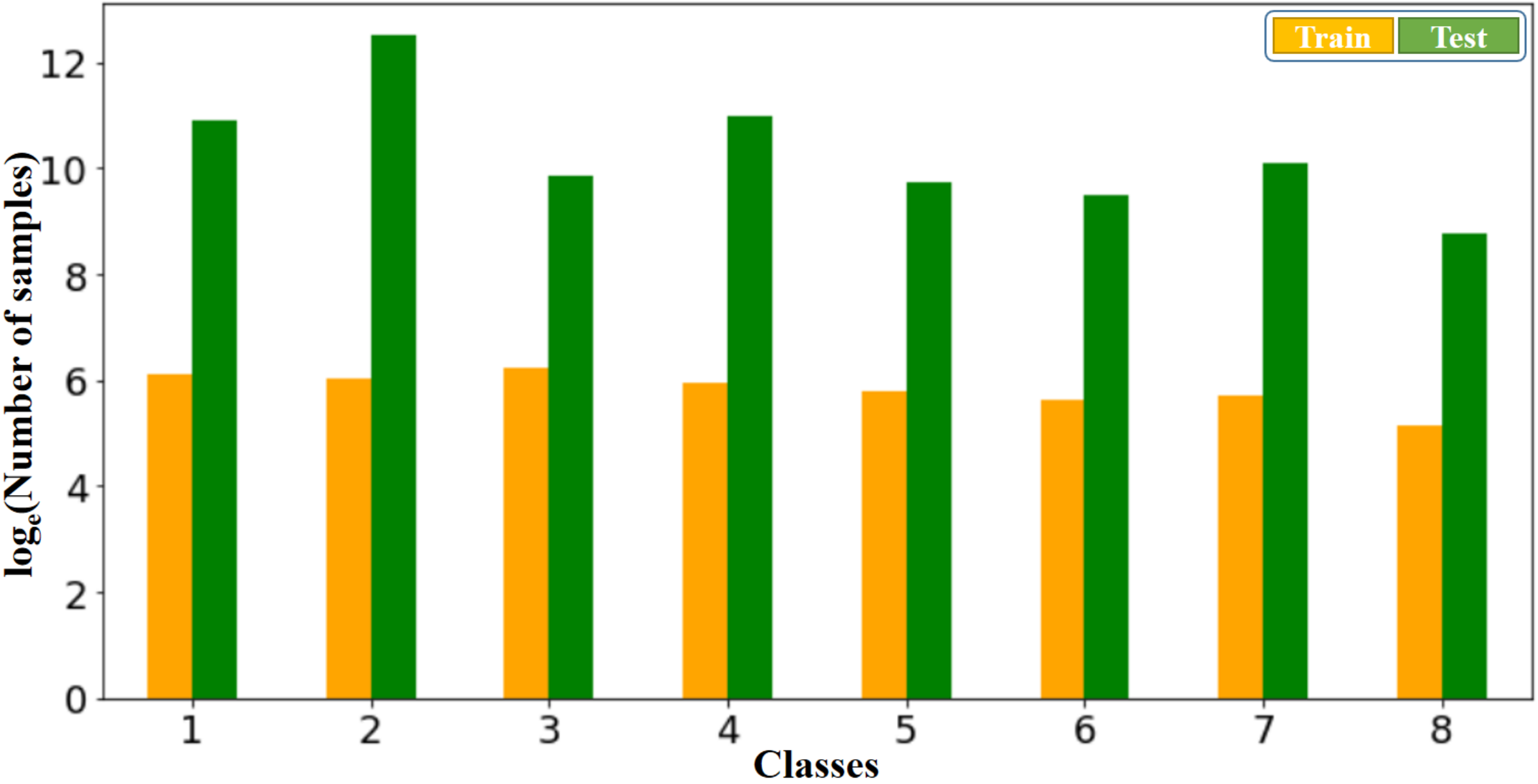}}
  \caption{\textcolor{black}{Logarithmic bar chart representing train and test samples for TU Berlin dataset.}}\medskip
  \vspace{-0.5cm}
  \label{fig:TUBplot}
\end{figure}

\noindent \textbf{TU Berlin}: The TU Berlin dataset includes an Environmental Mapping and Analysis Program (EnMAP) hyperspectral image (HSI) and a corresponding synthetic aperture radar (SAR) image obtained from the Sentinel-1 satellite. The dataset covers an urban area and its surrounding rural region in Berlin. Both the HSI and SAR images have spatial dimensions of 797$\times$220. The HSI consists of 244 spectral bands, while the SAR image has 4 bands (from dual-PolSAR). Figure \ref{fig:TUBds} displays the HSI and SAR images along with the ground truth samples \citep{hong2020x}. For consistency with the other datasets, we only utilize the first SAR band. The dataset provides two separate masks for training and testing. The training set contains 2,820 samples, and the test set includes a total of 461,851 samples. Both the train and test samples are classified into 8 land use/land cover (LULC) classes. These classes are listed in Table \ref{tab:TUB_data}, and their logarithmic plot is presented in Figure \ref{fig:TUBplot}.

\subsubsection{Training protocols}
For training and evaluation of all models, we set the patch size to 11$\times$11$\times B_1$ for HSI and 11$\times$11$\times B_2$ for LiDAR/SAR. The training process is conducted on \textit{Google Colaboratory} using \textit{TensorFlow 2.0} for deep models, while conventional machine learning models are trained using \textit{sklearn}. All models are trained for 500 epochs with the Adam optimizer and Nesterov momentum \citep{dozat2016incorporating}, starting with an initial learning rate of 0.00006 for HSI-LiDAR datasets and 0.00004 for the HSI-SAR dataset. The learning rate is gradually reduced by a factor of $e^{-0.005}$ after 20 epochs to optimize model performance. To effectively manage the number of parameters in both the proposed and comparison models, we employ rotation-based data augmentation during training. Each training sample is rotated by 90\textdegree, 180\textdegree, and 270\textdegree, increasing the training data four-fold. We compare the performance of our approach against algorithms from the conventional machine learning domain and state-of-the-art algorithms from the deep learning domain, which are discussed as follows:

\textbf{Random Forest}: We evaluate a random forest (RF) model with 200 decision trees. The model is trained and tested at the pixel level, where an HSI pixel vector is concatenated with a LiDAR/SAR pixel vector. This approach follows the methodology described by Breiman \citep{breiman2001random}.

\textbf{Canonical Correlation Forest}: For fusion-based classification, we utilize a canonical correlation forest (CCF) with 200 trees. Similar to RF, this model is trained and tested at the pixel level \citep{rainforth2015canonical}.

\textbf{2D CNN}: This approach follows a concatenation-based fusion paradigm but with patch-based input, incorporating spatial context. The model consists of 6 convolution modules with padding. Each convolution has a kernel size of 3$\times$3, and the number of channels increases as 32, 64, and 128 for every two blocks. After every second convolution layer, a max-pooling layer is applied, and a Global Average Pooling (GAP) layer is placed after the last max-pooling layer. The output from the GAP layer is then sent to a softmax-based classifier \citep{mou2019learning}.

\textbf{HybridSN}: This method adopts a hybrid approach, where the initial layers are based on 3D CNN for feature extraction, followed by 2D CNN layers. The output from the 2D CNN layers is then utilized for classification. Similar to other methods, HybridSN uses concatenation-based fusion of HSI and LiDAR/SAR patches \citep{roy2019hybridsn}.

\textbf{Two Branch CNN}: As one of the pioneering fusion methods in remote sensing, this network consists of a 1D CNN-based spectral branch for spectral feature extraction from HSI, a 2D CNN-based branch for spatial feature extraction from HSI, and another 2D CNN-based branch for spatial feature extraction from LiDAR-based DSM. The extracted features from all branches are concatenated and fed into a classifier \citep{xu2017multisource}.

\textbf{FusAtNet}: This method is a pioneer in remote sensing-based image fusion for HSI and LiDAR, utilizing attention mechanisms. The model consists of three CNN-based attention modules. The first two attention modules operate in parallel and are used for spectral and spatial feature extraction with specialized attention masks. The third attention module is employed to determine which modality contributes more to the classification \citep{mohla2020fusatnet}.

\begin{figure*}[t!]
  \centering
  \centerline{\includegraphics[width=14cm]{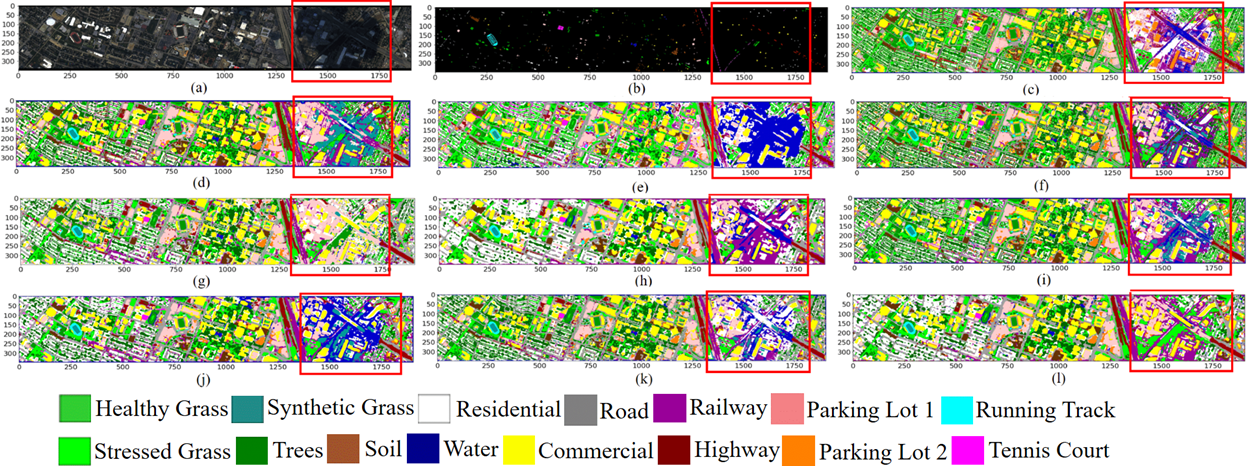}}
  \caption[Houston 2013 HSI-LiDAR dataset with RGB image, groundtruth and classification maps.]{\textcolor{black}{(a) Three band true colour composite of Houston 13 HSI. (b) Groundtruth of Houston 13 dataset. Classification maps using (c) CCF, (d) 2D CNN, (e) HybridSN, (f) Two Branch CNN, (g) FusAtNet (h) Two Headed Dragons (i) MAHiDFNet (j) Coupled CNNs (k) CCR-Net (l) Proposed model}}\medskip
  \vspace{-0.5cm}
  \label{fig:cmH13}
\end{figure*}
\begin{figure*}[t!]
  \centering
  \centerline{\includegraphics[width=14cm]{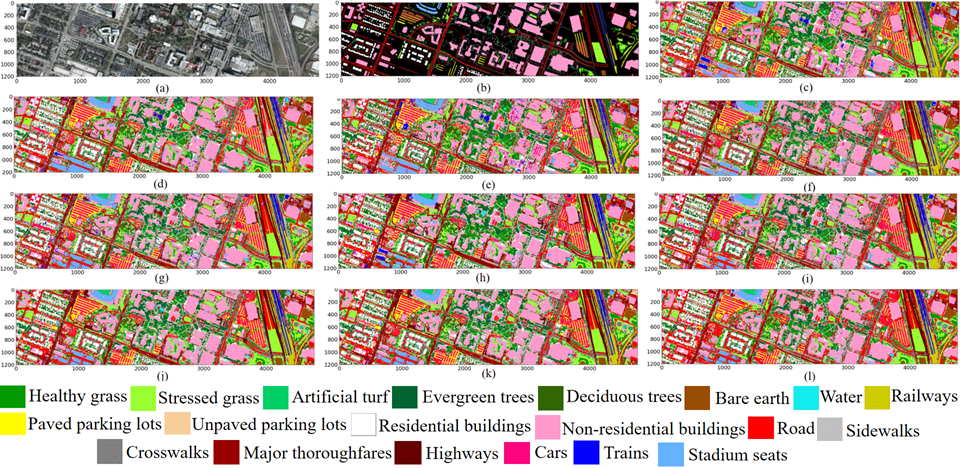}}
  \caption[Houston 2018 HSI-LiDAR dataset with RGB image, groundtruth and classification maps.]{\textcolor{black}{(a) Three band true colour composite of Houston 18 HSI. (b) Groundtruth of Houston 18 dataset. Classification maps using (c) CCF, (d) 2D CNN, (e) HybridSN, (f) Two Branch CNN, (g) FusAtNet (h) Two Headed Dragons (i) MAHiDFNet (j) Coupled CNNs (k) CCR-Net (l) Proposed model}}\medskip
  \vspace{-0.5cm}
  \label{fig:cmH18}
\end{figure*}
\begin{figure*}[t!]
  \centering
  \centerline{\includegraphics[width=14cm]{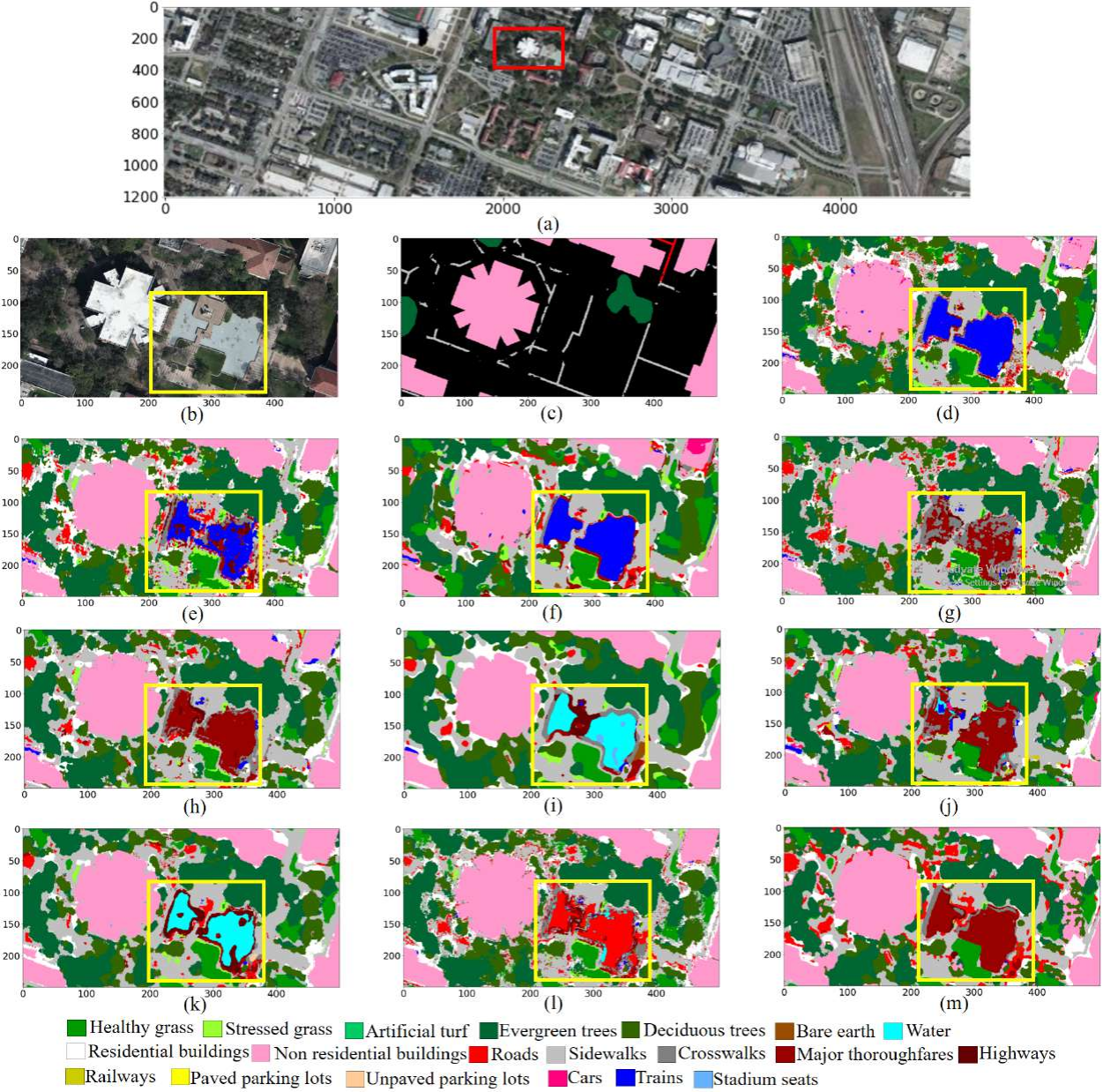}}
  \caption[High resolution subset of Houston 2018 HSI-LiDAR dataset with RGB image, groundtruth and classification maps.]{(a) Three band true colour composite of Houston 18 HSI. (b) Three band true colour composite of subset of Houston 18 HSI. (c) Groundtruth of subset of Houston 18 HSI. Classification maps using (d) CCF, (e) 2D CNN, (f) HybridSN, (g) Two Branch CNN, (h) FusAtNet (i) Two Headed Dragons (j) MAHiDFNet (k) Coupled CNNs (l) CCR-Net (m) Proposed model}\medskip
  \vspace{-0.5cm}
  \label{fig:cmH18_sub}
\end{figure*}
\begin{figure*}[t!]
  \centering
  \centerline{\includegraphics[width=14cm]{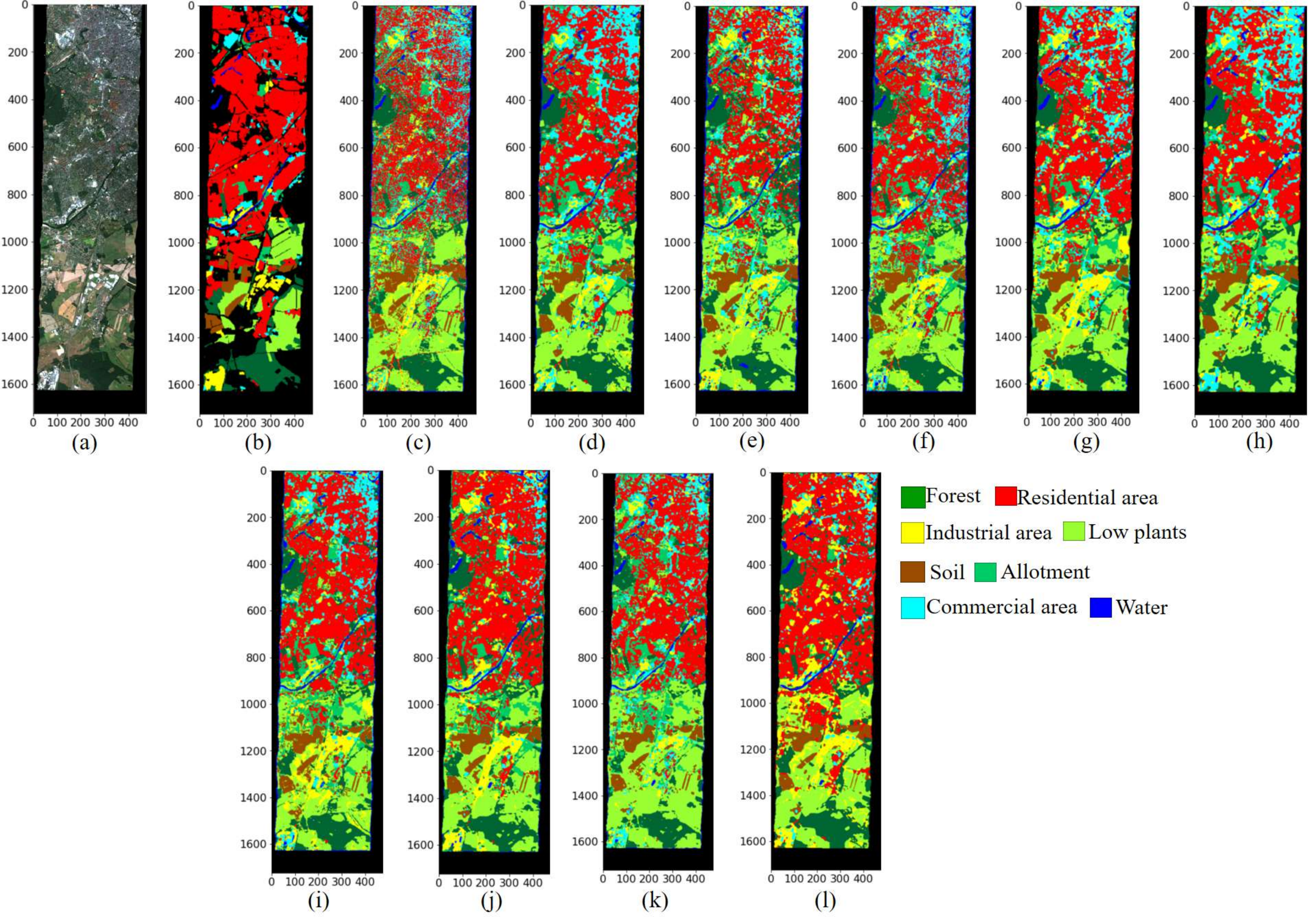}}
  \caption[TU Berlin HSI-SAR dataset with RGB image, groundtruth and classification maps.]{(a) Three band true colour composite of TU Berlin HSI. (b) Groundtruth of TU Berlin dataset. Classification maps using (c) CCF, (d) 2D CNN, (e) HybridSN, (f) Two Branch CNN, (g) FusAtNet (h) Two Headed Dragons (i) MAHiDFNet (j) Coupled CNNs (k) CCR-Net (l) Proposed model}\medskip
  \vspace{-0.5cm}
  \label{fig:cmTUB}
\end{figure*}

\textbf{Two Headed Dragons}: This work introduces transformers to the domain of image fusion in remote sensing. The model comprises four sequential convolution-based attention blocks, each utilizing self-attention and cross-attention for feature extraction from HSI and LiDAR. The outputs from all modules are concatenated and passed through a classifier \citep{bose2021two}.

\textbf{Multi-Attentive Hierarchical Dense Fusion (MAHiDFNet)}: This recent work focuses on HSI-LiDAR fusion and leverages the self-attention concept from transformers in a multiscale setting. Initially, CNN-based architectures are employed to extract spectral and spatial features from HSI and LiDAR. The extracted features are then fed into spectral and spatial attention modules, where self-attention is utilized for shallow feature fusion and a modality attention module highlights more specific features through deep fusion \citep{wang2022multi}.

\textbf{Coupled CNNs}: This method proposes weight sharing across the two branches corresponding to the HSI and LiDAR modalities. Each branch consists of three convolutional layers, with the last two layers sharing weights. Additionally, each branch has its own cross-entropy loss function in addition to the combined loss that acts on the fused HSI-LiDAR features. A decision-based fusion technique is employed for class label prediction \citep{hang2020classification}.

\textbf{CCRNet}: This method incorporates two separate feature extractors for the HSI and LiDAR modalities. The outputs from these extractors are concatenated and passed through a fully connected layer for classification. Furthermore, the output features from one modality are also used to reconstruct those from the other modality at the feature level, enhancing the diversity of the extracted features \citep{wu2021convolutional}. It should be noted that this approach differs significantly from our proposed approach, and we provide an explicit comparison between the two in Table \ref{tab:feat_rec}, where we replace our reconstruction module with that of CCRNet.

To assess the performance of the models, we utilize metrics such as overall accuracy (OA), Cohen's kappa ($\kappa$), and producer's accuracy/classwise accuracy (PA). Furthermore, to determine the statistical significance of the performance difference between our model and other models, we employ McNemar's test \citep{pembury2020effective}, as shown in Eqn. \ref{equation:MN_nn}.

\begin{equation}
z_{12} = \frac{|f_{12}-f_{21}|}{\sqrt{f_{12}+f_{21}}}
\label{equation:MN_nn}
\end{equation}

In the given equation, $z_{12}$ denotes the z-statistic, where ``1'' and ``2'' refer to the models being compared. The variable $f_{12}$ represents the number of test samples that are correctly classified by model 1 and incorrectly by model 2, while $f_{21}$ represents the number of test samples that are correctly classified by model 2 and incorrectly by model 1. In our experiment, the null hypothesis assumes that the two models do not differ significantly in performance. If, for a given pair of models, $z_{12}$ exceeds 1.96, the null hypothesis can be rejected.

\begin{sidewaystable*}[htbp]
  \centering{\scriptsize
  \caption{\label{tab:hous13_perf_nn}Accuracy analysis on the Houston 13 HSI-LiDAR dataset (in \%).}
    \begin{tabular}{|p{3.3 cm}|p{0.6cm}|p{0.6cm}|p{0.7cm}|p{1.0cm}|p{1.6cm}| p{1.3cm}|p{1.7cm}|p{1.3cm}|p{1.3cm}|p{1.3cm}|p{1.3cm}|}
 \hline
 Class name & RF & CCF & 2D CNN & Hybrid SN & Two Branch CNN & Fus At Net & Two Headed Dragons & MAHiD FNet & Coupled CNN  & CCR Net & Hyper Fuse Net\\
 \hline
    Healthy grass&82.53 & 83.10  & 87.08 & 82.05 & 83.10  & \textbf{96.87} & 82.24 & 83.10  & 83.10  & 82.24 & 83.10 \\
    Stressed grass&83.18 & 83.93 & 96.24 & 83.36 & 85.06 & 82.42 & 85.06 & 82.61 & 85.15 & 85.06 & \textbf{99.91} \\
    Synthetic grass&98.42 & \textbf{100} & \textbf{100} & 98.22 & \textbf{100} & \textbf{100} & 99.41 & \textbf{100} & \textbf{100} & \textbf{100} & \textbf{100} \\
    Trees&91.67 & 92.42 & 93.97 & 86.46 & \textbf{100} & 91.95 & 90.91 & 93.37 & 93.28 & 92.33 & 91.86 \\
    Soil&96.31 & 98.77 & 98.77 & 99.91 & 99.91 & 97.92 & \textbf{100} & 99.72 & 99.81 & 99.53 & \textbf{100} \\
    Water&98.60  & 99.30  & \textbf{100} & 95.80  & \textbf{100} & 90.91 & \textbf{100} & \textbf{100} & \textbf{100} & 99.30  & 99.30 \\
    Residential&79.48 & 84.42 & 84.79 & 82.84 & 90.30  & 92.91 & 88.71 & 92.26 & 93.28 & \textbf{95.90} & 94.50 \\
    Commercial&41.50  & 52.90  & 75.78 & 75.12 & 77.49 & 89.46 & 81.96 & 91.55 & 92.02 & 81.58 & \textbf{96.77} \\
    Road&71.10  & 76.02 & 83.19 & 79.98 & 84.70  & 82.06 & 85.55 & 86.02 & 88.67 & 86.69 & \textbf{93.11} \\
    Highway&45.17 & 67.18 & 48.17 & 55.12 & 68.44 & 66.60  & 67.86 & 66.31 & 67.28 & 67.08 & \textbf{71.53} \\
    Railway&73.81 & 84.44 & 85.10  & 73.62 & 91.84 & 80.36 & 98.29 & 95.92 & 91.27 & 91.08 & \textbf{99.24} \\
    Parking lot 1&68.01 & 92.80  & 89.82 & 90.97 & 94.14 & 92.41 & 92.60  & \textbf{94.24} & 84.63 & 93.85 & 92.89 \\
    Parking lot 2&64.56 & 76.49 & 92.28 & 75.09 & 92.28 & 92.63 & 91.93 & \textbf{98.60} & 92.28 & 91.58 & 87.72 \\
    Tennis court&\textbf{100} & 99.60 & \textbf{100} & \textbf{100} & 99.19 & \textbf{100} & \textbf{100} & \textbf{100} & \textbf{100} & \textbf{100} & \textbf{100} \\
    Running track&97.67 & 97.89 & 95.56 & 99.37 & 99.15 & 97.89 & 99.58 & \textbf{100} & \textbf{100} & 98.94 & \textbf{100} \\
    \hline
  Overall accuracy (OA)  &75.97 & 83.46 & 86.08 & 82.83 & 88.99 & 88.69 & 88.83 & 90.06 & 89.36 & 89.01 & \textbf{93.08} \\
  Average accuracy (AA)  &79.47 & 85.95 & 88.68 & 85.19 & 91.04 & 90.29 & 90.94 & 92.25 & 91.38 & 91.01 & \textbf{93.99}\\
  \textcolor{black}{Kappa ($\kappa$)} & \textcolor{black}{0.7416} & \textcolor{black}{0.8214} & \textcolor{black}{0.8493} & \textcolor{black}{0.8150}  & \textcolor{black}{0.8807} & \textcolor{black}{0.8772} & \textcolor{black}{0.8789} & \textcolor{black}{0.8924} & \textcolor{black}{0.8848} & \textcolor{black}{0.8810} & \textcolor{black}{\textbf{0.9249}} \\
     \hline
    \end{tabular}}
\end{sidewaystable*}

\begin{sidewaystable*}[htbp]
  \centering{\scriptsize
  \caption{\label{tab:hous18_perf_nn}Accuracy analysis on the Houston 18 HSI-LiDAR dataset (in \%).}
    \begin{tabular}{|p{3.3 cm}|p{0.6cm}|p{0.6cm}|p{0.7cm}|p{1.0cm}|p{1.6cm}| p{1.3cm}|p{1.7cm}|p{1.3cm}|p{1.3cm}|p{1.3cm}|p{1.3cm}|}
 \hline
 Class name & RF & CCF & 2D CNN  & Hybrid SN & Two Branch CNN & Fus At Net & Two Headed Dragons & MAHiD FNet & Coupled CNN & CCR Net & Hyper Fuse Net\\
 \hline
    Healthy grass&91.33 & 96.49 & 93.62 & 95.86 & 94.29 & 93.04 & 94.46 & 94.43 & 93.22 & 94.47 & \textbf{96.91} \\
    Stressed grass&9.90  & 92.02 & 89.65 & 88.49 & 89.95 & 91.35 & 90.66 & 90.16 & \textbf{91.55} & 86.26 & 87.63 \\
    Artificial turf&\textbf{100} & \textbf{100} & 99.96 & 99.85 & \textbf{100} & \textbf{100} & \textbf{100} & \textbf{100} & \textbf{100} & \textbf{100} & \textbf{100} \\
    Evergreen trees&92.78 & 98.09 & 97.67 & 96.36 & 97.99 & 98.54 & 98.80  & \textbf{99.29} & 98.93 & 98.06 & 99.08 \\
    Deciduous trees&55.85 & 90.08 & 95.57 & 91.32 & 93.16 & 95.85 & 93.85 & 96.52 & \textbf{97.95} & 97.47 & 90.01 \\
    Bare earth&0 & 99.05 & 99.87 & 97.79 & 99.83 & 99.53 & 99.93 & 99.64 & 99.41 & 99.28 & \textbf{100} \\
    Water&92.53 & 99.90  & \textbf{100} & \textbf{100} & 98.96 & \textbf{100} & \textbf{100} & \textbf{100} & \textbf{100} & 99.90  & \textbf{100} \\
    Residential buildings&72.27 & 84.36 & 88.49 & 74.23 & 86.27 & 89.26 & 88.86 & 87.75 & 84.63 & 86.82 & \textbf{88.88} \\
    Non-residential buildings&48.47 & 78.10  & 84.96 & 74.96 & 84.77 & 85.26 & 87.18 & 83.56 & 87.11 & 85.87 & \textbf{90.87} \\
    Roads&0 & 42.89 & 53.32 & 47.49 & 51.19 & 48.64 & 52.22 & 55.14 & 57.78 & 55.85 & \textbf{69.57} \\
    Sidewalks&0.04  & 53.11 & 60.83 & 50.79 & 62.49 & 62.13 & 64.33 & \textbf{67.04} & 65.12 & 58.42 & 63.58 \\
    Crosswalks&\textbf{86.64} & 67.19 & 71.22 & 67.33 & 81.84 & 80.08 & 75.97 & 78.45 & 75.75 & 83.72 & 71.96 \\
    Major thoroughfares&0.02  & 47.34 & 53.84 & 53.50  & 63.28 & 61.03 & \textbf{65.02} & 64.72 & 50.86 & 48.71 & 57.95 \\
    Highways&0 & 86.97 & 94.17 & 88.47 & 95.50  & 93.85 & 96.52 & 94.29 & \textbf{98.52} & 92.84 & 95.29 \\
    Railways&8.74  & 99.44 & 98.94 & 99.37 & 98.00 & 99.01 & \textbf{99.63} & 99.39 & 98.65 & 98.01 & 98.94 \\
    Paved parking lots&0 & 92.83 & 92.95 & 89.82 & 91.95 & 96.18 & 93.18 & 89.17 & 94.74 & \textbf{95.05} & 92.41 \\
    Unpaved parking lots&97.95 & 99.79 & \textbf{100} & \textbf{100} & 99.59 & \textbf{100} & \textbf{100} & \textbf{100} & \textbf{100} & \textbf{100} & 99.79 \\
    Cars&3.24  & 91.06 & 95.92 & 90.50  & 96.96 & \textbf{98.74} & 98.34 & 99.53 & 98.08 & 97.81 & 96.42 \\
    Trains&1.30   & 93.47 & 99.02 & 95.98 & 99.51 & 99.64 & \textbf{99.97} & 99.58 & 99.53 & 99.41 & 99.01 \\
    Stadium seats&3.49  & 98.08 & 99.69 & 97.52 & 99.89 & 99.88 & \textbf{100} & 99.90  & 99.96 & 99.33 & 99.91 \\
    \hline
    Overall accuracy (OA) &33.33 & 74.39 & 79.96 & 72.72 & 80.55 & 80.74 & 82.36 & 80.97 & 81.37 & 79.72 & \textbf{84.59} \\
    Average accuracy (AA) &38.23 & 85.51 & 88.48 & 84.98 & 89.27 & 89.60 & 89.94 & 89.58 & 88.64 & \textbf{89.92} & 89.91 \\
    \textcolor{black}{Kappa ($\kappa$)} &\textcolor{black}{0.2456} & \textcolor{black}{0.6840}  & \textcolor{black}{0.7487} & \textcolor{black}{0.6646} & \textcolor{black}{0.7562} & \textcolor{black}{0.7585} & \textcolor{black}{0.7776} & \textcolor{black}{0.7621} & \textcolor{black}{0.7653} & \textcolor{black}{0.7453} & \textcolor{black}{\textbf{0.8036}} \\
     \hline
    \end{tabular}}
\end{sidewaystable*}%

\begin{sidewaystable*}[htbp]
  \centering{\scriptsize
  \caption{\label{tab:tub_perf}Accuracy analysis on the TU Berlin HSI-SAR dataset (in \%).}
    \begin{tabular}{|p{3.3 cm}|p{0.6cm}|p{0.6cm}|p{0.7cm}|p{1.0cm}|p{1.6cm}| p{1.3cm}|p{1.7cm}|p{1.3cm}|p{1.3cm}|p{1.3cm}|p{1.3cm}|}
 \hline
 Class name & RF & CCF & 2D CNN & Hybrid SN & Two Branch CNN & Fus At Net & Two Headed Dragons & MAHiD FNet & Coupled CNN  & CCR Net & Hyper Fuse Net\\
 \hline
    Forest&81.90 & 77.66 & 81.61 & 75.34 & 82.34 & 82.49 & \textbf{85.98} & 82.08 & 75.08 & 60.49 & 84.93 \\
    Residential area &0     & 57.79 & 62.77 & 50.89 & 65.94 & 62.97 & 61.42 & 64.56 & 70.86 & 86.22 & \textbf{76.71} \\
    Industrial area &87.61 & 52.08 & 43.47 & 53.18 & 46.36 & 57.46 & 36.16 & 49.77 & 67.92 & 22.84 & \textbf{69.92} \\
    Low plants & 0     & 83.01 & 77.29 & 81.12 & 80.30  & 73.37 & \textbf{83.42} & 70.42 & 78.64 & 55.08 & 77.45 \\
    Soil & 68.27 & 80.13 & 84.88 & 87.97 & 80.88 & 85.48 & 87.65 & 86.76 & \textbf{87.82} & 69.29 & 75.89 \\
    Allotment & 0     & 69.51 & 80.33 & 75.78 & 67.93 & 77.13 & \textbf{82.30} & 81.60  & 80.05 & 11.63 & 39.34 \\
    Commercial area &0     & 34.09 & 50.44 & 28.35 & 41.02 & 39.77 & \textbf{52.93} & 41.36 & 15.45 & 13.96 & 16.99 \\
    Water & 61.50  & 68.69 & 75.12 & 73.25 & 72.27 & 76.76 & 80.36 & 73.70  & \textbf{84.84} & 28.98 & 74.55 \\
    \hline
    Overall accuracy (OA) & 16.67 & 63.05 & 66.88 & 58.91 & 68.40  & 66.56 & 67.46 & 66.96 & 70.37 &69.22  & \textbf{73.21} \\
    Average accuracy (AA) & 37.41 & 65.34 & 69.49 & 65.73 & 67.5  & 69.43 & 71.28 & 68.78 & 70.08 & 49.81 & \textbf{71.28} \\
    \textcolor{black}{Kappa ($\kappa$)}&\textcolor{black}{0.1195} & \textcolor{black}{0.5016} & \textcolor{black}{0.5470} & \textcolor{black}{0.4647} & \textcolor{black}{0.5607} & \textcolor{black}{0.5448} & \textcolor{black}{0.5595} & \textcolor{black}{0.5465} & \textcolor{black}{0.5800} & \textcolor{black}{0.4356} & \textbf{\textcolor{black}{0.6060}}\\
     \hline
    \end{tabular}}
\end{sidewaystable*}

\subsection{Results and Discussion}
\begin{table}[ht]
\centering{\scriptsize
 \caption{\label{tab:tab_mcnmr_nn}Z - values obtained from McNemar's test between proposed method and other models.}
\begin{tabular}{|p{3.0cm}|c|c|c|}
 \hline
Method& Houston 13 & Houston 18 & TU Berlin \\
\hline
    RF    & 44.32 & 999.42 & 493.55 \\
    CCF   & 31.29 & 343.95 & 129.58 \\
    2D CNN & 26.33 & 181.90 & 93.77 \\
    HybridSN & 34.41 & 381.03 & 190.49 \\
    Two Branch CNN & 17.23 & 164.48 & 73.21 \\
    FusAtNet  & 16.23 & 156.41 & 101.18 \\
    Two Headed Dragons & 22.20 & 98.09 & 81.70 \\
    MAHiDFNet & 18.57 & 146.00 & 93.61 \\
    CCNN  & 18.72 & 133.67 & 51.61 \\
    CCRNet & 18.31 & 194.04 & 114.85 \\
 \hline
\end{tabular}}
\end{table}

\begin{figure}[t!]
  \centering
  \centerline{\includegraphics[width=8.5cm]{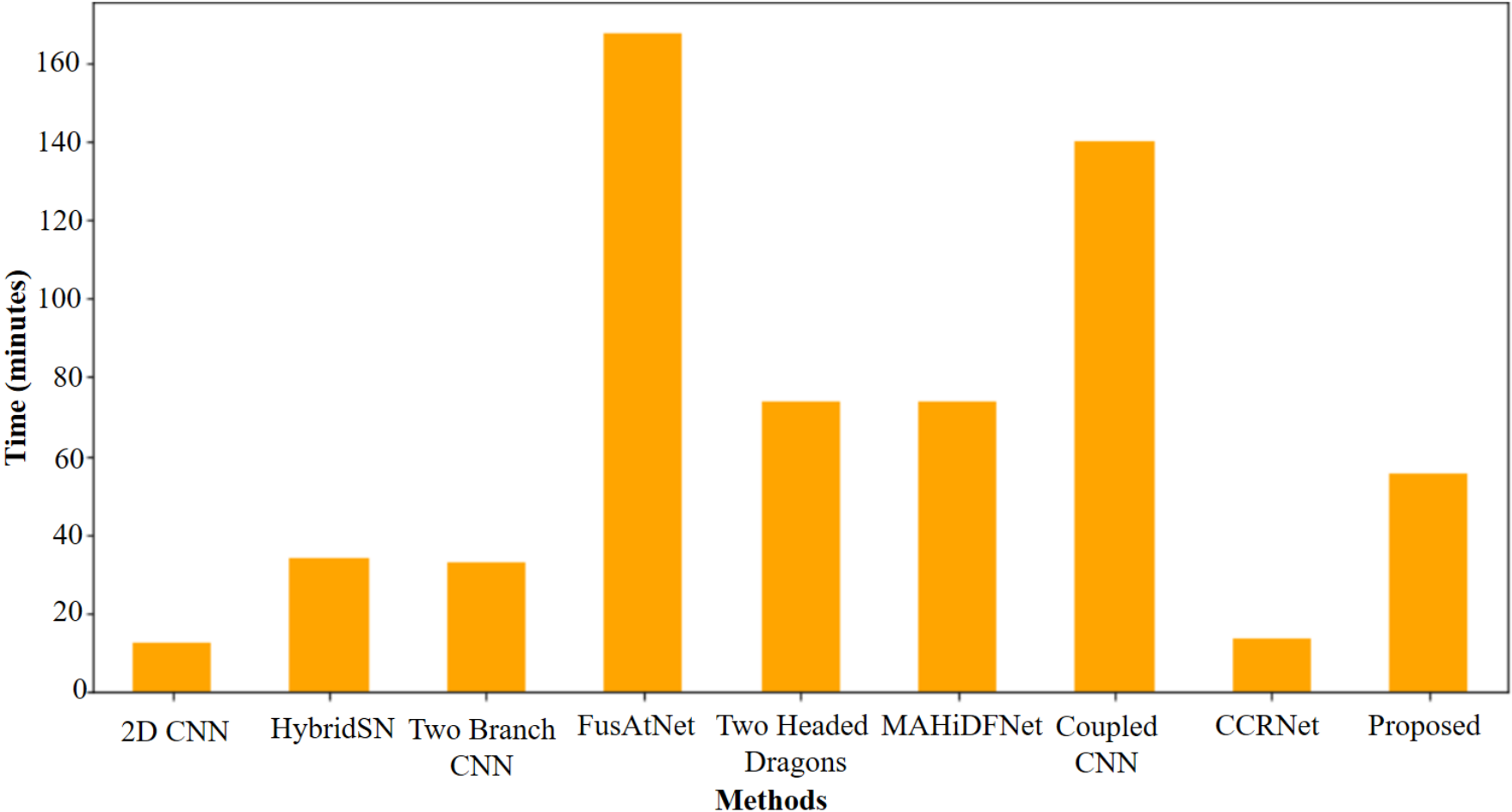}}
  \caption{\textcolor{black}{Time taken for training the deep learning models.}}\medskip
  \vspace{-0.5cm}
  \label{fig:time}
\end{figure}

The tables (Tables \ref{tab:hous13_perf_nn}, \ref{tab:hous18_perf_nn}, and \ref{tab:tub_perf}) present the overall accuracy (OA), classwise accuracy (PA), and Cohen's kappa ($\kappa$) for the Houston 13, Houston 18, and TU Berlin datasets, respectively. Our method demonstrates superior performance compared to other methods in terms of OA (93.08\%, 84.59\%, and 73.21\%) and $\kappa$ (0.9219, 0.8036, and 0.6060) for the three datasets, as evident from the tables. Regarding average accuracy, our model outperforms other models for the Houston 13 and TU Berlin datasets, while it exhibits similar performance to other models for the Houston 18 dataset. In terms of PA, our models consistently achieve satisfactory accuracy for most classes and outperform other models in nine classes (stressed grass, synthetic grass, soil, commercial, road, highway, railway, tennis court, and running track) for the Houston 13 dataset. Similarly, for the Houston 18 dataset, we achieve better performance for seven classes (healthy grass, artificial turf, bare earth, residential buildings, non-residential buildings, and roads), and for the TU Berlin dataset, the highest accuracy is obtained for two classes (residential area and industrial area). Furthermore, based on the values of the z-statistic (as obtained from Eqn. \ref{equation:MN_nn}) presented in Table \ref{tab:tab_mcnmr_nn}, our method statistically outperforms all the other models, as the z-statistic values significantly exceed 1.96 for all the models.

\begin{table}[ht]
\centering{\scriptsize
 \caption{\label{tab:para_nn}Approximate number of parameters for the deep models.}
\begin{tabular}{|p{3.0cm}|p{2.8cm}|}
 \hline
    Method & Approximate number of parameters (in 1000s)\\
    \hline
    CNN2D & 480 \\
    HybridSN & 1350 \\
    Two Branch CNN & 3000 \\
    FusAtNet & 38110 \\
    Two Headed Dragons & 3100 \\
    MAHiDFNet & 77700 \\
    Coupled CNN & 140 \\
    CCRNet & 72 \\
    Proposed & 1800 \\
 \hline
\end{tabular}}
\end{table}

We have conducted a qualitative analysis of the classification results by presenting the classification maps for the Houston 13, Houston 18, and TU-Berlin datasets in Figs. \ref{fig:cmH13}, \ref{fig:cmH18}, and \ref{fig:cmTUB}, respectively. In Figure \ref{fig:cmH13}, which corresponds to the Houston 13 dataset, the classification map obtained from our method exhibits smoother boundaries and minimal speckle compared to other methods. Additionally, we have highlighted a specific area (shown in red) in the image that includes a cloud shadow over a residential/commercial area (visibly identified in Figure \ref{fig:cmH13} (a)). Most methods classify this area as water, despite utilizing both spectral and depth information. However, our method correctly identifies the area as a combination of land-based classes, predominantly residential/commercial classes. This demonstrates that our method effectively captures the spectral intricacies of the images when combined with complementary depth information. 

For the Houston 18 dataset, we present the classification map for the entire image in Figure \ref{fig:cmH18}. Once again, the proposed method produces smoother classification boundaries. However, due to the large size of the image, we present the classification map of a subset of the Houston 18 image in Figure \ref{fig:cmH18_sub}. In Figure \ref{fig:cmH18_sub} (a), we highlight the specific area in red, and its zoomed-in version is shown in Figure \ref{fig:cmH18_sub} (b). In both cases, we can observe smooth delineation between multiple classes. Moreover, within the yellow rectangle in Figure \ref{fig:cmH18_sub} (b), we notice a field/ground class that is absent from the ground truth (Figure \ref{fig:cmH18_sub} (c)). The classification maps reveal that most methods incorrectly identify this class, labeling it as \textit{train} (Figures \ref{fig:cmH18_sub} (d), (e), and (f)) or \textit{water} (Figures \ref{fig:cmH18_sub} (i) and (k)). In contrast, our method consistently assigns it to a land-based class (\textit{highway}) based on spectral and depth information. 

Finally, we present the third classification map for the TU-Berlin dataset in Figure \ref{fig:cmTUB}. Similarly, the classification maps exhibit reduced speckle and sharper, more distinct boundaries compared to other methods.

\begin{table}[ht]
\centering{\scriptsize
 \caption[Ablation study by removing different modules in the proposed framework (all accuracies in \%).]{\label{tab:baseline}Ablation study by removing different modules in the proposed framework (all accuracies in \%). We observe that as the modules are removed, the accuracy show a decreasing trend for all the datasets.}
\begin{tabular}{|c|c|c|c|c|c|}
 \hline
Coupling& SL connection&CMR& Houston 13 & Houston 18 & TU Berlin\\
\hline
    \xmark     & \xmark     & \xmark     & 92.58 & 82.26 & 66.11 \\
    \xmark     & \xmark     & \cmark     & 91.49 & 82.23 & 72.91 \\
    \xmark     & \cmark     & \xmark     & 91.88 & 82.47 & 68.76 \\
    \cmark     & \xmark     & \xmark     & 91.19 & 83.31 & 66.82 \\
    \cmark     & \cmark     & \xmark     & 90.56 & 83.46 & 66.47 \\
    \cmark     & \xmark     & \cmark     & 92.88 & 83.95 & 66.08 \\
    \xmark     & \cmark     & \cmark     & 93.06 & 83.95 & 66.60 \\
    \cmark     & \cmark     & \cmark     & \textbf{93.08} & \textbf{84.59} & \textbf{73.21} \\
 \hline
\end{tabular}}
\end{table}

\begin{figure*}[t!]
  \centering
  \centerline{\includegraphics[width=14cm]{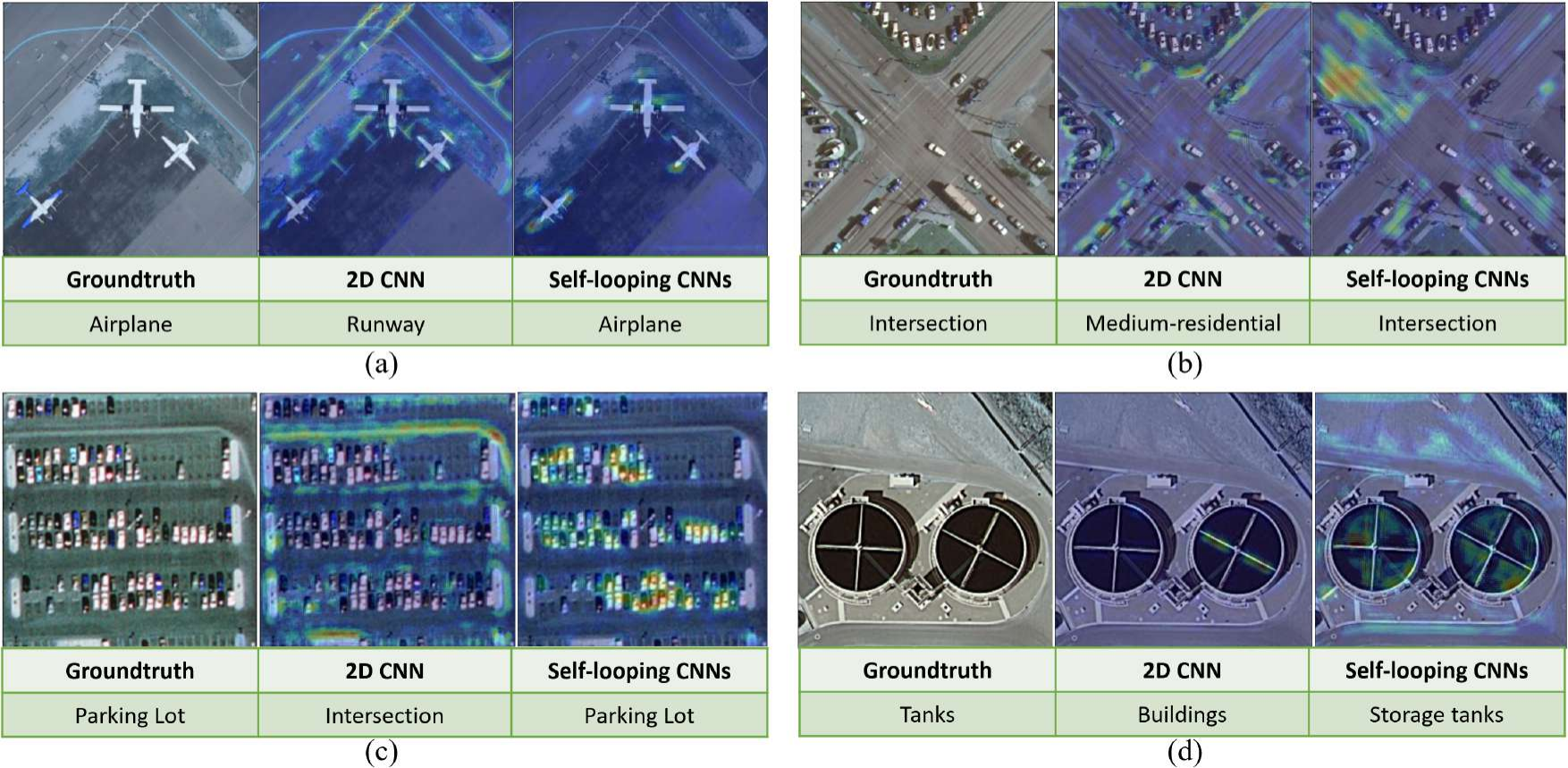}}
  \caption[GradCam representation on UC-Merced dataset to show the effect of self-looping convolutions in comparison to conventional convolution layers]{GradCam representation on UC-Merced dataset to show the effect of self-looping convolutions in comparison to conventional convolution layers (a) Airplane class (b) Intersection class (c) Parking-lot class (d) Storage tanks class.}\medskip
  \vspace{-0.5cm}
  \label{fig:gc}
\end{figure*}

\begin{figure}[t!]
  \centering
  \centerline{\includegraphics[width=8.5cm]{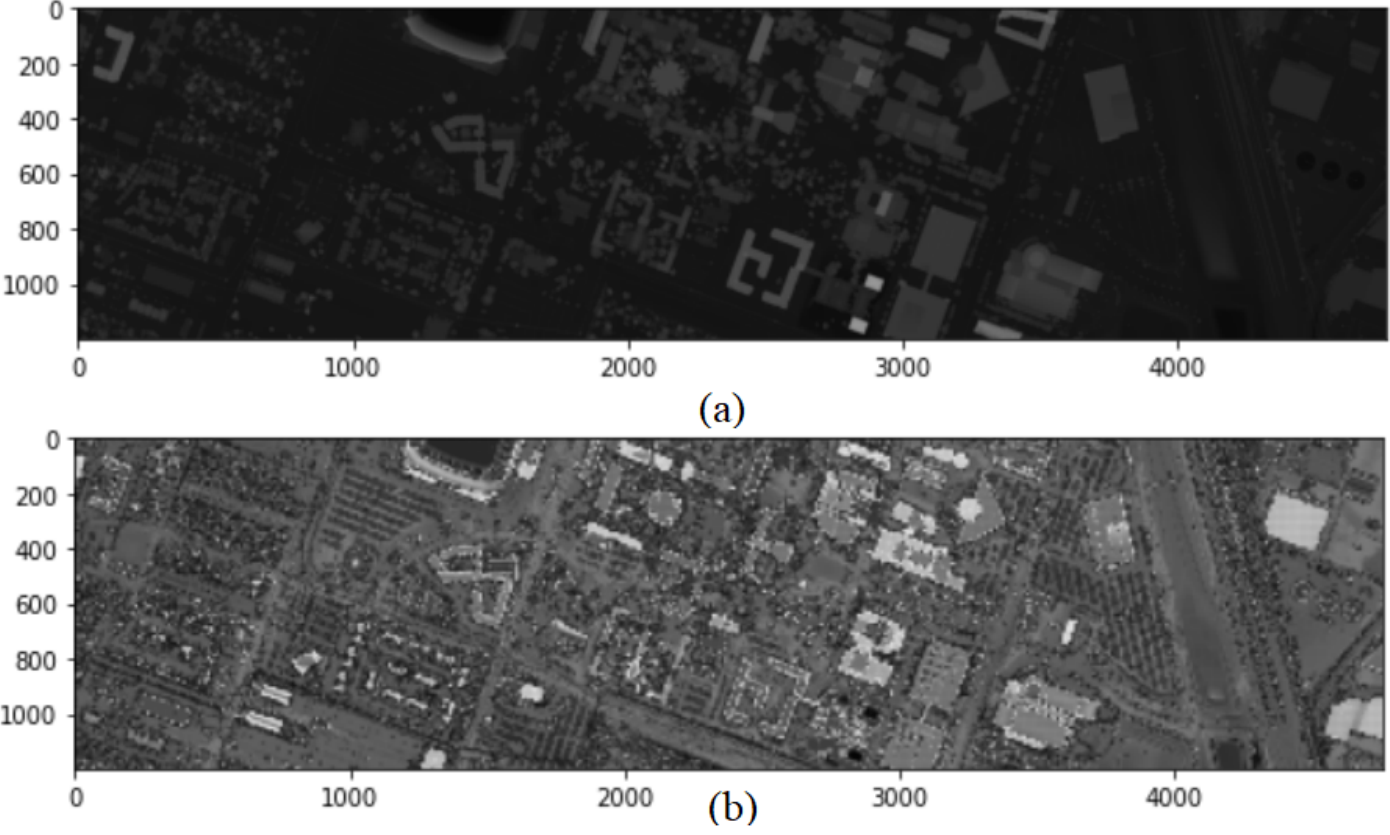}}
  \caption{\textcolor{black}{LiDAR digital surface model (DSM) reconstruction performance on Houston 2018 dataset. (a) DSM of LiDAR data. (b) Reconstructed DSM from the proposed model.}}\medskip
  \vspace{-0.5cm}
  \label{fig:rec_map}
\end{figure}

\begin{figure*}[t!]
  \centering
  \centerline{\includegraphics[width=14cm]{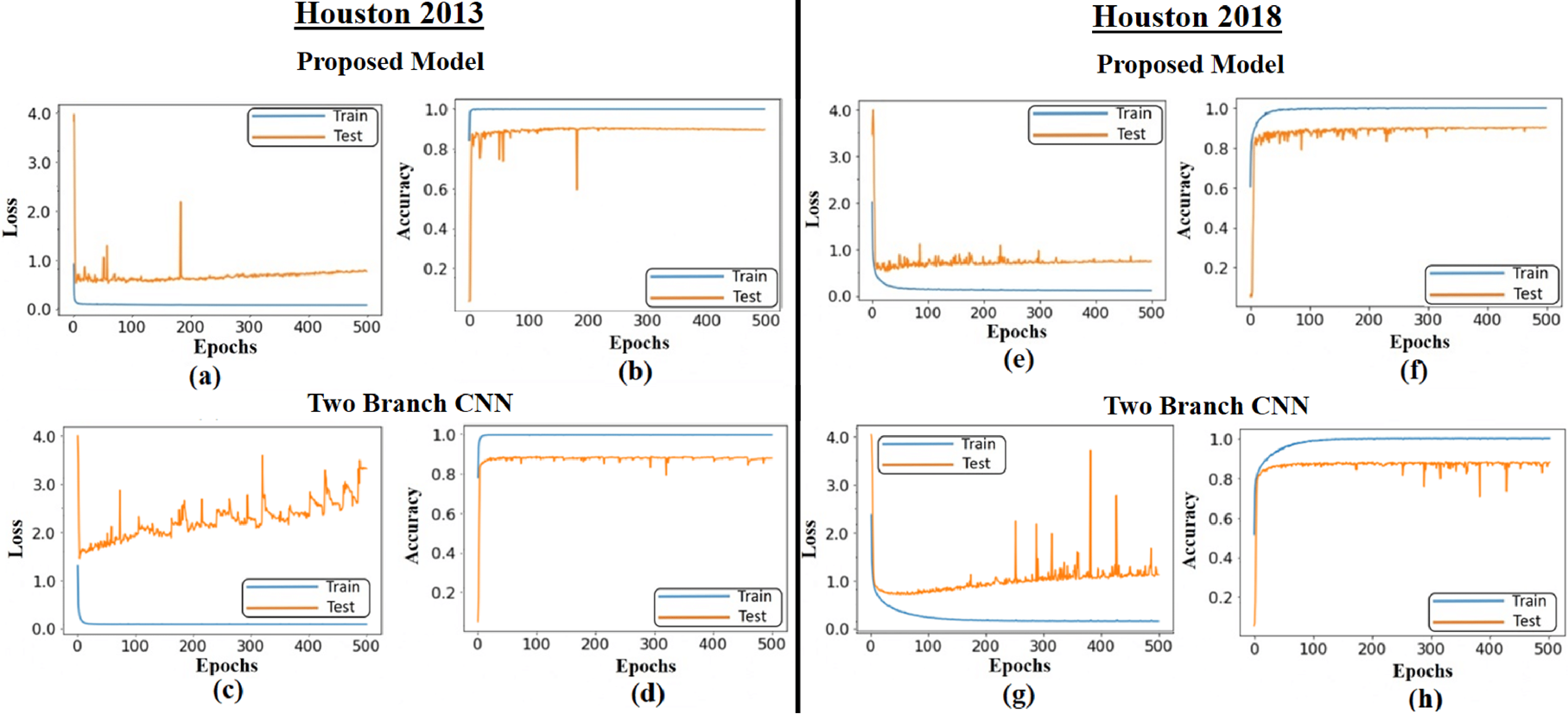}}
  \caption[Loss and accuracy curves for the proposed fusion method.]{\textcolor{black}{\textbf{Houston 2013 dataset.} (a) and (c) Displayed are the loss curves obtained using our proposed model and the Two Branch CNN model, correspondingly. (b) and (d) Represent the accuracy curves achieved using our proposed model and the Two Branch CNN model, correspondingly. The evaluation is conducted on the \textbf{Houston 2018 dataset}. (e) and (g) Illustrate the loss curves obtained using our proposed model and the Two Branch CNN model, respectively. (f) and (h) Show the accuracy curves obtained using our proposed model and the Two Branch CNN model, respectively.}}\medskip
  \vspace{-0.5cm}
  \label{fig:la_nn}
\end{figure*}

To further demonstrate the effectiveness of the \textit{self-looping} connections, we provide GradCAM visualizations \citep{selvaraju2016grad} in Figure \ref{fig:gc}. Since the hyperspectral images utilize a smaller patch size of 11$\times$11 where visualization is impractical, we present the GradCAM visualizations using the UC-Merced remote sensing dataset (with RGB channels) \citep{yang2010bag}. Each image in the dataset has dimensions of 256$\times$256$\times$3, and it consists of a total of 21 land-use/land-cover classes with 100 samples per class.

In Figure \ref{fig:gc} (a), we observe that the self-looping convolutions exhibit a stronger focus on the `airplanes' in the image, whereas the 2D CNN primarily focuses on the road and misidentifies the entire image as a `runway'. In Figure \ref{fig:gc} (b), the 2D CNN incorrectly labels the image as `medium-residential' instead of `intersection'. Figure \ref{fig:gc} (c) demonstrates that the 2D CNN predominantly focuses on the roads and misclassifies the image as an `intersection', while the self-looping convolutions pay attention to both cars and roads, leading to the correct identification of the image as a `parking-lot'. Similarly, in Figure \ref{fig:gc} (d), our model accurately captures the circular region in the image and correctly identifies it as `storage tanks'.

\begin{table}[ht]
\centering{\scriptsize
 \caption[Ablation study with different number of channels (all accuracies in \%).]{\label{tab:feat_abl_nn} Ablation study with different number of channels (all accuracies in \%). It can be observed that the accuracy has decreased as the number of channels decrease from 64 to 32.}
\begin{tabular}{|c|c|c|c|}
 \hline
& Houston 13 & Houston 18 & TU Berlin\\
\hline
32 channels per layer &90.32 & 81.44 & 67.05\\
64 channels per layer &\textbf{93.08} & \textbf{84.59} & \textbf{73.21}\\
 \hline
\end{tabular}}
\end{table}
\begin{table}[ht]
\centering{\scriptsize
 \caption[Ablation study to show difference in performance when parameters are not shared across time and modalities (all accuracies in \%).]{\label{tab:par_abl}\textcolor{black}{Ablation study to show difference in performance when parameters are not shared across time and modalities (all accuracies in \%). When there is no communication of future information to the past and across modalities using shared weights, the performance decreases.}}
\begin{tabular}{|c|c|c|c|}
 \hline
& Houston 13 & Houston 18 & TU Berlin\\
\hline
Without parameter sharing &91.49 & 82.23 & 72.91\\
With parameter sharing &\textbf{93.08} & \textbf{84.59} & \textbf{73.21}\\
 \hline
\end{tabular}}
\end{table}
\begin{table}[ht]
\centering{\scriptsize
 \caption[Ablation study for different degrees of augmentation (all accuracies in \%).]{\label{tab:aug_abl}Ablation study for different degrees of augmentation (all accuracies in \%). As the augmentation is reduced from full to partial to none, the accuracy also decreases.}
\begin{tabular}{|c|c|c|c|}
 \hline
& Houston 13 & Houston 18 & TU Berlin\\
\hline
No augmentation &88.38 & 81.50 & 68.85\\
Partial augmentation &90.36 & 83.69 & 72.28\\
Full augmentation &\textbf{93.08} & \textbf{84.59} & \textbf{73.21}\\
 \hline
\end{tabular}}
\end{table}
\begin{table}[ht]
\centering{\scriptsize
 \caption[Ablation study with different sizes of kernels for convolution (all accuracies in \%).]{\label{tab:aug_35}Ablation study with different sizes of kernels for convolution (all accuracies in \%). We observe that 3$\times$3 convolution is more contributing in better classification than 5$\times$5 convolutions.}
\begin{tabular}{|c|c|c|c|}
 \hline
& Houston 13 & Houston 18 & TU Berlin\\
\hline
Only 3$\times$3 convolution &91.36 & 83.46 & 71.32\\
Only 5$\times$5 convolution &90.52 & 82.40 & 67.53\\
Both convolutions&\textbf{93.08} & \textbf{84.59} & \textbf{73.21}\\
 \hline
\end{tabular}}
\end{table}
\begin{table}[ht]
\centering{\scriptsize
 \caption[Ablation study with different patch sizes (all accuracies in \%).]{\label{tab:patch_79}Ablation study with different patch sizes (all accuracies in \%). We observe that with higher patch size, the accuracy is higher for all the datasets on account of higher spatial information.}
\begin{tabular}{|c|c|c|c|}
 \hline
Patch size& Houston 13 & Houston 18 & TU Berlin\\
\hline
7$\times$7 &87.41 & 80.61 & 65.89\\
9$\times$9 &90.24 & 81.97 & 67.30\\
11$\times$11&\textbf{93.08} & \textbf{84.59} & \textbf{73.21}\\
 \hline
\end{tabular}}
\end{table}
\begin{table}[ht]
\centering{\scriptsize
 \caption[Ablation study for different values of $\lambda_1$ and $\lambda_2$ in the final loss function for Houston 13 dataset.]{\label{tab:aug_lam}Ablation study for different values of $\lambda_1$ and $\lambda_2$ in the final loss function for Houston 13 dataset. It is clearly visible that when both $\lambda_1$ and $\lambda_2$ are equal to 1, we have achieved maximum accuracy.}
\begin{tabular}{|c|c|c|}
 \hline
$\lambda_1$ & $\lambda_2$ & Accuracy (\%)\\
\hline
1&0.2&90.83\\
1&0.4&91.61\\
1&0.6&91.17\\
1&0.8&92.66\\
0.2&1&91.59\\
0.4&1&91.89\\
0.6&1&91.82\\
0.8&1&92.65\\
0.2&0.2&91.68\\
0.4&0.4&90.90\\
0.6&0.6&90.72\\
0.8&0.8&91.05\\
1&1&\textbf{93.08}\\
 \hline
\end{tabular}}
\end{table}

\subsubsection{Ablation studies}

Ablation studies for our proposed model are also included for all the datasets. Table \ref{tab:baseline} presents the ablation study where different components of the model are removed, and their performances are observed. It is evident that when all the components are included, the proposed model achieves the best performance for all the datasets. As the components are gradually removed, the decrease in accuracy values becomes more noticeable. For the Houston 13 dataset, it is observed that coupling is not as effective as other modules and can be compensated by the inclusion of self-looping connections and cross-modal reconstruction. A similar trend can be observed in other datasets as well. Furthermore, optimizing multiple modules simultaneously in the network is a challenging task that can lead to stagnation in a local minimum. This is observed in the Houston 13 dataset, where a decent accuracy of 92.29\% is achieved without any additional module. Similarly, for the TU-Berlin dataset, an accuracy of 72.91\% is obtained with only the self-supervised cross-modal reconstruction module.

An ablation study on changing the number of channels in the convolution blocks is presented in Table \ref{tab:feat_abl_nn}. It is observed that as the number of channels decreases to 32, the accuracy consistently decreases as well for all three datasets. \textcolor{black}{To emphasize the effect of incorporating shared parameters across time and modalities, the results are separately presented in Table \ref{tab:par_abl}. It is clearly observed that the models achieve the best performances on all three datasets when the same parameters are shared across time and modalities, facilitating maximum intra-model communication.}

The effect of changes in the degree of data augmentation is presented in Table \ref{tab:aug_abl}. As the augmentation is reduced from full to none, the accuracy uniformly decreases for all the datasets. Here, full augmentation refers to including all rotations (90\degree, 180\degree, and 270\degree) of the training samples, partial augmentation refers to rotating the training samples only by 180\degree, while no augmentation refers to no rotation of the training samples.

Additionally, an ablation study on the size of the convolution kernels in MCSL blocks is presented in Table \ref{tab:aug_35}. From the table, it can be observed that the contribution of the 3$\times$3 convolution is greater than that of the 5$\times$5 convolution. This could be attributed to the relatively lower resolution of the datasets. Table \ref{tab:patch_79} presents a complementary study highlighting the effect of patch sizes on the algorithm. It is evident that as the patch sizes decrease from 11$\times$11 to 9$\times$9 and 7$\times$7, the accuracy correspondingly decreases. This behavior is due to the inclusion of higher spatial information with increasing patch size.

\begin{table}[ht]
\centering{\scriptsize
 \caption[Ablation study for decreasing the number of samples and check model's performance.]{\label{tab:num_samp}Ablation study for decreasing the number of samples and check model's performance. It is visible that even when the number of samples are reduced to 50\% and 25\%, the model still performs satisfactorily well for Houston 13 dataset. For Houston 2018 dataset, the performance drop is significantly more.} 
\begin{tabular}{|c|c|c|c|}
 \hline
& 25\% & 50\% &100\% \\
\hline
Houston 13 &89.26 & 90.83 & \textbf{93.08}\\
Houston 18 &73.47 & 79.70 & \textbf{84.59}\\
\hline
\end{tabular}}
\end{table}

\begin{table}[ht]
\centering{\scriptsize
 \caption[Ablation study for self-reconstruction and cross-reconstruction (all accuracies in \%).]{\label{tab:self}Ablation study for self-reconstruction and cross-reconstruction (all accuracies in \%). It is evident that the accuracy is much less in self-reconstruction that shows the significance of cross-reconstruction for fusion.}
\begin{tabular}{|c|c|c|c|}
 \hline
& Houston 13 & Houston 18 & TU Berlin\\
\hline
Self-reconstruction &89.24 & 82.81 & 65.96\\
Cross-reconstruction &\textbf{93.08} & \textbf{84.59} & \textbf{73.21}\\
\hline
\end{tabular}}
\end{table}

\begin{table}[ht]
\centering{\scriptsize
 \caption[Ablation study for feature based cross-reconstruction and modality based cross-reconstruction (all accuracies in \%).]{\label{tab:feat_rec}Ablation study for feature based cross-reconstruction and modality based cross-reconstruction (all accuracies in \%). It is visible that our method surpasses the feature-based cross reconstruction module used in CCRNet \citep{wu2021convolutional}.}
\begin{tabular}{|c|c|c|c|}
 \hline
& Houston 13 & Houston 18 & TU Berlin\\
\hline
CCRNet module & 92.44 & 81.81 &70.56\\
Proposed &\textbf{93.08} & \textbf{84.59} & \textbf{73.21}\\
\hline
\end{tabular}}
\end{table}

Table \ref{tab:aug_lam} demonstrates the performance variation of the model when changing the values of $\lambda_1$ and $\lambda_2$ to balance the reconstruction-based losses (presented on the Houston 13 dataset). It is evident that the model performs best when both $\lambda_1$ and $\lambda_2$ are fixed at 1.

Furthermore, a sensitivity analysis regarding the number of training samples has been conducted in Table \ref{tab:num_samp}. It is clear that as the number of samples decreases from 100\% to 50\% and 25\%, the performance drop is less noticeable in the Houston 2013 dataset. However, the performance drop is more pronounced in the Houston 2018 dataset, which can be attributed to its higher spatial variation.

The significance of the cross-reconstruction approach is also demonstrated in comparison to the self-reconstruction approach (where the same modality is reconstructed as the input modality) in Table \ref{tab:self}. It is evident that the model's performance in cross-reconstruction is significantly better than in self-reconstruction.

Table \ref{tab:feat_rec} presents an ablation study that compares the cross-reconstruction module presented in the CCRNet paper \citep{wu2021convolutional} with the one in our proposed model. In the ablation study, our self-supervision-based reconstruction module is replaced with the one used in CCRNet, while keeping all other parameters constant. It is clearly visible that our proposed module still surpasses the former module, thereby reinforcing its novelty and superior performance.

\section{Summary}

This chapter focuses on processing hyperspectral images in the multimodal setting, with three research works primarily addressing the extrapolation of missing modalities and efficient fusion of HSI with LiDAR and SAR using modality-specific feature extractors. The first work deals with modality distillation in RS image classification. It considers the scenario where all modalities are available during training but some modalities are missing for test data. The teacher-student framework is followed, where the teacher model is trained on multi-modal information. The student model has two tasks: (i) learning to hallucinate missing modalities from available ones, and (ii) training the student's classifier using available and hallucinated modalities. A novel C-GAN based model is proposed for modality hallucination, ensuring the generation of distinctive samples. Extensive experiments on RS datasets demonstrate the efficacy of the model. The focus now shifts towards exploring zero-shot knowledge distillation, where pseudo samples are learned to train the student, as opposed to using the training samples deployed to train the target.

In the second work, we introduce a novel fusion network called FusAtNet, aimed at producing improved land-cover maps by combining HSI and LiDAR data. The proposed fusion network, FusAtNet, judiciously utilizes different attention learning modules to learn joint feature representations given both input modalities. To achieve this, we propose the concept of cross-attention, where the feature learning stream of one modality is influenced by the other modality. The results obtained from multiple datasets confirm the efficacy of the proposed fusion network. Due to the generic nature of FusAtNet, it can be easily extended to support a wide range of modalities with minimal overhead. In the future, we plan to expand the network to support more than two modalities and conduct rigorous model engineering to reduce the number of learnable parameters without compromising performance, such as incorporating the concept of dilated convolution in the attention modules effectively.

Finally, we introduce a novel framework for multimodal fusion in the remote sensing domain, utilizing self-supervision based cross-modal reconstruction and coupled self-looping convolutional blocks. The key strength of this approach lies in parameter sharing across both the temporal domain and modalities. This sharing of parameters enhances the robustness of convolutions during feature extraction by leveraging information from the future to augment information from the past. Additionally, by incorporating coupling, we can extract features that incorporate representations from both fused modalities. To further enhance feature robustness, we propose the concept of self-supervised learning. Through an auxiliary task of cross-modal reconstruction, we utilize pre-fusion features from one modality to reconstruct the other, resulting in more informative pre-fusion abstractions that reflect the joint characteristics of both modalities. we have evaluated the proposed approach on HSI-LiDAR datasets (Houston 2013 and Houston 2018) as well as a HSI-SAR dataset (TU-Berlin), where it outperforms existing state-of-the-art methods. Moving forward, the future plans include incorporating more than two modalities for fusion, optimizing the number of parameters for improved efficiency, and extending the work to handle unknown classes in open-set scenarios during the testing phase.

\section{Publications}

\begin{enumerate}
    \item Shivam Pande, Avinandan Banerjee, Biplab Banerjee, Subhasis Chaudhuri, An Adversarial Approach to Discriminative Modality Distillation for Remote Sensing Image Classification, \textit{CroMoL, ICCV workshop, 2019}.
    \item Satyam Mohla, Shivam Pande, Biplab Banerjee, Subhasis Chaudhuri, FusAtNet: Dual Attention Based SpectroSpatial Multimodal Fusion network for Hyperspectral and LiDAR Classification, \textit{PBVS, CVPR workshop, 2020}.
    \item Shivam Pande, Biplab Banerjee, Self-supervision Assisted Multimodal Remote Sensing Image Classification with Coupled Self-Looping Convolution Networks, \textit{Neural Networks, 2023}. 
\end{enumerate}
\chapter{Self-supervised learning in hyperspectral image analysis}

\section{Introduction}

The utilization of various modalities in remote sensing has increased alongside advancements in the field. Hyperspectral images (HSIs) are among these modalities frequently employed for landuse/land cover classification due to their high spectral resolution, enabling better reflectance recognition. However, the large number of bands in HSIs makes their processing burdensome, as they generate a substantial amount of data (commonly referred to as the ``curse of dimensionality'') \citep{rasti2020feature}. Currently, deep learning (DL) algorithms are extensively used for data processing tasks, especially on a large scale. Although DL techniques have shown exceptional performance with big data, many applications, particularly supervised learning, demand a significant number of annotated samples. Since manual annotation is labor-intensive, it is crucial to develop techniques that can achieve satisfactory performance with limited samples. Self-supervised and semi-supervised learning have emerged as effective learning paradigms for limited label learning, where models are trained by leveraging the inherent information in unlabeled data \citep{wang2022self}.

SSL techniques have been applied in remote sensing and hyperspectral imaging domains in various contexts. For instance, dimensionality reduction of HSIs has predominantly been addressed through autoencoder-based techniques. Discriminative learning-based approaches have also been employed, wherein models can be trained on pretext tasks during the pretraining stage. These tasks include matching image/patch augmentations for similar patches, identifying image/patch rotations, or rearranging image patches as a jigsaw puzzle \citep{wang2022self}.

From the perspective of generative learning, an influential work in SSL is presented in \citep{mou2017unsupervised}, where the authors propose a generative SSL approach for HSI classification. In this approach, the network is initially pretrained as an autoencoder to reconstruct the original 3D HSI patch. Subsequently, the encoder part of the pretrained model is utilized for classifying HSI samples. Another study by \citep{tao2015unsupervised} employs a sparse-stacked autoencoder (SSAE) to obtain a low-dimensional representation of HSI features. Similarly, \citep{mei2019unsupervised} introduces a method called 3DCAE, which utilizes an autoencoder based on 3D convolutions and deconvolutions to learn the spectral-spatial mapping from a high-dimensional space to a low-dimensional feature space.

To tackle the issue of high dimensionality, various methods from statistics, machine learning, and deep learning are extensively employed. Traditional techniques like principal component analysis (PCA), kernel principal component analysis (kPCA), and discriminant analysis feature extraction \citep{rasti2020feature} have conventionally been used for dimensionality reduction. In the present scenario, advanced techniques such as autoencoders are also being incorporated. These techniques leverage neural networks to project high-dimensional data onto a lower-dimensional subspace, and the low-dimensional representation is then used to reconstruct the original features. The success of dimensionality reduction is measured by the proximity between the original and reconstructed features \citep{pande2021attention}. To fulfill specific requirements, different types of autoencoders have been introduced, including those based on convolutional neural networks (CNNs), attention mechanisms, and recurrent neural networks (RNNs), among others \citep{pande2021attention}.

Among the aforementioned approaches, CNN-based autoencoders have demonstrated outstanding performance due to their ability to handle both spatial and spectral information. In particular, 1D CNNs have been specifically employed to process HSI bands, as they excel in handling sequential data \citep{chen2016deep}. However, they suffer from a limitation in which they assign equal importance to all bands or features, regardless of their contribution to image classification. To overcome this issue, researchers in the computer vision field introduced the concept of attention mechanisms for images \citep{woo2018cbam}. The idea behind attention mechanisms is to selectively highlight the most important regions of the images while suppressing the rest.

Another drawback of conventional CNNs is that they only have forward connections, which limits their ability to generate robust representations in later stages. This is because information from future layers is not effectively utilized to update the information in earlier layers. To address this issue, \citep{yang2018convolutional} proposed the inclusion of feedback connections in CNNs for image classification. Within the network, \textit{loopy} structures are created among CNN layers using shared weights, allowing information from later layers to be passed back to earlier layers for obtaining more refined features. The advantage of this approach is that, with the same number of weights, the features across layers can be significantly improved. This concept has also been extended to the domain of hyperspectral image classification, where \textit{loopy} convolutions have been utilized in a multi-scale setting \citep{pande2022hyperloopnet}.

Furthermore, in order to exploit the sequential nature of HSIs, recurrent neural network (RNN) based models have been extensively explored. However, conventional RNNs suffer from the problem of vanishing gradients when dealing with longer sequences, which hinders effective training \citep{pu2016variational}. Moreover, RNNs are limited in scope as they process data in a single direction and fail to capture information from future states \citep{schuster1997bidirectional}. The former issue is addressed by utilizing a variation of RNNs known as gated recurrent units (GRUs) \citep{pu2016variational}. GRUs incorporate gating mechanisms that determine the amount of information used to update the next cell state. The latter problem is tackled by introducing a bidirectional structure in RNNs \citep{schuster1997bidirectional}, where information flows both forward and backward simultaneously, enabling consideration of both past and future states.

From a discriminative learning perspective, recent works have also explored the idea of SSL for HSI classification. For instance, \citep{braham2022self} proposed a self-supervised learning method inspired by the Barlow Twins approach. Their method involves pretraining by minimizing the distance between the cross-correlation matrix of the embeddings from two augmented views and the ideal correlation matrix of HSI samples. Similarly, \citep{li2022robyol} applied the concept of ``bootstrap your own latent (BYOL)'' to HSI classification and introduced the use of random occlusion for augmentation in multiple views.

Most of the aforementioned works rely solely on the concept of self-supervised learning. However, \citep{assran2021semi} proposed the idea of providing additional guiding information for pretraining by leveraging the training samples available in the ground truth. Their approach focuses on minimizing a consistency loss between pseudo-labels assigned to different augmented views, based on the labeled samples from the ground truth (referred to as PAWS). Building upon this concept, we aim to explore the potential of semi-supervised learning for HSI classification in our current research work.

Based on the aforementioned challenges, our research work incorporates SSL-based techniques in both dimensionality reduction and discriminative learning aspects. Our contributions are as follows:

\begin{itemize}
    \item We devise a spectral-spatial autoencoder framework that leverages residual connections with 3D CNNs to reduce the dimensionality of HSIs. This is explained in section \ref{sec:SSDR} and the work is published as in \cite{pande2020dimensionality}.
    \item We introduce a novel approach called AttAE, where attention modules reinforce 1D convolutions. This enables the model to explicitly focus on important features, leading to improved low-dimensional representations (section \ref{sec:AttAE}), published as \cite{pande2021attention}.
    \item Additionally, we propose FBAE, a 1D convolution-based autoencoder with successive feedback convolution blocks. Within each block, feedback connections ensure that features from later layers refine earlier layers, resulting in more robust representations (section \ref{sec:FBAE}), published as \cite{pande2022feedback}.
    \item We also propose BiGRUAE, an autoencoder based on Bidirectional GRU, specifically for reducing HSI dimensions. To the best of our knowledge, this is the first application of Bidirectional GRUs for dimensionality reduction in HSI. The proposed model effectively processes HSIs by capitalizing on their sequential nature, offers an enhanced field of view by considering both past and future states, and maintains a continuous gradient flow due to gating mechanisms (section \ref{sec:BGAE}), published as \cite{pande2021bidirectional}.
    \item We extend the concept of PAWS to hyperspectral image classification and compare its performance with supervised and self-supervised algorithms. Additionally, skip connections inspired by a DenseNet structure are incorporated to ensure consistent gradient flow within the network during training (section \ref{sec:PAWS}), published as \cite{pande2023semi}.
    \item The proposed approaches are evaluated on several benchmark hyperspectral datasets, including Indian Pines 1992, Indian Pines 2010, Salinas Valley, and Houston 2013 datasets.
\end{itemize}

\section{Dimensionality reduction in hyperspectral images}

This section would focus on the dimensionality reduction techniques on hyperspectral images both from spectral and spatial-spectral perspective using autoencoders. The proposed autencoders utilize different deep learning based feature extractors. For spectral-spatial case, we have used residual connections based 3D CNNs, while for spectral feature extractors, different sequential methods like 1D attenional CNNs, bidirectional GRUs and feedback networks are used.  

\subsection{Spectral-spatial dimensionality reduction}
\label{sec:SSDR}
\subsubsection{Methodology}

\begin{figure*}[t!]
  \centering
  \centerline{\includegraphics[width=15cm]{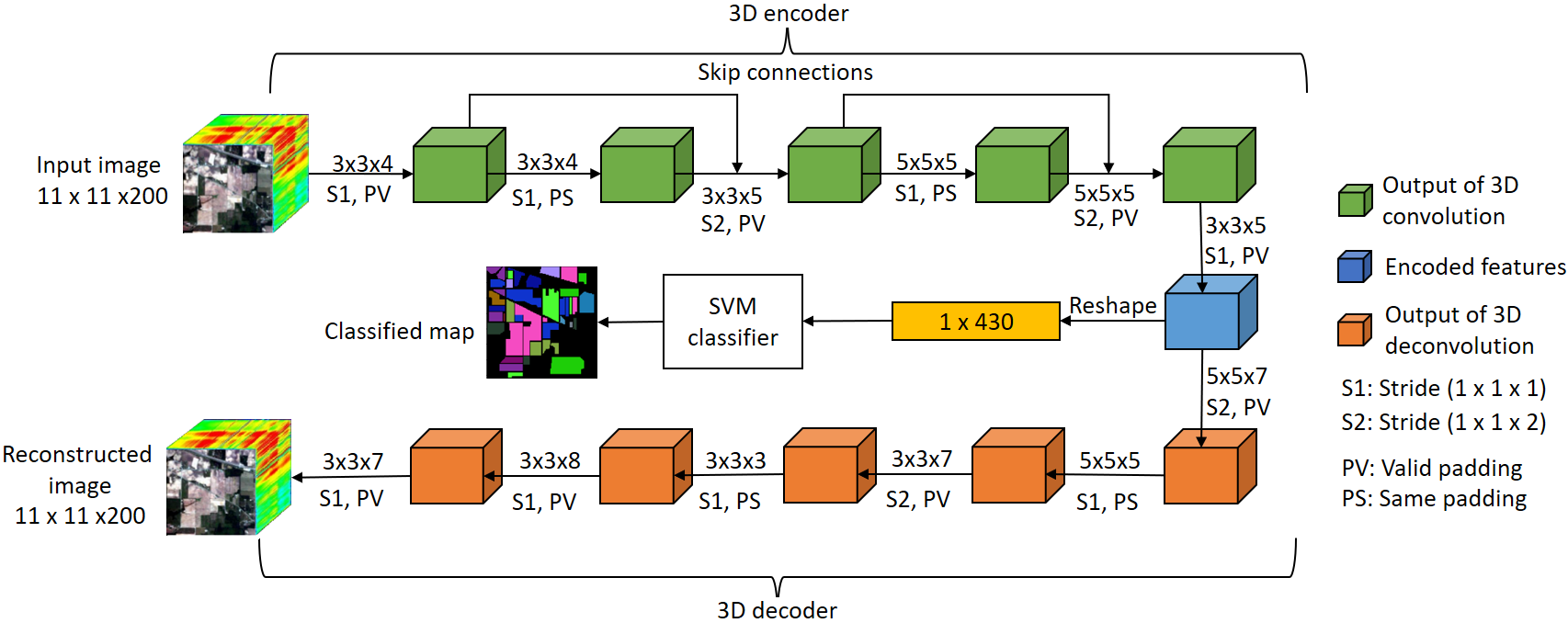}}
  \caption{Schematic of 3D residual autoencoder (3D ResAE), presented on Indian pines dataset.}\medskip
  \vspace{-0.5cm}
  \label{fig:Diagram_3DResAE}
\end{figure*}

This section describes the method of 3D CNNs and residual connections based autoencoders\footnote{Published as: Shivam Pande, Biplab Banerjee. \href{https://ieeexplore.ieee.org/abstract/document/9323359}{Dimensionality Reduction Using 3D Residual Autoencoder for Hyperspectral Image Classification}, In: International Geoscience and Remote Sensing Symposium, Sep., 2020, 2029-2032.}. The schematic of the proposed approach can be referred in Figure \ref{fig:Diagram_3DResAE}.

\noindent\textbf{Problem definition}

Let's consider the hyperspectral dataset $\mathcal{X}=\{\{\textbf{x}_{S}^i\}_{i=1}^n \cup \{\textbf{x}_{U}^j\}_{j=1}^m\}$ and $\mathcal{Y}=\{y^i\}_{i=1}^n$, where $\textbf{x}_{S}^i, \textbf{x}_{U}^j \in \mathbb{R}^{M\times N \times B \times T}$ and $\textbf{x}_{S} \cap \textbf{x}_{U} \neq 0$. Here, $\textbf{x}_{S}^i$ represents the $i^{th}$ patch from the imagery with an available ground truth label. $y^i$ refers to the corresponding label, and $n$ is the total number of such samples. $\textbf{x}_{U}^j$ represents the $j^{th}$ patch without any associated ground truth label, and $m$ is the number of such samples. The values of $M$ and $N$ represent the number of rows and columns respectively for each patch, while $B$ represents the number of bands, and $T$ represents the number of feature cubes. In the case of HSI, the value of $T$ is fixed at $1$ when the data is fed in its original shape. The 3D ResAE is trained on $\textbf{x}_{U}^j$, and then $\textbf{x}_{S}^i$ is input to the trained model to obtain the projected features, which are subsequently used for classification.

\noindent\textbf{Model Architecture}
The model consists of two main components: an encoder utilizing 3D convolutions and skip connections, and a decoder incorporating 3D deconvolutions. The specific characteristics of the model's architecture are elaborated upon in the following section:

\noindent\textit{\textbf{Encoder}} $E$: The encoder comprises a 3D CNN with six layers, where each layer consists of a 3D convolution \citep{mei2019unsupervised}, followed by a \textit{ReLU} activation function and a batch normalization (BN) operation (Figure \ref{fig:Diagram_3DResAE}). The convolution operation can be expressed as:
\begin{equation}
h_{kl}^{efg} = \sum_{t}^{T} \sum_{p=0}^{P-1} \sum_{q=0}^{Q-1} \sum_{r=0}^{R-1} w_{lt}^{pqr}z_{(k-1)t}^{(e.s_{e}+p)(f.s_{f}+q)(g.s_{g}+r)}
\label{equation:3DCNN}
\end{equation}
Here, $e$, $f$, and $g$ are the indices referring to the output cube $h$ in the three dimensions (for the input $z$). Similarly, $k$ and $l$ represent the indices corresponding to the convolution layer and the kernel within that layer, respectively. The dimensions of the 3D convolution kernel $w$ are denoted by $P$, $Q$, and $R$, with $p$, $q$, and $r$ representing the indices in each dimension. The variable $t$ represents the number of feature cubes $T$. Additionally, $s_{e}$, $s_{f}$, and $s_{g}$ denote the strides in the three dimensions. To establish skip connections, concatenation operations are employed after the second and fourth convolutions, which can be expressed as follows:
\begin{equation}
z_{k} =  \sigma(b+z_{k-1} \odot w) \oplus z_{k-1}
\label{equation:skip}
\end{equation}
In this context, $\odot$ and $\oplus$ represent the 3D convolution and concatenation operations, respectively. The symbol $b$ corresponds to the bias term, while $\sigma$ represents the combined \textit{ReLU} and BN operations. The output of the final convolution layer is denoted as $E(\textbf{x}_{U}^j)$, which represents the encoded features in a lower dimensional space.

\noindent\textit{\textbf{Decoder}} $D$: The network consists of a 3D deconvolution architecture with five layers. Each layer comprises a 3D deconvolution filter, followed by a \textit{ReLU} non-linearity and a BN operation (similar to $E$). Deconvolution refers to the process of upsampling an input sample from a low-dimensional space to a high-dimensional space \citep{mei2019unsupervised}. The output of $D$ corresponds to a reconstructed sample with the same dimensions and size as the input sample. To ensure that the low-dimensional encoded features accurately represent the spectral-spatial characteristics of the high-dimensional features, the entire autoencoder framework is trained end-to-end using a reconstruction loss based on mean squared error. This loss, denoted as $\mathcal{L}_R$, measures the discrepancy between the reconstructed features $D(E(\textbf{x}_{U}^j))$ and the original input.
\begin{equation}
\label{eqn:reconstruction}
{\mathcal{L}_R = \frac{1}{m}\sum_{j=1}^{m}{\|\textbf{x}_{U}^j}-D(E(\textbf{x}_{U}^j))}\|_2^2
\end{equation}

\noindent The trained 3D ResAE model receives patches $\textbf{x}_{S}$, which are encoded as $E(\textbf{x}_{S})$, and then forwarded for classification using a support vector machine (SVM) classifier with a radial basis function kernel \citep{melgani2004classification}. The complete network pipeline is illustrated in Figure \ref{fig:Diagram}. The values associated with the arrows represent the dimensions of the convolution-deconvolution kernels, the stride sizes, and the type of padding utilized. For the Indian pines dataset, the input patches have dimensions of 11$\times$11$\times$200 (24200 features) and are encoded into 430 dimensions. Conversely, for the Salinas dataset, the patches have a fixed size of 11$\times$11$\times$204 (24684 features), which are encoded into 220 dimensions. The selection of the number of encoded dimensions is determined empirically to achieve the highest possible accuracy while minimizing the dimensionality. Further details on the datasets can be found in Section \ref{sec:datasets}.

\subsubsection{Datasets and Experiments}
\label{sec:datasets}
This section provides a comprehensive explanation of the utilized datasets, the corresponding experimental procedures, and the results and discussions associated with them.

\noindent\textbf{Datasets}

Two benchmark hyperspectral datasets, obtained through the AVIRIS sensor, have been employed to validate the proposed model.

\begin{figure}[htb]
\begin{minipage}[b]{1.0\linewidth}
  \centering
  \centerline{\includegraphics[width=8.5cm]{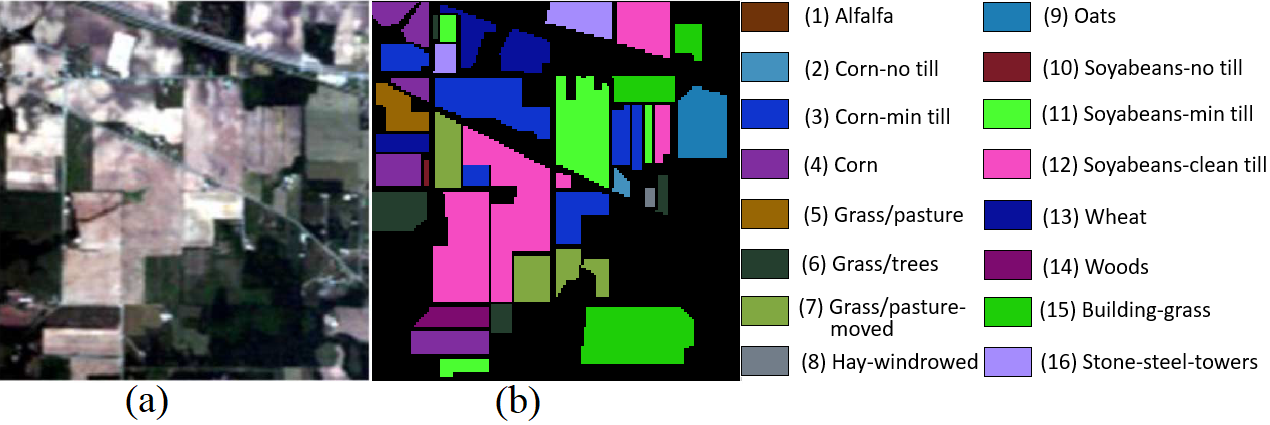}}
\end{minipage}
\caption{Indian pines hyperspectral dataset. (a) Colour composite of 3 bands from the image (b) Groundtruth with classes.}
\label{fig:ip_data}
\end{figure}

\noindent\textit{Indian pines (IP)}: The Indian Pines dataset has a spatial resolution of 20 meters and comprises 200 bands, with each band having a size of 145$\times$145 pixels. The ground truth information is available for 16 classes, encompassing a total of 10,249 pixels (Figure \ref{fig:ip_data}) \citep{mei2019unsupervised}.

\begin{figure}[htb]
\begin{minipage}[b]{1.0\linewidth}
  \centering
  \centerline{\includegraphics[width=8.5cm]{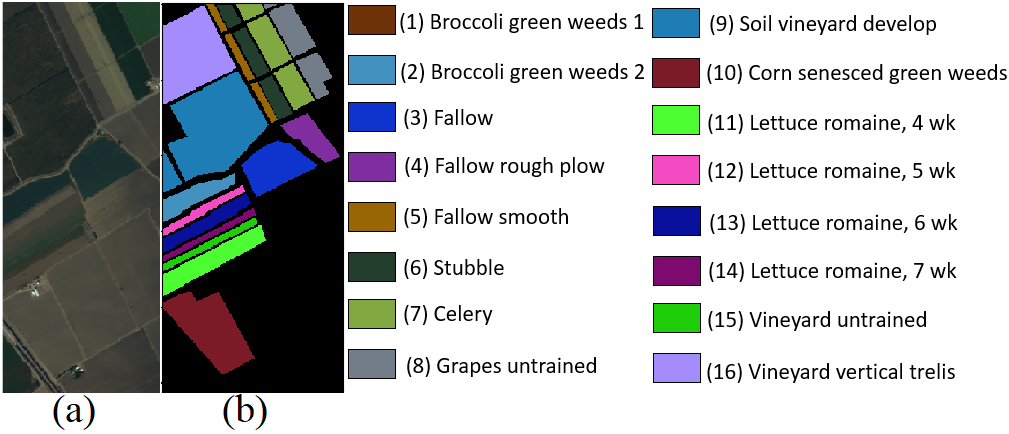}}
\end{minipage}
\caption{Salinas Valley hyperspectral dataset. (a) Colour composite of 3 bands from the image (b) Groundtruth with classes.}
\label{fig:sal_data_ch4}
\end{figure}

\noindent\textit{Salinas Valley}: The Salinas Valley dataset is composed of 204 bands with a spatial resolution of 3.7 meters. Each band consists of 512 rows and 217 columns. The dataset contains a total of 16 classes distributed over 54,129 pixels (Figure \ref{fig:sal_data_ch4}) \citep{mei2019unsupervised}.

\noindent\textbf{Training protocols and evaluation} 

The training of the proposed 3D ResAE model follows a semi-supervised approach, where patches are sampled from the image using a window size of 11$\times$11. For the IP dataset, a fixed stride of (1,1) is used, resulting in 18,225 patches, while for the Salinas dataset, a stride of (2,2) creates 26,104 patches. To prevent overfitting, an L2-norm regularizer with a coefficient of 0.01 is applied to all the weights. The training process utilizes the Adam optimizer with Nesterov momentum \citep{dozat2016incorporating} on the Keras API \citep{chollet2015keras} and is performed on an Nvidia RTX 2060 GPU. The learning rate is set to 0.0001, and the training is conducted for 40 epochs.

For classification purposes, a subset of the ground truth patches is utilized. Specifically, 10\% of the ground truth patches (1,024 samples) with dimensions of 11$\times$11$\times$200 are employed for the IP dataset, while for the Salinas dataset, 5\% of the ground truth patches (2,706 samples) with dimensions of 11$\times$11$\times$204 are selected. The classification task is accomplished using SVM on MATLAB 2019a \citep{MATLAB:2019} with 10-fold cross-validation.

The proposed model is compared to state-of-the-art models, including TPCA \citep{ren2017hyperspectral}, SSAE \citep{tao2015unsupervised}, EPLS \citep{romero2015unsupervised}, and 3DCAE \citep{mei2019unsupervised}. The evaluation metrics used include overall accuracy (OA), producer's accuracy (PA), and average accuracy (AA). Additionally, t-SNE plots of the original and encoded features are generated to visually examine the learned features.

\subsubsection{Results and Discussion}

\begin{figure}[htb]
\begin{minipage}[b]{1.0\linewidth}
  \centering
  \centerline{\includegraphics[width=8.0cm]{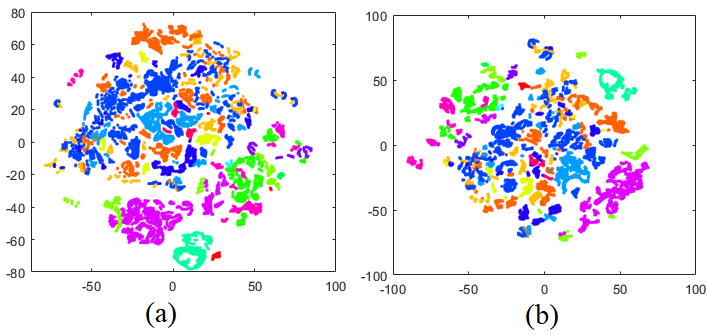}}
\end{minipage}
\caption{t-SNE of test samples of Indian pines dataset (a) original features (b) encoded features.}
\label{fig:tsne}
\end{figure}

Tables \ref{tab:ip_perf} and \ref{tab:sal_perf_ch4} provide an analysis of the performance of our proposed model compared to other state-of-the-art methods for dimensionality reduction \citep{mei2019unsupervised}. The results show that our approach achieves the highest overall accuracy (OA) and average accuracy (AA) for the Salinas dataset, reaching 97.25\% and 97.52\% respectively. For the IP dataset, our model achieves the highest OA (93.26\%), while the AA is second to that of the 3DCAE method. In terms of producer's accuracy (PA), our method attains the highest accuracy for 7 classes in the IP dataset and 5 classes in the Salinas dataset. Our model also demonstrates competitive performance for classes with relatively lower accuracy. Additionally, several experiments were conducted by varying the percentage of patches used in training the 3D ResAE model. Table \ref{tab:perc_patch} presents the performance of our model on the Indian pines dataset using 33\%, 50\%, 67\%, and 100\% of the patches. It can be observed that as the number of training samples decreases, the classification accuracy also decreases, indicating that the 3D ResAE model struggles to efficiently capture the entire data space with fewer samples. Furthermore, Figure \ref{fig:tsne} depicts the t-SNE visualization of the original and encoded features of the test samples from the IP dataset. The visualization demonstrates that even with reduced dimensions, similar classes are clustered together and distinct from other classes.

\begin{table}[ht]
\centering{\scriptsize
 \caption{\label{tab:ip_perf} Accuracy analysis on the IP dataset (\%).}
\begin{tabular}{ |p{0.7cm}| p{0.9cm} |p{0.9cm} |p{0.9cm} |p{1.0cm} |p{1.5cm}|}
 \hline
Class & TPCA & SSAE & EPLS & 3DCAE & 3D ResAE\\
 \hline
1   &  60.97 & 56.25 &  58.72 &90.48 & \textbf{95.12} \\
2   & 87.00  &69.58 &  59.91  & \textbf{92.49} & 90.82\\
3   & \textbf{94.51} & 75.36 & 71.34 & 90.37 & 88.22\\
4   &  79.34 & 64.58 &  74.31  & 86.90 & \textbf{87.79}\\
5   &  93.08 & 88.81 &  \textbf{97.95} & 94.25 & 94.02\\
6   &  96.34 & 87.00  & 96.44 & 97.07 & \textbf{97.11}\\
7   &  76.00 & 90.00 &  54.02  & \textbf{91.26} & 88.00\\
8   &  99.76  & 89.72 & 88.99  & 97.79 & \textbf{100.00}\\
9   & \textbf{100.00}  & \textbf{100.00} & 58.89  & 75.91 & 66.67\\
10  & 79.71 & 77.19 & 73.10  & \textbf{87.34} & 85.83\\
11  & 85.42 & 77.58 & 70.78 & 90.24 & \textbf{94.89}\\
12  & 84.24 & 72.00 &  57.51 & \textbf{95.76} & 87.27\\
13  & 98.91 & 87.80 & \textbf{99.25} & 97.49 & 97.84 \\
14  &  98.06 & 93.48 & 95.07  & 96.03 & \textbf{98.68}\\
15  &  87.31  & 72.36 & 91.26  & 90.48 & \textbf{98.56}\\
16  &  96.38  & 97.22 & 91.27 & \textbf{98.82} &94.05\\
\hline
OA  &  88.55  & 79.78&  77.18 & 92.35 & \textbf{93.26}\\
AA  &  89.31  & 81.18 & 77.43 & \textbf{92.04} & 91.55 \\
\hline
\end{tabular}}
\end{table}

\begin{table}[ht]
\centering{\scriptsize
 \caption{\label{tab:sal_perf_ch4} Accuracy analysis on the Salinas dataset (\%).}
\begin{tabular}{ |p{0.7cm}| p{0.9cm} |p{0.9cm} |p{0.9cm} |p{1.0cm} |p{1.5cm}|}
 \hline
Class & TPCA  & SSAE   & EPLS  & 3DCAE & 3D ResAE\\
\hline
1  &  99.88 & \textbf{100.00} &  99.99 & \textbf{100.00} & 96.60 \\
2  & 99.49  & 99.52 &  \textbf{99.92}  & 99.29 & 99.24\\
3  & 99.04 & 94.24 & 98.75 & 97.13 & \textbf{99.41}\\
4  & \textbf{99.84} & 99.17 &  98.52  & 97.91 & 98.34\\
5  & 98.96 & 98.82 &  98.83 & 98.26 & \textbf{99.45}\\
6  & 99.80 & \textbf{100.00}  & 99.92 & 99.98 & \textbf{100.00}\\
7  & 99.84 & \textbf{99.94} &  97.69  & 99.64 & 97.47\\
8  & 84.11  & 80.73 & 78.86  & 91.58 & \textbf{95.90}\\
9  & \textbf{99.60}  & 99.47 & 99.54  & 99.28 & 98.74\\
10 & 95.76 & 92.12 & 95.98  & \textbf{96.65} & 96.63\\
11 & 96.14 & 96.62 & \textbf{98.60} & 97.74 & 93.69\\
12 & 99.07 & 97.75 &  \textbf{99.44} & 98.84 & 96.12\\
13 & \textbf{100.00} & 95.81 & 98.85 & 99.26 & 96.90 \\
14 &  95.74 & 96.65 & \textbf{98.56}  & 97.49 & 97.93\\
15 &  79.54  & 79.73 & 83.13  &87.85 & \textbf{94.57}\\
16 &  98.40  & 99.12 & \textbf{99.50} & 98.34 & 99.36\\
\hline
OA &  96.57  & 92.11 &  92.35 & 95.81 & \textbf{97.25}\\
AA &  93.24  & 95.61 & 96.55 & 97.45 & \textbf{97.52} \\
 \hline
\end{tabular}}
\end{table}

\begin{table}[ht]
\centering{\scriptsize
 \caption{\label{tab:perc_patch} Comparison of accuracies with respect to percentage of training patches on Indian pines dataset.}
\begin{tabular}{|p{3.3cm}| p{0.7cm} |p{0.7cm} |p{0.7cm} |p{0.7cm}|}
 \hline
Fraction of train patches ($\%$) & 33  & 50  & 67  & 100\\
 \hline
 \hline
 OA ($\%$) &  89.96  & 90.33 & 92.40 & \textbf{93.26}\\
 \hline
\end{tabular}}
\end{table}

\subsection{Spectral dimensionality reduction}
\label{sec:SDR}
\subsubsection{Methodology}

\noindent\textbf{Problem Definition}
\begin{figure*}[t!]
  \centering
  \centerline{\includegraphics[width=14cm]{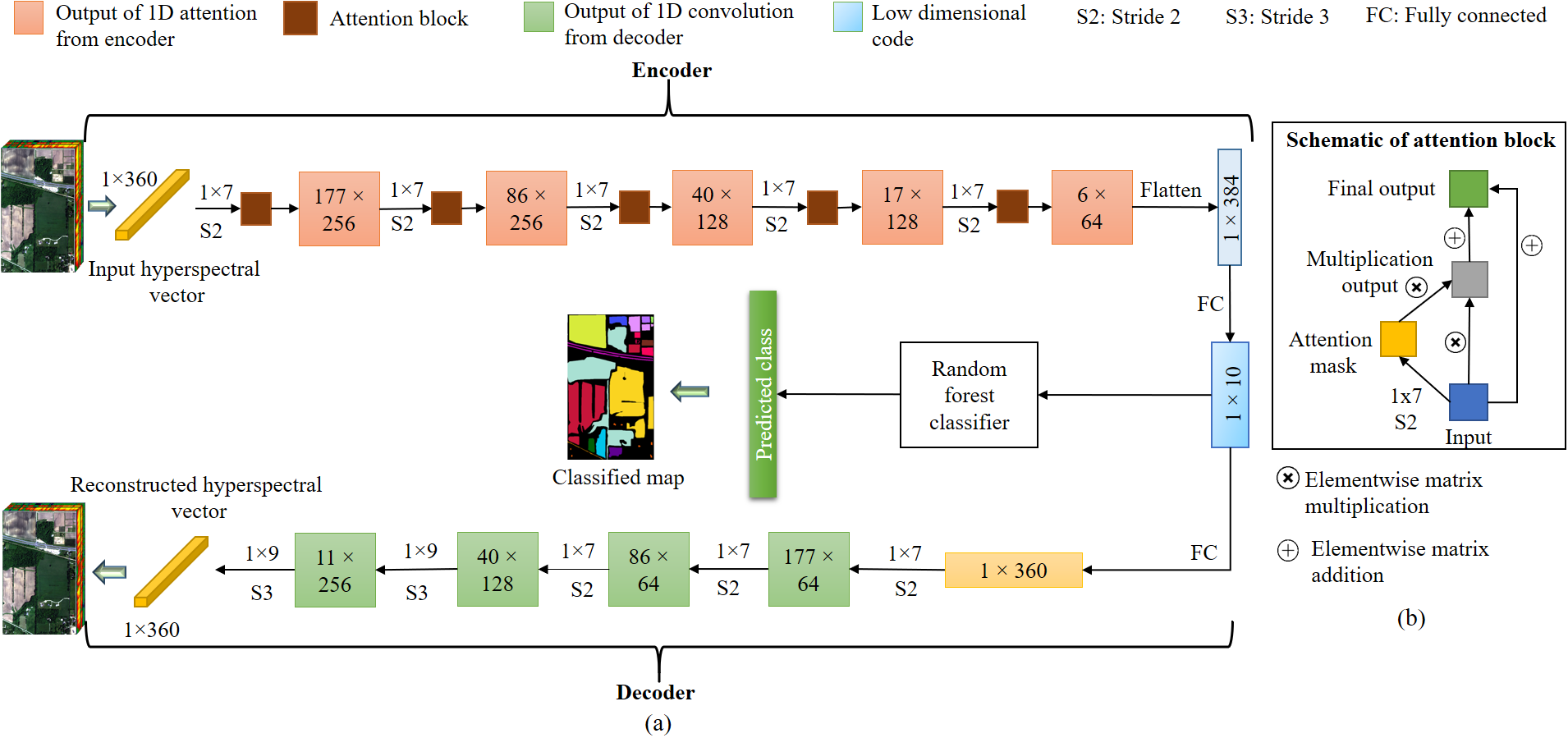}}

  \caption[Schematic of attention autoencoder (AttAE), presented on IP2010 dataset]{Schematic of attention autoencoder (AttAE), presented on IP2010 dataset. The convolution operations are denoted by the arrows, with kernel size mentioned above them. The output size is written inside the respective output block.}\medskip
  \vspace{-0.5cm}
  \label{fig:DiagramAttAE}
\end{figure*}
Let's assume we have a hyperspectral vector $\mathcal{X}$ represented as $\mathcal{X}=\{\textbf{x}^i\}_{i=1}^n$, where $\mathcal{X}\in \mathbb{R}^{1 \times B}$. The corresponding ground truth labels are denoted as $\mathcal{Y}=\{y^i\}_{i=1}^n$. In this context, $B$ represents the number of bands in the hyperspectral image (HSI), $i$ is the index of each sample, and $n$ denotes the total number of samples.

\subsubsection{Attention based autoencoder}
In this section, we will see the 1D CNNs based autoencoders enhanced with attention mechanism (called AttAE) for diemsnionality reduction HSIs. The framework is composed of encoder with attention enhanced CNNs to map the high dimensional spectral vector to a low dimensional representation, and a 1D CNN based decoder to reconstruct the original HSI vector.\footnote{Published as: Shivam Pande, Biplab Banerjee. \href{https://ieeexplore.ieee.org/abstract/document/9553019}{Attention Based Convolution Autoencoder for Dimensionality Reduction in Hyperspectral Images}, In: International Geoscience and Remote Sensing Symposium, Jul., 2021, 2727-2730.}.

\label{sec:AttAE}
\noindent\textbf{Model Architecture}
The depicted model, referred to as AttAE, is illustrated in Figure \ref{fig:DiagramAttAE}. It primarily consists of two components: an encoder responsible for converting the input features into a lower-dimensional code, and a decoder that reconstructs the original features using the encoded representation. The main objective is to ensure the similarity between the original and reconstructed features. The details of these components are elucidated individually below.

\noindent\underline{\textit{Encoder}} $E$: The encoder module comprises five modules that include 1D convolutions followed by ReLU activation and an attention block. The output of the fifth attention block is then passed through a fully connected layer to obtain the low-dimensional embedding. The operation of the 1D convolution is described by Equation \ref{equation:1DCNN}.
\begin{equation}
h_{ef}^{k} = \sigma \Bigg(b_{ef}+ \sum_{m} \sum_{p=0}^{P-1} w_{efm}^{p}a_{(e-1)m}^{(k+p)}\Bigg)
\label{equation:1DCNN}
\end{equation}

In Eq. \ref{equation:1DCNN}, $w$ and $b$ represent the convolution kernel and bias respectively, $h$ represents the output of convolution, $m$ and $k$ denote the index of input and output convolution maps respectively, $e$ and $f$ respectively represent the index of layer and kernel in that layer, $P$ stands for the size of convolution kernel while $a$ represents the output from the previous attention block. In addition, $\sigma$ is the ReLU activation function. 

The attention block takes the output of the previous convolution layer as input. Then, it applies its own 1D convolution to generate the attention mask of same size as that of the input. The resultant attention mask is then multiplied to the input features to get the attention enhanced features. These attention enhanced features are then added to the input features, so that the characteristics of the latter are incorporated in them as well. The resultant features are sent to the next convolution layer. The schematic of the attention block can be referred in Figure \ref{fig:DiagramAttAE} and the respective equation is shown in Eq. \ref{equation:att}, where, $\otimes$ indicates element-wise matrix multiplication while $\oplus$ denotes element-wise matrix addition. 
\begin{equation}
a_{e} = (h_{e} \otimes g_{e}({h_{e}})) \oplus h_{e}
\label{equation:att}
\end{equation}
In this context, the attention mask is denoted as $g_{e}({h_{e}})$, where $g_e$ represents the convolution operator within the $e^{th}$ attention block. The resulting low-dimensional encoded features can be represented as $E(\textbf{x}^i)$.

\noindent \underline{\textit{Decoder}} $D$: The decoder begins with a fully connected layer, followed by five 1D convolution layers. Both the fully connected and convolution layers utilize the ReLU non-linearity. The purpose of the decoder is to reconstruct the sample with the same dimension as the input. The output of the decoder is denoted as $D(E(\textbf{x}^i))$. To ensure similarity between the input and the reconstructed sample, our model is trained using the mean squared error loss, as depicted in Eq. \ref{eqn:reconstructionAtt}.
\begin{equation}
\label{eqn:reconstructionAtt}
{\mathcal{L}_R = \frac{1}{n}\sum_{i=1}^{n}{\|\textbf{x}^i}-D(E(\textbf{x}^i))}\|_2^2
\end{equation}


\subsubsection{BiDirectional GRU based autoencoder}
\label{sec:BGAE}
\begin{figure*}[t!]
  \centering
  \centerline{\includegraphics[width=14cm]{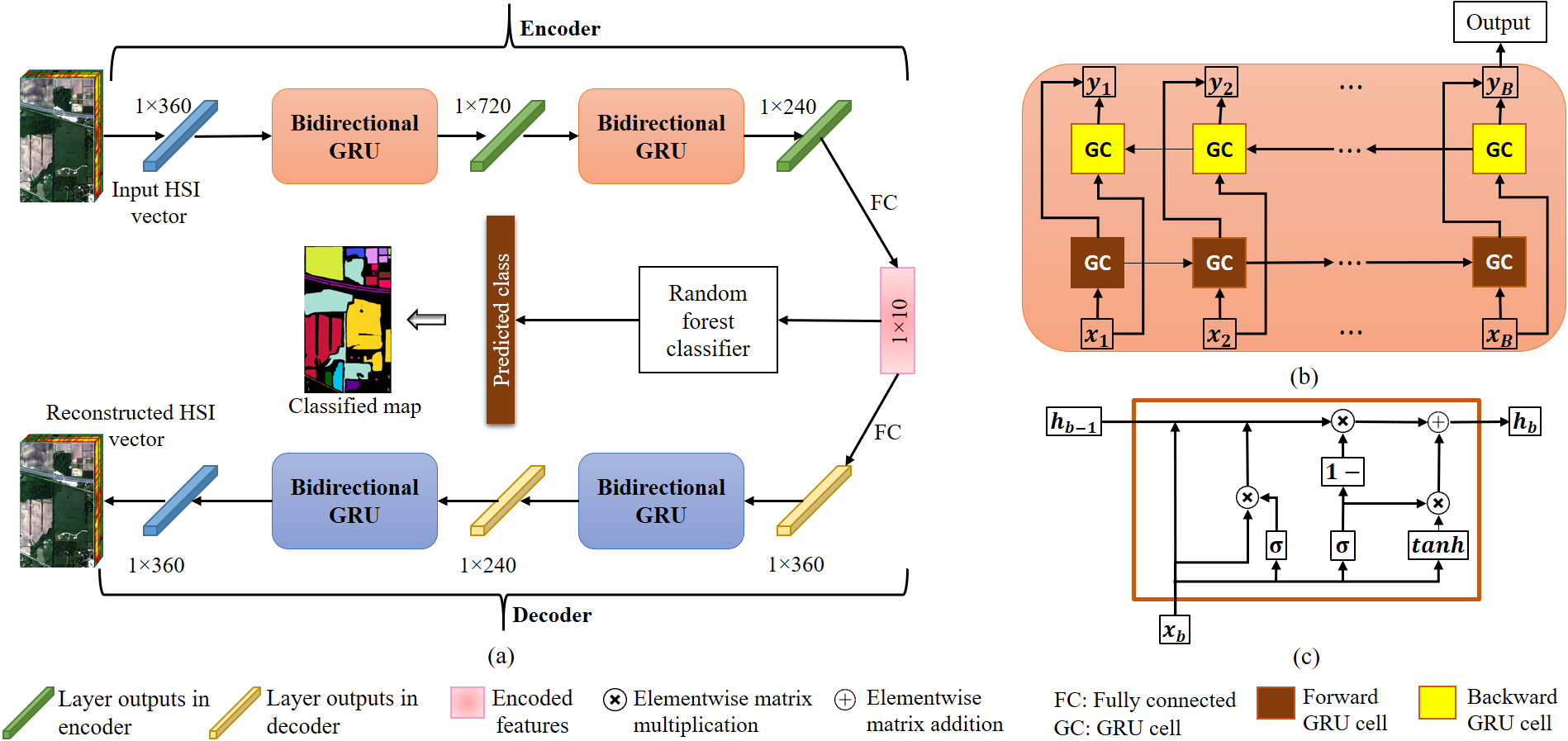}}
  \caption[Schematic of Bidirectional GRU based autoencoder (BiGRUAE), presented on IP2010 dataset]{(a) Schematic of Bidirectional GRU based autoencoder (BiGRUAE), presented on IP2010 dataset. The output size is written around the respective output. (b) Schematic of BiDirectional GRU. $y$ is the concatenated output of forward and backward states. (c) Schematic of a GRU cell. The $\sigma$ denotes \textit{sigmoid} activation.}\medskip
  \vspace{-0.5cm}
  \label{fig:DiagramBG}
\end{figure*}

The complete framework consists of a BiDirectional GRU based encoder, which maps the original HSI features to a low-dimensional code, and a decoder, which reconstructs the original features\footnote{Published as: Shivam Pande, Biplab Banerjee. \href{https://ieeexplore.ieee.org/abstract/document/9555048}{Bidirectional GRU Based Autoencoder for Dimensionality Reduction in Hyperspectral Images}, In: International Geoscience and Remote Sensing Symposium, Jul., 2021, 2731-2734.}. The details of these components are described in the following sections, and the architecture is depicted in Figure (\ref{fig:DiagramBG}).
\noindent\textbf{Model Architecture}

\noindent\underline{\textit{Encoder}} $E$: The encoder consists of two consecutive Bidirectional GRU (BiGRU) units, followed by a fully connected (FC) layer. The BiGRU utilizes the hyperbolic tangent activation function (\textit{tanh}) for the hidden states and the sigmoid activation function ($\sigma$) for the reset and update gates. The FC layer employs the Rectified Linear Unit activation function (\textit{ReLU}). The equations (\ref{equation:reset}) to (\ref{equation:hfinal}) demonstrate the operation of the Bidirectional GRU in the forward direction (indicated by the arrow at the top). In these equations, $x_b$ and $\overrightarrow{h}_{b-1}$ represent the values at the $b^{th}$ band and the $(b-1)^{th}$ state, respectively, while $\overrightarrow{r_b}$ and $\overrightarrow{u_b}$ denote the reset and update gates. The reset gate, as shown in Equation (\ref{equation:reset}), determines the amount of information from the previous state to retain, while the update gate, as depicted in Equation (\ref{equation:update}), determines the amount of information to pass to the next state. The reset gate produces a candidate state $\overrightarrow{c_{b}}$ (Equation (\ref{equation:cs})), which represents the resulting memory after retaining relevant information from the previous state. Finally, the previous state and candidate state are combined using the update gate to obtain the resulting state $\overrightarrow{h_b}$ (Equation (\ref{equation:hfinal})). The terms starting with $W$, $U$, and $v$ correspond to the weights for the current input, weights for the previous state, and bias for the respective equations, while $\otimes$ represents element-wise multiplication.
\begin{equation}
\overrightarrow{r_b} = \sigma(\overrightarrow{W_r}x_b+\overrightarrow{U_r}\overrightarrow{h}_{b-1}+\overrightarrow{v_r})
\label{equation:reset}
\end{equation}
\begin{equation}
\overrightarrow{u_b} = \sigma(\overrightarrow{W_u}x_b+\overrightarrow{U_u}\overrightarrow{h}_{b-1}+\overrightarrow{v_u})
\label{equation:update}
\end{equation}
\begin{equation}
\overrightarrow{c_b} = tanh(\overrightarrow{W_c}x_b+\overrightarrow{U_c}(\overrightarrow{r_b} \otimes \overrightarrow{h}_{b-1})+\overrightarrow{v_c})
\label{equation:cs}
\end{equation}
\begin{equation}
\overrightarrow{h_b} = \overrightarrow{u_b} \otimes \overrightarrow{h}_{b-1} + (1-\overrightarrow{u_b}) \otimes \overrightarrow{c_b}
\label{equation:hfinal}
\end{equation}
The equations following the aforementioned pattern can be employed to obtain the equations for the backward direction. The $b-1$ terms will be substituted with $b+1$ terms, and the direction of the arrows will be reversed. The resulting states will be concatenated as $h_b = [\overrightarrow{h_b}; \overleftarrow{h_b}]$ and forwarded to the subsequent layer. The output from the second layer is passed to the FC layer, which produces the resulting low-dimensional representation $E(\textbf{x}^i)$.

\noindent\underline{\textit{Decoder}} $D$: The decoder consists of a FC layer followed by two GRU layers, which are similar to the ones employed in the encoder. The output of the decoder is a hyperspectral vector represented as $D(E(\textbf{x}^i))$, possessing the same dimensions as the original vector. The entire BiGRUAE model is trained using the mean squared error loss function to ensure the resemblance between the original and reconstructed hyperspectral vectors (Eq. (\ref{eqn:reconstructionBG})), where $\mathcal{L}_R$ denotes the loss function. This training objective guarantees the preservation of informative low-dimensional representations.
\begin{equation}
\label{eqn:reconstructionBG}
{\mathcal{L}_R = \frac{1}{n}\sum_{i=1}^{n}{\|\textbf{x}^i}-D(E(\textbf{x}^i))}\|_2^2
\end{equation}

\subsubsection{Feedback networks based autoencoder}

The section presents the framework of feddback connection based autoencoder (FBAE), where we have an encoder where 1D CNNs are enhanced with feedback connection, and a 1D CNN based decoder to reconstruct the original features from the low dimensional projected features\footnote{Published as: Shivam Pande, Biplab Banerjee. \href{https://ieeexplore.ieee.org/abstract/document/9883594}{Feedback Convolution Based Autoencoder for Dimensionality Reduction in Hyperspectral Images}, In: International Geoscience and Remote Sensing Symposium, Jul., 2022, pp: 147-150.}.

\label{sec:FBAE}
\begin{figure*}[t!]
  \centering
  \centerline{\includegraphics[width=14cm]{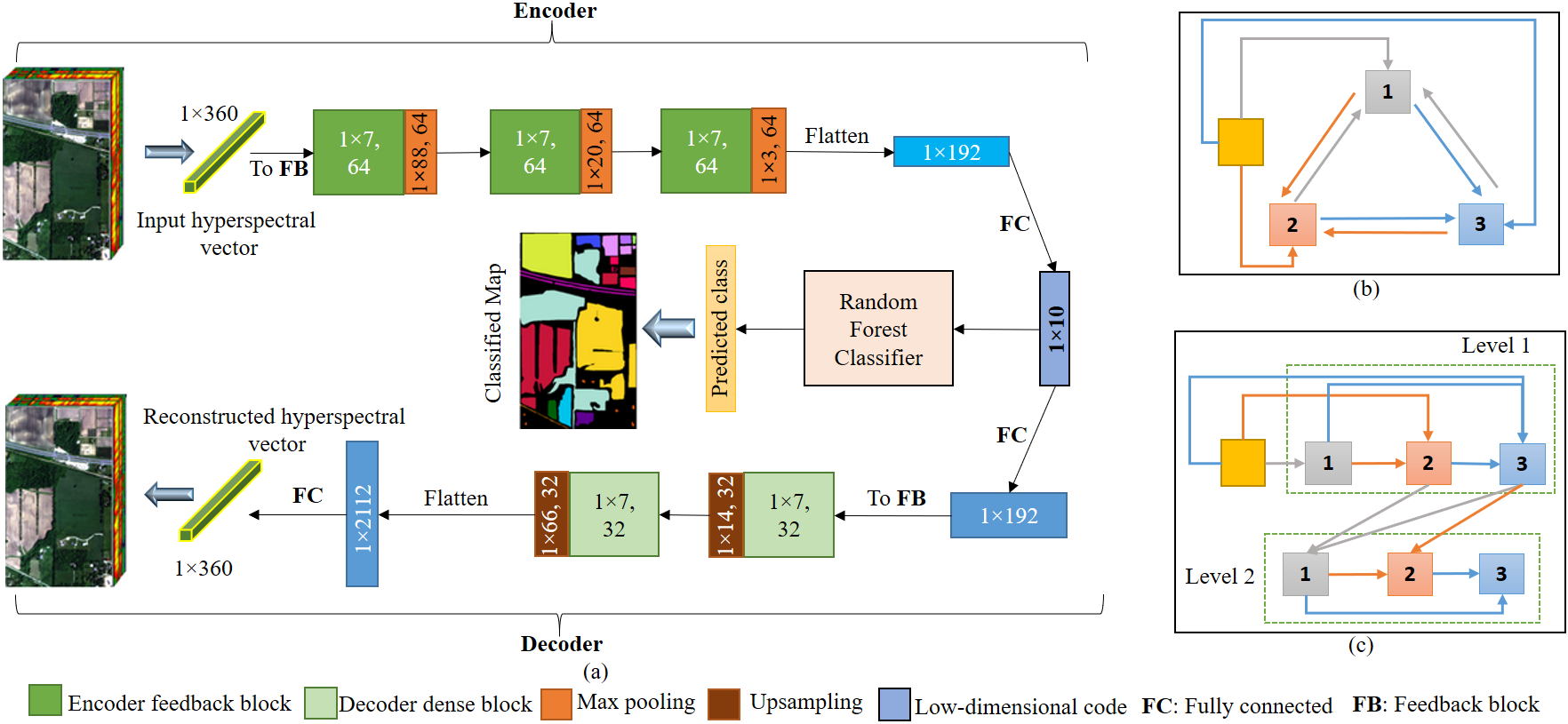}}
  \caption[Schematic of feedback autoencoder (FBAE), presented on IP2010 dataset]{(a) Schematic of feedback autoencoder (FBAE), presented on IP2010 dataset. The dimensions and numbers of filters in convolution layers are written in each feedback and dense block, and the output size is written in the max pooling and upsampling block. (b) Rolled representation of a feedback block. (c) Unrolled representation of a feedback block. The output from the last convolution layer after feedback mechanism is sent to the maxpooling layer. In (b) and (c), the colours of arrows denote the layer to which the input is being sent.}\medskip
  \vspace{-0.5cm}
  \label{fig:DiagramFB}
\end{figure*}

\noindent\textbf{Model Architecture}

The structure of FBAE is depicted in Figure (\ref{fig:DiagramFB}). FBAE primarily consists of an encoder that converts the features into a low-dimensional space and a decoder that reconstructs the original features. The network is trained to minimize the dissimilarity between the input and reconstructed features. Detailed explanations of the individual components will be provided in the following sections.

\noindent\underline{\textit{Encoder}} $E$: The encoder comprises three feedback blocks (FB) followed by max-pooling layers. Each feedback block consists of four 1D-convolution layers. The first convolution layer is utilized to reduce the number of feature maps to a desired size, empirically set as 64. This ensures compatibility in dimensions during weight sharing. The 1D convolution operation is represented by Equation (\ref{eq:cv1}) and will be denoted as $\odot$ in subsequent equations. In Equation (\ref{eq:cv1}), $w$ and $b$ represent the weights and biases in a convolution layer, respectively. $\sigma$ denotes the ReLU activation function, and $M$ represents the convolution output. The indices $p$, $q$, $t$, and $r$ correspond to the layer, kernel, input feature map, and output feature map, respectively. $E$ denotes the kernel size, and $c$ represents the input to the convolution layer.
\begin{equation}
{M}_{pq}^{r} = \sigma \Bigg(b_{pq}+ \sum_{t} \sum_{e=0}^{E-1} w_{pqt}^{e}c_{(p-1)t}^{(r+e)}\Bigg)
\label{eq:cv1}
\end{equation} 
The remaining three layers in the two levels are interconnected using feedback connections, as described in the following sections.

\begin{enumerate}
    \item \textbf{\textit{Level 1}}: In the first level (refer to Figure (\ref{fig:DiagramFB} (c))), a dense architecture is employed where the outputs from all previous layers are passed to the next layer through addition (to manage the number of features). Equations (\ref{eq:L1_inputFB}) and (\ref{eq:L1_outputFB}) illustrate the input and output of the convolutional layers in \textit{level 1}. In these equations, $W$ and $\beta$ represent the weights and biases in the layer, respectively, while $k$ denotes the index of the convolutional layer in the feedback block. $c_j$ denotes the output from the $j^{th}$ layer, and $c_0$ represents the output from the first convolutional layer (depicted as the yellow-coloured block in Figure (\ref{fig:DiagramFB} (c))).
        \begin{equation}
        \alpha_k = c_0 + \sum_{j=1}^{k-1}c_j
        \label{eq:L1_inputFB}
        \end{equation}
        \begin{equation}
        c_k = \sigma(W_k \odot \alpha_k + \beta_k)
        \label{eq:L1_outputFB}
        \end{equation}
    
    \item \textbf{\textit{Level 2}}: The feedback mechanism is established in this level by recycling the outputs of the later layers back to the initial layers using the same weights (refer to Figure (\ref{fig:Diagram} (c))). The recycled inputs are represented in Equation (\ref{eq:L2_input}). Here, $K$ denotes the number of convolutional layers in the feedback block. The output is computed in the same manner as in Equation (\ref{eq:L1_outputFB}), utilizing the same weights to facilitate weight recycling for the feedback mechanism. It should be noted that the weights can be recycled multiple times; however, for the sake of computational efficiency, we have implemented a single recycling iteration.
        \begin{equation}
        x_k = \sum_{j=1, j \neq k}^{K}c_j
        \label{eq:L2_input}
        \end{equation}
    The features obtained from the final max pooling layer are reshaped into a flattened form and passed through a fully connected layer to obtain the low-dimensional features. The output of the encoder is denoted as $E(\textbf{x})$.
\end{enumerate}

\noindent\underline{\textit{Decoder}} $D$: The decoder comprises two dense blocks, each followed by an upsampling layer. The dense block shares a similar structure to that of the encoder, but only includes level one features and replaces the first convolution layer with a transposed convolution layer to increase the dimensions for reconstruction. The low-dimensional features are initially passed through a fully connected layer, and the output is then forwarded to the first dense block. The output from the second upsampling layer is directed to the fully connected layer, resulting in the HSI vector, which can be denoted as $D(E(\textbf{x}))$. The original and reconstructed vectors are compared using the mean squared error loss function (refer to Equation (\ref{eq:mse})).
\begin{equation}
\label{eq:mse}
{\mathcal{L}_R = \frac{1}{n}\sum_{i=1}^{n}{\|\textbf{x}^i}-D(E(\textbf{x}^i))}\|_2^2
\end{equation}

For all the proposed autoencoder methods, once the training process is finished, both the training and test samples are fed into the classifier to obtain their respective low-dimensional representations. Subsequently, a random forest model \citep{breiman2001random} with 200 decision trees is trained using the low-dimensional training samples and evaluated using the low-dimensional test samples. The accuracy achieved on the test set serves as a metric to assess the performance of the model.

\subsubsection{Datasets and Experiments}

\noindent\textbf{Datasets}

\begin{figure}[htb]
\begin{minipage}[b]{1.0\linewidth}
  \centering
  \centerline{\includegraphics[width=8.5cm]{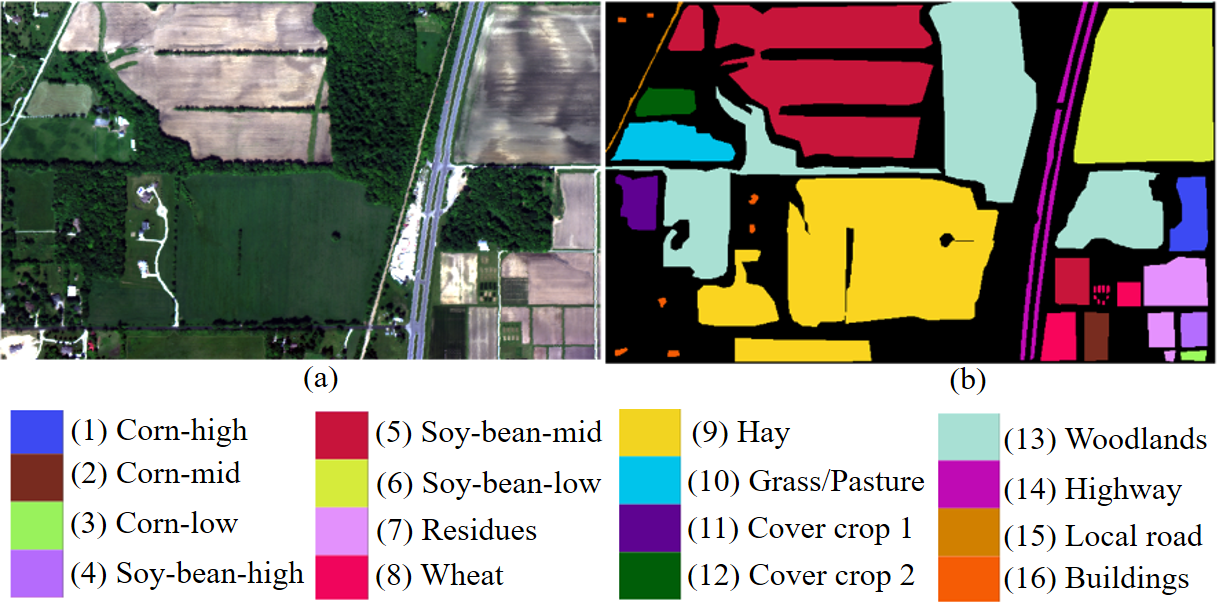}}
\end{minipage}
\caption{Salinas Valley hyperspectral dataset. (a) Colour composite of 3 bands from the image (b) Groundtruth with classes.}
\label{fig:ip2010_data}
\end{figure}

\noindent\underline{\textit{Indian pines 2010 (IP2010)}}: This hyperspectral image (HSI) was captured in May 2010 using ProSpecTIR in the vicinity of Purdue University, Indiana. The image has a spatial extent of 445$\times$750 pixels and consists of 360 channels \citep{rasti2020feature}. It contains 16 classes, with 10,954 samples assigned to the training set and 187,120 samples assigned to the test set (Figure (\ref{fig:ip2010_data})).

\noindent\underline{\textit{Indian pines 1992 (IP1992)}}: The HSI was acquired using the AVIRIS sensor and has dimensions of 145$\times$145$\times$200 (i.e., 200 bands) \citep{pande2021attention}. The image comprises 10,249 labeled samples classified into 16 classes, with 1,024 samples used for training and 9,225 samples for testing (Figure (\ref{fig:ip_data})).

\noindent\underline{\textit{Salinas}}: This image, obtained using the AVIRIS sensor, covers the Salinas Valley in California and consists of 204 bands, with each band having a size of 512$\times$217 pixels. The image encompasses 16 classes and includes a total of 54,129 ground truth samples, with 10,825 samples assigned to the training set and 43,304 samples assigned to the test set (Figure (\ref{fig:sal_data})) \citep{zheng2020fpga}.

\noindent\textbf{Training Protocols}

All the proposed autoencoders are trained using semi-supervised learning with only the samples from the training set, where the hyperspectral vector is used as input. The dimension of the low-dimensional representation is fixed at 1$\times$10 for the IP2010 and Salinas datasets, and 1$\times$5 for the IP1992 dataset. In our comparison, we include other methods such as PCA, FCAE, RNNAE, GruAE, and CNNAE \citep{pande2021attention}. During training, we utilize the Adam optimizer \citep{kingma2014adam} with a learning rate of 0.00025. The number of epochs is set to 500 for IP2010 and 250 for IP1992. The training process is conducted using the \textit{Tensorflow 2.0} framework on \textit{Google Colaboratory}. To evaluate the performance, we use metrics such as overall accuracy (OA), class-wise accuracy, and the percentage kappa coefficient (100$\times \kappa$). Additionally, for visual comparison, we generate t-SNE plots before and after dimensionality reduction.

\subsubsection{Results and Discussion}

\begin{table}[ht]
\centering{\scriptsize
 \caption{\label{tab:ip10_perf} Accuracy analysis on the IP2010 dataset (\%).}
\begin{tabular}{|c|c|c|c|c|c|c|c|}
 \hline
Class & PCA & FCAE & RNNAE & CNNAE & AttAE & BiGRUAE& FBAE\\
 \hline
1  & 91.47 &91.43& 94.66 & \textbf{94.02} & 92.45 &90.15&91.17\\
2  & \textbf{97.88} &95.92& 89.41 & 86.67 & 94.67&92.00&94.82\\
3  & \textbf{97.59} &95.17& 84.48 & 83.45 & 91.38&97.24&79.66\\
4  & 16.14 &68.88&67.05 & 70.12 & \textbf{76.95}&78.10&71.37\\
5  & 77.16 &76.40& 72.17 & 81.77 & \textbf{87.58}&85.53&83.96\\
6  & 72.46 &88.05& 82.62 & 78.82 & 84.69&\textbf{92.97}&88.16\\
7  & \textbf{72.18} &54.65& 37.66 & 43.27 & 48.06&49.94&46.38\\
8  & \textbf{32.26} &25.80& 24.21 & 24.31 & 24.24&24.77&23.96\\
9  & \textbf{78.49} &65.61& 67.45 & 73.29 & 75.18&77.59&75.38\\
10 & 69.57 &\textbf{84.25}& 73.83 &76.45 & 78.23&79.99&71.95\\
11 & \textbf{79.24} &61.51& 48.93 & 57.46 & 57.20&65.35&58.44\\
12 & \textbf{100.00} &\textbf{100.00}& \textbf{100.00} & 99.85 & \textbf{100.00}& \textbf{100.00}&92.10\\
13 & 90.18 &97.85& 92.34 & 96.44 & 97.21&98.77&\textbf{98.63}\\
14 & \textbf{90.31} &87.94& 89.85 & 86.80 & 88.40&90.48&90.27\\
15 & 93.33 & \textbf{93.56} & 89.33 & 86.67 & 89.56&93.56&86.22\\
16 & 64.82 & 70.75 & \textbf{81.23} & 4.74 & 12.65&74.31&6.72\\
 \hline
OA &  79.60  &80.16& 76.73 & 80.55 & 83.53&\textbf{85.79}&83.44\\
$\kappa$ (\%) & 0.7527 &0.7581& 0.7165 & 0.7629 & 0.7981 &\textbf{0.8263}&0.7972\\
 \hline
\end{tabular}}
\end{table}

\begin{table}[ht]
\centering{\scriptsize
 \caption{\label{tab:ip92_perf} Accuracy analysis on the IP1992 dataset (\%).}
\begin{tabular}{|c|c|c|c|c|c|c|}
 \hline
Class & PCA & FCAE & RNNAE & CNNAE & AttAE & FBAE\\
 \hline
1  & 4.88 & 24.39 &  17.07 & \textbf{26.83} & 24.39&24.39\\
2  & 48.87  & 46.30 &  42.41  & 56.19 & \textbf{58.60}&53.31\\
3  & \textbf{47.93} & 41.37 & 41.23 & 46.05 & 42.97&46.05\\
4  & 28.17 & 35.21 &  30.99  & 23.94 & 34.74&\textbf{38.97}\\
5  & 52.41 & 52.64 &  42.53 & 60.69 & \textbf{62.53}&61.15\\
6  & \textbf{90.41} & 87.52  & 81.13 & 86.45 & 90.11& 82.19\\
7  & 60.00 & \textbf{76.00} &  72.00  & 68.00 & 44.00&40.00\\
8  &  97.21  & 97.21 & 96.51  & 97.91 & \textbf{98.60}&97.21\\
9  & 0.00 & 0.00 & 0.00  & 0.00 & 0.00 & 0.00\\
10 & 57.03 & 53.94 & 55.20  & 61.49 & \textbf{64.91}&63.31\\
11 & 76.83 & 77.06 & 74.84 & 77.47 & 78.10&\textbf{78.96}\\
12 & \textbf{34.08} & 27.72 &  20.22 & 32.58 & 33.71&32.02\\
13 & 94.59 & 85.41 & 86.49 & \textbf{96.22} & 89.73 &90.81\\
14 & \textbf{94.56} & 85.43 & 85.60  & 87.53& 88.41&88.94\\
15 & 23.63  & 26.80 & 20.75  & 18.73 & \textbf{27.95}&22.48\\
16 & 83.33 & 82.14 & 83.33 & 83.33 & 79.76&\textbf{85.71}\\
 \hline
OA &  65.97  &63.37 &  60.69 & 66.47& \textbf{67.95}&66.74\\
$\kappa$ (\%) & 60.75 & 57.79 & 54.67 & 0.6139 & \textbf{0.6315}&0.6174\\
 \hline
\end{tabular}}
\end{table}

\begin{table}[ht]
\centering{\scriptsize
 \caption{\label{tab:sal_perf_1D} Accuracy analysis on the Salinas Valley dataset (\%).}
\begin{tabular}{|c|c|c|c|c|c|}
 \hline
Class & PCA & FCAE & CNNAE & GRUAE & BiGRUAE\\
 \hline
    1     & 98.57 & 97.95 & 98.01 & 98.07 &\textbf{98.63} \\
    2     & \textbf{99.80}  & 99.23 & 99.53 & 99.70  & 99.73 \\
    3     & 98.61 & 97.98 & 98.61 & 99.05 &\textbf{99.24} \\
    4     & 99.64 & 99.37 & 99.64 & \textbf{99.73} & 99.64 \\
    5     & 98.41 & 98.51 & 98.41 & \textbf{98.79} & \textbf{98.79}\\
    6     & \textbf{99.78} & 99.75 & \textbf{99.78} & \textbf{99.78} & \textbf{99.78} \\
    7     & 99.23 & 99.09 & \textbf{99.41} & 99.09 & 99.37 \\
    8     & \textbf{86.49} & 83.71 & 85.07 & 85.41 & 85.84 \\
    9     & 99.64 & 99.54 & 99.80  & 99.82 & \textbf{99.92} \\
    10    & 92.18 & 91.95 & 93.67 & 93.75 & \textbf{95.16} \\
    11    & 94.85 & 92.51 & 94.85 & 94.26 & \textbf{98.36} \\
    12    & 99.48 & 99.22 & 99.22 & \textbf{99.74} &\textbf{99.74} \\
    13    &\textbf{98.77} & 98.36 & 98.36 & 98.50  & 98.23 \\
    14    & 94.63 & 94.86 & 94.28 & \textbf{96.03} & 95.09 \\
    15    & 64.32 & 68.48 & \textbf{70.89} & 68.67 & 69.20 \\
    16    &\textbf{98.27}& 97.37 & 98.06 & 97.79 & \textbf{98.27} \\
    \hline
    OA    & 91.30  & 91.07 & 91.94 & 91.78 & \textbf{92.16} \\
    $\kappa$ (\%)  & 0.9030  & 0.9005 & 0.9102 & 0.9084 & \textbf{0.9127} \\
 \hline
\end{tabular}}
\end{table}

\begin{table}[ht]
\centering{\scriptsize
\caption{\label{tab:abl_low_CNN} Comparison of accuracies with respect to number of reduced dimensions on IP2010 dataset.}
\begin{tabular}{|c|c|c|c|}
\hline
Dimensions & 4  & 7 & 10\\
\hline
\hline
OA (AttAE) ($\%$) & 76.19 & 81.13 & \textbf{83.39}\\
OA (FBAE) ($\%$) & 79.25 & 82.39 & \textbf{83.44}\\
\hline
\end{tabular}}
\end{table}

\begin{table}[ht]
\centering{\scriptsize
 \caption{\label{tab:abl_low_BGAE} Comparison of accuracies with respect to number of reduced dimensions on IP2010 dataset.}
\begin{tabular}{|c|c|c|c|c|}
 \hline
Dimensions & 3 & 5 & 7 & 10\\
 \hline
 \hline
 OA ($\%$) & 78.92 & 81.90 & 83.11 & \textbf{85.79}\\
 \hline
\end{tabular}}
\end{table}

The results for all the methods are provided in Table \ref{tab:ip10_perf} (for IP2010 dataset). For Indian Pines 1992 dataset, results are provided in Table \ref{tab:ip92_perf} for AttAE and FBAE. The results for BiGRUAE methods are simultaneously evaluated on Salinas dataset and presented in Table \ref{tab:sal_perf_1D}. 

We see that our proposed approaches have surpassed the previous methods of dimensionality reduction in HSI. The best performance has been observed by BiGRUAE, (with $\kappa$ value of 0.8263, on IP2010 dataset), followed by AttAE and FBAE, with respective $\kappa$ values of 79.81\% and 79.72\%. Similarly, on IP1992 dataset, we see that proposed methods AttAE and FBAE give 63.15\% and 61.74\% performance in terms of $\kappa$, while for Salinas, we see that highest $\kappa$ is obtained by BiGRUAE (0.9127). Furthermore, our method exhibits consistently homogeneous performance across all the classes for all the three datasets. We further see that, there are certain classes that have continuously low performance for most of the methods. For instance, in IP2010, for class `wheat', the performance is subpar for all the models involved. Similarly, in IP1992, it could be seen that the performance of all the models is 0.00\% for `oats' class. This shows, that in order to have better self-supervised training for dimensionality tasks, it is required to have a good number of training samples. 

\begin{figure}[htb]
\begin{minipage}[b]{1.0\linewidth}
  \centering
  \centerline{\includegraphics[width=8.0cm]{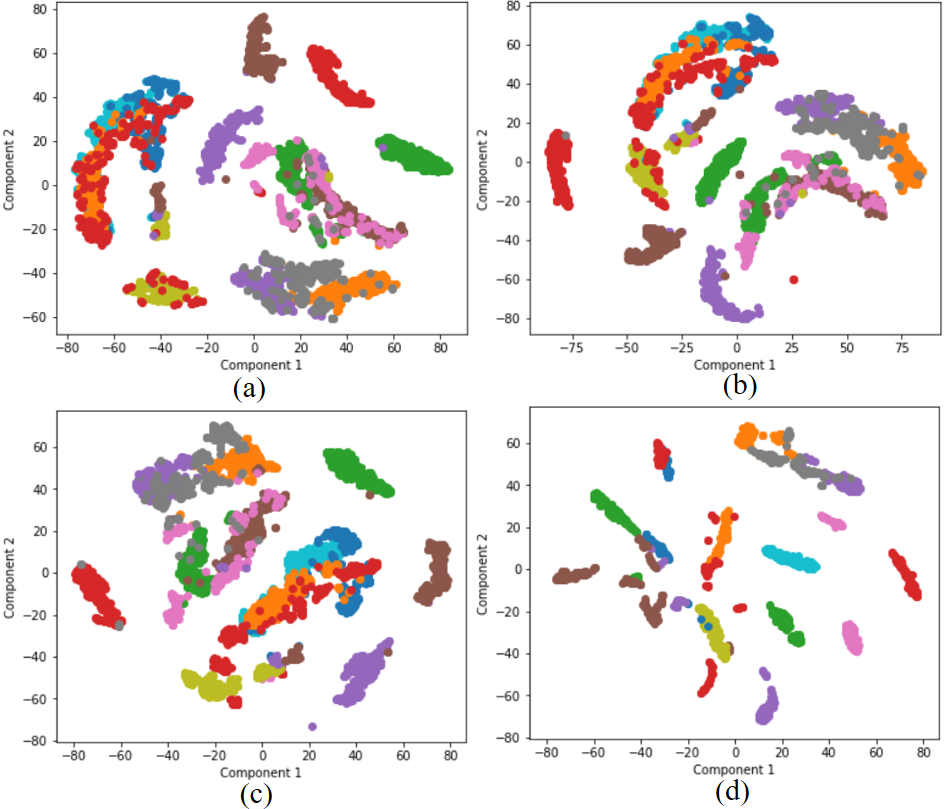}}
\end{minipage}
\caption{t-SNE of test samples of Indian pines 2010 dataset (a) original features (b) encoded features by CNNAE (c) encoded features by BiGRUAE (d) encoded features by FBAE.}
\label{fig:tsne_AE}
\end{figure}

We have presented the ablation study by decreasing the number of dimensions in our datasets from 10 to 7 and 4 for AttAE and FBAE (see Table \ref{tab:abl_low_CNN}). It is visible that as the dimensions reduce, the performance starts to reduce as well. However, even with 7 dimensions, our methods have shown to perform better than other methods. Similar trends are also visible in Table \ref{tab:abl_low_BGAE} where a similar ablation performed with BiGRUAE on IP2010 dataset. In Figure \ref{fig:tsne_AE}, the t-SNE visualization of the features before and after dimensionality reduction using are proposed techniques presented. It is evident that similar clusters are observed for both overlapping and non-overlapping classes, suggesting that the intrinsic information of the classes is predominantly preserved following the reduction in dimensionality.

\section{Label assisted self-supervised learning for hyperspectral image classification}
\label{sec:PAWS}
In this section, a semi-supervised approach is presented for the task of LULC classification in HSIs. Unlike conventional discriminative SSL methods, the proposed approach utilizes the unannotated samples alongwith the few available annotated samples from the dataset. The idea is to match the probability of an augmented unannotated sample to belong to the labelled sample with that of the other augmented unannotated sample. Thus, the conventional SSL based method is guided using an extra supervision from the available annotated samples, thus resulting in better performance. The details are highlighted in the subsequent sections\footnote{Published as: Shivam Pande, Nassim Ait Ali Braham, Yi Wang, Conrad N. Albrecht, Biplab Banerjee, Xiaoxiang Zhu. \href{https://arxiv.org/abs/2306.10955}{Semi-Supervised Learning for hyperspectral images by non-parametrically predicting view assignment}, In: International Geoscience and Remote Sensing Symposium, Jul., 2023.}. 

\subsection{Methodology}

\begin{figure*}[ht!]
  \centering
  \centerline{\includegraphics[width=14cm]{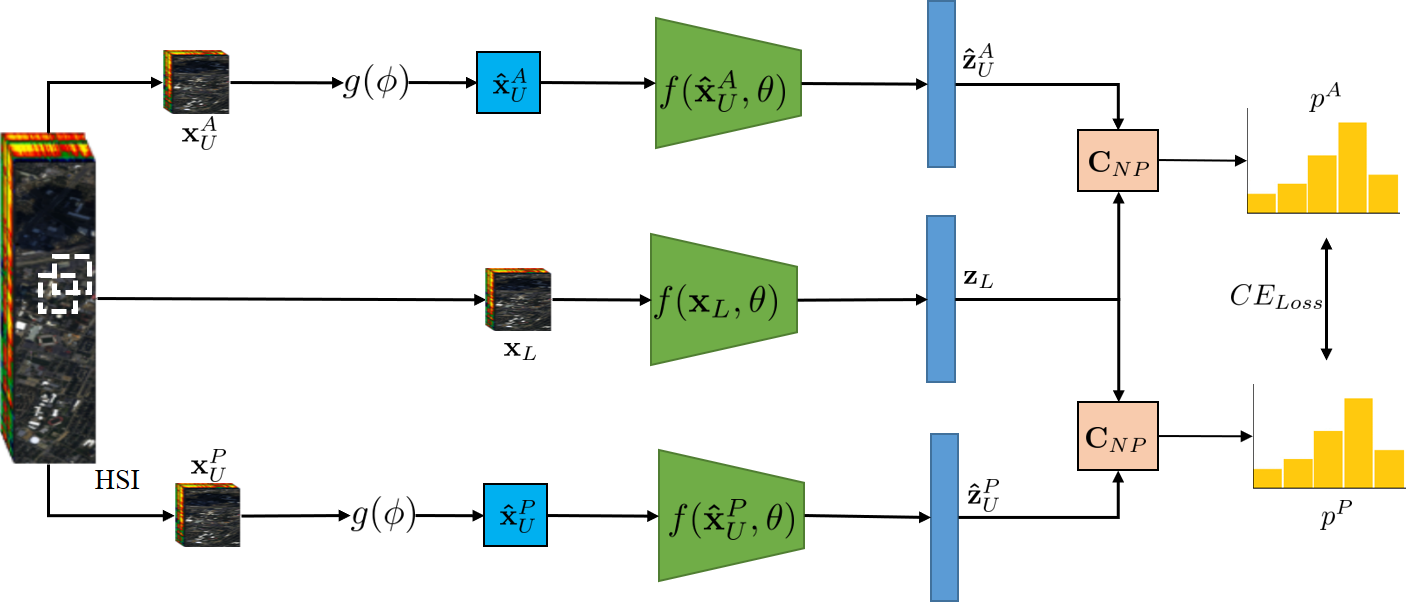}}
\caption[Schematic of the PAWS model used for semi-supervised pretraining]{The figure illustrates the schematic of the PAWS model used for semi-supervised pretraining. Starting from the hyperspectral image (HSI), we extract overlapping patches for the unlabelled anchor and positive sets. These patches are then passed through the augmentation function $g(\phi)$ to obtain the augmented representations. The augmented anchor and positive samples are fed into the encoder $f(\theta)$, which produces the embeddings $\hat{z}_U^A$ and $\hat{z}_U^P$. Similarly, for the labelled sample, the embedding $z_L$ is obtained. The embeddings $\hat{z}_U^A$ and $\hat{z}_U^P$ are compared to $z_L$ using a non-parametric classifier $C_{NP}$. The resulting pseudo-label probabilities are then used to calculate the cross-entropy loss, which is used to train the model.}\medskip
  \vspace{-0.5cm}
\label{fig:paws}
\end{figure*}

Let $\mathcal{X} \in \mathbb{R}^{M \times N \times B}$ represent the hyperspectral image, where $M$, $N$, and $B$ denote the number of rows, columns, and channels, respectively. $\mathbf{x}_U^{j} \in \mathbb{R}^{p \times p \times B}$ represents the unlabelled samples, while $\mathbf{x}^{i}_L \in \mathbb{R}^{p \times p \times B}$ represents the labelled samples, where each sample has the labels of the central pixel assigned to it. Here, $p$ represents the spatial dimension of the sample. The labels associated with the training samples are denoted as $y^{i}_L \in \mathcal{Y}_L$, where $i\in{1,2,...N_L}$ and $j\in{1,2,...N_U}$. Here, $N_L$ and $N_U$ indicate the number of labelled and unlabelled samples, respectively. The training process consists of two stages: \textit{pretraining} and \textit{classification}, which will be discussed in the subsequent sections. The schematic of the proposed approach is presented in Figure \ref{fig:paws}.

\subsubsection{Pretraining}

This section presents the semi-supervised pretraining process of the proposed PAWS model. Let $(x^i)^A_U \in \mathbb{R}^{p \times p \times B}$ denote the anchor view and $(x^i)^P_U \in \mathbb{R}^{p \times p \times B}$ denote the positive view, which have spatial overlap. These views are passed through a non-trainable probabilistic augmentation function $g(\phi)$ to obtain the augmented views $(\hat{x}^i)^A_U$ and $(\hat{x}^i)^P_U$, respectively. Here, $\phi$ is the probability vector used for selecting the augmentation. The encoder $f(\theta)$ maps the given hyperspectral patch to a hidden representation ($f(\theta): \mathbb{R}^{p \times p \times B} \xrightarrow{} \mathbb{R}^d$). Two types of encoders have been used in the experiments: one derived from a WideResNet architecture \citep{zagoruyko2016wide} with modifications to accommodate hyperspectral data, and the other being a fully convolutional framework with an initial 3D convolution layer and followed by three depthwise separable convolution layers, resulting in the $d$-dimensional vectors $(\hat{z}^i)^A_U$ and $(\hat{z}^i)^P_U$. Simultaneously, for a labelled sample $(x^j)_L \in \mathbb{R}^{p \times p \times B}$ (obtained from the labelled support set), we obtain the representation $(z^j)_L$.

The support representation $(z^i)_L$ is then compared with the anchor and positive representations $(\hat{x}^i)^A_U$ and $(\hat{x}^i)^P_U$ using a cosine similarity-based non-parametric soft nearest neighbour (SNN) classifier \citep{salakhutdinov2007learning} denoted as $C_{NP}$, given by:
\begin{equation}
(\hat{p}^j)^A = C_{NP}({z^j_U, \mathbf{z}_L}) = \sigma_\tau((z^j)^A_U\mathbf{z}^T_L)\mathcal{Y}_L
\end{equation}

Similarly, $(\hat{p}^j)^P$ can also be computed. In this equation, $\sigma$ represents the softmax activation function and $\tau (>0)$ denotes the temperature parameter. The resulting probabilities are then sharpened to prevent representation collapse and obtain confident predictions. This sharpening operation is defined by the following function:

\begin{equation}
[\rho(\hat{p}^j)]^k = \frac{([p^j]_k)^{\frac{1}{T}}}{\sum_{t=1}^K([p^t]_k)^{\frac{1}{T}}} 
\end{equation}

Here, $k \in \{1, 2, ..., K\}$, where $K$ represents the number of classes. During model training, we incorporate two cross entropy loss terms simultaneously, comparing $(\hat{p}^j)^P$ with $(\hat{p}^j)^A$. Moreover, we include an auto-entropy based regularizer that is derived from the average of sharpened predictions. This regularizer helps ensure that the average prediction closely resembles the mean distribution. The complete loss function is defined as follows:
\begin{equation}
\mathcal{L}_{SSL} = \frac{1}{2n}\sum_{i = 1}^{n}(\mathcal{E}(\rho((\hat{p}^j)^A), (\hat{p}^j)^P) + \mathcal{E}(\rho((\hat{p}^j)^P), (\hat{p}^j)^A)) - \mathcal{E}(\bar{p}) 
\end{equation}
In this context, the function $\mathcal{E}$ represents the entropy calculation, and $\bar{p}$ corresponds to the average probability of the predictions. The average probability is defined as follows:
\begin{equation}
\bar{p} = \frac{1}{2n}\sum_{i = 1}^{n}(\rho((\hat{p}^j)^P), \rho((\hat{p}^j)^A))
\end{equation}
During the training process, when using the cross entropy loss, the gradients are not computed with respect to the sharpened targets.

\subsection{Classification}
\label{sec:classification}
\begin{figure}[ht!]
  \centering
  \centerline{\includegraphics[width=8.5cm]{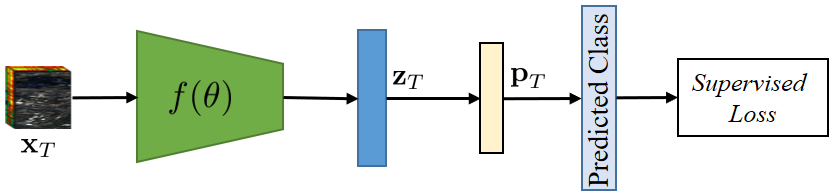}}
\caption{Schematic of linear classification and finetuning after PAWS based pretraining.}\medskip
  \vspace{-0.5cm}
\label{fig:ft}
\end{figure}

After the pretraining phase, the model is assessed for land-use/land-cover classification task using three different techniques as benchmarks:

\begin{enumerate}
\item \textbf{Linear layer}: A fully connected layer is added on top of the existing model. The pretrained model's weights are kept fixed, while only the weights of the final layer are updated using the training samples. The model is then evaluated on the test samples.

\item \textbf{Fine-tuning}: Similar to the linear layer approach, all the weights of the model are modified simultaneously using the training samples instead of keeping them frozen. The model is then evaluated on the test samples. The schematic representation of linear classification and fine-tuning can be observed in Figure \ref{fig:ft}.

\item \textbf{Soft nearest neighbor (SNN) classifier}: In this technique, the trained model serves as a feature extractor without any further optimization downstream. Both the training and test samples are passed through the pretrained model. The test samples are then compared to the training samples using a cosine similarity-based classifier, such as the $C_{NP}$ classifier.
\end{enumerate}

\subsection{Datasets and Experiments}
\label{sec:experiments}
\subsubsection{Datasets}
\begin{figure}[ht!]
  \centering
  \centerline{\includegraphics[width=8.5cm]{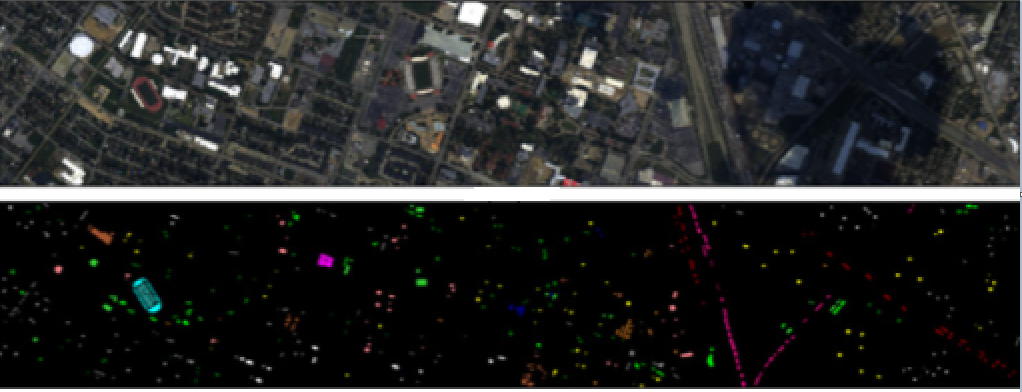}}
\caption{Houston 2013 dataset. False colour composite (Left). Groundtruth (Right)}\medskip
  \vspace{-0.5cm}
\label{fig:h13_data}
\end{figure}

We conducted our experiments on the following two datasets:

\noindent\textbf{Houston dataset}: This dataset is derived from the data fusion contest held in $2013$. The hyperspectral image (HSI) has a spatial extent of $349 \times 1905$ pixels and consists of $144$ spectral bands. It contains a total of $2,832$ training samples and $12,197$ test samples, distributed across $15$ land use/land cover classes. For our experiments, we selected $100$ samples per class from the training set, resulting in a support set of $1,500$ patches in total. We also sampled $71,416$ patches as anchor and positive points. The HSI and its corresponding ground truth are illustrated in Figure \ref{fig:h13_data} \citep{debes2014hyperspectral}.

\begin{figure}[ht!]
  \centering
  \centerline{\includegraphics[width=8.5cm]{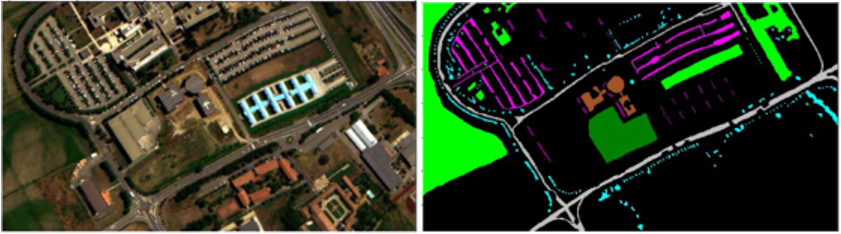}}
\caption{Pavia University dataset. False colour composite (Left). Groundtruth (Right)}\medskip
  \vspace{-0.5cm}
\label{fig:PU_data}
\end{figure}

\noindent\textbf{Pavia University dataset}: This dataset comprises a hyperspectral image of size $610 \times 340 \times 103$. It contains a total of $42,776$ labeled samples. For the anchor and positive points, we sampled $99,000$ points. The support set consists of $900$ patches ($100$ per class), and the test set contains $41,876$ points. The patch size used is $9$. Figure \ref{fig:PU_data} presents the HSI and the corresponding ground truth \citep{braham2022self}.

In both datasets, the anchor and positive patches have a $67\%$ overlap.


\subsubsection{Training Protocols}
\label{ssec:tp}
For the pretraining stage of all deep learning models, we utilized the LARS optimizer \citep{you2019large}. During the classification phase, both with the linear layer and fine-tuning, we used the stochastic gradient descent optimizer. In our experiments, we fixed the values of temperatures $\tau$ (for softmax) and $T$ (during sharpening) to $0.25$ and $0.10$ respectively. The number of epochs was set to 50 for model pretraining, linear classification, and fine-tuning. We applied the following augmentations to each unlabeled sample:
\begin{enumerate}
\item \textbf{Spectral}: Random channel swapping, random channel dropping, channel intensity suppression, random channel averaging for a fixed number of channels.
\item \textbf{Spatial}: Flipping (horizontal, vertical, and mirror), random cropping, random rotation.
\item \textbf{Spectral-spatial}: Pixel vector removal, noise introduction across channels and spatial dimensions.
\end{enumerate}

\begin{table}[htbp]
  \centering{\scriptsize
  \caption{Accuracy analysis for the proposed semi-supervised PAWS model on Houston 2013 dataset.}
    \begin{tabular}{|c|c|c|c|}
    \hline
    Encoder & Augmentations & Classifier & \multicolumn{1}{l|}{Accuracy} \\
    \hline
    \multicolumn{4}{|c|}{} \\
    \hline
    CNN   & No    & Supervised & 79.10\% \\
    Barlow Twins & Yes   & Fine Tune & 77.14\% \\
    None  & No    & Linear & 64.21\% \\
    PAWS trained & Yes   & Linear & 80.92\% \\
    PAWS trained & Yes   & Fine Tune & \textbf{83.12}\% \\
    \hline
    \multicolumn{4}{|c|}{} \\
    \hline
    None & No & SNN & 32.63\% \\
    Untrained & No  & SNN & 34.28\% \\
    PAWS trained & Yes   & SNN   & \textbf{75.31}\% \\
    \hline
    \end{tabular}
  \label{tab:results_h13}}
\end{table}

\begin{table}[htbp]
  \centering{\scriptsize
  \caption{Accuracy analysis for the proposed semi-supervised PAWS model on Pavia University dataset.}
    \begin{tabular}{|c|c|c|c|}
    \hline
    Encoder & Augmentations & Classifier & \multicolumn{1}{l|}{Accuracy} \\
    \hline
    \multicolumn{4}{|c|}{} \\
    \hline
    CNN & No & Supervised & 21.38\% \\
    Barlow Twins & Yes & Linear & 82.44\% \\
    None & No & Linear & 74.12\% \\
    PAWS trained & Yes & Linear & \textbf{83.64}\% \\
    \hline
    \multicolumn{4}{|c|}{} \\
    \hline
    None  & No & SNN & 50.12\% \\
    Untrained & No & SNN & 46.33\% \\
    PAWS trained & Yes & SNN & \textbf{70.34}\% \\
    \hline
    \end{tabular}
    \label{tab:results_pavia}}
\end{table}

\section{Results and Discussion}

The results for the Houston 13 and Pavia University datasets are presented in Tables \ref{tab:results_h13} and \ref{tab:results_pavia} respectively. For the Houston 13 dataset, a custom CNN model was used, while the results for the Pavia University dataset were obtained when training on WideResNet. In Column 1, the specific model configurations are indicated. Column 2 indicates whether augmentations were applied. Column 3 shows the choice of classifier for downstream land cover classification, and the corresponding overall accuracy values are listed in Column 4. It can be observed that our proposed semi-supervised approach outperforms the supervised approach in both the linear layer and fine-tuning settings. Additionally, our method demonstrates better performance compared to a recently attempted self-supervised approach based on Barlow Twins loss. As a baseline benchmark, the SNN classification performance improves when the classifier is trained on features extracted from our proposed approach. From Table \ref{tab:results_pavia}, it can also be seen that the accuracy of the supervised classification is relatively lower compared to the other models. This discrepancy could be attributed to the larger number of parameters in the WideResNet model compared to the custom CNN model. The best performances are highlighted in bold.

\section{Summary}

In this chapter, we delve into the application of self-supervised learning (SSL) techniques in the processing and classification of hyperspectral images. We explore both generative and discriminative settings where SSL can be effectively incorporated. Specifically, we focus on leveraging the generative aspect of SSL to address the challenge of dimensionality reduction in hyperspectral imaging. In this context, we propose several autoencoder-based frameworks to tackle this problem.

We propose several novel approaches to address the problem of dimensionality reduction in hyperspectral image classification. First, we introduce a 3D residual autoencoder (3D ResAE) that incorporates skip connections over 3D convolutions to overcome the vanishing gradient problem and project high-dimensional hyperspectral patches to a lower-dimensional feature space. By reconstructing the original features from the projected ones under mean squared error (MSE) loss, we maintain the spectral-spatial integrity of the original features in the projected space. The effectiveness of the proposed method is extensively evaluated on Indian pines and Salinas datasets, clearly demonstrating the superiority of skip connections over 3D convolutions in achieving better dimensionality reduction and classification performance.

Furthermore, we propose a dimensionality reduction technique utilizing Bidirectional Gated Recurrent Units (BiGRUAE) to leverage the contiguous characteristic of hyperspectral images. By treating HSI features as a sequence from both forward and backward directions, BiGRUs in the encoder incorporate future information in the sequences. The presence of gating mechanisms in GRUs effectively addresses the vanishing gradient issue, resulting in more stable training. Evaluating our approach on Indian pines 2010 and Salinas hyperspectral datasets demonstrates its superiority over benchmark methods. 

In addition, we present an attention mechanism-driven and 1D convolution-based autoencoder (AttAE) specifically designed for hyperspectral images. The 1D convolutions process the spectral information while retaining its sequential nature, and the attention modules selectively highlight the most relevant features for the low-dimensional embedding. Reconstruction of the original hyperspectral vector using a 1D CNN-based decoder ensures similarity between the original and reconstructed features. Lastly, we introduce the feedback autoencoder (FBAE) as a novel method of dimensionality reduction. Our proposed model utilizes 1D convolutions to efficiently process hyperspectral images due to their sequential nature. We incorporate feedback connections in the 1D convolutional layers, enabling the model to utilize information from future layers to update the past. Moreover, we reinforce the layers with skip connections to ensure better gradient flow. As a result, the low-dimensional features obtained are more robust and informative for classification. Evaluations on Indian pines 2010 and Indian pines 1992 datasets demonstrate the superiority of the proposed methods over previous approaches. 

For future work, we consider extending this approach from pixels to patches to incorporate spatial information for all the pixel-based autoencoders. Additionally, we also aim to incorporate better parallelization techniques in the networks for faster training and convergence. From a discriminative SSL perspective, we apply the semi-supervised learning method, specifically based on the PAWS approach, for hyperspectral image classification. Our method utilizes the labelled samples available in the downstream task to effectively guide the self-supervised pretraining process prior to classification. Through experimentation, we observe that the proposed proposed model surpasses both the supervised and self-supervised classification tasks, demonstrating superior performance in both parametric and non-parametric settings. Encouraged by these results, we plan to extend the experiments to larger-scale missions, such as DLR's EnMAP satellite that was launched in April 2021 and has recently transitioned into the operational phase.

\section{Publications}

\begin{enumerate}
    \item Shivam Pande, Biplab Banerjee. Dimensionality Reduction Using 3D Residual Autoencoder for Hyperspectral Image Classification, In: International Geoscience and Remote Sensing Symposium, Sep., 2020, 2029-2032.
    \item Shivam Pande, Biplab Banerjee. Attention Based Convolution Autoencoder for Dimensionality Reduction in Hyperspectral Images, In: International Geoscience and Remote Sensing Symposium, Jul., 2021, 2727-2730.
    \item Shivam Pande, Biplab Banerjee. Bidirectional GRU Based Autoencoder for Dimensionality Reduction in Hyperspectral Images, In: International Geoscience and Remote Sensing Symposium, Jul., 2021, 2731-2734.
    \item Shivam Pande, Biplab Banerjee. Feedback Convolution Based Autoencoder for Dimensionality Reduction in Hyperspectral Images, In: International Geoscience and Remote Sensing Symposium, Jul., 2022, pp: 147-150.
    \item Shivam Pande,  Nassim Ait Ali Braham, Yi Wang, Conrad M. Albrecht, Biplab Banerjee, Xiao Xiang Zhu, Semi-supervised Learning for Hyperspectral Images by Non-Parametrically predicting View Assignment, In: International Geoscience and Remote Sensing Symposium, Jul., 2023.
\end{enumerate}
\chapter{Conclusions and future works}

\section{Conclusions}

This thesis presents several research works targetted at analysing and extracting information from hyperspectral images, primarily for the task of LULC classification. Three different avenues and experimental settings are identified to realize this goal, namely single-modal setting, multimodal setting and self-supervision based learning scenario.

In first setting, we firstly identify the challenge of domain adaptation in the remote sensing domain for hyperspectral images. To address the issue we setup a learning frame work based on adversarial learning, which successfully helps in mitigating the domain gap between the training and test data. The research work continues in the development of efficient feature extractors for HSIs. In the first research work, a novel and hybrid attention mechanism is designed in the CNN based architecture to selectively highlight the spectral and spatial properties of HSIs with respect to the classification task. Additionally, the classification performance is further enhanced using Wasserstein loss in conjugation with the cross-entropy loss. In the second part, we address the limitations in the forward nature of CNNs, and to mitigate that, we design a feedback connections oriented CNN framework with multiscale convolution filter banks. All our approaches are tested on several HSI benchmark datasets, where they achieve state of the art results.

In the multimodal setting, we start with the problem of missing modality identification for MS-PAN remote sensing image, alongwith the hyperspectral images. To realise this, we design a hallucination network based on conditional GANs, that generates the features of hidden modality from those of available modalities. In addition, we design a knowledge distillation based framework to transfer the knowledge from full-modal setting (during pre-training stage), to partial modal setting (during deployment stage). The next two works focus on joint classification of HSIs with complementary modalities such as LiDAR and SAR. The first research work in this regards utilizes modality specific attention mechanism to identify which modalities are more in line with the available classes for the classification task. In the second work, we introduce the notion of cross-modal and intra-modal communication using shared weights. Additionally, we also enhance the model robustness using the idea of self-supervision across the participating modalities. Our model performs exemplarily on several benchmark HSI-LiDAR and HSI-SAR datasets for LULC classification.  

Thirdly, we exploit the paradigm of self-supervised learning in the domain of HSI classification to address the issue of `curse of dimensionality' and limited training samples. To this end, we employ the notion of generative SSL and contrastive SSL for HSI analysis. From the generative perspective, we address the problem of dimensionality reduction in HSIs. We design several autoencoders, that address the sequential and spatial characteristics in the HSI using 1D CNNs, attention mechanism, feedback connections and bidirectional GRUs. We also design a contrastive learning based framework to pre-train a CNN based model using guidance from a few labelled samples, and then deploy the model on the downstream classification tasks. All our approaches gave decent performance on several HSI datasets.

In the subsequent sections, we will see different directions in which we can further explore the problem of HSI classification. 

\section{Future directions}

\subsection{Few shot HSI and LiDAR fusion using meta learning (FSL)}

\begin{figure}[htb]
  \centering
  \centerline{\includegraphics[width=14.0cm]{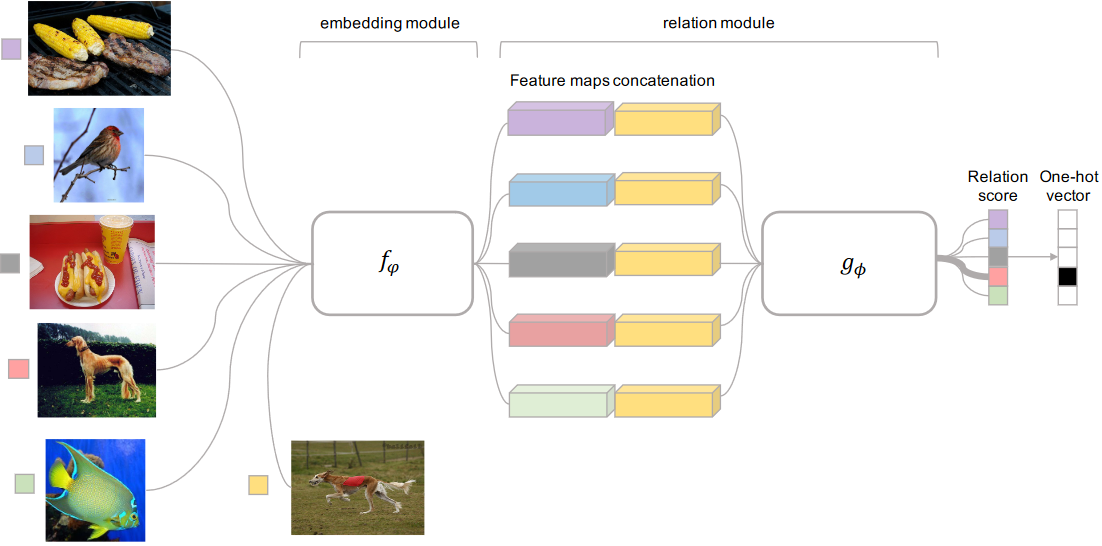}}
\caption[Schematic of Relational Network.]{Schematic of Relational Network for few shot learning \citep{sung2018learning}}
\label{fig:RLN}
\end{figure}

It has been observed in deep learning techniques that even though the methods give reliably good performance in the task of classification, there is a downside of requirement of a high amount of data owing to the problem of overfitting. In case of geospatial domain, collecting the real time data can often become a time consuming and tedious since it requires specialised equipment and onsite presence of the surveyor. To address this problem, a meta learning approach in the deep learning domain has been developed that comes under the umbrella term of few shot learning. Several successful methods such as matching networks \citep{vinyals2016matching}, prototypical networks \citep{snell2017prototypical}, relational networks \citep{sung2018learning} and the more recent ones such as reptile \citep{clouatre2019figr}. The idea here is to have a model (or set of models) whose weights are updated on a dataset that has a good number of training samples. This model is then deployed on the dataset where the groundtruth is available is only for a few samples, and prediction has to be made on the remaining samples. The prediction on an unknown sample is then made by comparing its embedding (obtained from the trained model) with the embedding of the groundtruth sample (again, obtained from the trained model). The sample is assigned the class which is closest to it in the embedding space. This kind of learning paradigm is also referred as meta learning, since the learning has been carried out on entirely different set of classes while predictions are made on different set of classes. A schematic of relational network is provided in Figure \ref{fig:RLN}.

In the further research problem, the idea of few shot learning with meta-learning regime would be explored in the problem of multimodal fusion of HSI and LiDAR data. This is further owing to the reason that in multimodal learning scenario, the data collection has to be carried using multiple specialised equipment. This make the task of data collection even more tedious an may also require more trained professionals. Furthermore, the problem of HSI and LiDAR fusion is still not tackled in the FSL scenario (even though single modal HSI classification has been attempted such as in \citep{tang2019spatial} and \citep{xi2020deep}), which makes it a good opportunity to establish novel benchmarks in this area. 

\subsection{Patch-free self-supervised learning for hyperspectral image classification}

\begin{figure}[htb]
  \centering
  \centerline{\includegraphics[width=14.0cm]{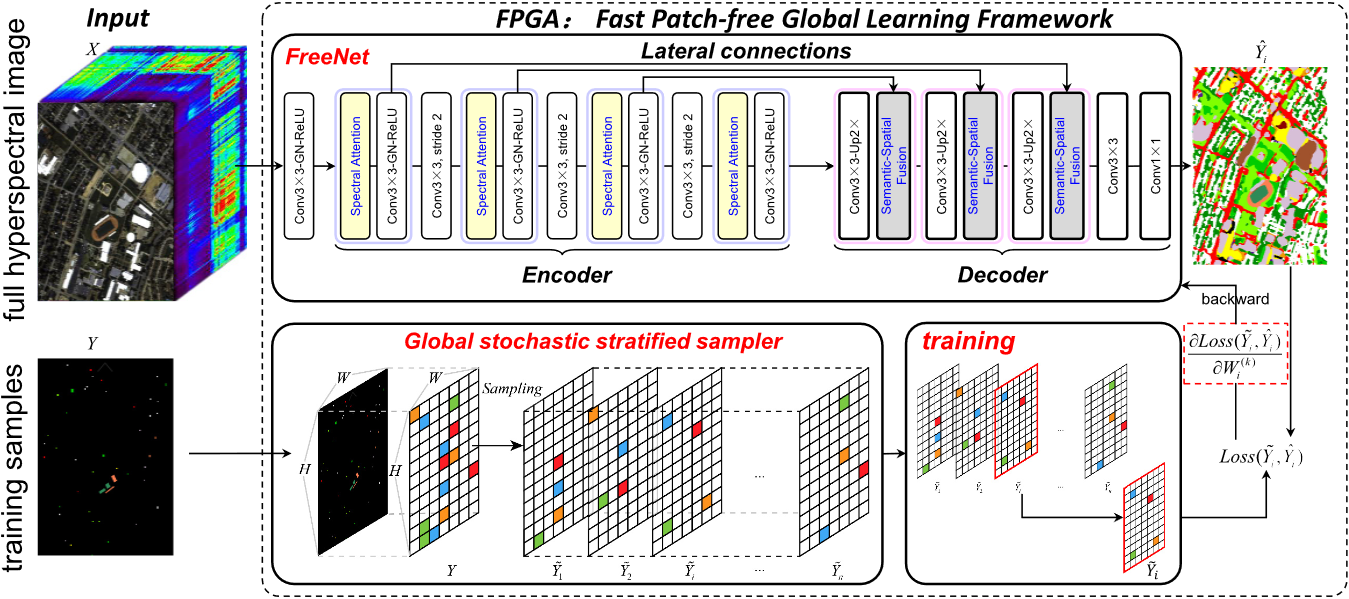}}
\caption[Schematic of FreeNet for HSI classification.]{Schematic of FreeNet for HSI classification \citep{zheng2020fpga}.}
\label{fig:fpga}
\end{figure}

In HSI classification, most of the methods till now follow a patch based approach for training the classification models. Even though the approach has largely been successful in incorporating both spectral and spatial information in the learning process, there are certain problems associated with the process. First of all, in most of the cases, when there is only single image for training and validation, there is always the possibility of overlapping of the training patches and test patches. Secondly, overlapping patches share a lot of geographical area, leading to redundant computations in a deep learning model. Recently, there have been approaches that try to move away the conventional path of patch based learning, and incorporate a patch-free learning framework for HSI classification. For instance, \citep{zheng2020fpga} introduced a method that approaches the problem of pixel based classification in HSI in a semantic segmentation based approach. The proposed method FreeNet utilizes a global stochastic stratified sampling strategy for sampling out the groundtruth masks for the image, that is passed through the network in one shot. Similar works have also been presented in \citep{liu2021patch} and \citep{yu2022cross}, where a segmentation based approach has been proposed for HSI classification. This way, the training time of the network is highly reduced. The schematic of the approach is presented in Figure \ref{fig:fpga}. 

However, since only a single is available for training the network, it becomes intuitively difficult to apply conventional self-supervised learning techniques in such a learning framework, which was otherwise applicable in patch-based scenarios. Therefore, to enhance the feature representations in such a setting, it is an open problem to design effective SSL pretraining techniques for such a scenario.

\subsection{Multimodal point-cloud processing in remote sensing}

In the last decade, remote sensing field has seen a lot of work from the perspective of image based modalities (MSI, HSI, SAR etc.) both in the single-modal and multimodal scenario. Owing to this, the existing research works focussed on these modalities have approached the problem statements from several learning paradigms such as SSL, supervised learning, meta-learning and several others. However, because of limited data available for LiDAR sensors/Laser scanners, the development in that areas have been relatively slow. And even if LiDAR data/point clouds are used with the other sensors, it is mostly in the form of digital surface models (DSM). This is on account of different data structuring in point clouds (which are of geometrical/unstructured nature) and images, which are gridded and structured. 
\begin{figure}[htb!]
  \centering
  \centerline{\includegraphics[width=14.0cm]{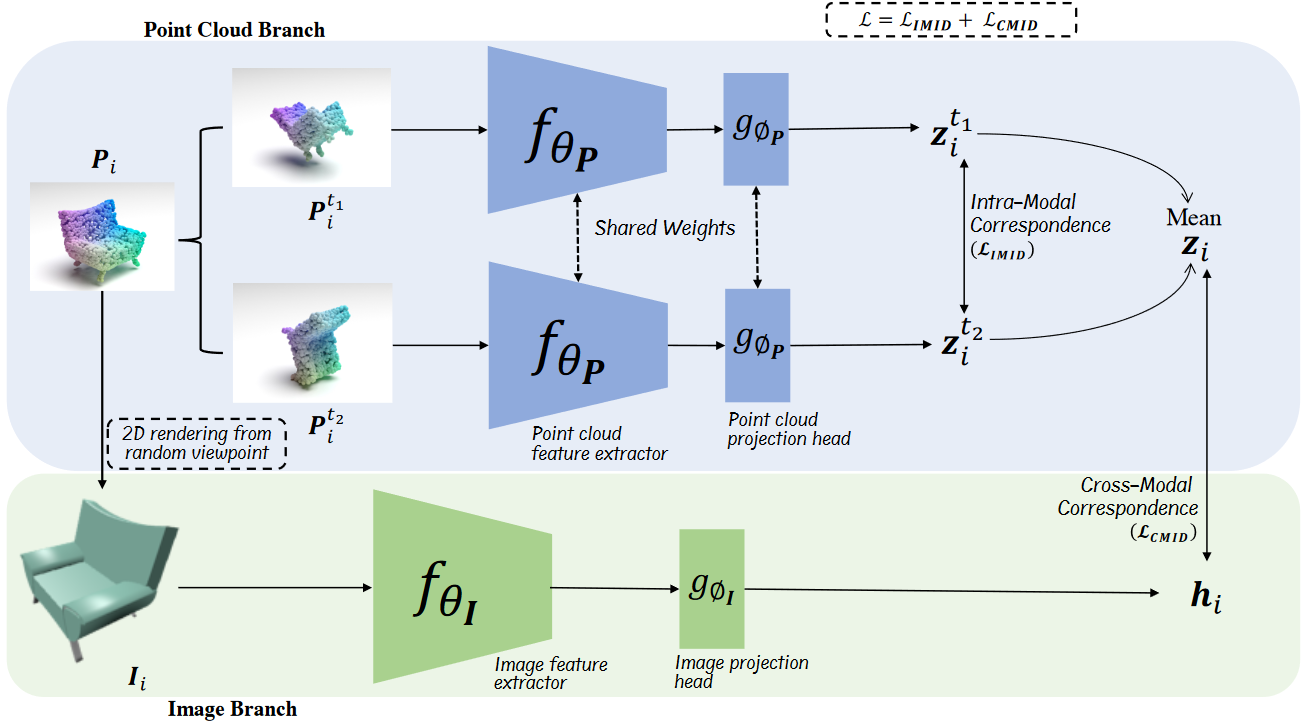}}
\caption[Schematic of CrossPoint, contrastive learning framework for point clouds and images.]{Schematic of CrossPoint, contrastive learning framework for point clouds and images \citep{afham2022crosspoint}.}
\label{fig:pcssl}
\end{figure}
However, recently point clouds and laser scanners are finding more and more utility in remote sensing domain, owing to their effectiveness in 3D modelling of the terrain and urban landscapes, and providing elevation information. There have been some limited works on multimodal point cloud analysis in RS domain. For instance, \citep{brell20193d} proposed an method of hyperspectral point cloud generation by fusion the spectrum information for HSI and the co-ordinate information from LiDAR data. \citep{poliyapram2019point} proposed an end-to-end network for 3D point cloud fusion with aerial images for point cloud segmentation. Therefore, a large research gaps exists in remote sensing domain in utilising point-cloud data with structured modalities and non-image based modalities. 

Furthermore, building on this idea, there is also a vast scope of implementing the techniques of SSL and few-shot learning in conjuction with the multimodal pointcloud analysis. Recently, in vision domain, there have been some parallel works that have been attempting to bridge the gap between the image and point cloud data. For instance, \citep{afham2022crosspoint} and \citep{wu2023self} use contrastive learning based pretraining to match the point-cloud objects with the corresponding rendered images, and then used the point-cloud specific branch for segmentation tasks. This can be seen in Figure \ref{fig:pcssl}. So, to this end, my further research work will focus on the multimodal processing of point cloud data for remote sensing in weakly supervised/limited label setting.



\newpage
\setlength{\parskip}{5mm}
\titlespacing{\chapter}{0cm}{0mm}{0mm}
\titleformat{\chapter}[display]
  {\normalfont\huge\bfseries}
  {\chaptertitlename\ \thechapter}{20pt}{\Huge}

\bibliographystyle{References/elsarticle-harv}
\bibliography{References/mybibfile}

\chapter*{List of Publications}
\label{ch:pub}
\addcontentsline{toc}{chapter}{\nameref{ch:pub}}

\section*{Journals}
\addcontentsline{toc}{section}{Journals}

\begin{enumerate}
\item \textbf{S. Pande}, B. Banerjee, Self-supervision assisted multimodal remote sensing image classification with coupled self-looping convolution networks. Neural Networks, 164 (2023), 1-20. 
\item \textbf{S. Pande}, B. Banerjee, HyperLoopNet: Hyperspectral image classification using multiscale self-looping convolutional networks, ISPRS Journal of Photogrammetry and Remote Sensing, 183 (2022) 422–438.
\item \textbf{S. Pande}, B. Banerjee, Adaptive hybrid attention network for hyperspectral image classification, Pattern Recognition Letters, 144 (2021) 6–12.
\item A. K. Patel, J. K. Ghosh, \textbf{S. Pande}, S. U. Sayyad, Deep-learning-based approach for estimation of fractional abundance of nitrogen in soil from hyperspectral data, IEEE Journal of Selected Topics in Applied Earth Observations and Remote Sensing, 13 (2020) 6495–6511.
\end{enumerate}


\section*{Conferences}
\addcontentsline{toc}{section}{Conferences}

\begin{enumerate}
\item	H. Pal, R. Khandelwal, \textbf{S. Pande}, S. Karanam, B. Banerjee, Cross Domain 3D Shape Retrieval from Monocular Images, Winter Conference on Applications of Computer Vision, 2024.
\item \textbf{S. Pande}, N. A. A. Braham, Y. Wang, Conrad M. Albrecht, B. Banerjee, X. X. Zhu, Semi-supervised learning for hyperspectral images by non-parametrically predicting view assignments, accepted in: IEEE International Geoscience and Remote Sensing Symposium, 2023.
\item J. Songara, \textbf{S. Pande}, S. Choudhury, B. Banerjee, R. Velmurugan, Visual question answering in remote sensing with cross-attention and multimodal information bottleneck, accepted in: IEEE International Geoscience and Remote Sensing Symposium, 2023.
\item \textbf{S. Pande}, B. Banerjee, Feedback convolution based autoencoder for dimensionality reduction in hyperspectral images, in: IGARSS 2022-2022 IEEE International Geoscience and Remote Sensing Symposium, IEEE, 2022.
\item Kumar, D. Tamboli, \textbf{S. Pande}, B. Banerjee, RSINet: Inpainting remotely sensed images using triple GAN framework, in: IGARSS 2022-2022 IEEE International Geoscience and Remote Sensing Symposium, IEEE, 2022.
\item R. Bose, \textbf{S. Pande}, B. Banerjee, Two headed dragons: Multimodal fusion and cross modal transactions, in: 2021 IEEE International Conference on Image Processing (ICIP), IEEE, 2021, pp. 2893–2897.
\item \textbf{S. Pande}, B. Banerjee, Attention based convolution autoencoder for dimensionality reduction in hyperspectral images, in: 2021 IEEE International Geoscience and Remote Sensing Symposium IGARSS, IEEE, 2021, pp. 2727–2730.
\item \textbf{S. Pande}, B. Banerjee, Bidirectional GRU based autoencoder for dimensionality reduction in hyperspectral images, in: 2021 IEEE International Geoscience and Remote Sensing Symposium IGARSS, IEEE, 2021, pp. 2731–2734.
\item A. Jha, A. Kumar, \textbf{S. Pande}, B. Banerjee, S. Chaudhuri, MT-UNet: A novel U-Net based multi-task architecture for visual scene understanding, in: 2020 IEEE International Conference on Image Processing (ICIP), IEEE, 2020, pp. 2191–2195.
\item \textbf{S. Pande}, B. Banerjee, Dimensionality reduction using 3D residual autoencoder for hyperspectral image classification, in: IGARSS 2020-2020 IEEE International Geoscience and Remote Sensing Symposium, IEEE, 2020, pp. 2029–2032.
\item S. Mohla, \textbf{S. Pande}, B. Banerjee, S. Chaudhuri, FusAtNet: Dual attention based spectrospatial multimodal fusion network for hyperspectral and LiDAR classification, in: Proceedings of the IEEE/CVF Conference on Computer Vision and Pattern Recognition (CVPR) Workshops, 2020, pp. 92–93.
\item \textbf{S. Pande}, B. Banerjee, Aleksandra Pi\v{z}urica, Class reconstruction driven adversarial domain adaptation for hyperspectral image classification, in: Iberian Conference on Pattern Recognition and Image Analysis, Springer, 2019, pp. 472–484.
\item \textbf{S. Pande}, A. Banerjee, S. Kumar, B. Banerjee, S. Chaudhuri, An adversarial approach to discriminative modality distillation for remote sensing image classification, in: Proceedings of the IEEE/CVF international conference on computer vision workshops, 2019.
\end{enumerate}


\newpage
\setlength{\parskip}{5mm}
\titlespacing{\chapter}{0cm}{0mm}{0mm}
\titleformat{\chapter}[display]
  {\normalfont\huge\bfseries \centering}
  {\chaptertitlename\ \thechapter}{20pt}{\Huge}

\chapter*{Acknowledgements}
\label{ch:Acknowledgements}
\addcontentsline{toc}{chapter}{\nameref{ch:Acknowledgements}}
\vspace{10mm}
I would like to express my sincere gratitude to \textbf{Prof. Biplab Banerjee}, my esteemed supervisor, whose guidance and mentorship have been invaluable throughout my PhD journey. Without his unwavering support, the completion of my research work would not have been possible. I am also deeply thankful to the faculty members who imparted their knowledge and expertise in the subjects directly relevant to my research area. Their teachings have significantly contributed to shaping my understanding and expanding my horizons in the field. I would also like to extend my gratitude to \textbf{Prof. Xiao Xiang Zhu}, who gave me an opportunity to work in International Futures Lab, TU Munich, Germany as a Beyond Fellow, and \textbf{Dr. Conrad M. Albrecht} and \textbf{Prof. Sudipan Saha}, who provided research supervision and moral support during my stay in Germany. I also extend heartfelt gratitude to \textbf{Prof. Aleksandra Pi{\v{z}}urica} for her guidance and collaboration in one of my research works on domain adaptation during my PhD tenure.

Additionally, I am also thankful to the members of the RPC, \textbf{Prof. Krishna Mohan Buddhiraju} and \textbf{Prof. Indu Jaya}, for graciously agreeing to evaluate my research work. Their expertise and insights have played a crucial role in refining the quality of my study. 

I would also like to thank my several student co-authors and collaborators in the research work, namely \textbf{Avinandan Banerjee}, \textbf{Saurabh Kumar}, \textbf{Satyam Mohla}, \textbf{Rupak Bose}, \textbf{Dipesh Tamboli}, \textbf{Jayesh Songara}, \textbf{Advait Kumar} and \textbf{Harsh Pal} from IIT Bombay, and \textbf{Yi Wang} and \textbf{Nassim Ait Ali Braham} from TU Munich. I am also thankful to my labmates and colleagues from IIT Bombay and TU Munich for their moral support, and technical and conceptual assistance during PhD. The names are in no particular order. IIT Bombay: \textbf{Prof. Subhadip Dey}, \textbf{Dr. Ushasi Chaudhuri}, \textbf{Dr. Rajat Shinde}, \textbf{Maitreya Mohan Sahoo}, \textbf{Ankit Jha}, \textbf{Sayan Rakshit}, \textbf{Shabnam Choudhury}, \textbf{Debabrata Pal}, \textbf{Priyanka Talwar}, \textbf{Pratyush Talreja}, \textbf{Awanish Kumar}, \textbf{Valay Bundele}, \textbf{Mainak Singha}, \textbf{Sehajpal Singh} and \textbf{Shabana Ibrahim}. TU Munich: \textbf{Shahriar Kabir}, \textbf{Qasim Khan}, \textbf{Dawood Wasif}, \textbf{Shan Zhao}, \textbf{Xiangyu Zhao}, \textbf{Lixian Zhang}, \textbf{Dr. Miguel-Ángel Fernández-Torres}, \textbf{Chenying Liu}, \textbf{Fahong Zhang} and \textbf{Julia Kollofrath}. I am also indebted to the administrative staff of CSRE and IIT Bombay, alongwith the hostel and mess staff for their support and assistance. 

I am also thankful to the Almighty to grant me the means to pursue and finish my PhD, and my parents, sister, family members and friends for their unwavering support.

\end{document}